\documentclass{article}

\usepackage{microtype}
\usepackage{graphicx}
\usepackage{subcaption}
\usepackage{booktabs} %

\usepackage{hyperref}

\usepackage{url}
\usepackage{amsmath}
\usepackage{amssymb}
\usepackage{bm}
\usepackage{bbm}
\usepackage{xspace}
\usepackage{minitoc}
\usepackage{graphicx}
\usepackage{subcaption}
\usepackage[percent]{overpic}
\usepackage{fontawesome}
\usepackage[dvipsnames]{xcolor}
\usepackage{placeins}

\usepackage[accepted]{icml2025}

\usepackage{mathtools}
\usepackage{amsthm}

\usepackage[capitalize,noabbrev]{cleveref}

\theoremstyle{plain}

\theoremstyle{definition}

\theoremstyle{remark}

\usepackage{xspace}

\newcommand{\change}[1]{#1}

\newcommand{\cl}[1]{#1}

\newcommand*{\ie}{i.e.,\@\xspace}
\newcommand*{\eg}{e.g.,\@\xspace}
\newcommand*{\etc}{etc.\@\xspace}

\newcommand{\search}{\faSearch\@\xspace}

\newcommand{\modelname}{FS-SINR\@\xspace}

\icmltitlerunning{Feedforward Few-shot Species Range Estimation}

\begin{document}

\twocolumn[
\icmltitle{Feedforward Few-shot Species Range Estimation}

\icmlsetsymbol{equal}{*}

\begin{icmlauthorlist}
\icmlauthor{Christian Lange}{eee}
\icmlauthor{Max Hamilton}{xxx}
\icmlauthor{Elijah Cole}{zzz}
\icmlauthor{Alexander Shepard}{nnn}
\icmlauthor{Samuel Heinrich}{ccc}
\icmlauthor{Angela Zhu}{xxx}
\icmlauthor{Subhransu Maji}{xxx}
\icmlauthor{Grant Van Horn}{xxx}
\icmlauthor{Oisin Mac Aodha}{eee}
\end{icmlauthorlist}

\icmlaffiliation{xxx}{UMass Amherst}
\icmlaffiliation{nnn}{iNaturalist}
\icmlaffiliation{ccc}{Cornell}
\icmlaffiliation{zzz}{GenBio AI}
\icmlaffiliation{eee}{University of Edinburgh}

\icmlcorrespondingauthor{Christian Lange}{c.p.lange@sms.ed.ac.uk}

\icmlkeywords{species distribution modeling, SDM, species range modelling, spatial implicit neural representation, SINR, low-shot learning, few-shot learning}

\vskip 0.3in
]

\printAffiliationsAndNotice{}  %

\begin{abstract}
Knowing where a particular species can or cannot be found on Earth is crucial for ecological research and conservation efforts. 
By mapping the spatial ranges of all species, we would obtain deeper insights into how global biodiversity is affected by climate change and habitat loss. 
However, accurate range estimates are only available for a relatively small proportion of all known species. 
For the majority of the remaining species, we typically only have a small number of records denoting the spatial locations where they have previously been observed. 
We outline a new approach for few-shot species range estimation to address the challenge of accurately estimating the range of a species from limited data. 
During inference, our model takes a set of spatial locations as input, along with optional metadata such as text or an image, and outputs a species encoding that can be used to predict the range of a previously unseen species in a feedforward manner. 
We evaluate our approach on two challenging benchmarks, where we obtain state-of-the-art range estimation performance, in a fraction of the compute time, compared to recent alternative approaches. 
\end{abstract}

\section{Introduction}
\label{Introduction}

Understanding the spatial distribution of plant and animal species is essential to mitigate the ongoing decline in global biodiversity~\citep{jetz2019essential}. 
Monitoring these distributions over time allows us to quantify the impacts of climate change, habitat loss, and conservation interventions~\citep{mantyka2012interactions}. 
Estimating a species' spatial distribution typically starts with collecting a set of observations that denote the locations where the species has been confirmed to be present or absent. 
Traditionally, this data is used to train models that can then generate detailed predictions over a spatial region of interest~\citep{elith2006novel,beery2021species}.
When sufficient data is available, these models enable practitioners to estimate important quantities such as the spatial range (\ie where a species can be found) or abundance (\ie the total number of individuals) of a species, in addition to quantifying how these quantities are changing over time.

\begin{figure}[t]
\centering
\includegraphics[width=\columnwidth]{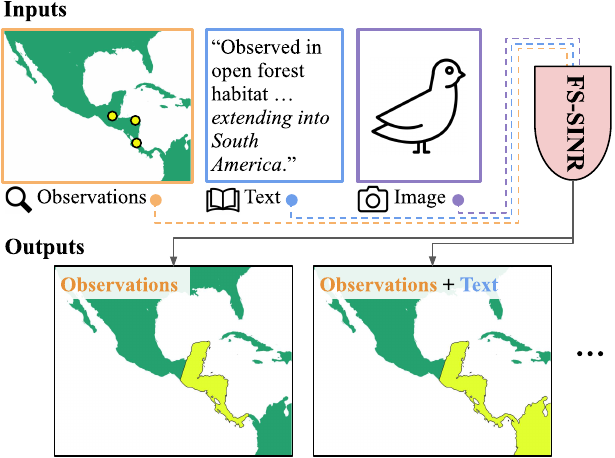} 
\vspace{-20pt}
\caption{ %
{\bf Few-shot species range estimation with FS-SINR}.
Our \modelname approach is trained on citizen science collected species observation data (\ie locations where a species has been observed), and once trained,  can  estimate the spatial range of a previously \emph{unseen} species with a single forward pass through the model, with no retraining required at inference time. 
It supports different input modalities such as variable length sequences of location observations,  in addition to other metadata such as text or images. 
In this illustration, we show two different range predictions: one using only location observations (bottom left) and the other using observations and text (bottom right). 
}
\vspace{-10pt}
\label{fig:overview}
\end{figure}

Despite the availability of well-established modeling techniques, our current understanding of species' distributions is extremely limited as little or no observational data is available for most species.
For example, iNaturalist,  one of the largest citizen science platform documenting global biodiversity, has collected over 130 million research quality observations for approximately 373,000 species globally~\citep{inatWeb}. 
However, the data is severely long-tailed, \ie a small percentage of common species account for the majority of the observations, while many species have very few observations. 
In fact, over half of the 373,000 species cataloged by iNaturalist have been observed fewer than ten times to date. 
This data limitation is amplified by the fact that the vast majority of the several million species that are thought to exist have not yet even been documented by science~\citep{mora2011many}. 
Identifying locations where under-observed species can be found is a time-consuming and laborious process, often requiring long expeditions to remote locations to search for species that are hard to find. 
Consequently, there is a pressing need for computational methods that can reliably estimate the spatial distributions of species using only a small number of observations.

Knowing the range of one species can help predict the range of another due to shared ecological, environmental, and geographic contexts.
Recent advances in range estimation, \cl{such as Spatial Implicit Neural Representations (SINR)}, have leveraged this idea by training  on millions of observations, across tens of thousands of species, inside one model~\citep{cole2023spatial}.
However, these models still rely on relatively large numbers of training observations for individual species, which limits their applicability to species with limited observations.
In this work, we introduce \cl{Few-shot Spatial Implicit Neural Representations} (\modelname), a novel Transformer-based model that overcomes this limitation and offers two key advantages over previous approaches.
First, we obtain \cl{improved} performance in the few-shot regime,  a scenario that represents the reality for the majority of species, yet remains underexplored in prior work. 
Second, we make accurate predictions for species not present in the training set without any additional training, which can enable interactive exploration and modeling.
At inference time, we only require a set of observed locations for the unseen species to generate reliable range estimates. Furthermore, we show we can flexibly incorporate additional non-geographic context information (\eg a text summary of the species' habitat or range preferences or an image of the species) to further improve prediction quality.
\cref{fig:overview} illustrates how \modelname can be used at inference time.

In summary, we make the following  contributions: 
(i)~We introduce \modelname, a new approach for few-shot species range estimation. \modelname has novel capabilities, including the ability to predict the spatial range of a previously unseen species at inference time without requiring any retraining. 
(ii)~We demonstrate that \modelname achieves state-of-the-art performance in the few-shot setting on the challenging IUCN and S\&T benchmark datasets. 
(iii) We provide detailed ablation studies and visualizations to highlight the benefits of integrating observational data with textual and visual context, as well as to compare our approach with alternative methods.

\section{Related Work}
{\bf Species Distribution Modeling}. 
Estimating the spatial distribution of a species is a widely explored topic in both statistical ecology and machine learning~\citep{beery2021species}. 
The goal is to develop models that can predict the distribution of species over space, and possibly time, given sparse observation data. 
Different machine learning approaches, initially using traditional techniques, such as decision trees among others have been extensively explored, \eg~\citep{phillips2004maximum,elith2006novel}. 
More recently,  deep learning-based methods have been introduced~\citep{botella2018deep,mac2019presence,cole2023spatial,kellenberger2024performance}. 
One of the strengths of these deep methods is that they can jointly represent thousands of different species within the same model and have been shown to improve as more training data is added, even when the data is from different species~\citep{cole2023spatial}. 

There has also been work investigating different approaches to address some of the challenges associated with training and evaluating these models.  
Examples include attempts to address imbalances across species in the training observation data~\citep{zbinden2024imbalance}, sampling pseudo-absence data~\citep{zbinden2024selection}, biases in training locations~\citep{chen2019bias}, representing location information~\citep{russwurm2023geographic}, discretizing continuous model predictions~\citep{binary_maps_cv4e_2024}, active learning approaches~\citep{lange2024active}, using additional metadata such as images~\citep{teng2023bird,dollinger2024sat,picek2024geoplant} or text~\citep{sastry2023ld,sastry2025taxabind,hamilton2024}, and designing new evaluation datasets to benchmark performance~\citep{cole2023spatial,picek2024geoplant}. 
In our work, we investigate the underexplored \emph{few-shot} setting, where only limited observations (\eg fewer than ten) are available for each species at training time.

{\bf Few-shot Species Range Estimation}. 
There are several aspects of the species range estimation task in the low-data regime that make it different from other few-shot problems more commonly explored in the literature~\citep{parnami2022learning,wang2020generalizing}. 
For example, the input domain is fixed (\ie all locations on earth), each location can support more than one species (\ie multi-label instead of multi-class), the label space is much larger (\ie tens of thousands of species as opposed to hundreds of classes in image classification), and only partial supervision is available (\eg presence-only data, with no confirmed absences).

\begin{figure*}[t]
\centering
\includegraphics[width=1.0\textwidth]{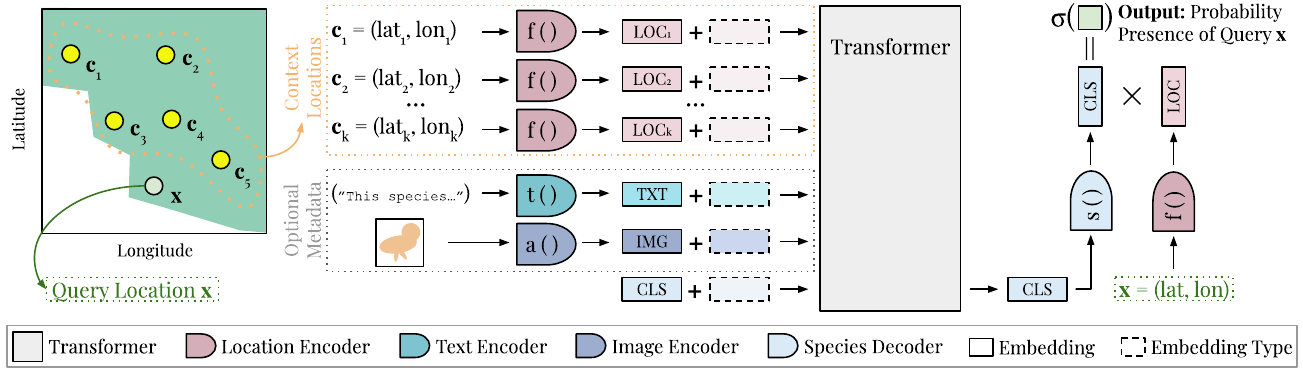}
\vspace{-15pt}
\caption{{\bf \modelname overview}. Here we depict our few-shot species range estimation model. 
The input consists of an arbitrary number of  context locations $\mathcal{C}^t$ for target species $t$ that are each independently tokenized using a location encoder $f_{\bm{\theta}}()$, and optional auxiliary context information like text or an image. 
A class token (\texttt{CLS}) is also appended to the input. 
All input tokens are processed by a Transformer $m_{\bm{\psi}}()$. 
Given the set of input context locations, we estimate the probability that a species is present at a query location $\mathbf{x}$ by multiplying the location encoder's embedding of $\mathbf{x}$ with the projected embedding of the $\texttt{CLS}$ token which is output from the species decoder. 
}
\label{fig:method}
\vspace{-10pt}
\end{figure*}

\citet{lange2024active} introduced an active learning-based approach for species range estimation which makes predictions based on linear combinations of learned species embeddings and showed its effectiveness in the few-shot regime. 
LE-SINR~\citep{hamilton2024} showed that internet sourced free-form text descriptions of species' ranges can be used when training models for zero-shot range estimation. 
They applied their approach to the few-shot setting, but it requires retraining a classifier for each new species observation added. 
In our evaluation, we demonstrate that our \modelname approach, which can incorporate additional metadata at training time and does not require retraining during inference, outperforms existing methods.

\section{Methods}
We first set up the species range estimation problem and then describe our approach for few-shot range estimation.

\subsection{Species Range Estimation}
We start by describing the SINR approach from~\citet{cole2023spatial}. 
Let $\bm{x} = (\text{lat}, \text{lon}) \in \mathcal{X}$ be a location of interest sampled from a spatial domain $\mathcal{X}$ (\eg a location on earth). 
Our goal is to train a model $g(): \mathcal{X} \to [0, 1]^s$ to predict the probabilities of $s$ different species of interest occurring at $\bm{x}$. 
We let $\hat{\bm{y}} = g(\bm{x})$, where $\hat{y}_j \in [0, 1]$ (\ie the $j^{th}$ entry of $\hat{\bm{y}}$) represents the probability that species $j$ occurs at location~$\bm{x}$. 

We can decompose the model as $g() = h_{\bm{\phi}}() \circ f_{\bm{\theta}}()$, where $f_{\bm{\theta}}(): \mathcal{X} \rightarrow \mathbb{R}^d$ is a location encoder with parameters $\bm{\theta}$ and $h_{\bm{\phi}}(): \mathbb{R}^d \rightarrow [0, 1]^s$ is a multi-label classifier with parameters $\bm{\phi}$. 
The location encoder $f_{\bm{\theta}}()$ maps a location $\bm{x}$ to a $d$-dimensional latent embedding $f_{\bm{\theta}}(\bm{x})$. 
The multi-label classifier $h()$ is implemented as a per-species linear projection followed by an element-wise sigmoid non-linearity, meaning that $\hat{\bm{y}}  = \sigma(f_{\bm{\theta}}(\bm{x})\bm{W})$, where $\bm{W} \in \mathbb{R}^{d \times s}$ (\ie $h_{\bm{\phi}}() = \bm{\phi} = \bm{W}$) and $\sigma()$ is the sigmoid function. Thus, each column vector $\bm{w}_j$ of $\bm{W}$ can be viewed as a species embedding, which we can combine with a location embedding $f_{\bm{\theta}}(\bm{x})$ via an inner product to compute the probability that the species $j$ is present at $\bm{x}$. Importantly, the location embedding is shared across all species.  
Once trained, it is possible to generate a prediction for a given species for all locations of interest by evaluating the model for all locations (\ie $\bm{x} \in \mathcal{X}$). 

One of the main challenges associated with training models for species range estimation is that there is a dramatic asymmetry in the available training data.
Specifically, it is much easier to collect presence observations (\ie confirmed sightings of a species) compared to absence observations (\ie confirmation that a species is not present at a specific location). 
As a result, many methods have been developed to train models using \emph{presence-only} data. 
In the presence-only setting, we have access to training pairs $(\bm{x}, z)$, where $\bm{x}$ is a geographic location, and $z \in \{1,\dots,s\}$ is an integer indicating which species was observed there. 
To overcome the lack of confirmed absence data, one common approach is to generate \emph{pseudo-absences} by sampling random locations on the surface of the earth~\citep{phillips2009sample}.   
Given these pseudo-absences, the parameters of $g()$ can be trained in an end-to-end manner using variants of the cross-entropy loss. 
Specifically, we use \emph{full assume negative loss} %
from \citet{cole2023spatial} to train the SINR baseline: 
\begin{multline}
\mathcal{L}_{\text{AN-full}}(\hat{\bm{y}}, z) = -\frac{1}{s} \sum_{j=1}^{s} [ \mathbbm{1}_{[z=j]} \lambda \log(\hat{y}_{j}) + \\
 \mathbbm{1}_{[z \neq j]} \log(1 - \hat{y}_{j}) + \log(1 - \hat{y}_j') ],
\end{multline}

where $z$ is the index of the species present for a given training instance, $\hat{y}_{j}$ is the predicted probability of the presence of species $j$,  $\hat{y}_j'$ is the model prediction for a randomly sampled pseudo-absence location, and the hyperparameter $\lambda$ balances the presence and pseudo-absence loss terms.

\subsection{Few-shot Range Estimation} 
For the SINR model to make predictions for a new species, it is necessary to learn a new embedding vector $\bm{w}_j$ for that species. %
If additional location data is later observed for that species, the model must be updated again. 
However, the number of observations for rarer species can be limited and thus it is necessary to have methods that can be updated efficiently with less training data. 

We address this challenge by proposing a new approach for few-shot species range estimation called \modelname.
Our model can predict the probability of presence for a previously unobserved species directly at inference time given only the set of confirmed presence locations available, without any retraining or parameter updates. 
At inference time, we assume we have access to a set of context locations
$\mathcal{C}^t = \{\bm{c}_1, \dots, \bm{c}_k\}$, which represent a set of $k$ locations where the species $j$ has been confirmed to be present. 
Each entry in this set denotes a geographic location, \ie $\bm{c} = (\text{lat}, \text{lon})$. 
Like SINR, our model is also conditioned on a location $\bm{x}$ of interest (\ie the `query' location), but uses the context locations to inform the prediction for the query location. 
Note, the context locations can come from a species not previously observed  during training. 

We represent our model as $g(\bm{x}) = m_{\bm{\psi}}(f_{\bm{\theta}}(\bm{x}), \mathcal{C}^t)$. 
Unlike in SINR, where the classifier head $h_{\bm{\phi}}()$ is a simple multi-label classifier and sigmoid non-linearity, in our case, the `head' of the model $m_{\bm{\psi}}()$ is a Transformer-based encoder~\citep{vaswani2017attention}. 
\modelname  takes an unordered set of context locations $\mathcal{C}^t$ as input, where each location is encoded into an embedding vector (\ie a token) via a SINR-style multi-layer perceptron location encoder -- see \cref{fig:method} for an illustration. 
Importantly, our model \cl{can accept a variable number of context locations and} is invariant to their ordering as we do not append any positional embeddings.
This flexibility ensures that it can process a variable number of context locations during inference. 
We also append an additional register token (\texttt{REG}) as in~\citet{darcet2023vision} to provide the model with an additional token to `store' information. 
Given that the input sequence is unordered and may or may not include additional context information, we add learned `embedding type' vectors to each token such that the Transformer knows if a given input token is a location, register, text, image, \etc 

We represent the species embedding vector (\ie $\bm{w}_j$ in SINR) as the class token \texttt{CLS} of the Transformer after passing it through a small species decoder MLP $s()$. 
To make a final prediction, we simply compute the inner product between the location embedding of the query location $\bm{x}$ and the species embedding vector, and pass it through a sigmoid. 
Our approach is computationally efficient in that once the species embedding is generated it can then be efficiently multiplied by the embeddings for all locations of interest to generate a prediction for a species' range. 

\modelname uses a similar training loss to $\mathcal{L}_\mathrm{AN-full}$. 
However, since it has no equivalent to $h_{\bm{\phi}}()$ we cannot easily include all species in the loss, and instead consider only those within the same batch of training examples of size $s^{b}$. 
We obtain a predicted species embedding vector for a given species during the forward pass which can be used to estimate the probabilities of presence of that species for all locations sampled in the batch. 
We denote this new loss as $\mathcal{L}_\mathrm{AN-full-b}$, which indicates that we are considering only those elements contained within the current batch $b$:
\begin{multline}
\mathcal{L}_{\text{AN-full-b}}(\hat{\bm{y}}, z^{b})  = -\frac{1}{s^{b}} \sum_{j=1}^{s^{b}} [ \mathbbm{1}_{[z^{b}=j]} \lambda \log(\hat{y}_{j}) + \\ 
\mathbbm{1}_{[z^{b} \neq j]} \log(1 - \hat{y}_{j}) + \log(1 - \hat{y}_j') ].
\end{multline}

\subsubsection{Additional Context Information} 
The design of \modelname is flexible in that we can also provide additional context information to the model if it is available. 
For example, if there is additional text (\eg a range description) or visual (\ie images) information available for a novel species, it can be added to the context, assuming that such information was also available at training time for other species. 
This observation is inspired by recent work that also uses language-derived information to improve range predictions~\citep{sastry2023ld,hamilton2024} and work that uses species images and observations~\citep{sastry2025taxabind}.   
This additional information can provide a rich source of metadata encoding aspects of a species' habitat preferences, even when there might only be a limited number of location observations available for it.  
We can represent the expanded contextual input tokens as $\{\bm{t}_j, \bm{a}_j, f_{\bm{\theta}}(\bm{c}_1), \dots, f_{\bm{\theta}}(\bm{c}_k)\}$, where $\bm{t}_j$ denotes a fixed-length text embedding from a large language model and $\bm{a}_j$ an image embedding obtained from a pre-trained vision model for species $j$  -- see \cref{fig:method}.  
Note that we train \modelname so that it can use arbitrary subsets, including none, of these input tokens during inference.

\section{Experiments}
Here we evaluate \modelname on the task of species range estimation and compare it to alternative methods. 

\subsection{Implementation Details}

{\bf Architecture}.  Our location encoders use the same fully connected neural network with residual connections as in~\citet{cole2023spatial}.  
Each of the context locations is processed by the same shared location encoder which is first pre-trained as in SINR after which the multi-label classifier head is discarded. 
Importantly, this pre-trained encoder is only trained on species from the training set, and does not observe any data from the evaluation species during training. 
The text embedding backbone is a frozen GritLM~\citep{muennighoff2024generative} and the default image embedding backbone is a frozen EVA-02 ViT~\citep{FANG2024105171} pre-trained on the iNaturalist species image classification dataset~\citep{van2021benchmarking}. 
Both backbones provide a fixed length embedding vector, and we train two-layer fully connected text and image encoders to transform these embeddings into their context tokens. 
\modelname's Transformer contains four encoder layers and the parameters are updated jointly with the location, text, and image encoders and species decoder during training. 
In total, \modelname has 8.2M learnable parameters compared to 11.9M for SINR.  
This reduction is due to the fact that we do not have to learn a per-species embedding vector as in SINR. 
We train with a batch size of 2,048 instances and randomly drop-out text/image or location tokens during training with a probability of 0.5 and 0.1 respectively to enhance robustness. 
See \cref{supp:fs_impl_details} for more details. Code for \modelname is available at: \url{https://github.com/Chris-lange/fs-sinr}

\begin{figure*}[t]
    \centering
    \includegraphics[width=0.90\textwidth]{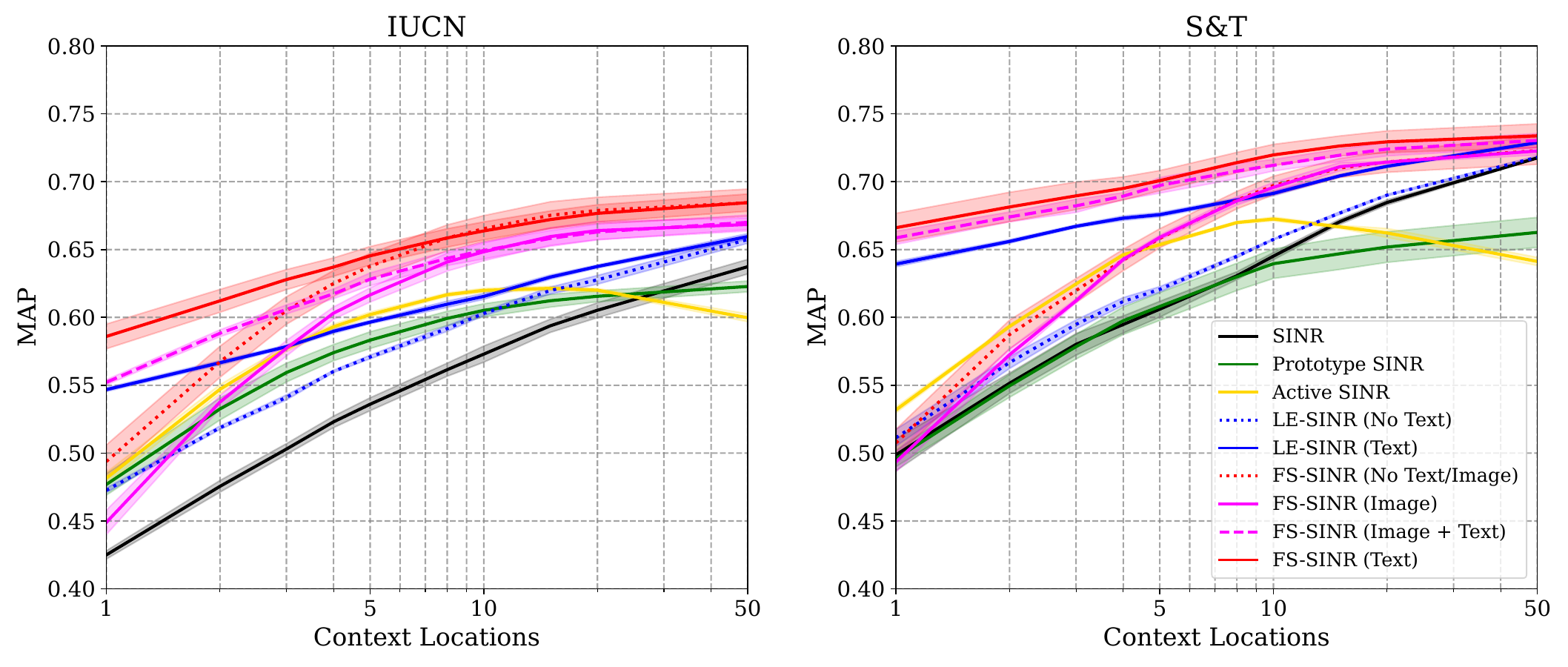}
    \vspace{-10pt}
    \caption{{\bf Few-shot results}. Here we evaluate different models on the task of species range estimation on the IUCN (left) and S\&T (right) datasets. 
    On the x-axis we vary the number of context locations seen at inference time for the held-out evaluation species. 
    The y-axis represents Mean Average Precision (MAP), where higher values are better. 
    The error bars display the standard deviation of three different runs. 
    Our \modelname approach outperforms existing methods, especially in the \cl{very low-data setting} (\ie $<$ five context locations).  
    Note that LE-SINR and SINR need to be retrained during evaluation when more observations are provided. \cl{\cref{tab:main_results_table_iucn,tab:main_results_table_snt} report expanded results including larger numbers of context locations.}}
    \label{fig:low_shot}
    \vspace{-10pt}
\end{figure*}

{\bf Data}.  We train \modelname on the presence-only dataset from~\citet{cole2023spatial}, which comprises 35.5 million citizen-science records—each annotated with latitude, longitude, and species label—for 47,375 diverse species including plants, fungi, and animals from the iNaturalist platform~\citep{inatWeb}. We also leverage 127 thousand text descriptions of these species used in ~\citet{hamilton2024} and 200 thousand images obtained from iNaturalist~\citep{inatWeb}.
The text provided during training is composed of sections of Wikipedia~\citep{wikipedia} articles of the target species. 
During training, we supply \modelname with 20 context locations per training example, although we find  that model performance is  robust to changes in the number of context locations provided during training. %

We evaluate models using the IUCN and S\&T datasets also from~\citet{cole2023spatial}, which contain expert and model-derived range maps for 2,418 and 535 different species, respectively. 
The IUCN dataset is more globally distributed and contains a larger variation in range size and more diverse animal species, while the S\&T dataset only contains bird species that are found primarily, but not always, in North America and have a larger average range size.
\cl{We follow the same preprocessing steps for these datasets as in \citet{cole2023spatial}.}
While not perfect, these datasets represent the best evaluation data currently available and contain large variety in terms of range sizes and locations. 
The text used during evaluation consists of pre-trained large language model generated summaries of the range or habitat of the target species as used in \citet{hamilton2024}. 
Importantly, we hold out any species from the union of these two datasets from the training set so that species from the evaluation set are not observed during training.  
As a result, by default, \modelname is trained on data from 44,422 species.  
Performance is reported as mean average precision (MAP) for different numbers of input (\ie context) locations.  

{\bf Baselines}.  Generating a species' range from \modelname for a held-out species at inference time only requires a single forward pass through the model to obtain an embedding vector for the species. 
Current methods (\eg LE-SINR or SINR) cannot be used in such a feedforward manner and need to be retrained for each species that was not observed at training time. 
To obtain an equivalent embedding for the SINR and LE-SINR baselines we train a per-species binary logistic regression classifier using any few-shot presence observations that are available, in addition to adding 10,000 uniformly random and 10,000 target (\ie in locations where species are) pseudo-absences as in LE-SINR. 
\cl{We also compare to the species embedding combination method from~\citet{lange2024active} and a Prototypical Network-style baseline~\citep{prototypes}, denoted Active SINR and Prototype SINR, respectively.}
These baselines do not require retraining. 
For fairness, we use the same presence observations across each method, and the larger number of presences are supersets of the smaller ones.  
Implementation details of the baseline methods can be found in \cref{sec:appendix_baselines}.

\subsection{Few-shot Evaluation} 
First, we evaluate how effective different range estimation models are at few-shot range estimation. 
The goal for each model is to generate a plausible prediction for a previously unseen species' range given limited location observations.  
Quantitative results are presented in \cref{fig:low_shot}, \cl{and additional results can be found in \cref{sec:addtional_experiments}}.

The SINR baseline performs poorly in the low-data regime, but as more data is added performance improves. 
As noted earlier, here a per-species embedding vector is learned using logistic regression using the provided presence locations and generated pseudo-absences. 
The recently introduced LE-SINR approach extends the basic SINR model to use text information (here range text), when available, at inference time. LE-SINR tends to outperform SINR, particularly when text data is available. 
Like our \modelname approach, neither \cl{the Active SINR or Prototype SINR baselines} require retraining at inference time, but perform much worse than \modelname.

In all instances, when the same metadata is available, \modelname outperforms existing methods. 
Furthermore, we also outperform SINR in the larger data regime (\ie when 50 observations are available).  
In general, we observe that image information is not as informative as text \cl{and does not help on average outside of the zero-shot case.
A range text description provides much more context than an image of a previously unseen species.
However, range estimates for some species benefit significantly from using images, but other species actually see no change or decreased performance.
Making use of images to improve zero-shot range estimates is rewarded during training, but this can harm performance for some species at inference time by biasing the ranges produced toward the rough estimates made using images, which can provide limited information.}
Importantly, unlike SINR and LE-SINR, \modelname does not need to be retrained at inference time. 
Instead, it can make predictions in a feedforward manner irrespective of the context data available. 
This is advantageous in interactive settings, whereby the model can compute the location embeddings for all query locations on earth once, and then a user could experiment by adding different context information interactively.
\cl{Removing the retraining step also allows \modelname to produce estimated ranges in a fraction of the compute time compared to other approaches.
Compared to the publicly released implementation of LE-SINR on the same hardware, \modelname generates range estimates from one context location and text for all species in the IUCN and S\&T datasets in 2\% of the time on CPU, and 6\% of the time on GPU.}

\begin{figure*}[t] 
\vspace{5pt}
\centering
\begin{overpic}[trim=480 170 1100 320,clip,width=0.23\textwidth]{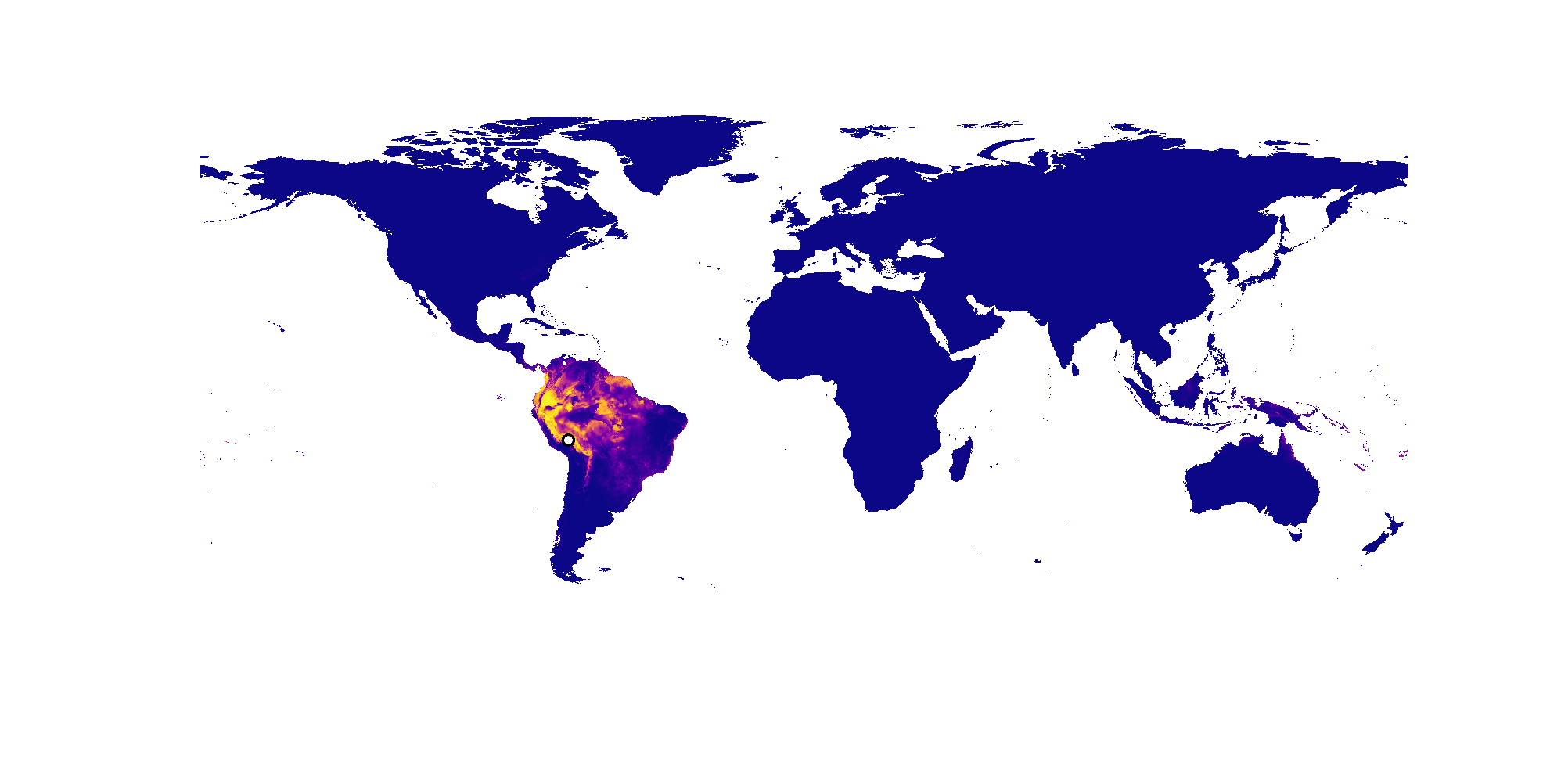}
 \put(0,12){
    \fcolorbox{black}{gray!30}{\strut\small  [No Text]}
  }
\end{overpic}
\hspace{2pt}
\begin{overpic}[trim=480 170 1100 320,clip,width=0.23\textwidth]{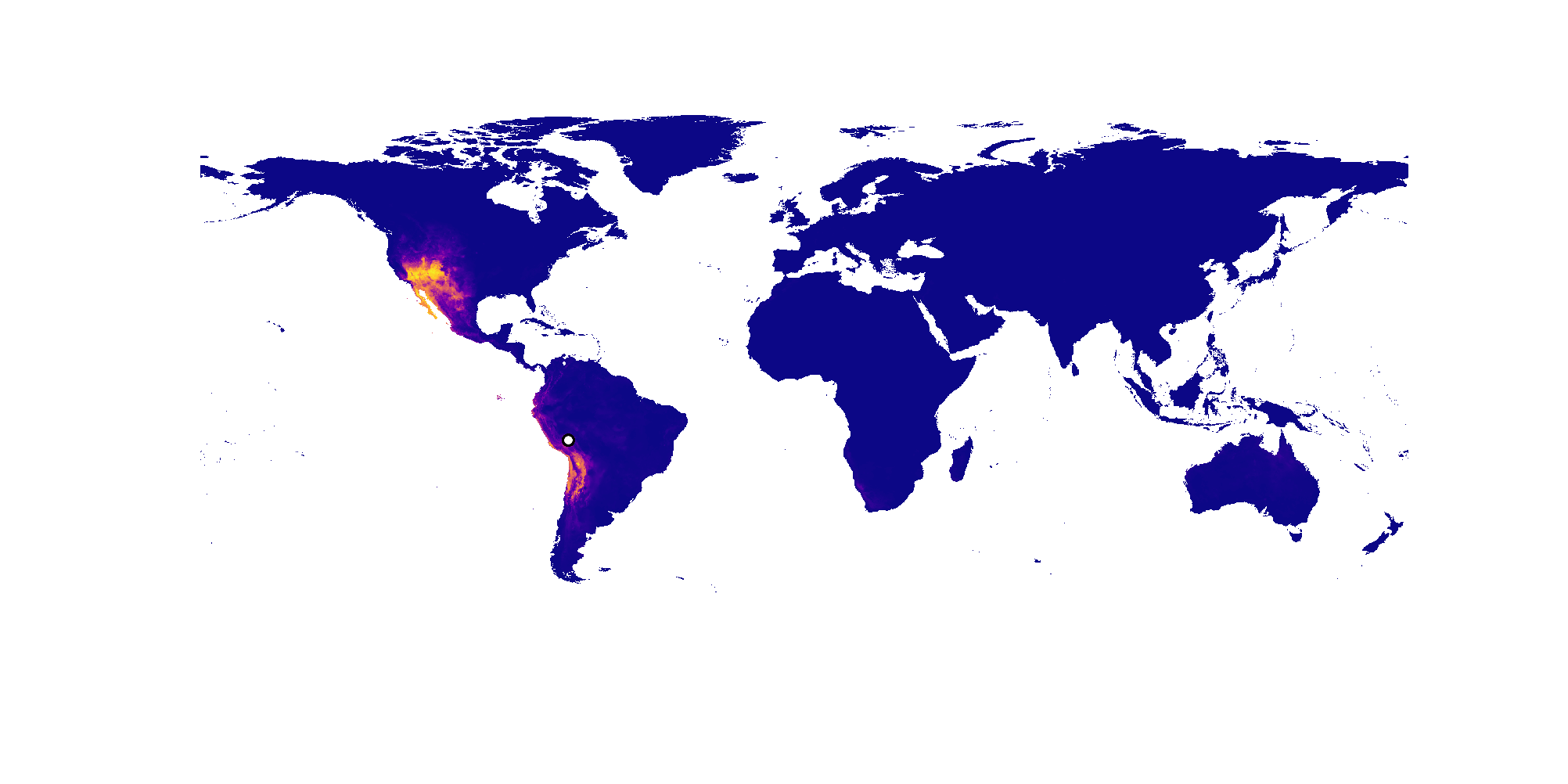}
 \put(0,12){
    \fcolorbox{black}{gray!30}{\strut\small  \search ``desert''}
  }
\end{overpic}
\hspace{2pt}
\begin{overpic}[trim=480 170 1100 320,clip,width=0.23\textwidth]{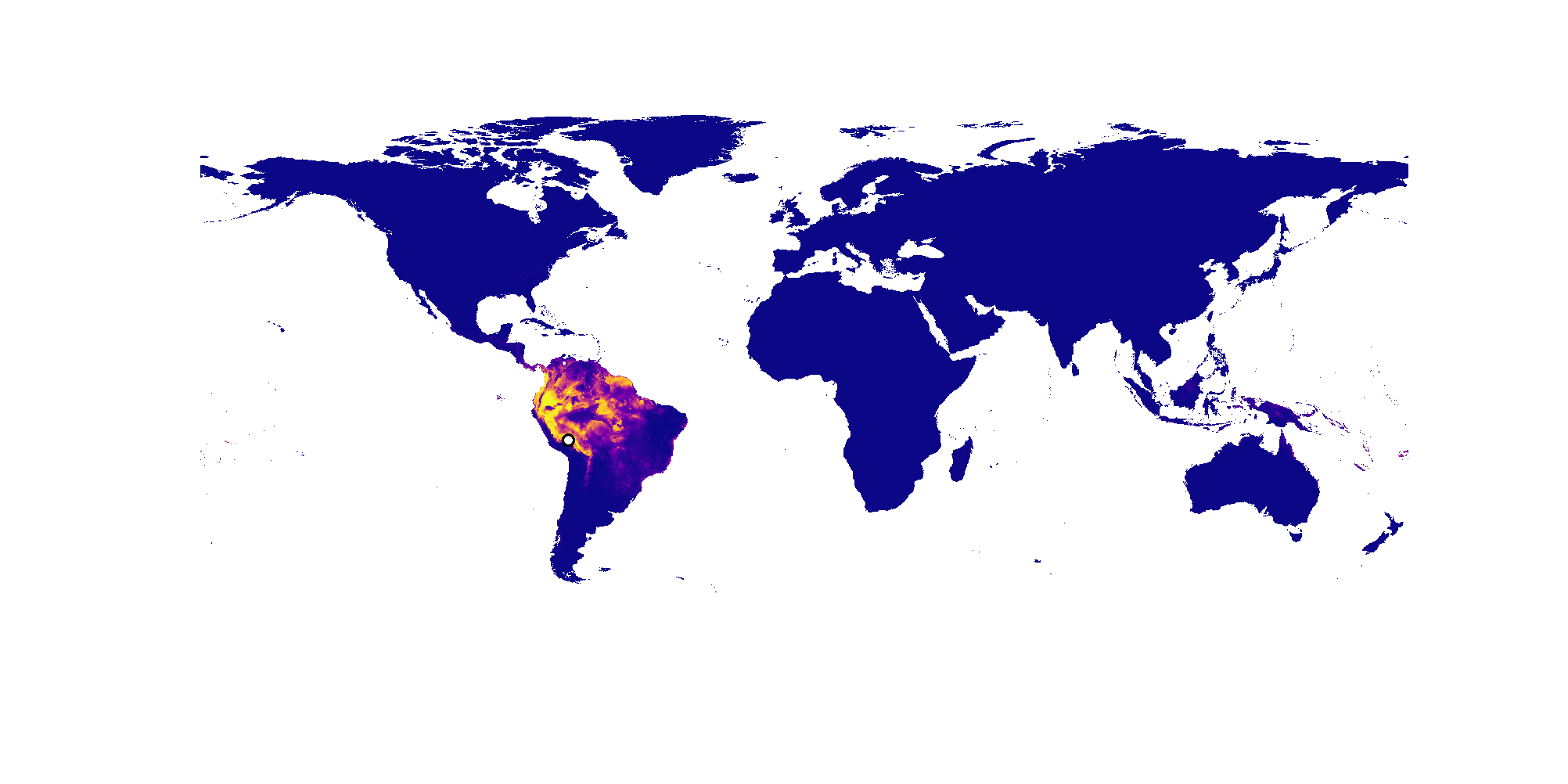}
 \put(0,12){
    \fcolorbox{black}{gray!30}{\strut\small  \search ``rainforest''}
  }
\end{overpic}
\hspace{2pt}
\begin{overpic}[trim=480 170 1100 320,clip,width=0.23\textwidth]{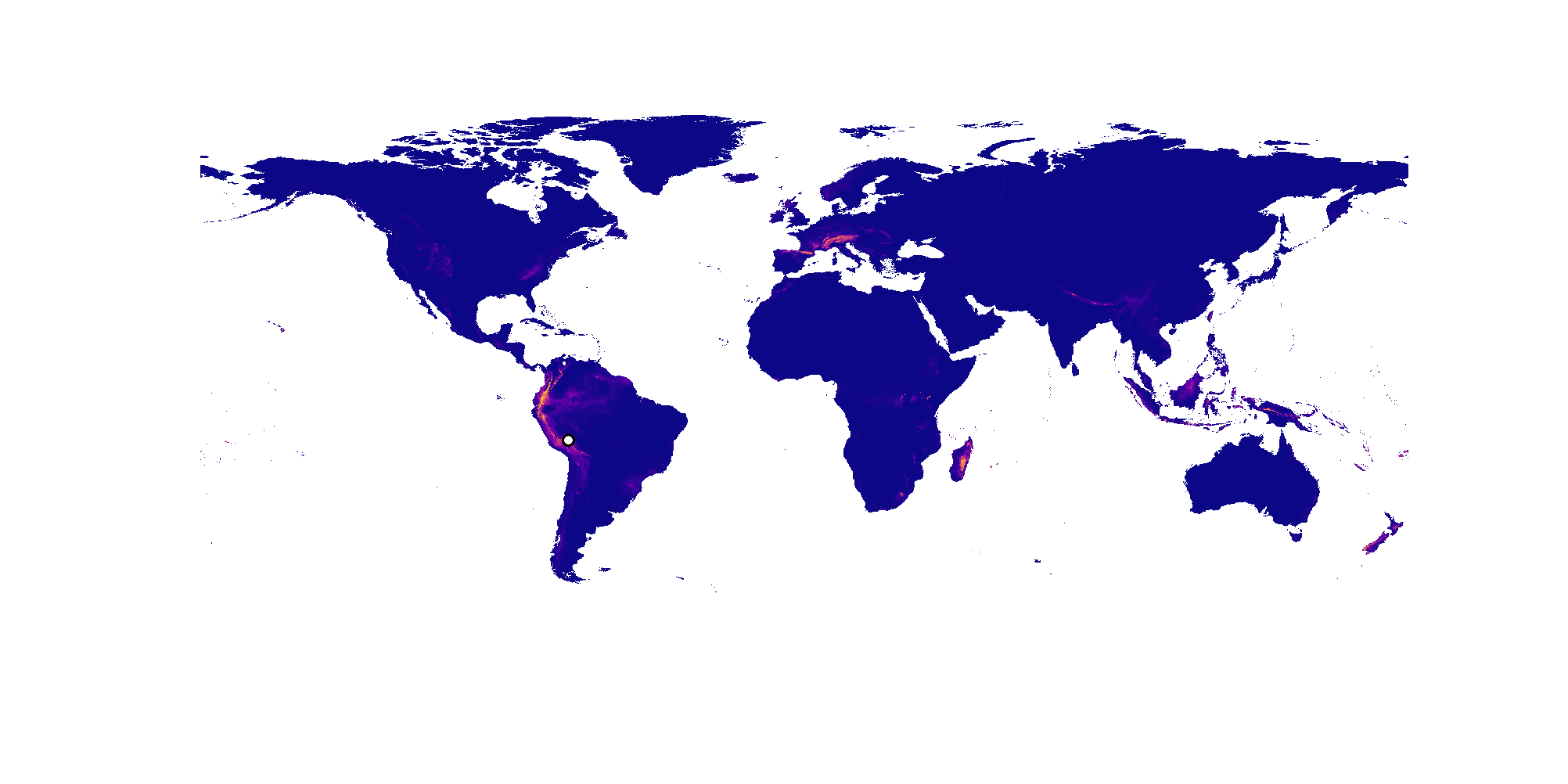}
 \put(0,12){
    \fcolorbox{black}{gray!30}{\strut\small  \search ``high altitude mountains''}
  }
\end{overpic}
\vspace{-15pt}
\caption{{\bf Controlling range predictions using a single context location with different text}. 
Given the same single context location, denoted as `$\circ$', \modelname can generate significantly different range predictions depending on the text provided. 
This example illustrates a use case where a user may have limited observations but some additional knowledge that can be encoded via text regarding the type of habitat a species of interest could be found in.  
Note, while `no text' and `rainforest' look similar, they are actually subtly different.  
}
\vspace{-5pt}
\label{fig:model_vis_south_america}
\end{figure*}

\begin{figure*}[t]
\centering
\includegraphics[width=\textwidth]{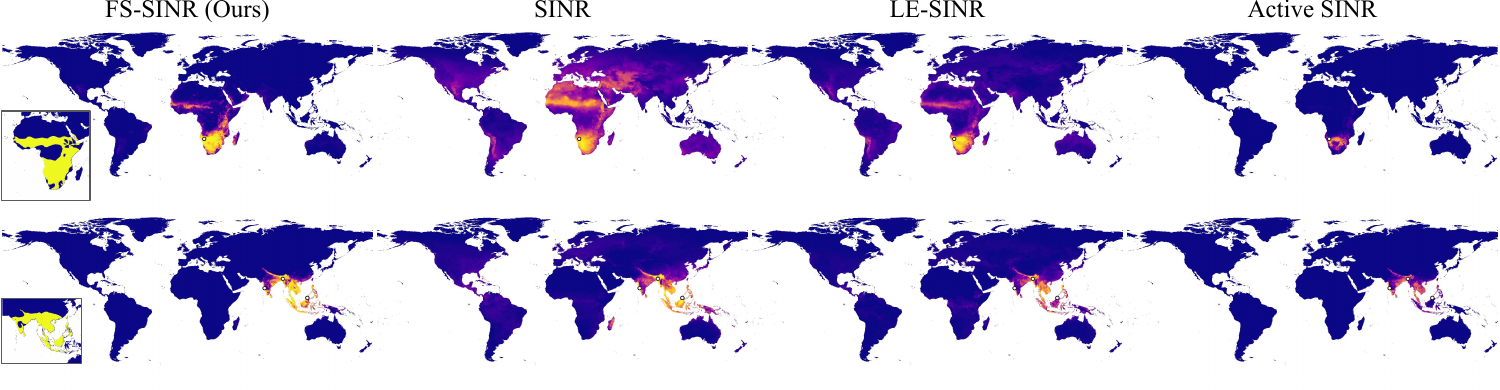} 
\vspace{-18pt}
\caption{{\bf Qualitative comparison of range predictions for different methods}. (Top) Predicted ranges from a single context location denoted as ‘◦’ and no additional metadata for the \texttt{Gabar Goshawk}. (Bottom) Predicted ranges from three context locations  for the \texttt{Black-naped Monarch}. From left to right, FS-SINR (ours) with expert range inset, SINR, LE-SINR, and Active SINR. Please zoom in to see details. 
}
\vspace{-5pt}
\label{fig:compare_others}
\end{figure*}

\begin{figure*}[t]
    \centering
    \renewcommand{\arraystretch}{1}
    \begin{tabular}{ccc}
        {\bf Common Kingfisher} & 
        {\bf European Robin} & 
        {\bf Black and White Warbler} \\

        \begin{overpic}[trim={0 1cm 0 0},clip,width=0.31\textwidth]{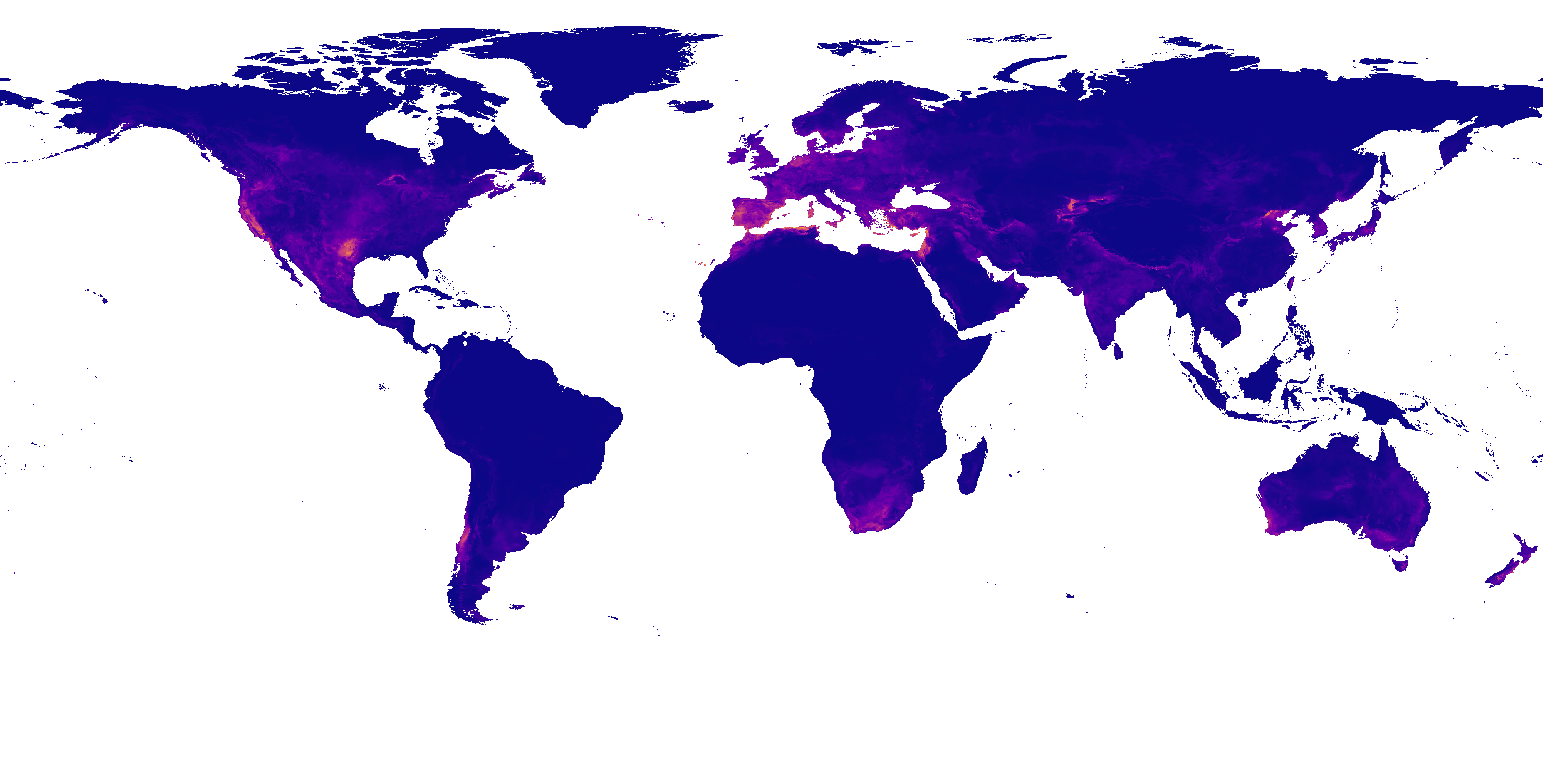}
            \put(0,0){%
              {%
                \setlength\fboxsep{0pt}%
                \setlength\fboxrule{1pt}%
                \fbox{%
                  \includegraphics[width=0.08\textwidth]{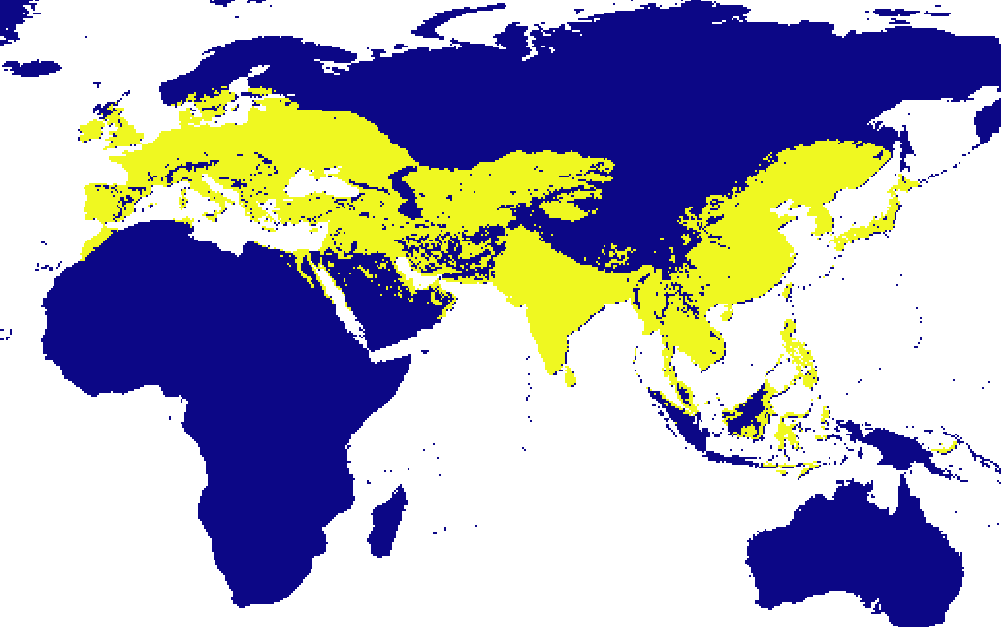}%
                }%
              }%
            }
        \end{overpic}
        &
        \begin{overpic}[trim={0 1cm 0 0},clip,width=0.31\textwidth]{figs/range_est/base_no_context_sm.png}
            \put(0,0){%
              {%
                \setlength\fboxsep{0pt}%
                \setlength\fboxrule{1pt}%
                \fbox{%
                  \includegraphics[width=0.08\textwidth]{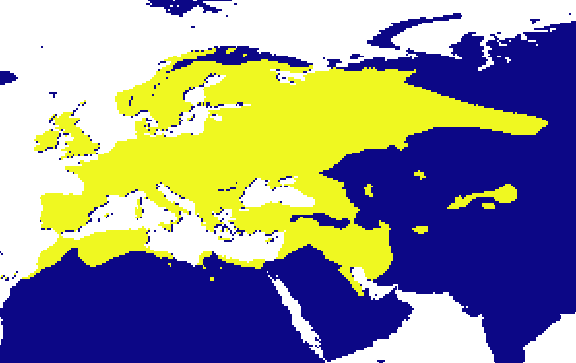}%
                }%
              }%
            }
        \end{overpic}
        &
        \begin{overpic}[trim={0 1cm 0 0},clip,width=0.31\textwidth]{figs/range_est/base_no_context_sm.png}
            \put(0,0){%
              {%
                \setlength\fboxsep{0pt}%
                \setlength\fboxrule{1pt}%
                \fbox{%
                  \includegraphics[width=0.08\textwidth]{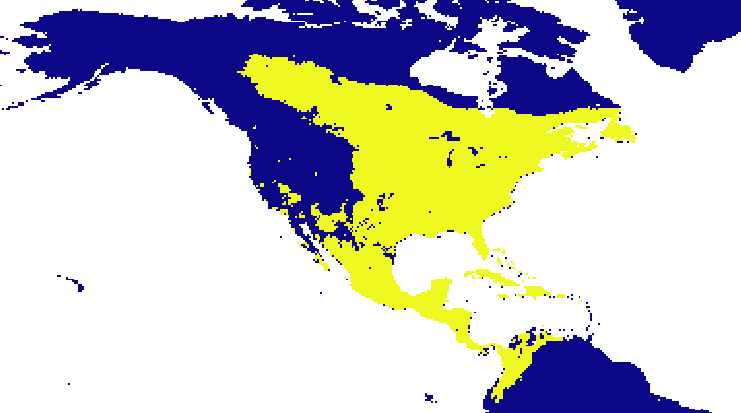}%
                }%
              }%
            }
        \end{overpic}
        \\
        
        \includegraphics[trim={0 1cm 0 0},clip,width=0.31\textwidth]{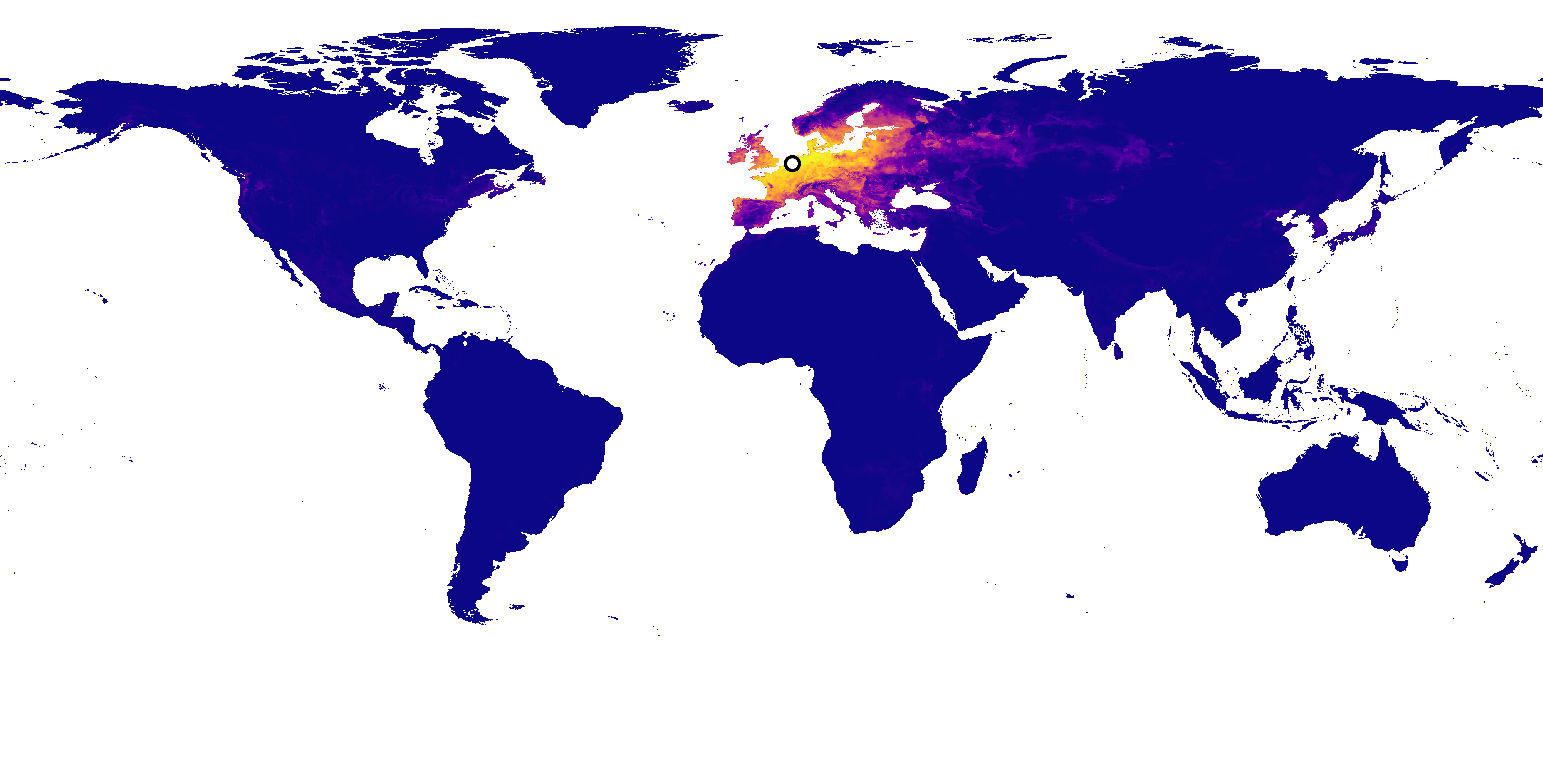} & 
        \includegraphics[trim={0 1cm 0 0},clip,width=0.31\textwidth]{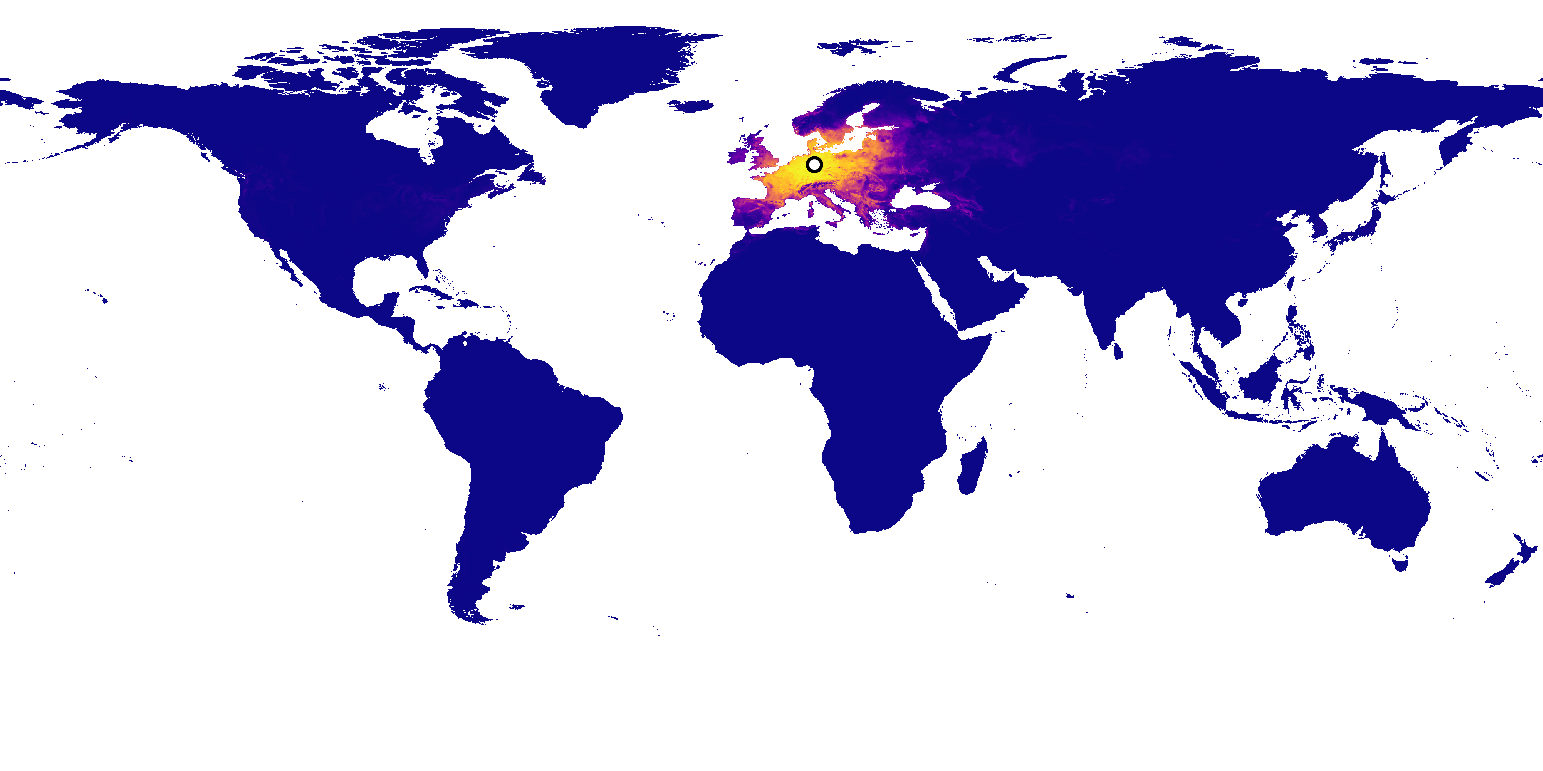} & 
        \includegraphics[trim={0 1cm 0 0},clip,width=0.31\textwidth]{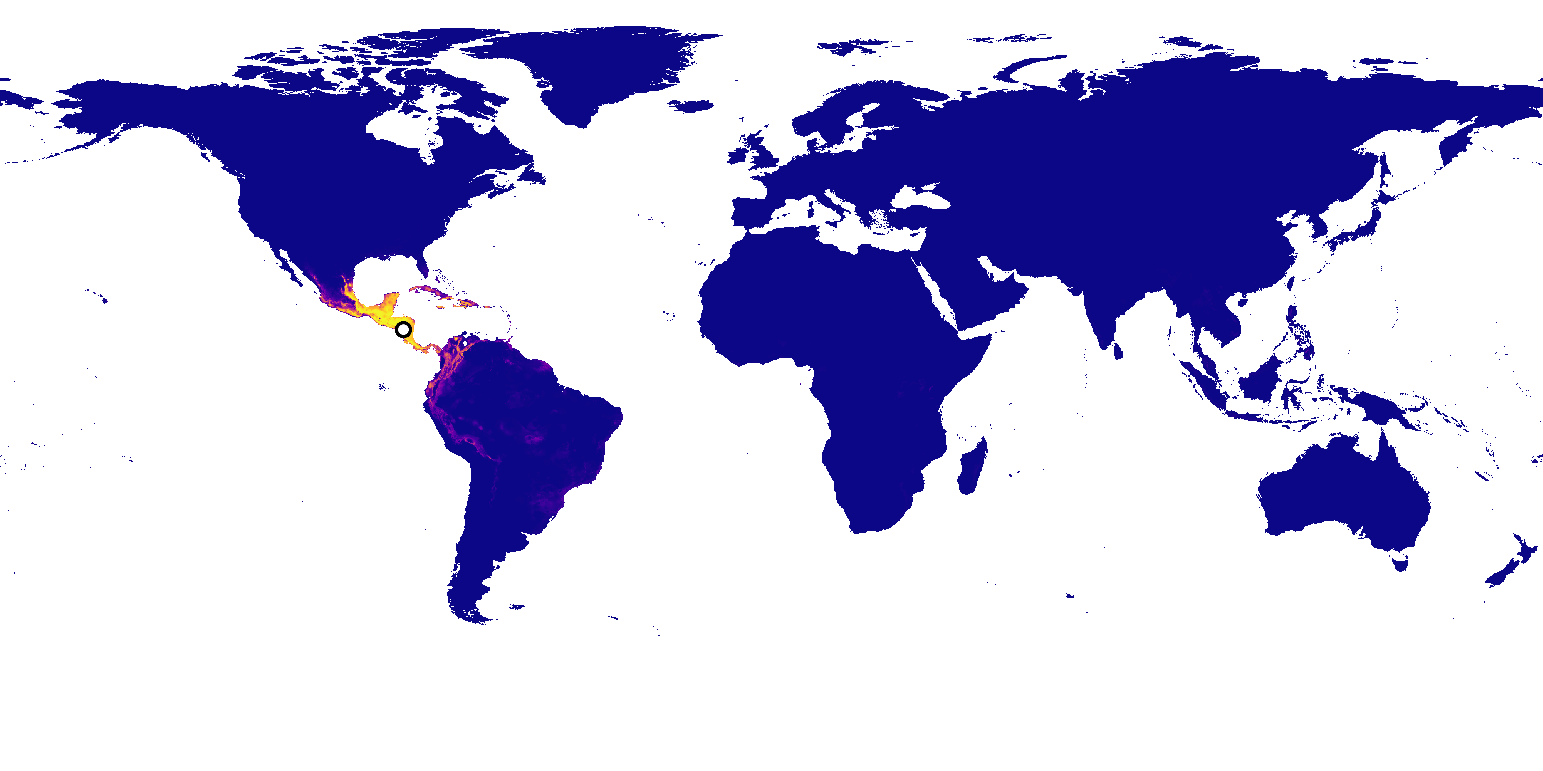} \\
        
        \includegraphics[trim={0 1cm 0 0},clip,width=0.31\textwidth]{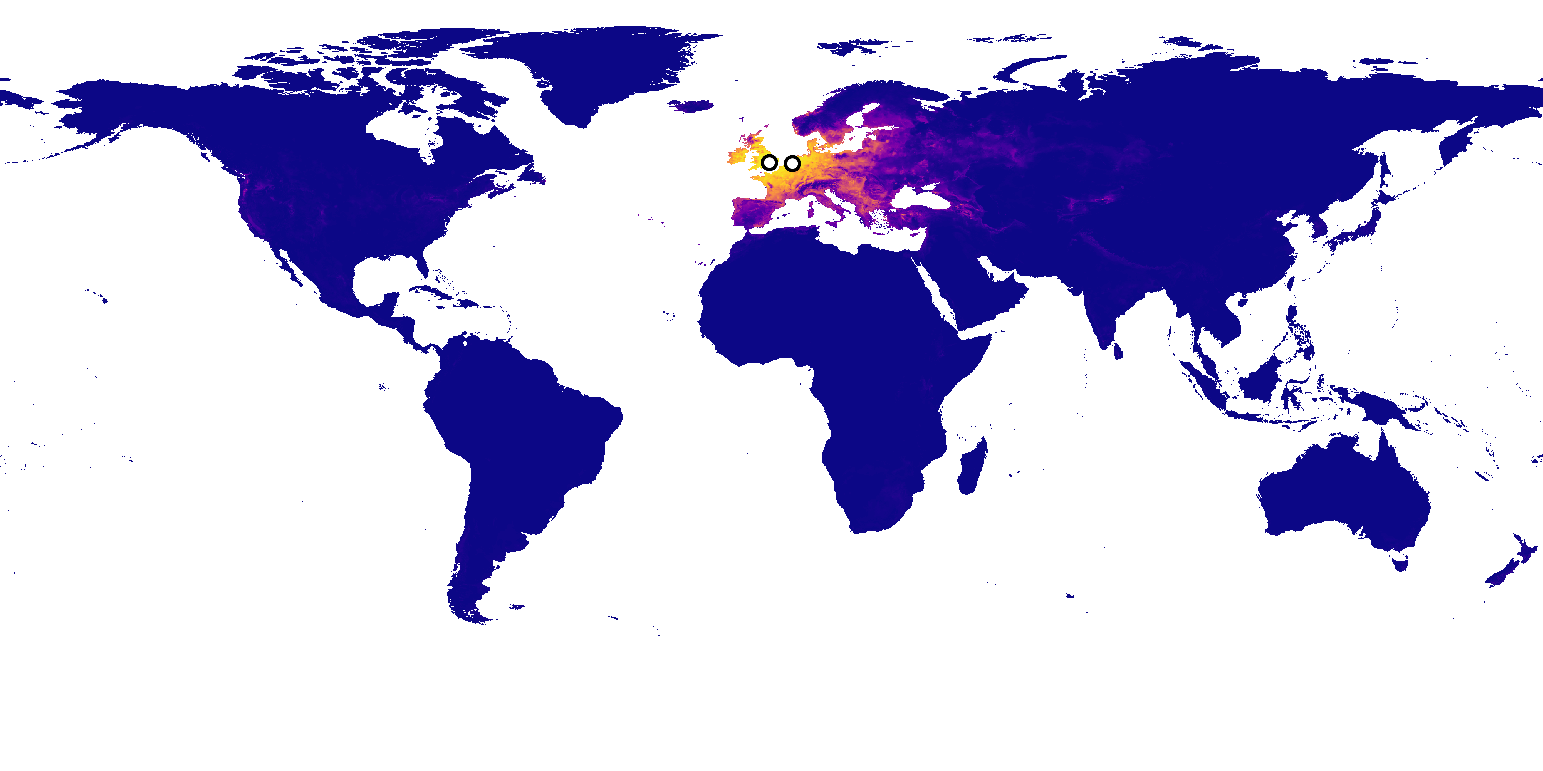} &
        \includegraphics[trim={0 1cm 0 0},clip,width=0.31\textwidth]{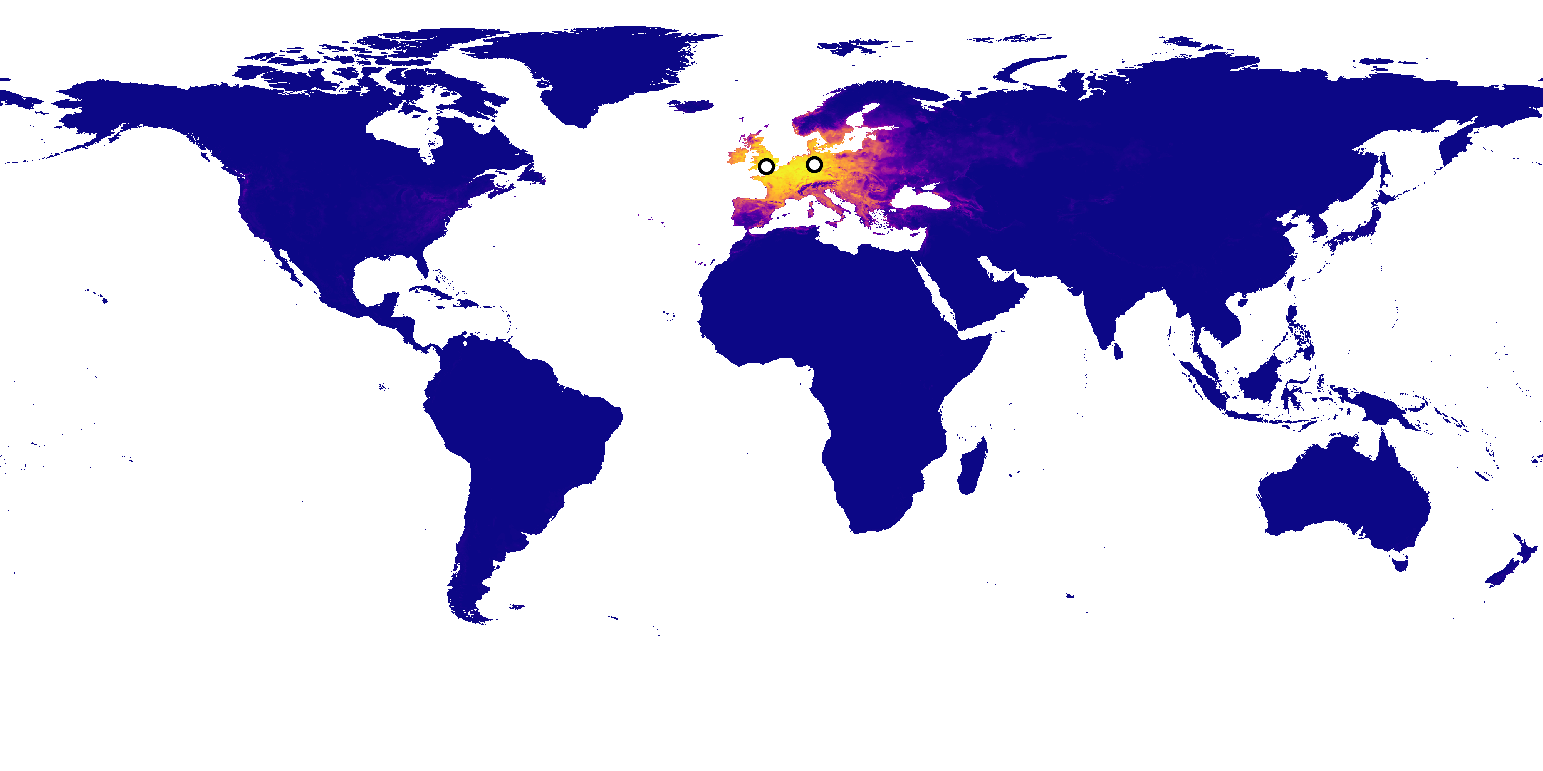} & 
        \includegraphics[trim={0 1cm 0 0},clip,width=0.31\textwidth]{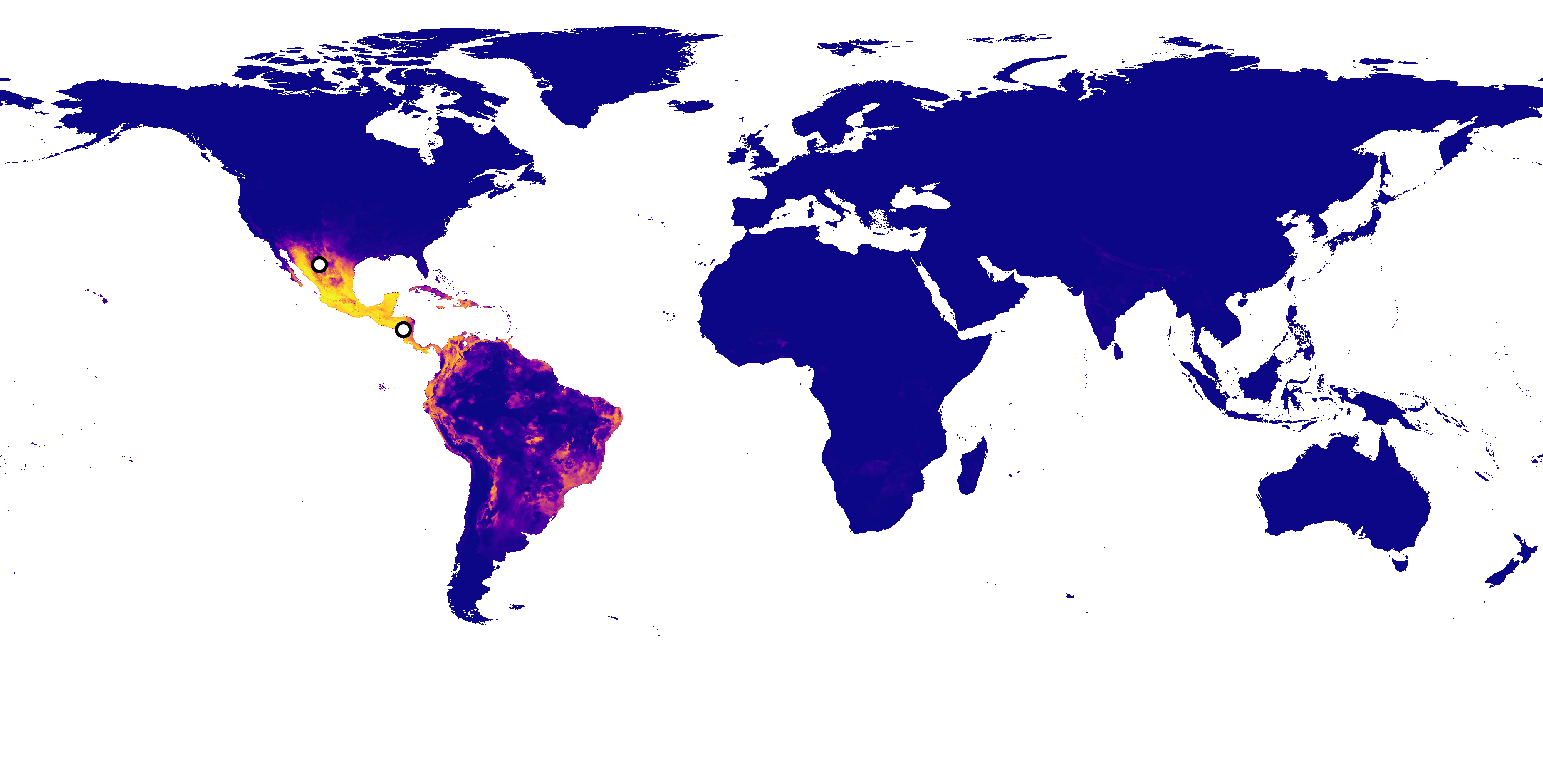} \\
        
        \includegraphics[trim={0 1cm 0 0},clip,width=0.31\textwidth]{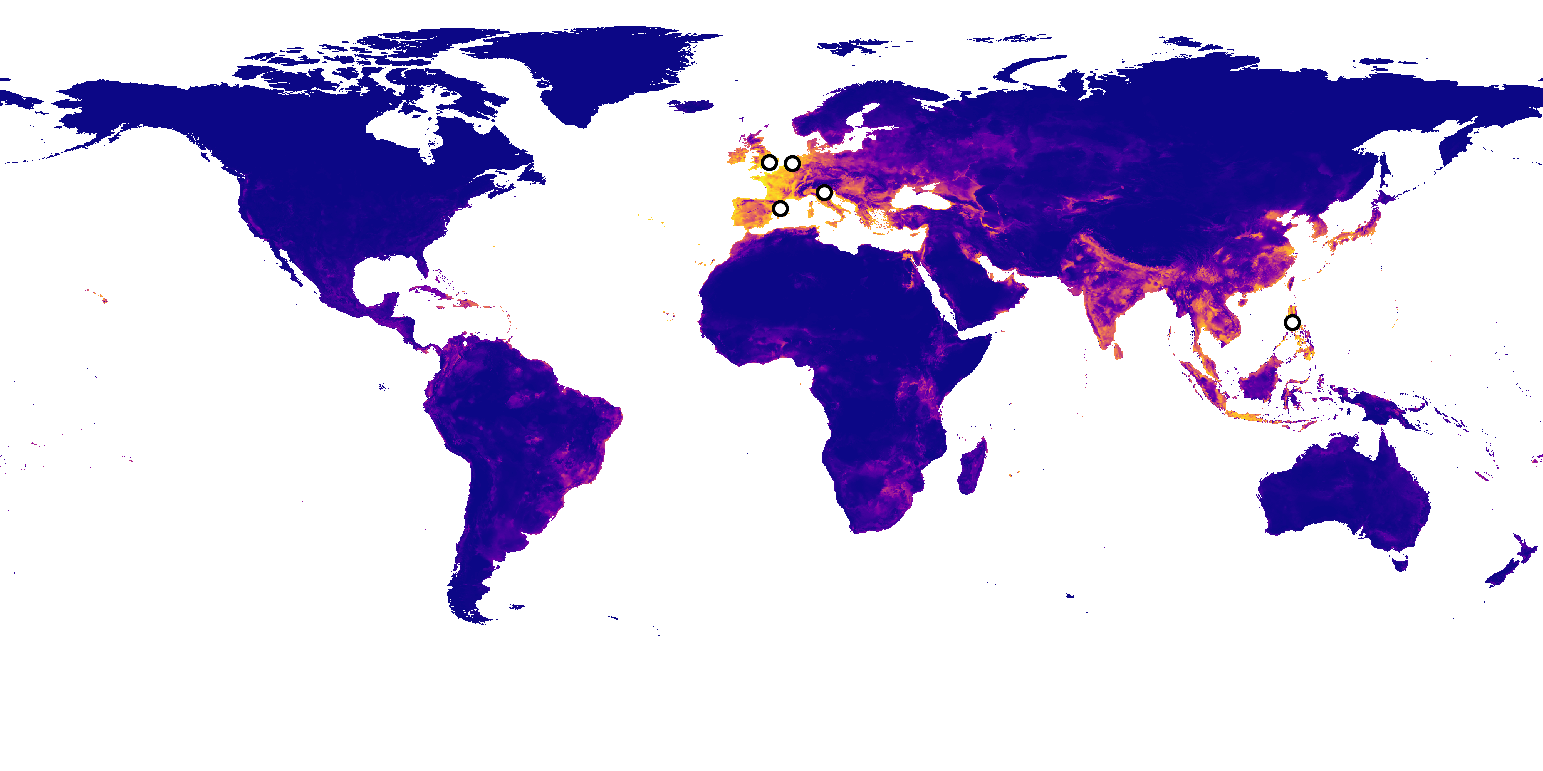} &
        \includegraphics[trim={0 1cm 0 0},clip,width=0.31\textwidth]{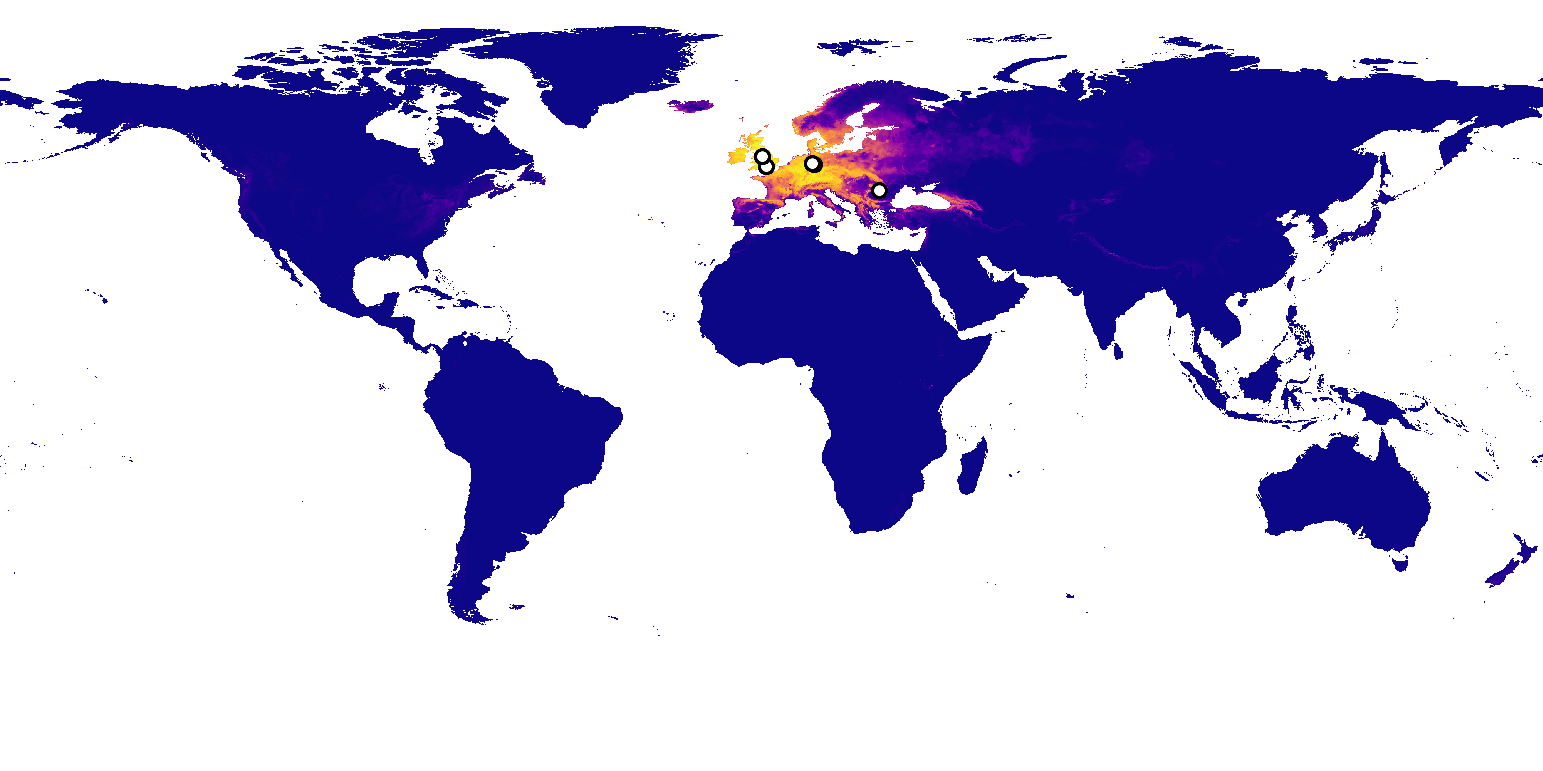} & 
        \includegraphics[trim={0 1cm 0 0},clip,width=0.31\textwidth]{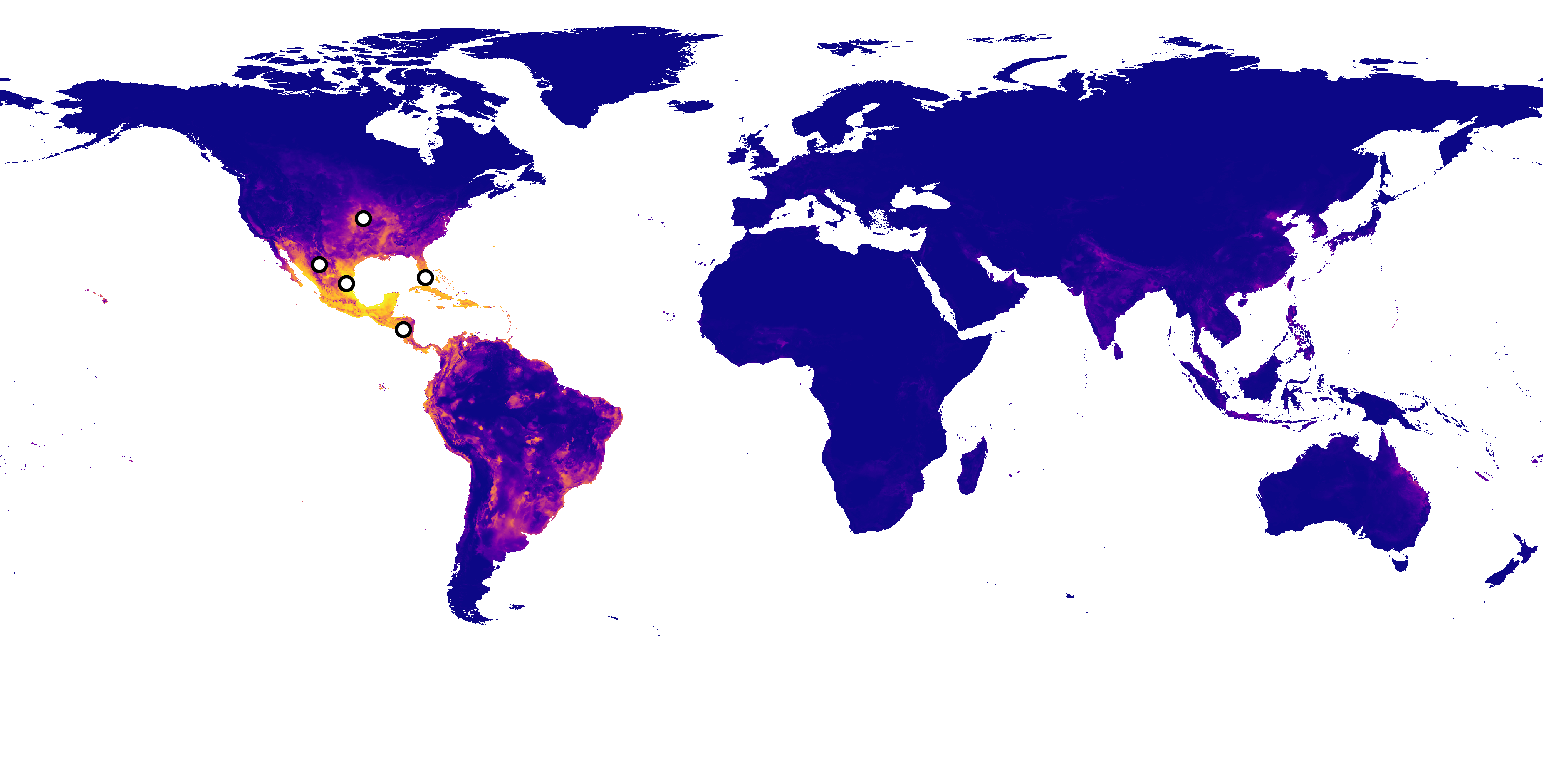} \\

        \includegraphics[trim={0 1cm 0 0},clip,width=0.31\textwidth]{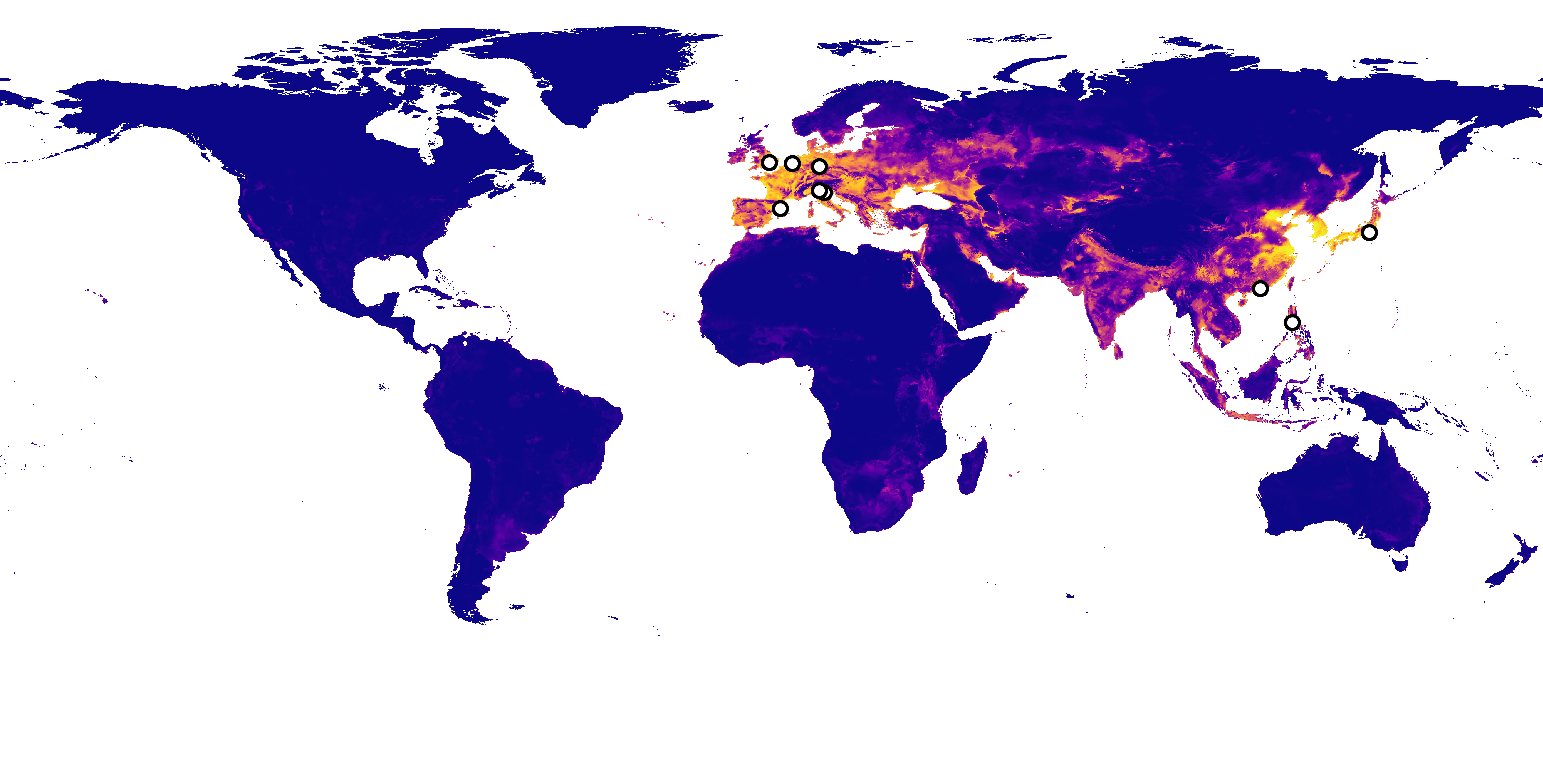} &
        \includegraphics[trim={0 1cm 0 0},clip,width=0.31\textwidth]{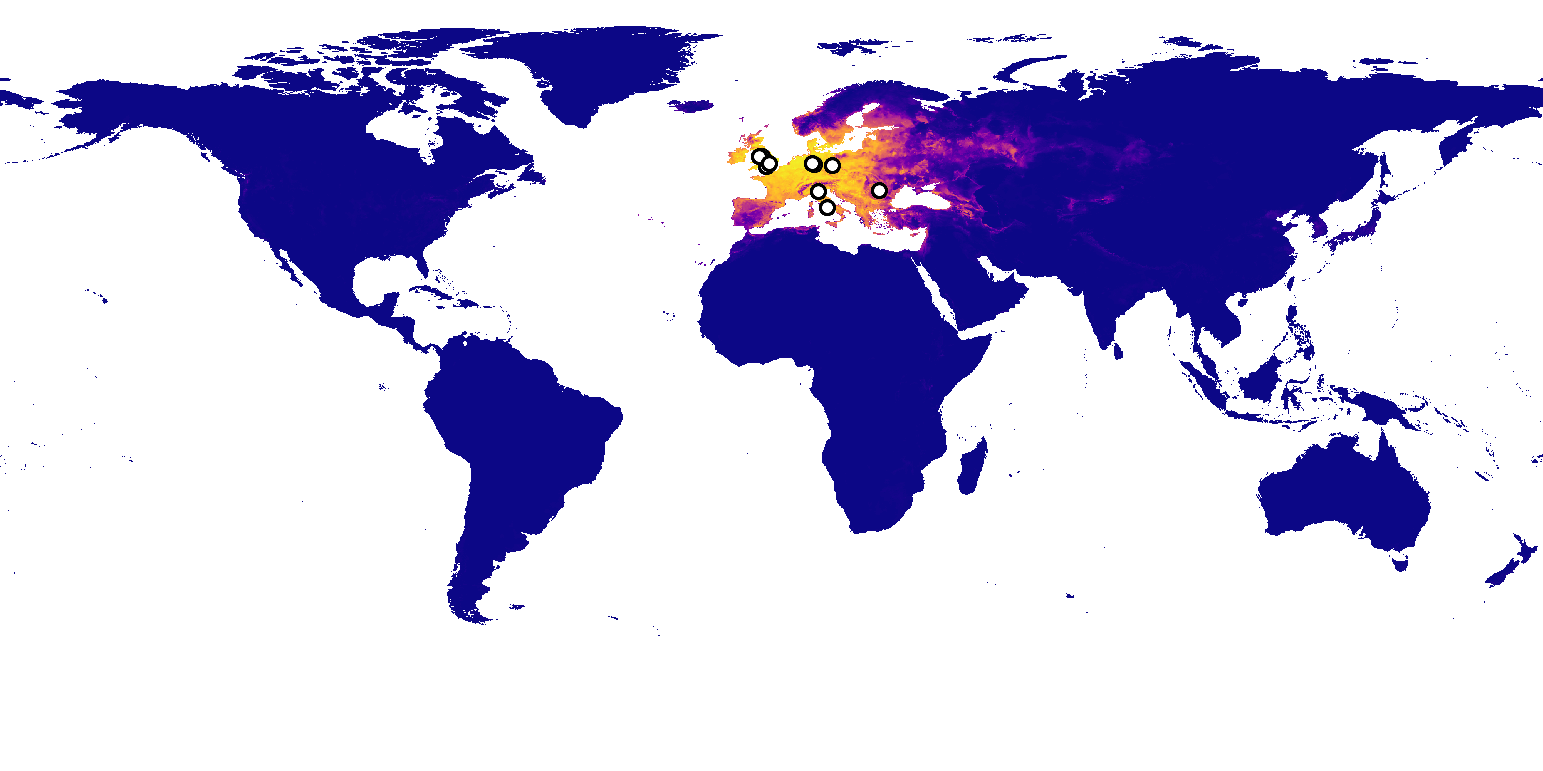} &    
        \includegraphics[trim={0 1cm 0 0},clip,width=0.31\textwidth]{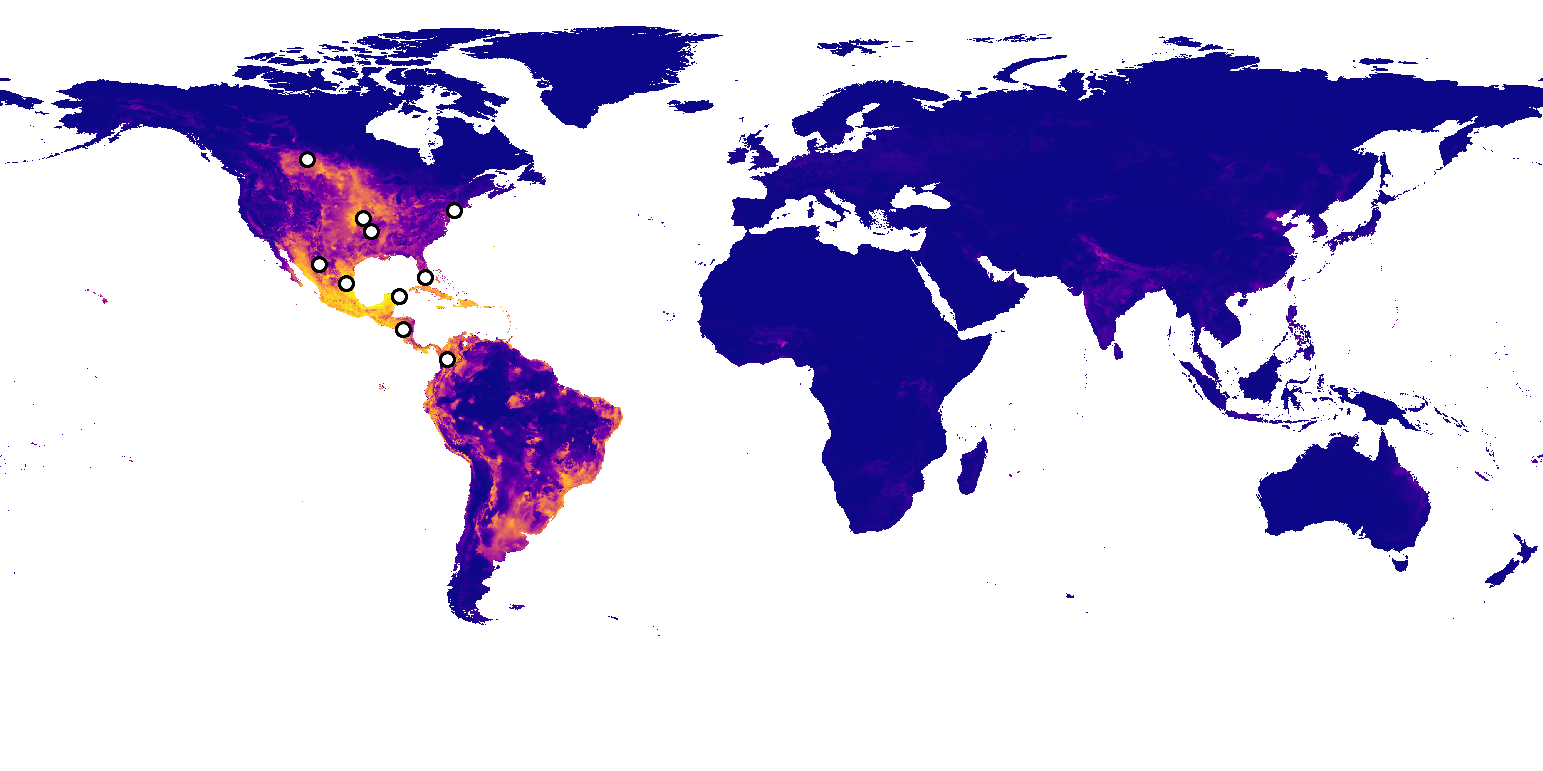} \\
        
    \end{tabular}
    \vspace{-10pt}
    \caption{{\bf Few-shot range estimation with increasing context locations}. 
    Here we illustrate few-shot range predictions from \modelname given an increasing number of context locations $\{0,1,2,5,10\}$ and no other context information for the \texttt{Common Kingfisher} (left), \texttt{European Robin} (center), and the \texttt{Black and White Warbler} (right).
    In the first row, we show the expert-derived range inset and the prediction for the model when no context locations are provided (which is the same for all species). 
    Then, in the remaining rows we increase the number of context locations, denoted as `$\circ$'.
    Please zoom in to see the context locations. 
    As we increase the number of context locations, the predictions become closer to the expert ranges.
    }
    \vspace{-5pt}
    \label{fig:qualitative_time}
\end{figure*}

We present qualitative results for three different species in~\cref{fig:qualitative_time} where we visualize \modelname's predictions as we change the number of context locations. 
Given only a single context location, the model does a sensible job of localizing the species on Earth. 
This supports the findings from~\cref{fig:low_shot} where we observe strong performance even when only one context location is available. 
When more information is provided, the predicted range more closely resembles the expert-derived range shown in the first row. 
However, we do note that the model can still make mistakes in our low data setting, such as the erroneous predictions for the `Black and White Warbler' in South America. 
In~\cref{fig:model_vis_south_america} we illustrate some examples of how text information, when paired with one single context location, can influence the model predictions.  
We observe dramatically different predicted ranges when the text prompt encourages the model to focus on different habitat types. 
We note that each of the predicted ranges is still consistent with the location of the single context location provided. 
Finally, in \cref{fig:compare_others} we compare \modelname range predictions to other approaches, namely SINR, LS-SINR, and Active SINR. We see that for this species \modelname more closely resembles the expert range when only three context locations are provided.
Additional qualitative examples are provided in \cref{sec:app_qual_results}. 
\subsection{Zero-shot Evaluation} 
In addition to being able to generate range predictions in the few-shot setting when limited location observations are provided, \modelname can also make predictions when no location information is provided but only additional metadata such as an image or text describing a previously unseen species is given, \ie the \emph{zero}-shot setting.  
These zero-shot results are presented in~\cref{tab:zeros_shot} for both the IUCN and S\&T datasets.  

\begin{table}[t]
\vspace{-5pt}
\caption{\textbf{Zero-shot results}.  
We report zero-shot performance where no location information is provided to each model, \cl{only additional metadata,} comparing to SINR~\citep{cole2023spatial} and LE-SINR~\citep{hamilton2024}. 
We denote additional metadata used by models as RT for `Range Text', HT for `Habitat Text', and `I' for `Image'. 
TST represents `Test Species in Train', indicating that a model uses location observations for the evaluation species at training time (\eg SINR which provides an upper bound on performance), unlike other models where these species are excluded. 
TRT models are trained using `Taxonomic Rank Text' as in~\citet{sastry2023ld}, which are also provided with the full taxonomic description from `class' to `species' during evaluation. 
Results are presented as MAP, where higher is better.  
}
\vspace{5pt}
\centering
\resizebox{0.75\columnwidth}{!}{%
\begin{tabular}{cl|l|ll}
ID & Method  & Variant & IUCN & S\&T \\ \hline
1 & SINR & TST &   0.67   &   0.77   \\ 
2 & FS-SINR  & HT, TST &   0.38   &  0.59  \\ 
3 & FS-SINR  & RT, TST &   0.55  &  0.67 \\ \hline
4 & FS-SINR &  &  0.05   &  0.18  \\ 
5 & FS-SINR & TRT &   0.21   &  0.34  \\ \hline
6 & LE-SINR & HT &   0.28   &  0.52    \\ 
7 & FS-SINR  & HT     &   0.33   &  0.53  \\ \hline
8 & LE-SINR & RT &   0.48   &   0.60   \\
9 & FS-SINR & RT &   0.52   &  0.64  \\ \hline
10 & FS-SINR & I &  0.19  & 0.38  \\
11 & FS-SINR & I + RT &  0.46  & 0.64  \\
\end{tabular}
}
\label{tab:zeros_shot}
\vspace{-10pt}
\end{table}

\begin{figure*}[t]
\centering
\vspace{5pt}
\includegraphics[width=\textwidth]{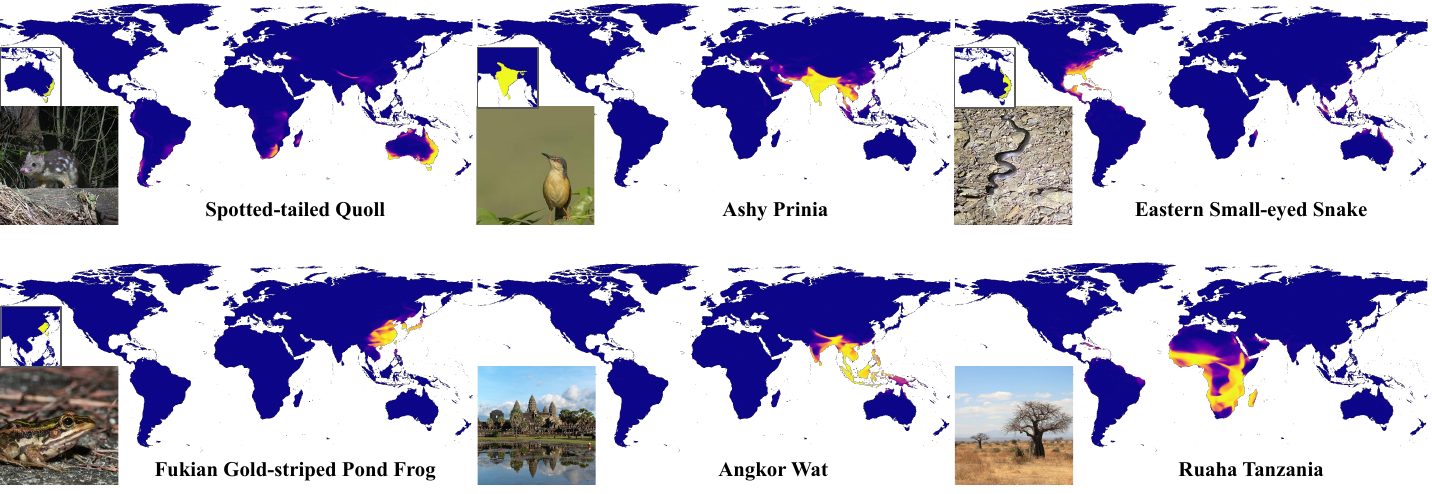} 
\vspace{-18pt}
\caption{{\bf Range predictions with a single context image as input}. 
We can condition \modelname on an arbitrary input image with no context locations or text, \eg a held-out species (top row and bottom left), famous landmarks (bottom middle), or landscape images (bottom right).}
\vspace{-10pt}
\label{fig:compare_vision}
\end{figure*}

We report results for several variants of \modelname where different types of metadata are used. 
As a baseline, we also present the performance of SINR (row 1) where the evaluation species are part of its training set \cl{\ie not zero-shot}. 
We can also add data from these species to the training set of our approach which unsurprisingly boosts performance (\ie row 3 vs.~9), though unlike SINR, \modelname does not have weights associated with individual species and so the impact of seeing evaluation species during training is fairly small.  
As a trivial baseline, we also report performance of \modelname (row 4) when no location or text metadata is provided, \ie this is simply the output of the class token. 
As expected, this model performs poorly, but interestingly it seems to have learned some spatial prior that results in non-trivial predictions on S\&T which contains bird species mostly concentrated in North America.  
We also compare to a version of \modelname (row 5) where we use taxonomic text (TRT) as in LD-SDM~\citep{sastry2023ld} (see \cref{sec:trt_section} for further details).  

In all instances, our \modelname approach outperforms LE-SINR, even when both models are provided with the same information at inference and training time (\ie row 6 vs.~7 or row 8 vs.~9).  
Confirming observations from LE-SINR, we see that range text (RT) is more informative than habitat text (HT) (\ie row 7 vs.~9). 
Additionally, image information provides some non-trivial signal (\ie row 4 vs.~10), but it is not as informative as text (\ie row 9 vs.~10), \cl{and can negatively impact performance when more informative sources are provided (\ie row 9 vs.~11 for the harder IUCN dataset), as the model may overfit to incorrect spurious features in the image.}
As we can see in \cref{fig:compare_vision} (with additional examples in \cref{fig:viz_zero_shot}), zero-shot image predictions can be sensible, but predicting an unobserved species' range from a single input image is ill posed. 
Text descriptions of range or habitat preferences are simply much more informative than a single image.

\subsection{\cl{Additional Results and Ablations}}  
In \cref{sec:addtional_experiments} we provide additional experimental results for \modelname. 
There we investigate uncertainty quantification to see how well calibrated the model predictions are. 
We also report results using a `distance-weighted MAP' metric which penalizes errors more the further they are in distance away from the actual range. 
This metric more closely aligns human judgment of predicted range quality.

\cl{We provide a more ecologically relevant breakdown of results in \cref{sec:eco_results}, where we find multiple potential sources of bias in our training data toward North America and Europe, and report higher evaluation performance in these regions for \modelname and LE-SINR.
Similarly, we see that biases in the text data potentially leads to increased performance for charismatic and well-studied mammals compared to other taxonomic groups, though providing more context locations reduces this gap.
\modelname is somewhat robust to these biases and outperforms other approaches across almost all categories.
We also find that for all approaches tested, estimating very small ranges is difficult and performance varies strongly with range sizes, though \modelname again shows comparatively good performance.}

\cl{We provide additional ablation experiments for \modelname in  \cref{sec:app_ablations}, where  we evaluate the impact of different input features, location encoders, and the amount of training data used, and explore architectural modifications such as removing the final species decoder that operates on the output of the Transformer.}
We observe that \modelname is robust to these changes, justifying its design decisions.

\section{Limitations} 
\label{sec:limitations}
While \modelname \cl{outperforms other zero and few-shot approaches for species range estimation}, there are some limitations. 
First, given a set of input context locations \modelname is deterministic in that it will always generate the same output range map. 
In practice, in the few-shot regime, the same set of points could actually be representative of many different possible range maps. 
An obvious extension of our work is to introduce stochasticity into the model outputs, \eg by treating class token output from the Transformer as a latent embedding for an additional sampling step.  
In \cref{fig:different_seeds} we observe that initializing \modelname with different random seeds during training results in diverse range predictions across the different models, \cl{and in \cref{sec:addtional_experiments} we show that this can be exploited to give estimates of the uncertainty of predictions when using an ensemble of \modelname models}. 
We leave further exploration of this for future work. 
Second, at inference time, users may wish to provide example locations indicating where a specific species has \emph{not} been found, \ie confirmed absences.  
Currently, our model is trained using presence-only data but could be adapted to use absence information, if available, which could be denoted via a different embedding type vector, to be learned during training alongside our existing token type embeddings. 
However, obtaining reliable large-scale  absence data for tens of thousands of species is a challenging problem. 

Finally, \cl{biodiversity data and particularly} global-scale citizen science datasets like the one we use to train \modelname can contain large biases~\citep{geldmann2016determines, hughessamplingbias}, \eg location, temporal, and taxonomic biases, among others. 
We do not explicitly account for these biases during training, \cl{though we make some attempt to evaluate the impact of them in \cref{sec:eco_results}}, and thus we would caution the use of the predictions of our model in any applications that would use our range predictions in the context of biodiversity assessments.
However, we note that we outperform existing and recent state-of-the-art range estimation methods, especially in the low observation data setting, and do not require any retraining at inference time.

\section{Conclusion}
\cl{The scientific community} has limited knowledge on the geographical distributions of the majority of species on Earth.
This lack of understanding is further hampered by the fact that we also have insufficient data to train models to estimate their ranges.    
To address this problem, we introduced \modelname, a new approach for few-shot species range estimation. 
We demonstrated that \modelname is able to fuse data from different modalities at inference time in a feedforward manner to efficiently make plausible range predictions for previously unobserved species. 
Our quantitative analysis, using expert-derived range maps, shows a 5-10\% performance improvement compared to current approaches in the few-shot setting, \ie when the number of observations equals ten,  for previously unseen species.  
In addition, we also outperform existing methods in the zero-shot setting. 
While our results are promising, they also indicate that there are many open challenges in this important task. 

{\bf Acknowledgements.} We thank the iNaturalist community for making the species observation data available.  
OMA was supported by a Royal Society Research Grant. 
MH and SM were supported by NSF grants 2329927 and 2406687.

\newpage

\section*{Impact Statement}
Given the limited observations available for most species, there is a great need for reliable machine-learning based solutions for estimating their ranges. 
Such methods would provide us with unprecedented insight into how biodiversity is distributed on our planet and how it is changing over time. 
However, there are potential negative consequences associated with inaccurate range predictions generated by automated methods, \eg a downstream conservation decision could be made based on an erroneous range map, resulting in wasted resources. 
Thus, it is important for practitioners to scrutinize the outputs of models such as ours.  

Another issue associated with training models on species observation data is that there is a risk that sensitive information (\eg the locations of protected species) could be leaked or extracted from the models. 
To respect this concern, the models in this work were trained using only publicly available information which does not include any sensitive observations. 
Finally, our approach integrates predictions from pre-trained large language models. 
These models are known to be biased and capable of hallucinating and fabricating outputs. 
Spatially localizing the outputs of such models runs the risk of amplifying such biases if used inappropriately.

\bibliography{main}
\bibliographystyle{icml2025}

\newpage
\appendix
\onecolumn

\setcounter{table}{0}   
\renewcommand{\thetable}{A\arabic{table}}
\setcounter{figure}{0}
\renewcommand{\thefigure}{A\arabic{figure}}

{\LARGE Appendix}\\

In this appendix we provide additional quantitative and qualitative results, analysis, implementation details, and ablations.

In \cref{sec:addtional_experiments}, we provide additional results for uncertainty quantification, use of taxonomic rank text, a `distance weighted' MAP metric, and show expanded results from \cref{fig:low_shot} for the non-low-shot setting. In \cref{sec:eco_results}, we perform an ecologically relevant analysis of our results, showing how performance varies with region, range size, and taxonomic group. In \cref{supp:impl_details}, we provide details on the implementation of \modelname and the baseline approaches, and of the training and evaluation procedure. In \cref{sec:app_ablations}, we provide additional ablations of \modelname, investigating the impact of training data, different input features, and modifications to the architecture. Finally, in \cref{sec:app_qual_results}, we show additional qualitative results, including visualizing zero-shot and few-shot ranges for species and non-species concepts, and comparisons to ranges produced by LE-SINR and SINR approaches.

\section{Additional Quantitative Results}
\label{sec:addtional_experiments}

\subsection{Uncertainty Quantification}

Here, we report results for an ensemble based on \modelname and quantify the uncertainty in the ensemble's predictions using methods adapted from \citet{Poggi_CVPR_2020}.
We create an ensemble by averaging the predictions of three \modelname models trained with different random seeds.
\cref{fig:different_seeds} shows range estimates from three such models, where we can see that each model can produce significantly different outputs.
We take the average of these models as the ensemble prediction and treat the variance between individual model predictions as an estimate of the uncertainty of the ensemble.
If all models agree that a species is either present or absent at a location, then the uncertainty will be low, while if models have different predictions, then the uncertainty will be high.
in \cref{tab:uncertainty}, comparing `Ensemble' MAP to `Model' MAP (repeated from \modelname `Text' MAP in \cref{tab:main_results_table_snt} for convenience), we can see that creating an ensemble increases MAP by 0.01 to 0.02, agreeing with typical findings that ensembling can improve performance relative to individual models \citep{ensembling}.

In order to quantify uncertainty of our ensemble we follow an approach used in \citet{Poggi_CVPR_2020}. 
We iteratively remove locations from the evaluation dataset that have the highest estimated uncertainty and recalculate the MAP without these locations.
In our case we remove 2\% of the data at each step.
If the ensemble uncertainty aligns with how likely it is to be incorrect, then the MAP at each step will increase as the data with higher estimated uncertainty is removed from the evaluation.
We can then plot the MAP against the fraction of data used for the evaluation.
Taking the area under the curve generated doing this gives us the estimated `Sparsification Error AUC' (SEAUC).
If, instead, we remove 2\% of locations randomly at each step then the MAP for each evaluation will remain approximately equal to the MAP when using the entire evaluation dataset.
The `random' SEAUC in this case is effectively equal numerically to the MAP using all evaluation data and can be estimated as such.
Taking the difference between the estimated SEARC and the MAP then gives us the `Area Under the Random Gain' (AURG).
A positive AURG shows that the ensemble's estimate of how uncertain its predictions are is better than random guessing.
In \cref{tab:uncertainty} we see that the AURG is positive for all number of context locations and increases as more context locations are provided, showing that \modelname ensembles can provide a useful estimate of how certain they are about a prediction and that providing more context locations allows the ensemble to be more accurate in this uncertainty estimate.

\begin{figure}[h!]
    \centering
    \begin{minipage}{0.04\textwidth}
    \end{minipage}%
    \hspace{0.5em}
    \begin{minipage}{0.45\textwidth}
        \centering \textbf{Range (Mean)}
    \end{minipage}%
    \hspace{0.5em}
    \begin{minipage}{0.45\textwidth}
        \centering \textbf{Uncertainty (Variance)}
    \end{minipage}%
    
    \vspace{1em}

    \begin{minipage}{0.04\textwidth}
        \rotatebox{90}{\textbf{Range}}
    \end{minipage}%
    \hspace{0.5em}
    \begin{minipage}{0.45\textwidth}
        \centering
        \includegraphics[width=\linewidth]{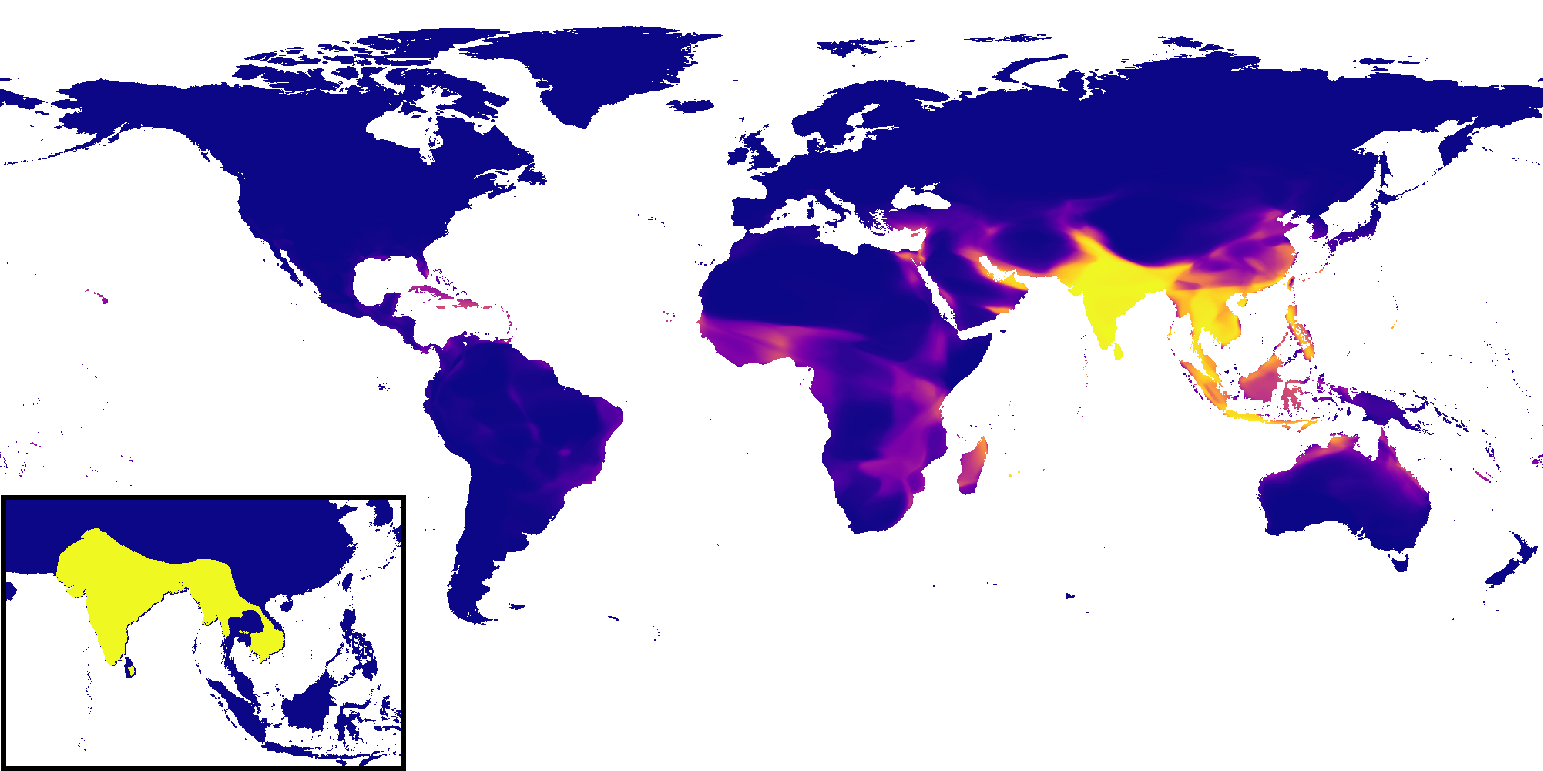}
    \end{minipage}%
    \hspace{0.5em}
    \begin{minipage}{0.45\textwidth}
        \centering
        \includegraphics[width=\linewidth]{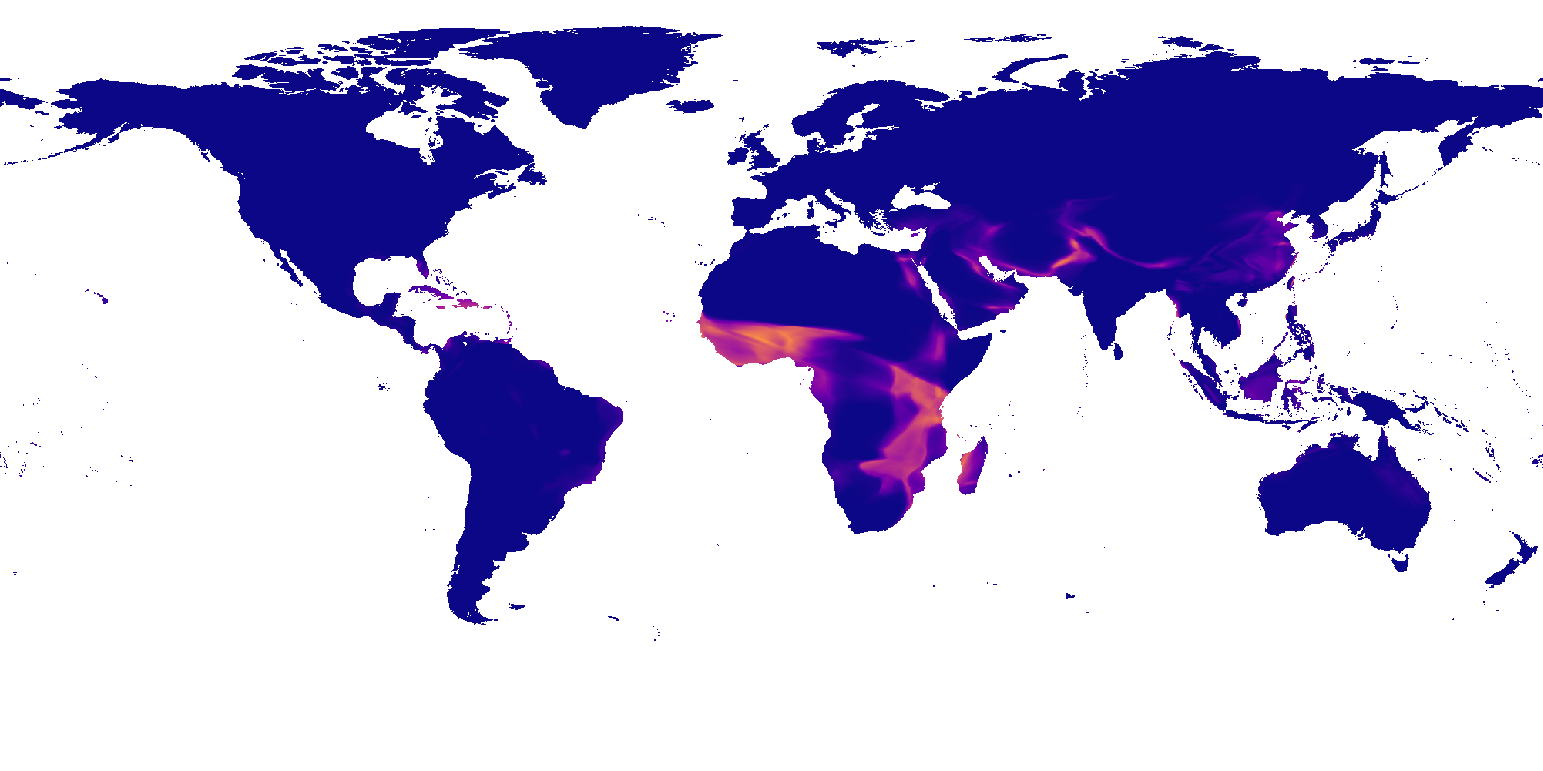}
    \end{minipage}%

    \vspace{1em}

    \begin{minipage}{0.04\textwidth}
        \rotatebox{90}{\textbf{Habitat}}
    \end{minipage}%
    \hspace{0.5em}
    \begin{minipage}{0.45\textwidth}
        \centering
        \includegraphics[width=\linewidth]{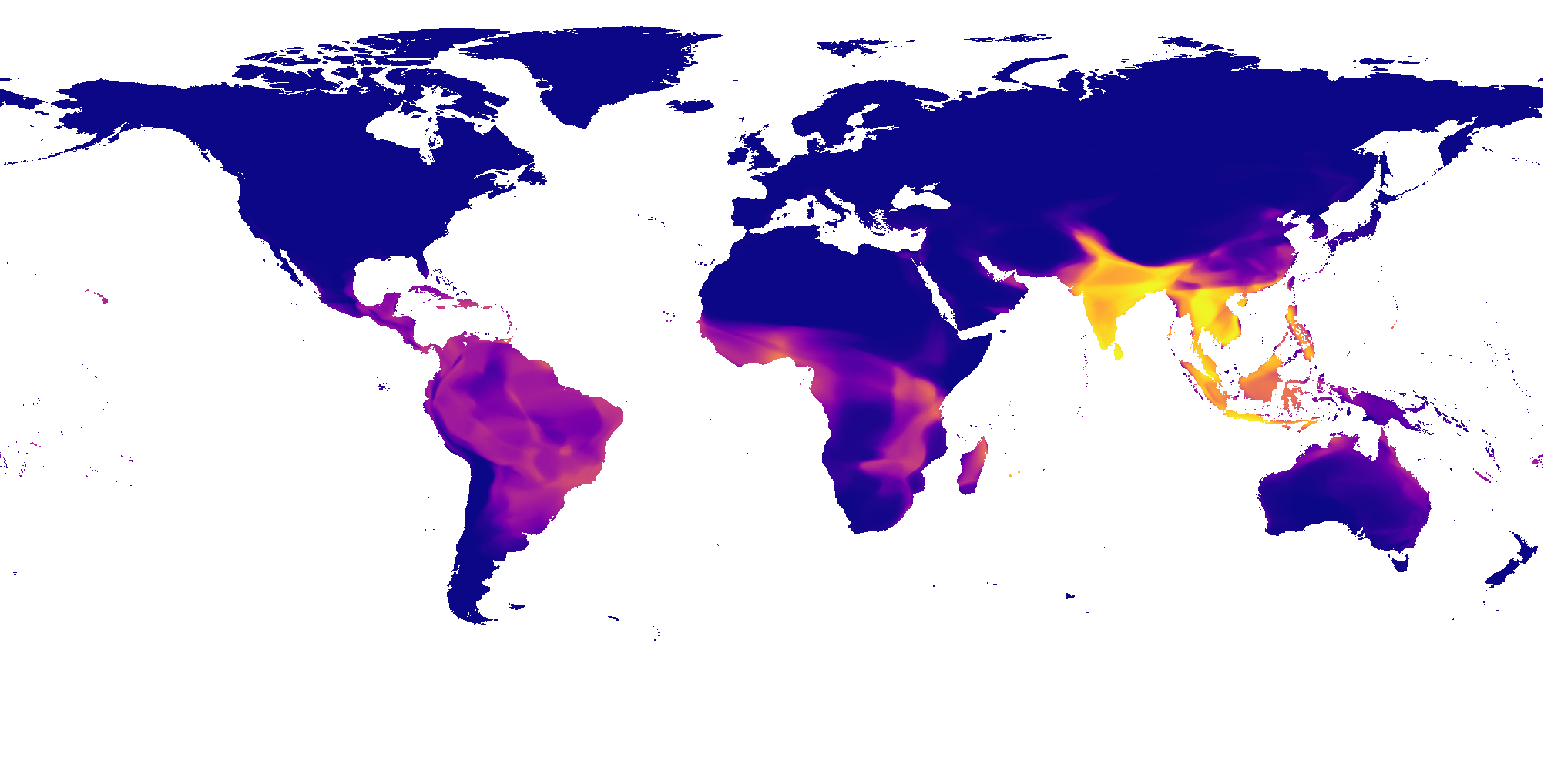}
    \end{minipage}%
    \hspace{0.5em}
    \begin{minipage}{0.45\textwidth}
        \centering
        \includegraphics[width=\linewidth]{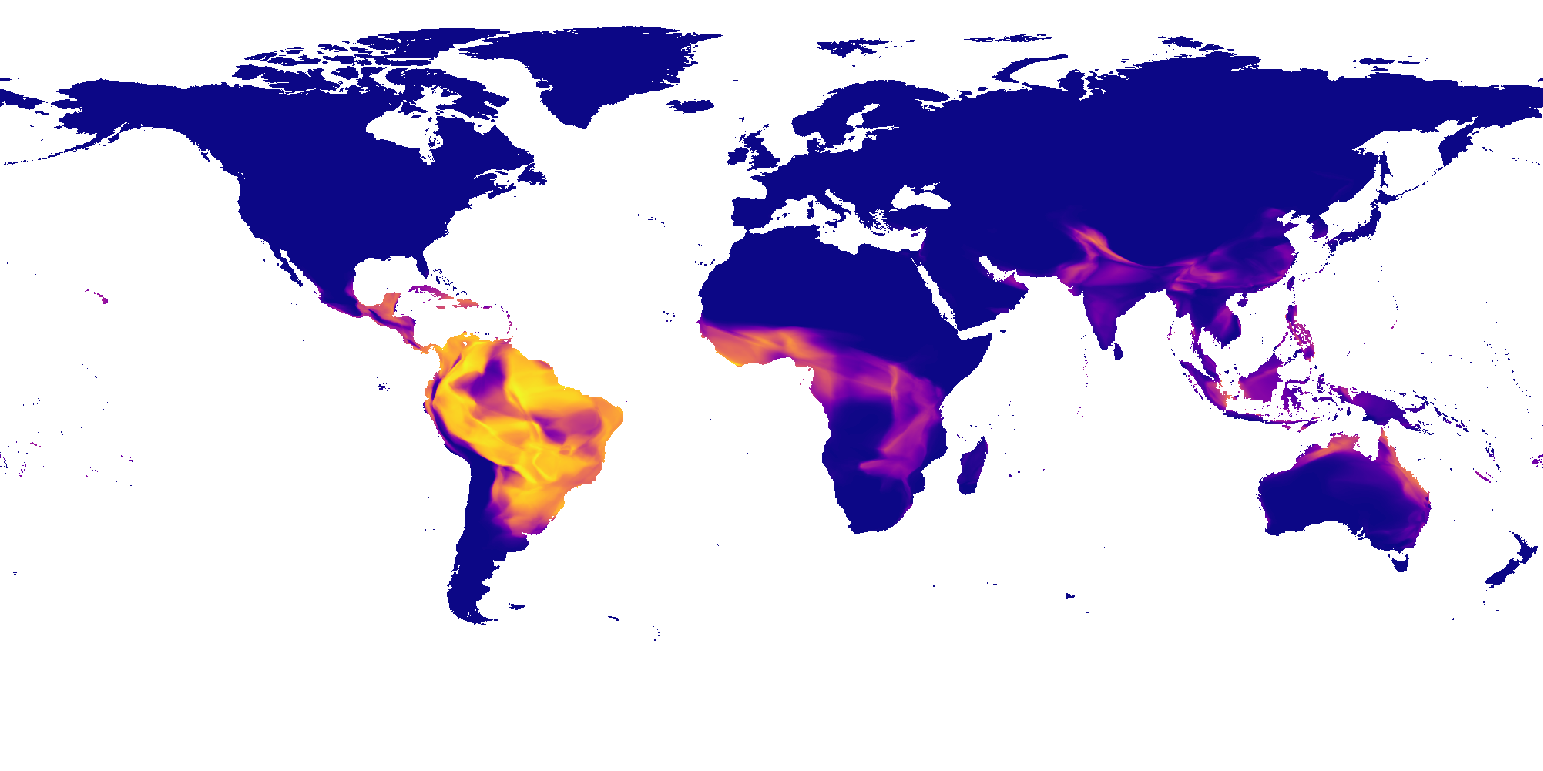}
    \end{minipage}%

    \vspace{1em}

    \begin{minipage}{0.04\textwidth}
        \rotatebox{90}{\textbf{No Text}}
    \end{minipage}%
    \hspace{0.5em}
    \begin{minipage}{0.45\textwidth}
        \centering
        \includegraphics[width=\linewidth]{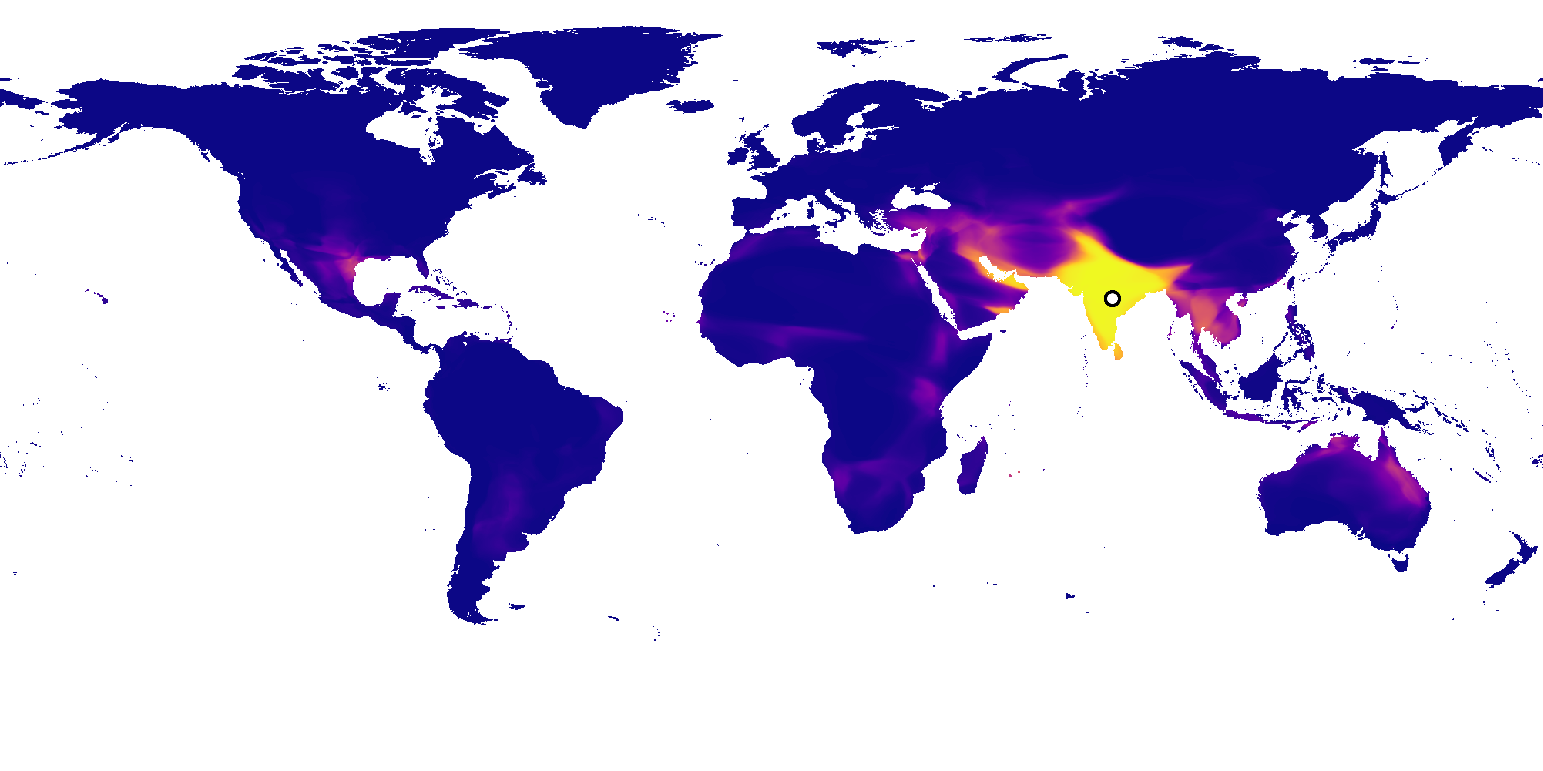}
    \end{minipage}%
    \hspace{0.5em}
    \begin{minipage}{0.45\textwidth}
        \centering
        \includegraphics[width=\linewidth]{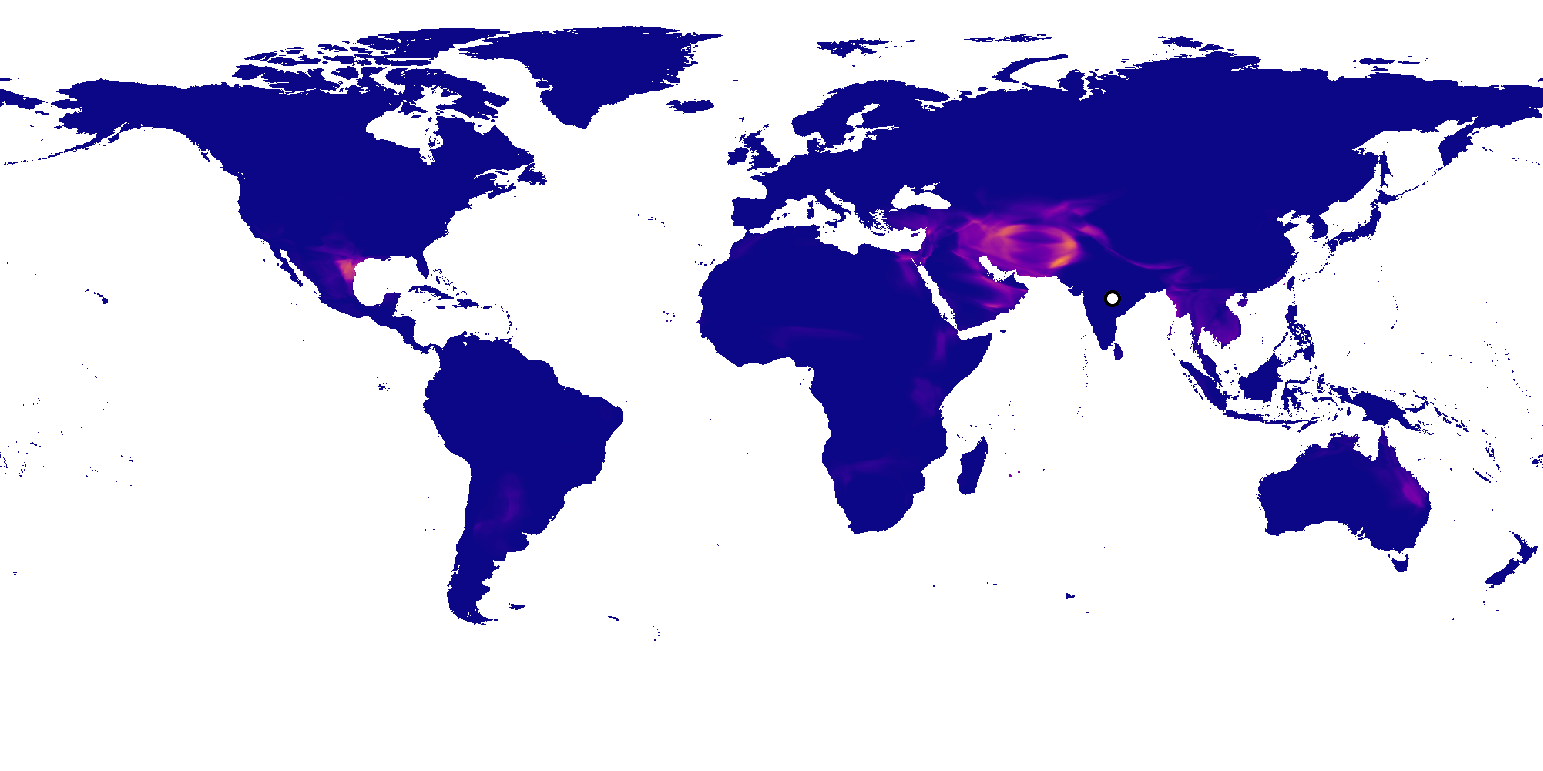}
    \end{minipage}%
    \vspace{-15pt}
    \caption{\change{\textbf{Range estimates and visualized uncertainty estimates for a \modelname ensemble.}
    We display range estimates and uncertainties for the \texttt{Yellow-footed Green Pigeon} from an ensemble of three \modelname models.
    Zero-shot estimates are based on `Range' text (top) and `Habitat' text (middle). A few-shot estimate using no text and a single context location (bottom) is also shown.
    Range estimates for the ensemble (left) are a mean average of individual model predictions, while uncertainties (right) are estimated using the variances of the model predictions.
    The uncertainty is lower in the region of the expert-derived range where all models agree, higher at the edges of this region where different models have different estimates of the extent of the range, and high in areas far from the true range where single models have incorrectly predicted presence.\\ 
    \textit{Range Text:} ``The yellow-footed green pigeon is found in the Indian subcontinent and parts of Southeast Asia. It is the state bird of Maharashtra."\\
    \textit{Habitat Text:} ``The species is a habitat generalist, preferring dense forest areas with emergent trees, especially Banyan trees, but can also be spotted in natural remnants in urban areas. They forage in flocks and are often seen sunning on the tops of trees in the early morning.'' 
    }}
    \label{fig:uncertainty_qual}
\end{figure}

In \cref{fig:uncertainty_qual} we visualize the ranges (means) and uncertainties (variances) for our \modelname ensemble for the \texttt{Yellow-footed Green Pigeon}, using `Range' text, `Habitat' text, or a single context location.
We observe that the mean is high in the region of the expert-derived range, and lower in areas far from the range where a single model has erroneously predicted presence.
The variance is generally lower in the region of the expert-derived range where all models agree, higher at the edges of this region where different models have different estimates of the extent of the range, and high in areas far from the true range where single models have incorrectly predicted presence, such as South America when `Habitat' text is provided, or parts of Africa when `Range' text is provided.

\begin{table}[h]
\caption{\textbf{Uncertainty quantification with ensembles.} We show MAP on the S\&T dataset for an ensemble of three \modelname models using `Range' text and differing numbers of context locations (Ensemble MAP), with metrics for uncertainty quantification adapted from \citet{Poggi_CVPR_2020}.
We report `Sparsification Error AUC' (SEAUC) and `Area Under the Random Gain' (AURG) for the ensemble.
Positive AURG shows the ensemble is performing better than random chance at estimating its uncertainty.
We also present results for the same text and context locations for individual \modelname models for comparison (Model MAP). 
Ensembling slightly improves performance for all numbers of context locations.}
\centering
  \begin{tabular}{c|c|c|c|c}
    \# Context & Model MAP & Ensemble MAP & SEAUC & AURG \\
    \hline
    0  & 0.64 & 0.66 & 0.68 & 0.03 \\
    1  & 0.66 & 0.68 & 0.71 & 0.03 \\
    2  & 0.67 & 0.69 & 0.73 & 0.03 \\
    3  & 0.68 & 0.70 & 0.74 & 0.04 \\
    4  & 0.69 & 0.71 & 0.75 & 0.04 \\
    5  & 0.70 & 0.71 & 0.75 & 0.04 \\
    8  & 0.71 & 0.72 & 0.76 & 0.04 \\
    10 & 0.72 & 0.73 & 0.77 & 0.04 \\
    15 & 0.72 & 0.73 & 0.78 & 0.05 \\
    20 & 0.72 & 0.74 & 0.79 & 0.05 \\
    50 & 0.73 & 0.74 & 0.78 & 0.05 \\
  \end{tabular}%
\label{tab:uncertainty}
\vspace{-10pt}
\end{table}

\subsection{Taxonomic Understanding} 
\label{sec:trt_section}

\begin{figure}
\centering
  \centering
    \rotatebox{90}{\hspace{-15pt}\parbox{20mm}{\small Class \newline \href{https://www.inaturalist.org/taxa/3-Aves}{\texttt{Aves}}}}
        \begin{minipage}{0.42\textwidth}
        \begin{overpic}[trim={0 0.5cm 0 0},clip,width=\linewidth]{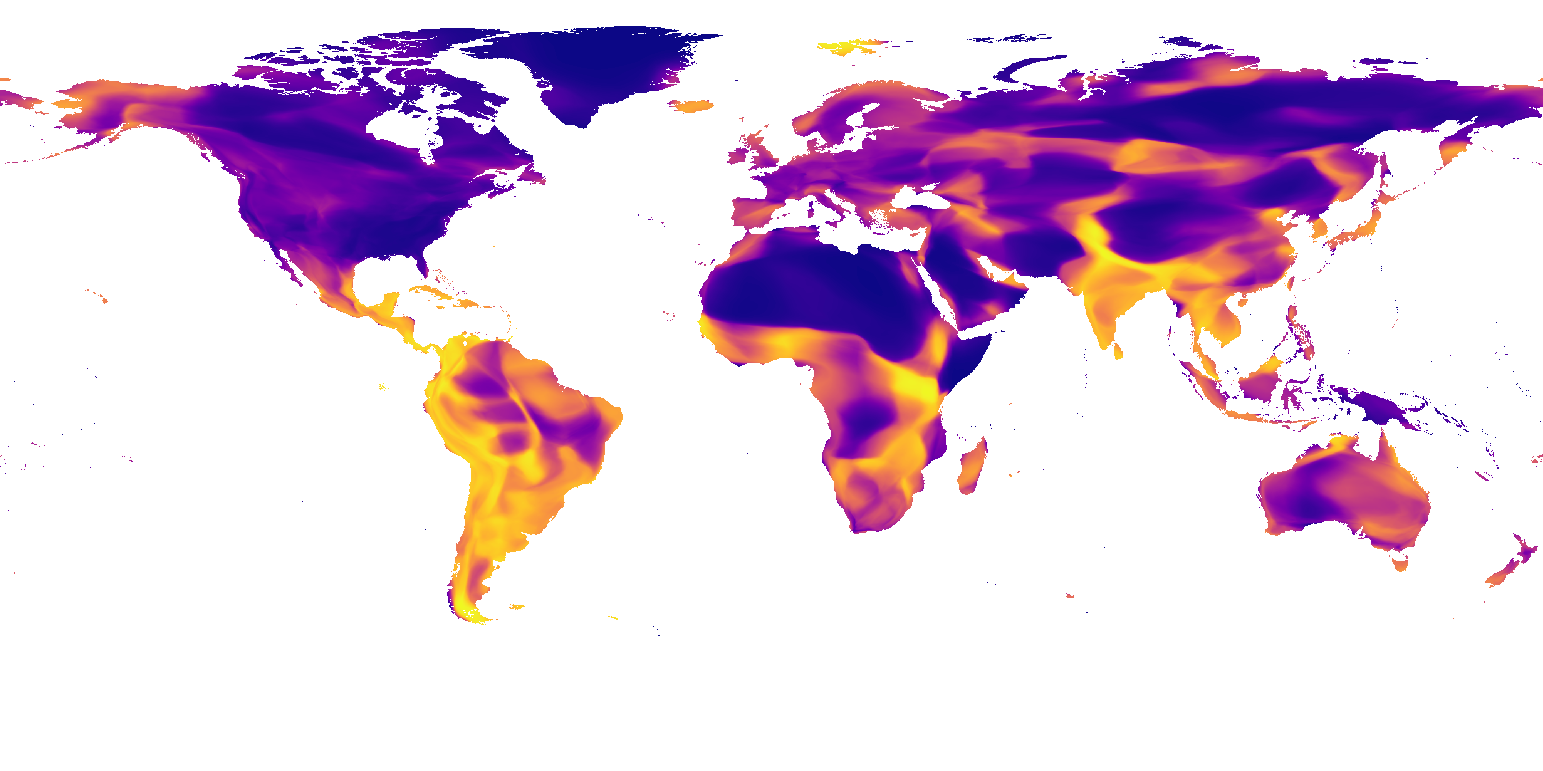}
            \put(0,0){%
              {%
                \setlength\fboxsep{0pt}%
                \setlength\fboxrule{1pt}%
                \fbox{%
                  \includegraphics[width=0.26\textwidth]{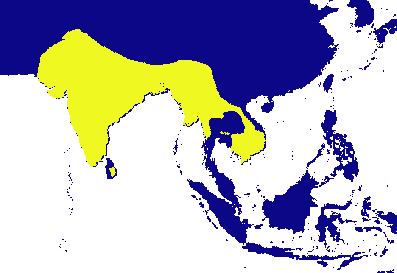}%
                }%
              }%
            }
        \end{overpic}
        \end{minipage}
    \begin{minipage}{0.42\textwidth}
        \includegraphics[width=\linewidth]{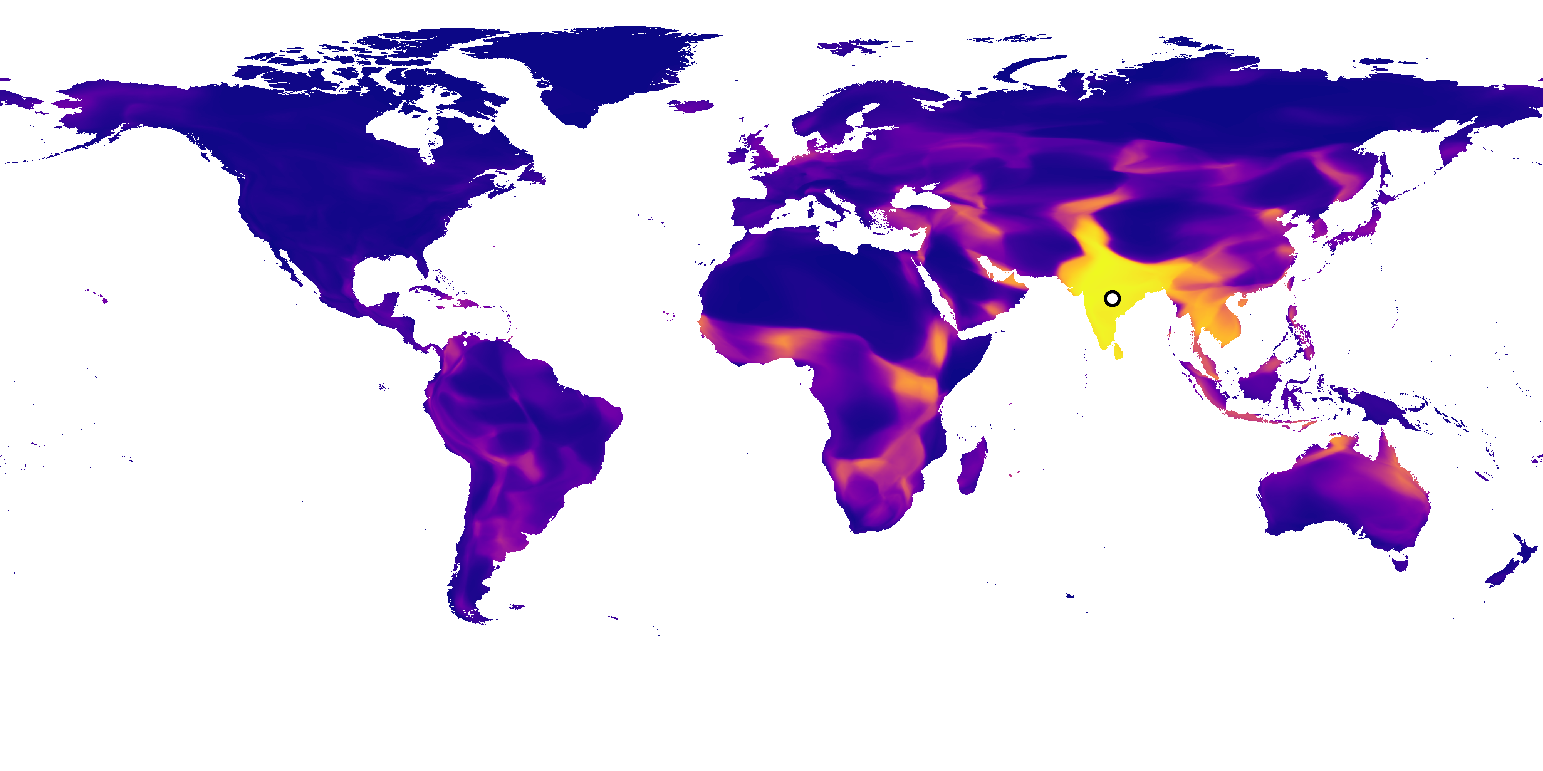} 
    \end{minipage}

    \rotatebox{90}{\hspace{-15pt}\parbox{20mm}{\small Order \newline \href{https://www.inaturalist.org/taxa/2708-Columbiformes}{\texttt{Columbiformes}}}}
    \begin{minipage}{0.42\textwidth}
        \includegraphics[width=\linewidth]{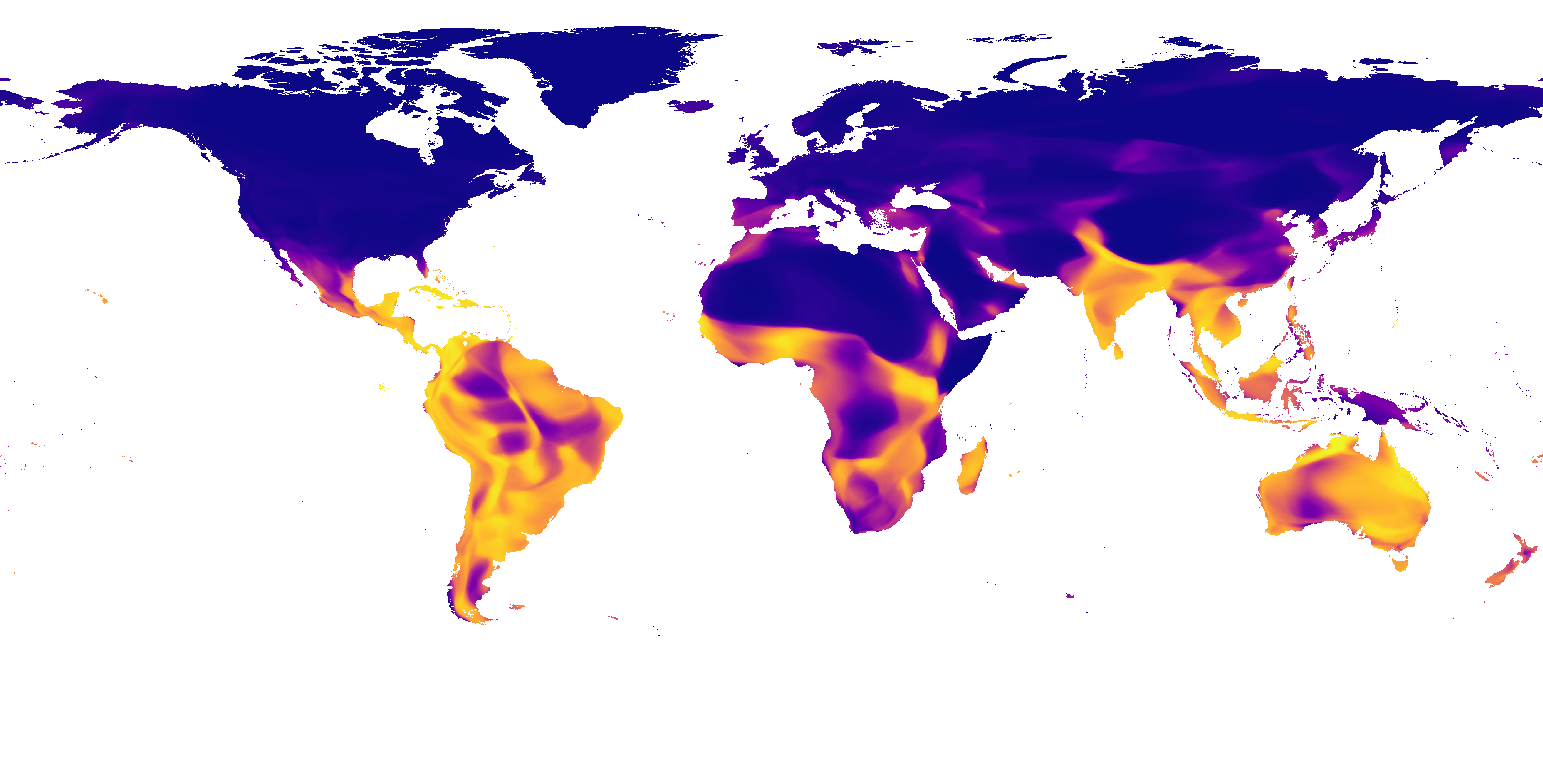} 
    \end{minipage}
    \begin{minipage}{0.42\textwidth}
        \includegraphics[width=\linewidth]{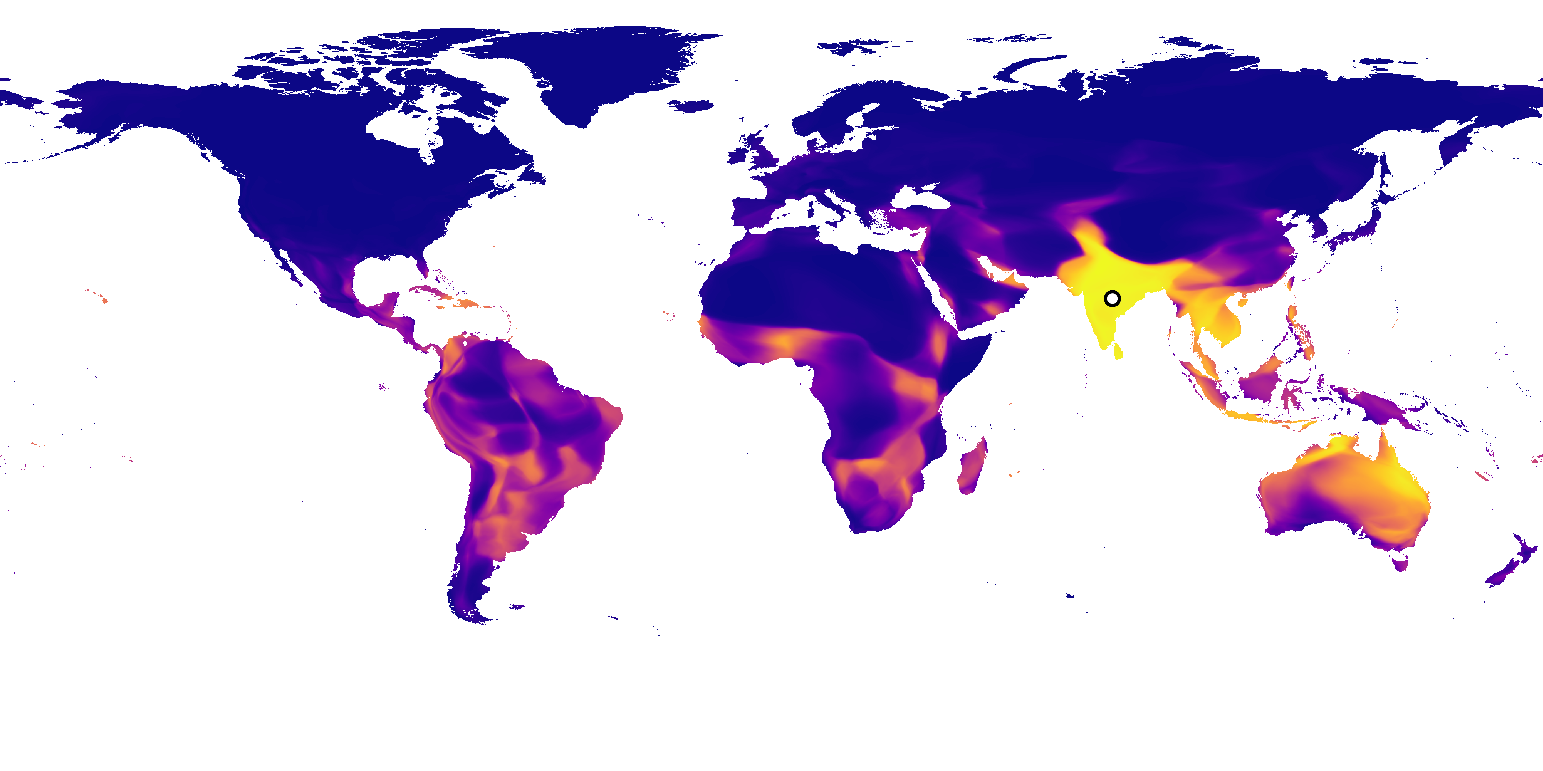} 
    \end{minipage}

    \rotatebox{90}{\hspace{-15pt}\parbox{20mm}{\small Family \newline \href{https://www.inaturalist.org/taxa/2715-Columbidae}{\texttt{Columbidae}}}}
    \begin{minipage}{0.42\textwidth}
        \includegraphics[width=\linewidth]{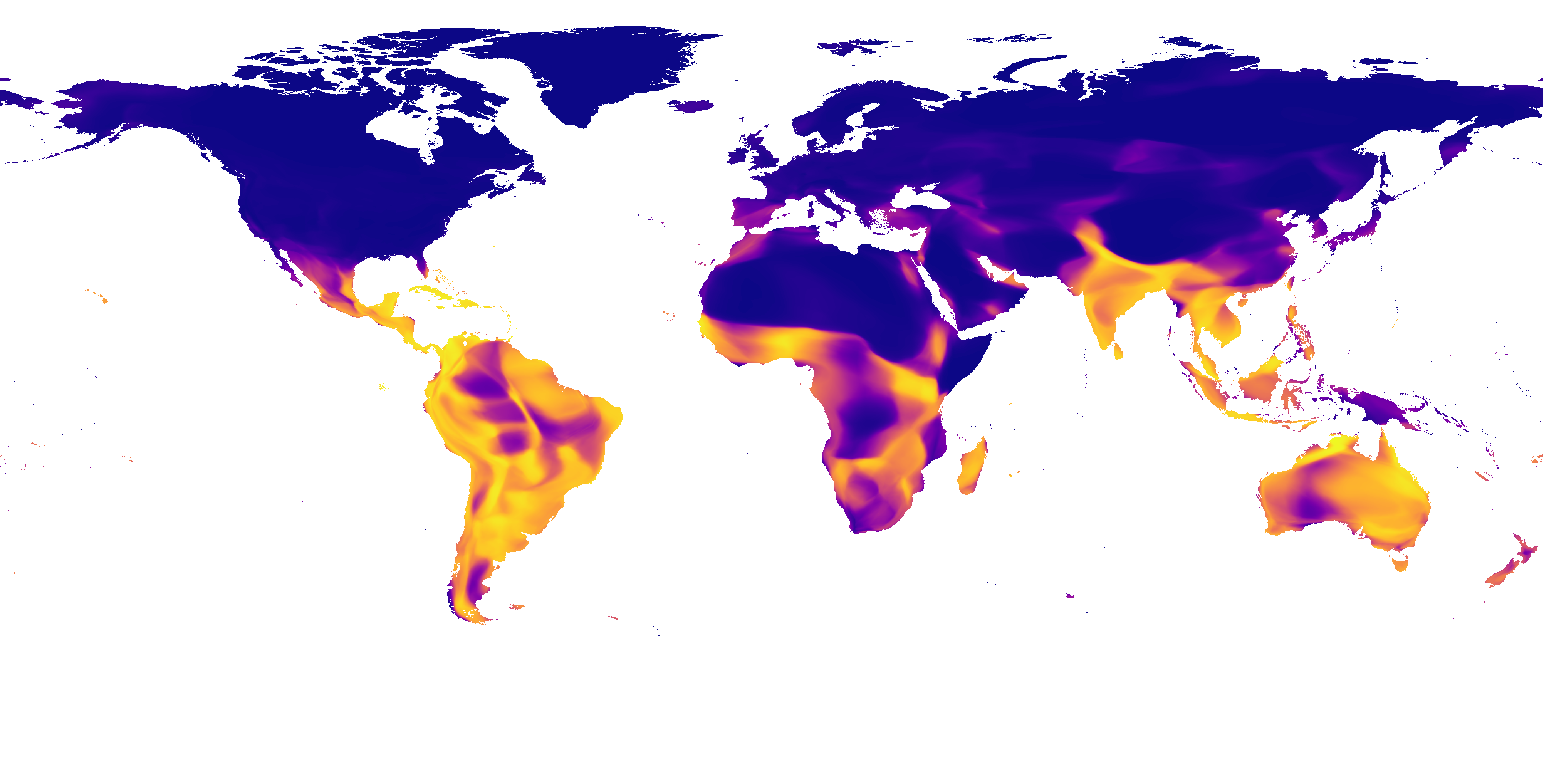} 
    \end{minipage}
    \begin{minipage}{0.42\textwidth}
        \includegraphics[width=\linewidth]{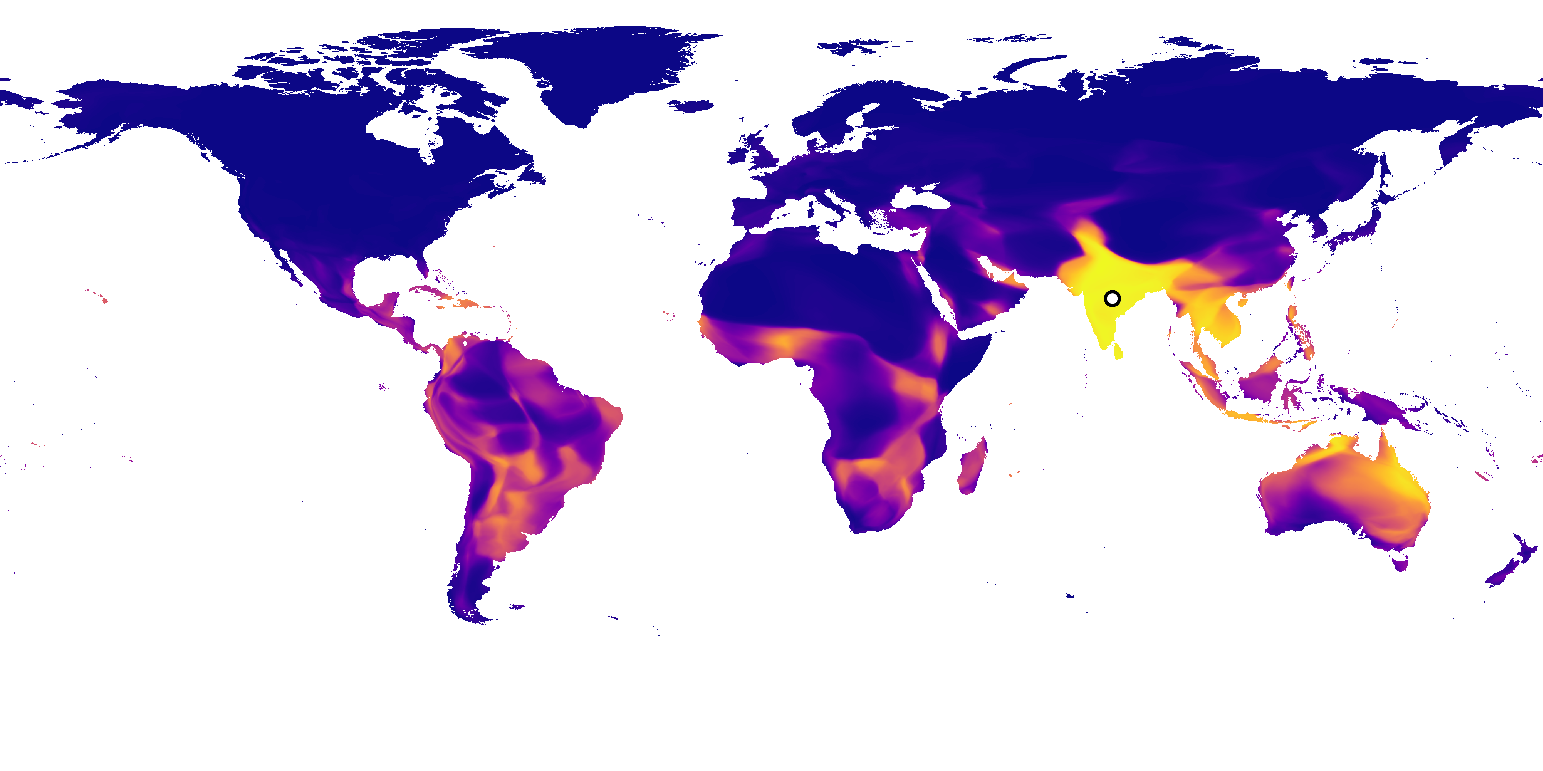} 
    \end{minipage}

    \rotatebox{90}{\hspace{-15pt}\parbox{20mm}{\small Genus \newline \href{https://www.inaturalist.org/taxa/3339-Treron}{\texttt{Treron}}}}
    \begin{minipage}{0.42\textwidth}
        \includegraphics[width=\linewidth]{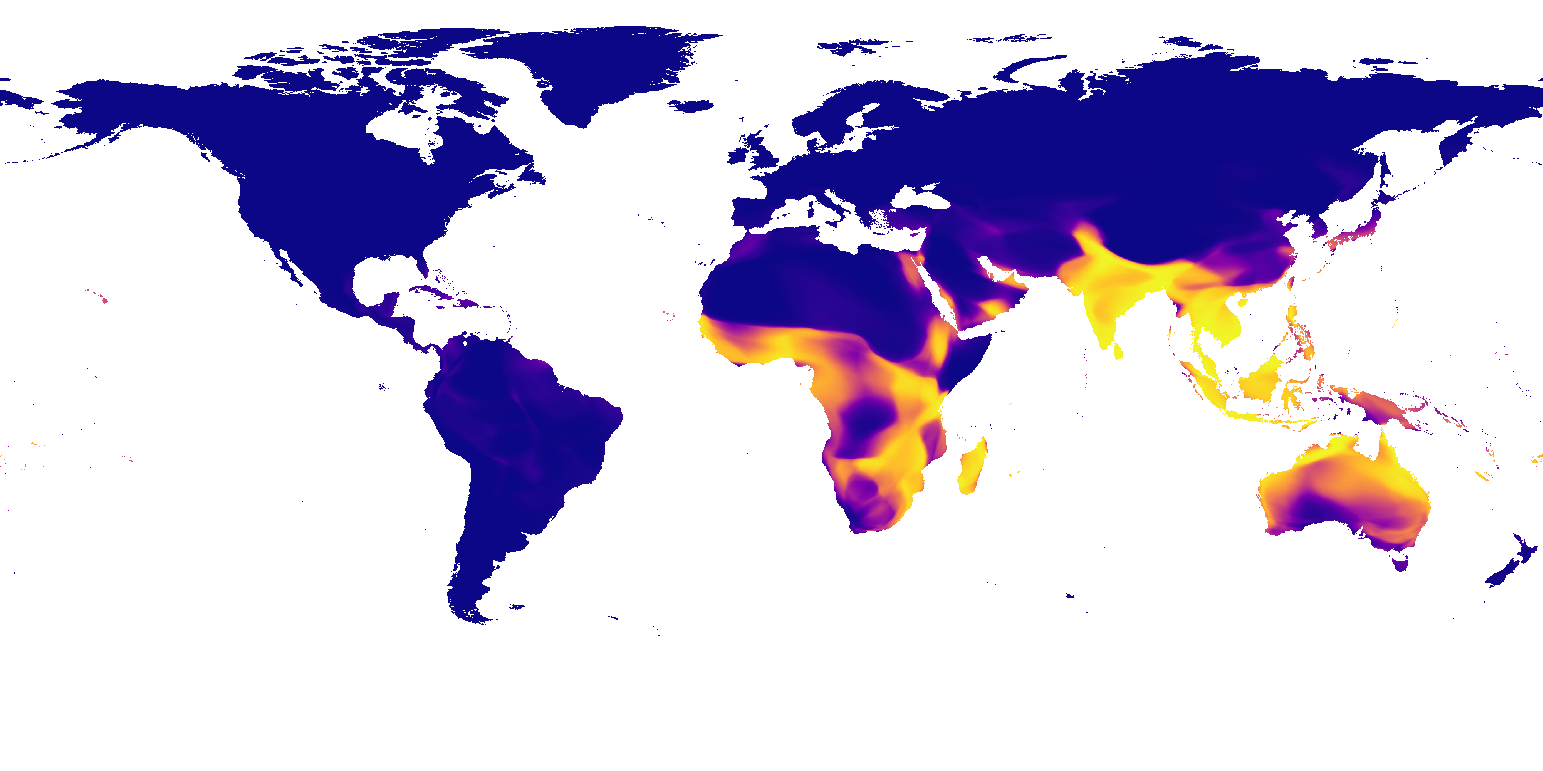} 
    \end{minipage}
    \begin{minipage}{0.42\textwidth}
        \includegraphics[width=\linewidth]{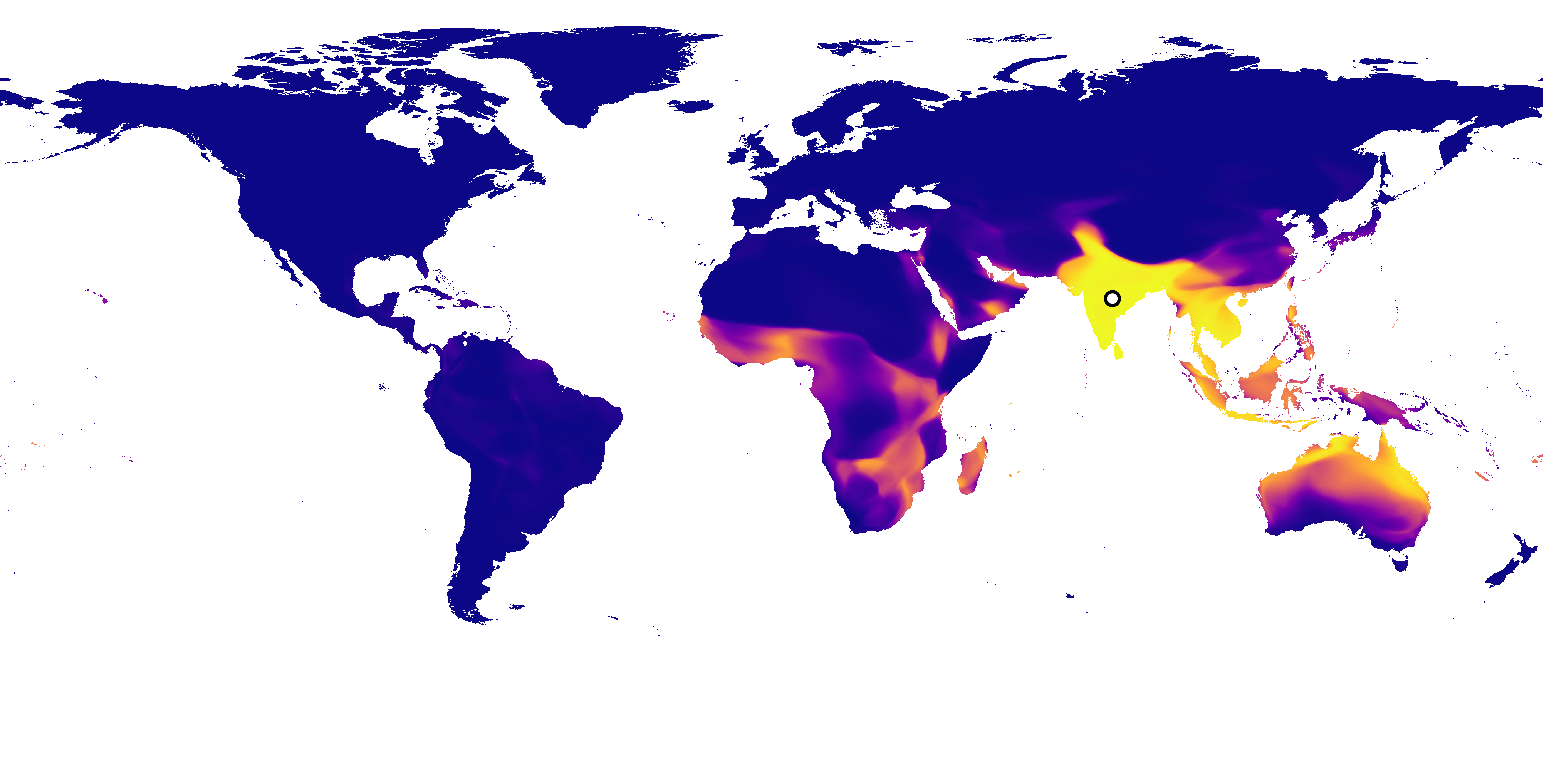} 
    \end{minipage}

    \rotatebox{90}{\hspace{-15pt}\parbox{20mm}{\small Species \newline \href{https://www.inaturalist.org/taxa/3352-Treron-phoenicopterus}{\texttt{Phoenicopterus}}}}
    \begin{minipage}{0.42\textwidth}
        \includegraphics[width=\linewidth]{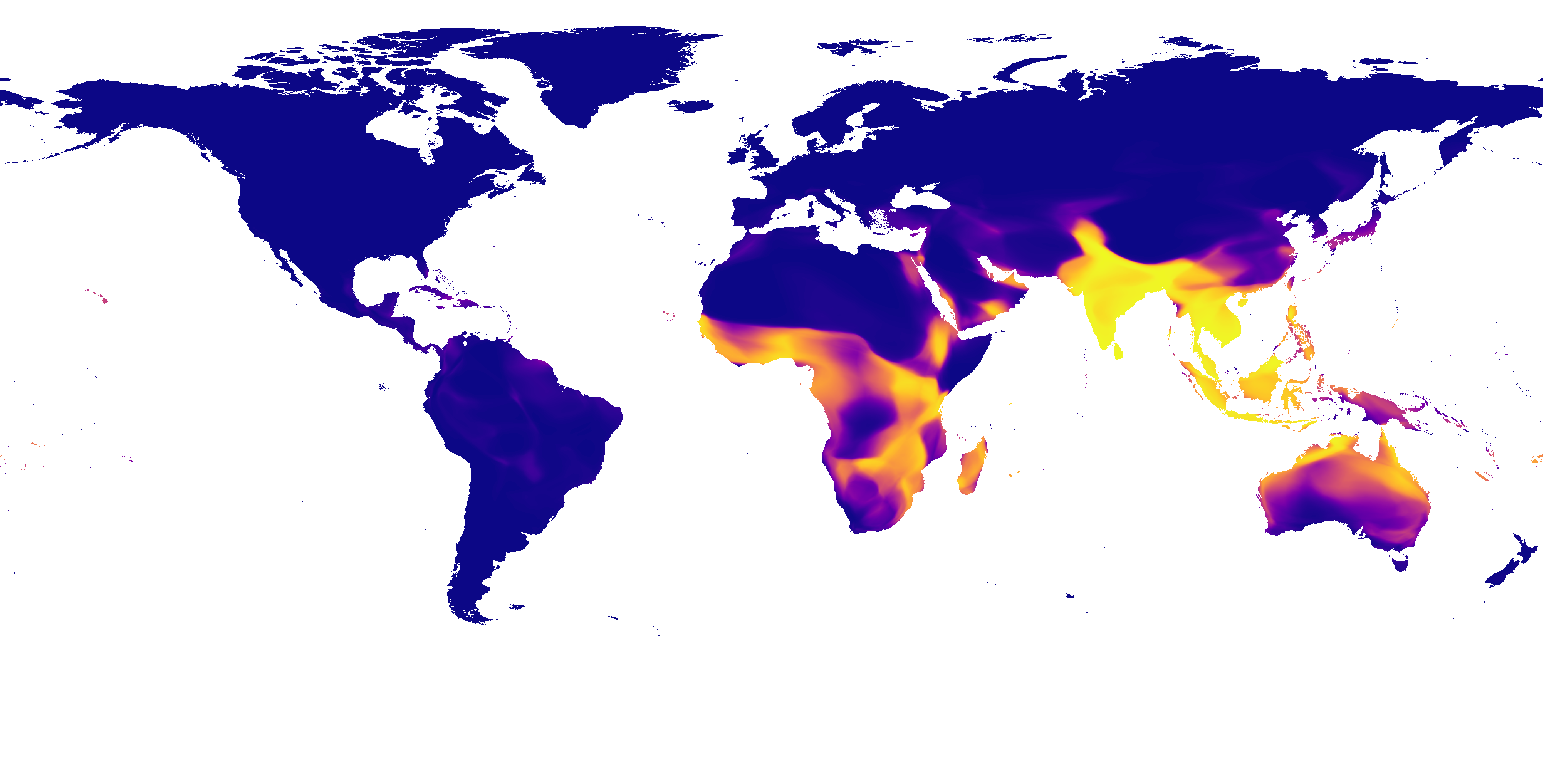} 
        \centering{\fcolorbox{black}{gray!30}{\strut\small [TRT Text]}}
    \end{minipage}
    \begin{minipage}{0.42\textwidth}
        \includegraphics[width=\linewidth]{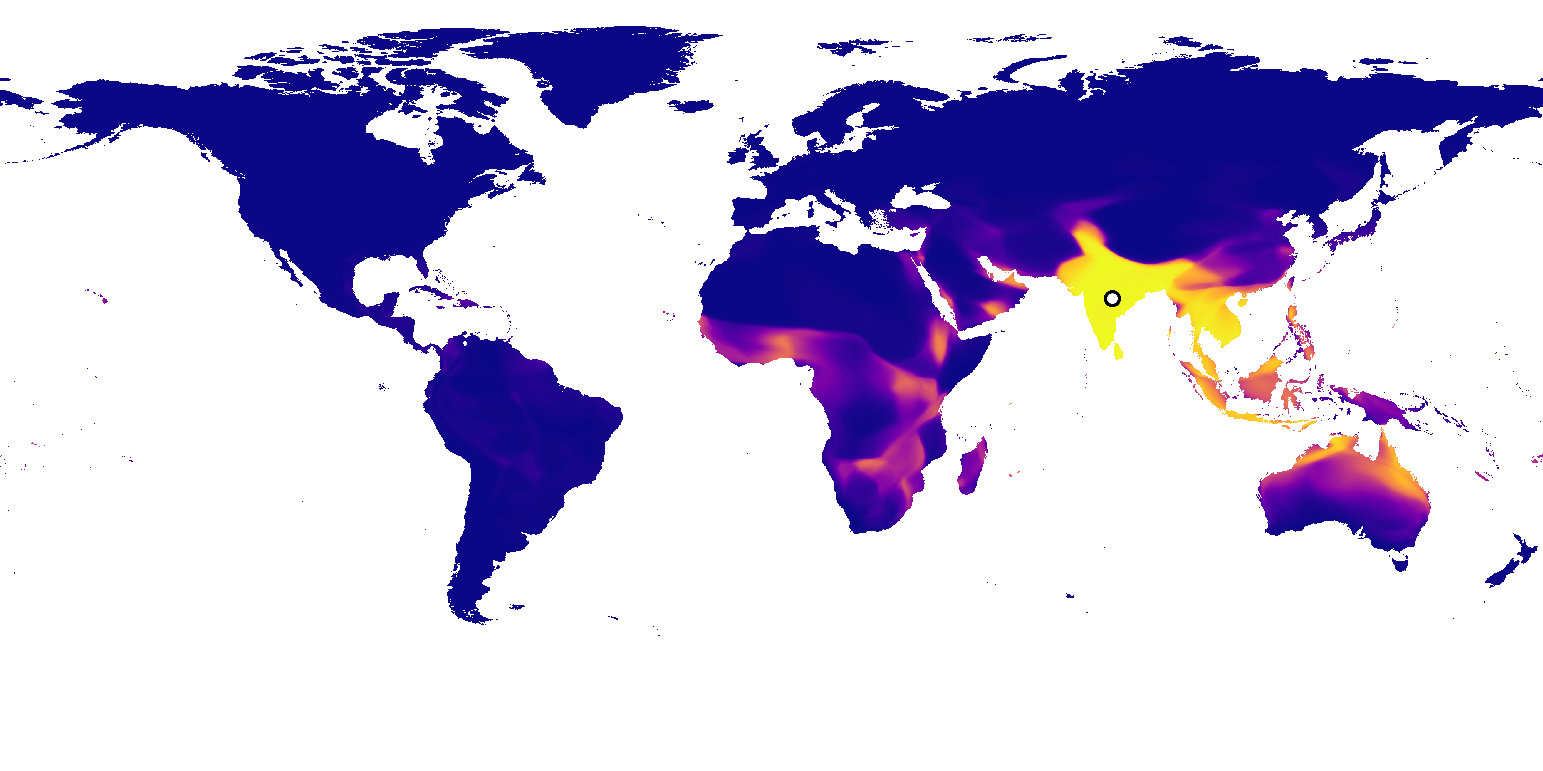} 
        \centering{\fcolorbox{black}{gray!30}{\strut\small [TRT Text] + 1 Context Location}}
    \end{minipage}
    \caption{\change{\textbf{Zero-shot and one-shot range estimation using Taxonomic Rank Text (TRT).} 
    Range predictions for the species \href{https://www.inaturalist.org/taxa/3352-Treron-phoenicopterus}{\texttt{Yellow-footed Green Pigeon}} from \modelname model trained on taxonomic rank text as in LD-SDM~\citep{sastry2023ld}, with expert-derived range inset. 
    As seen in \cref{fig:ld_sdm_plot} and \cref{tab:ld_sdm_zeros_shot}, the text-based zero-shot predictions seem to more closely match the expert-derived range as more of the taxonomic rank text of the species is provided.
    Taxonomic rank text allows the model to somewhat localize predictions to areas where species sharing the provided taxonomy ranks are present in the training set.
    For example, \href{https://www.inaturalist.org/taxa/3-Aves}{\texttt{Birds}} are globally distributed and we see the model attempt to output this in the zero-shot `Class' visualization. 
    \href{https://www.inaturalist.org/taxa/2715-Columbidae}{\texttt{Pigeons and Doves}} are not found in the extreme north and providing these ranks reduces predictions in these areas (and much of the northern hemisphere).
    The model mostly manages to identify that \href{https://www.inaturalist.org/taxa/3339-Treron}{\texttt{Green Pigeons}} are found only in Africa and parts of Asia.
    A single observation significantly contracts the predicted ranges, particularly when less taxonomic information is provided.
    Click on the taxonomic rank names to visit the \href{https://www.inaturalist.org/}{iNaturalist} page for that taxa where the geographic distribution of observations for it can be observed.}}
\label{fig:taxonomy viz}
\end{figure}

Here, we investigate the impact of providing \modelname with an understanding of the species taxonomy. For this we provide `Taxonomic Rank Text' (TRT) instead of the Wikipedia-based free-form descriptions of a species that are used for our standard \modelname approach. 
This text gives the taxonomy of the species in decreasing taxonomic rank, in the form `class order family genus species', so for a dog we would give the text `\texttt{Mammalia Carnivora Canidae Canis Familiaris}'. 
During training, we select a rank uniformly at random and remove all ranks beneath that.
We hope that this process will force the model to learn an understanding of the distributions of not only individual species, but also genera, families, \etc.
This may be helpful when facing unseen species as knowledge of the genus or family may provide clues about where this species may be found.
This is similar to the approach used by LD-SDM~\citep{sastry2023ld}.

In \cref{tab:ld_sdm_zeros_shot} we show zero-shot performance for \modelname models trained on TRT on the IUCN and S\&T evaluation tasks.
We see that as we provide additional taxonomic information zero-shot performance improves, though it is still much worse than using habitat or range text.
This implies that the model has managed to develop some understanding of the distributions of genera \etc and can use this to help map a novel species that shares higher order taxonomy with species in the training set.

In \cref{fig:taxonomy viz} we provide some qualitative zero-shot and few-shot results showing the impact of training on taxonomic text.
We see that the model appears to narrow down on the correct range as more specific taxonomy is revealed to it, from predicting across the entire globe when just the class \texttt{Aves} is provided, to removing northern latitudes as 
the family \texttt{Columbidae} is added, and finally removing the new world when the genus is provided.
This broadly matches the actual distribution of these taxonomic ranks. 
Note that the relationship between taxonomic hierarchy and species range is likely complex as many speciation events (\ie when a species splits into two or more new ones) can be the result of physical geographic barriers separating populations over time. 

\begin{figure}[h]
    \centering
        \includegraphics[width=0.42\textwidth]{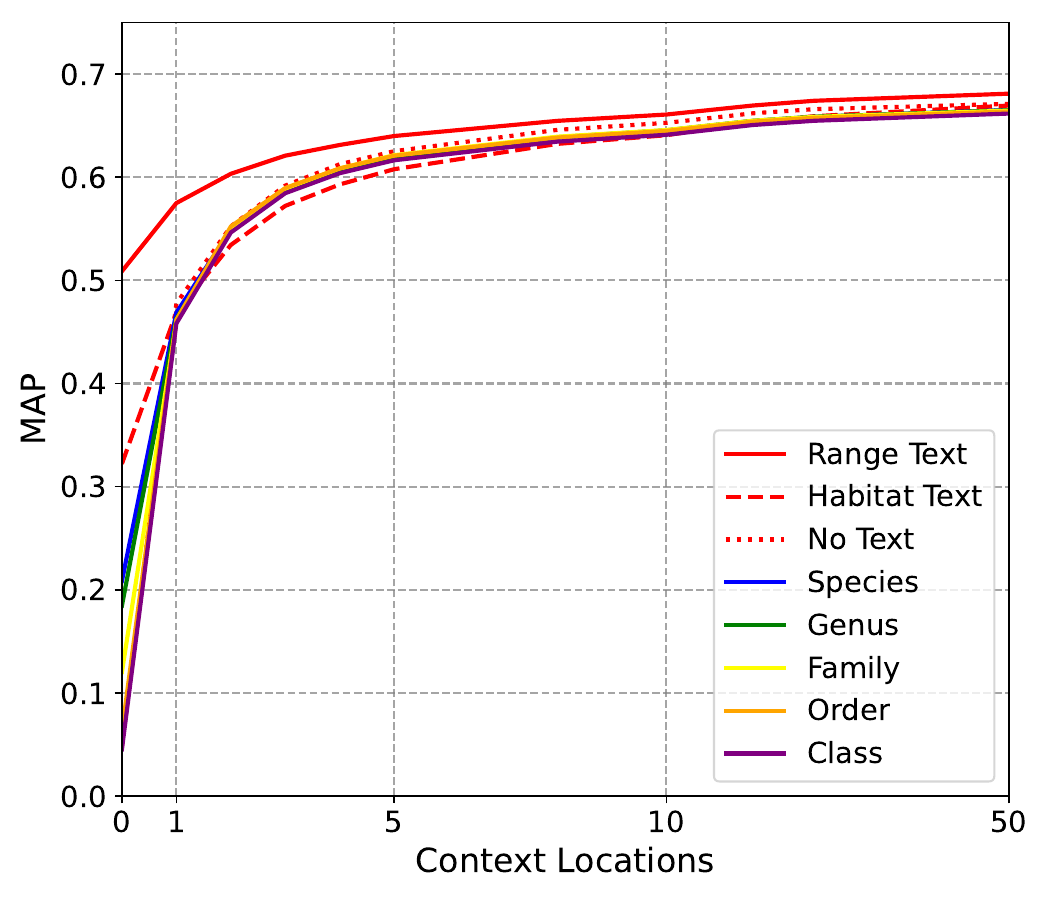}
        \hspace{15pt}
        \includegraphics[width=0.42\textwidth]{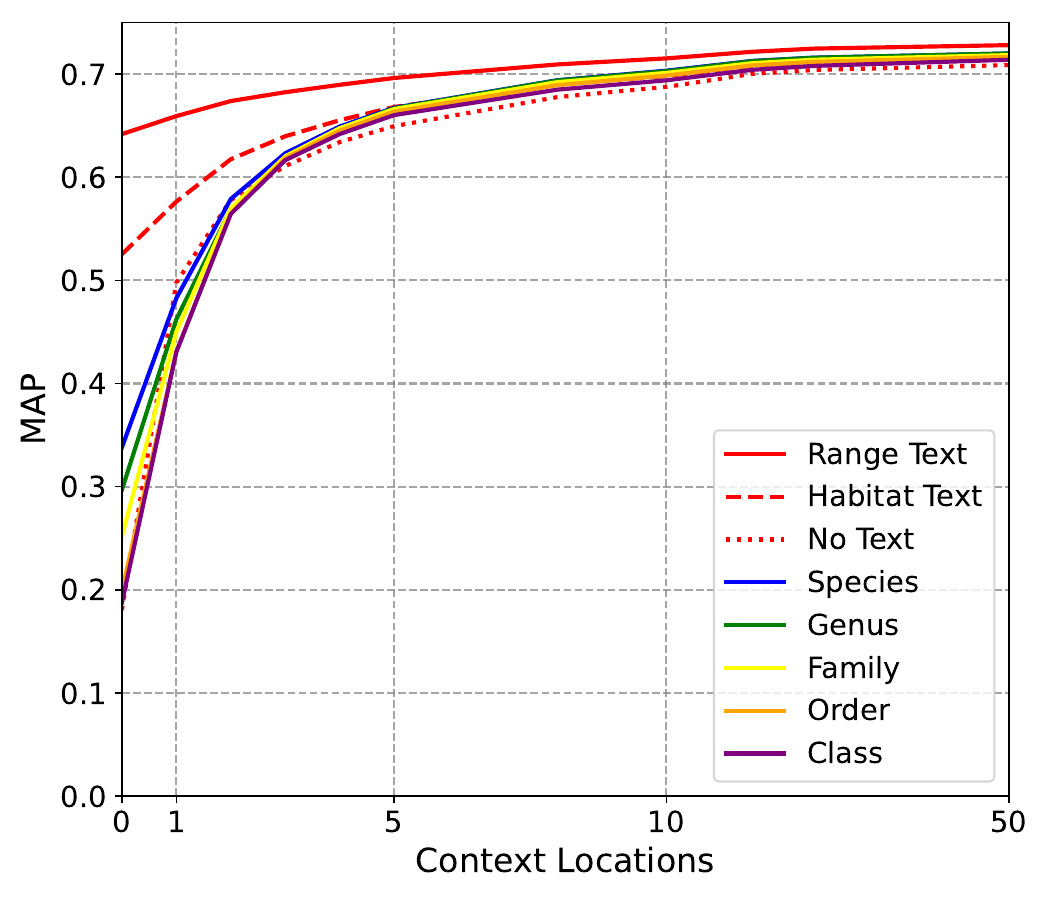} 
        \vspace{-15pt}
    \caption{\change{\textbf{Impact of training and evaluating with Taxonomic Rank Text.}
    Here we evaluate FS-SINR models trained using different context information on the IUCN dataset (left), and the S\&T dataset (right).
    `Class' indicates that only the taxonomic class of the species is provided as text during evaluation.
    `Order' indicates that the taxonomic class followed by the order is provided as a text string during evaluation, and so on, such that `Species' indicates that a text string in the format `class order family genus species' is provided during evaluation.
    Providing more specific taxonomic text increases zero-shot performance.
    This is also presented \cref{tab:ld_sdm_zeros_shot}.
    However we see that even the full taxonomy does not provide as much signal as habitat and range text for zero-shot range mapping.
    These more detailed texts provide more useful information for zero-shot range mapping - either actually mentioning geographic locations in the case of range text, or allowing the model to narrow predictions down to areas with specific features such as mountains and forests in the case of habitat text.
    When a single context location is provided, the choice of taxonomy text no longer seems to impact performance at all.
    It is possible that training on these less informative tokens means the model learns to pay less `attention' to these text tokens compared to the Wikipedia-based text tokens usually used during training.
    This could explain why different rank taxonomy text tokens seemingly provide no benefit when any context locations are provided to the model.}}
    \label{fig:ld_sdm_plot}
\end{figure}

In \cref{fig:ld_sdm_plot} we show few-shot results for \modelname models trained on TRT on the IUCN and SNT evaluation datasets.
Zero-shot performance improvement with increasing taxonomic information is evident, but after very few provided locations this effect seems to disappear.

\begin{table}[h]
\caption{{\bf Zero-shot results with taxonomy rank text}. 
We denote additional metadata used by models as RT for `Range Text' and HT for `Habitat Text'.
`Species', `Genus', `Family', `Order', `Class' refer to models trained and evaluated using taxonomic rank text.
Taxonomic information up to and including the specified rank is provided during evaluation.
}

\centering
\begin{tabular}{l|l|ll}
\textbf{Method}  & \textbf{Variant} & \textbf{IUCN} & \textbf{S\&T} \\ \hline
FS-SINR &  &  0.05   &  0.18  \\
FS-SINR  & HT     &   0.33   &  0.53  \\
FS-SINR & RT &   0.52   &  0.64  \\ \hline
FS-SINR & Class &  0.05  &  0.19 \\
FS-SINR & Order &  0.06 &  0.20 \\
FS-SINR & Family &  0.12  &  0.25 \\
FS-SINR & Genus &  0.18  &  0.30 \\
FS-SINR & Species & 0.21  &  0.34 \\
\end{tabular}
\label{tab:ld_sdm_zeros_shot}
\end{table}

\begin{figure}[h]
    \centering
    \begin{minipage}{0.31\textwidth}
        \centering \small\textbf{Unweighted MAP, $h = 0$}
    \end{minipage}%
    \hspace{0.5em}
    \begin{minipage}{0.31\textwidth}
        \centering \small\textbf{Weighted MAP, $h = 9$}
    \end{minipage}%
    \hspace{0.5em}
    \begin{minipage}{0.31\textwidth}
        \centering \small\textbf{Weighted MAP, $h = 99$}
    \end{minipage}
    
    \vspace{1em}

    \begin{minipage}{0.31\textwidth}
        \centering
        \includegraphics[width=\linewidth]{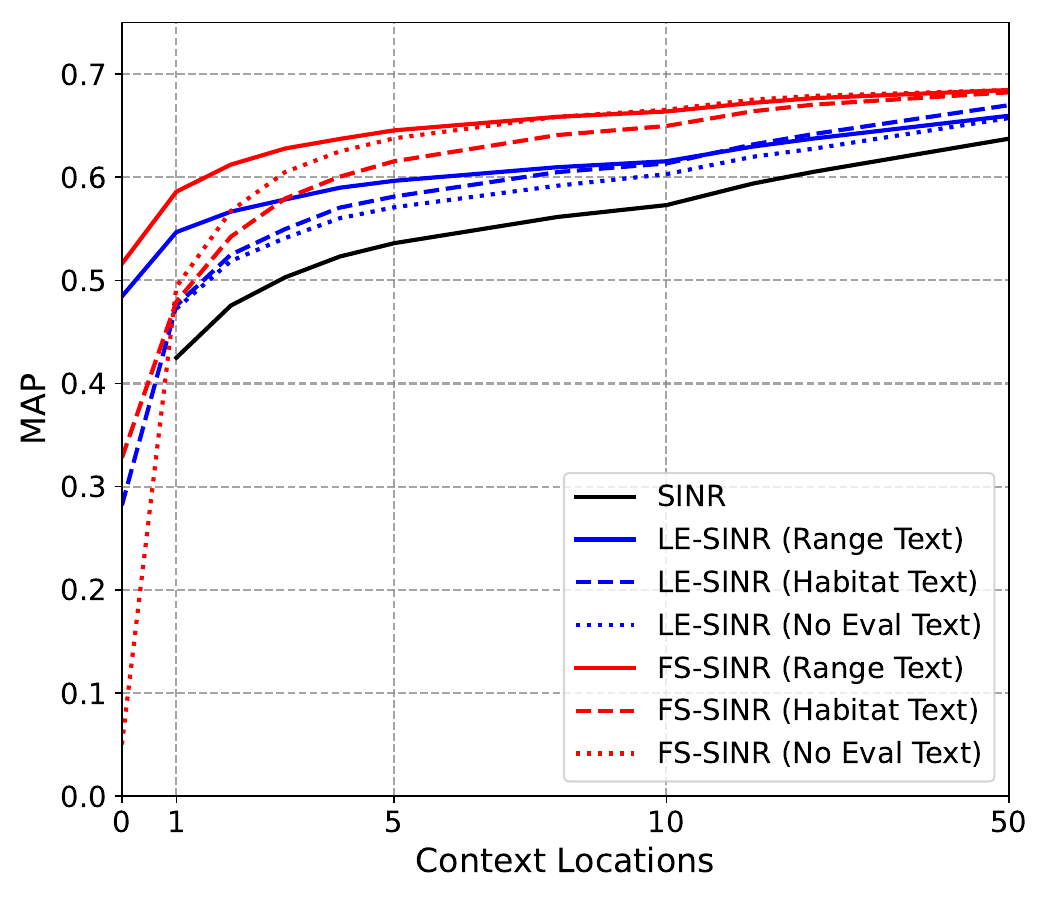}
    \end{minipage}%
    \hspace{0.5em}
    \begin{minipage}{0.31\textwidth}
        \centering
        \includegraphics[width=\linewidth]{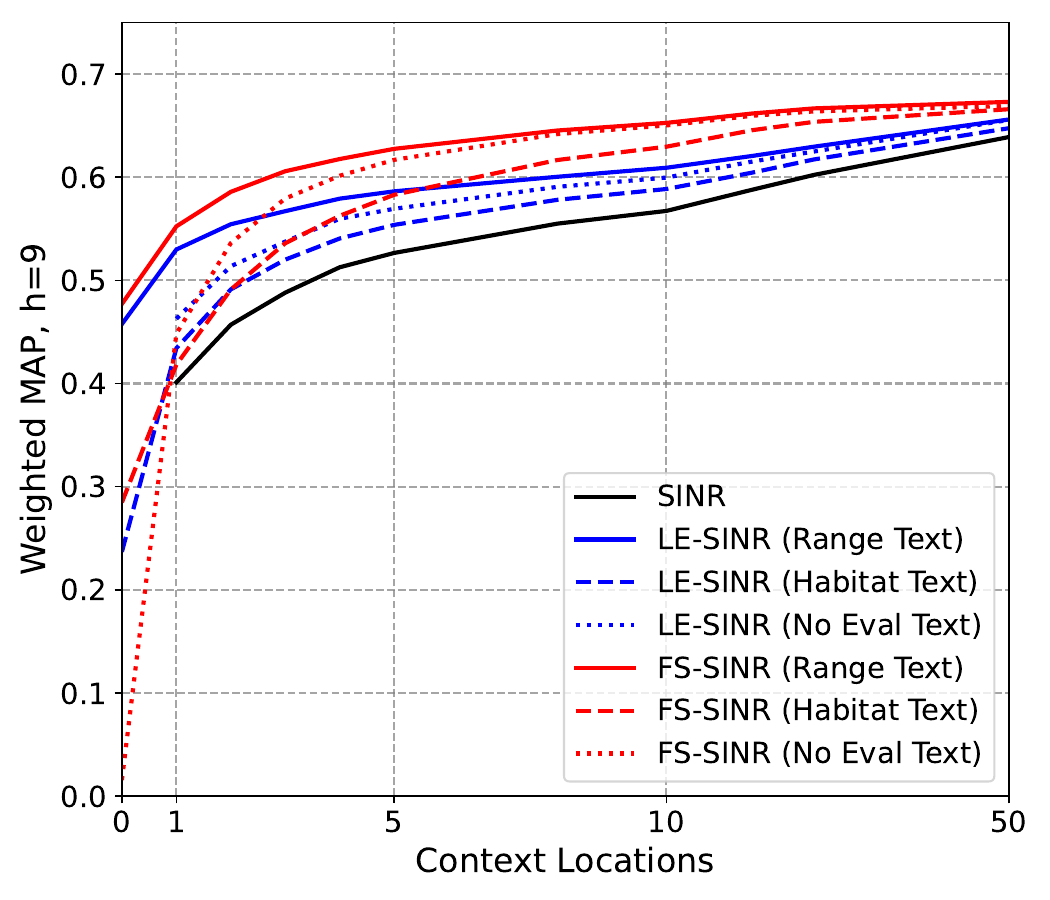}
    \end{minipage}%
    \hspace{0.5em}
    \begin{minipage}{0.31\textwidth}
        \centering
        \includegraphics[width=\linewidth]{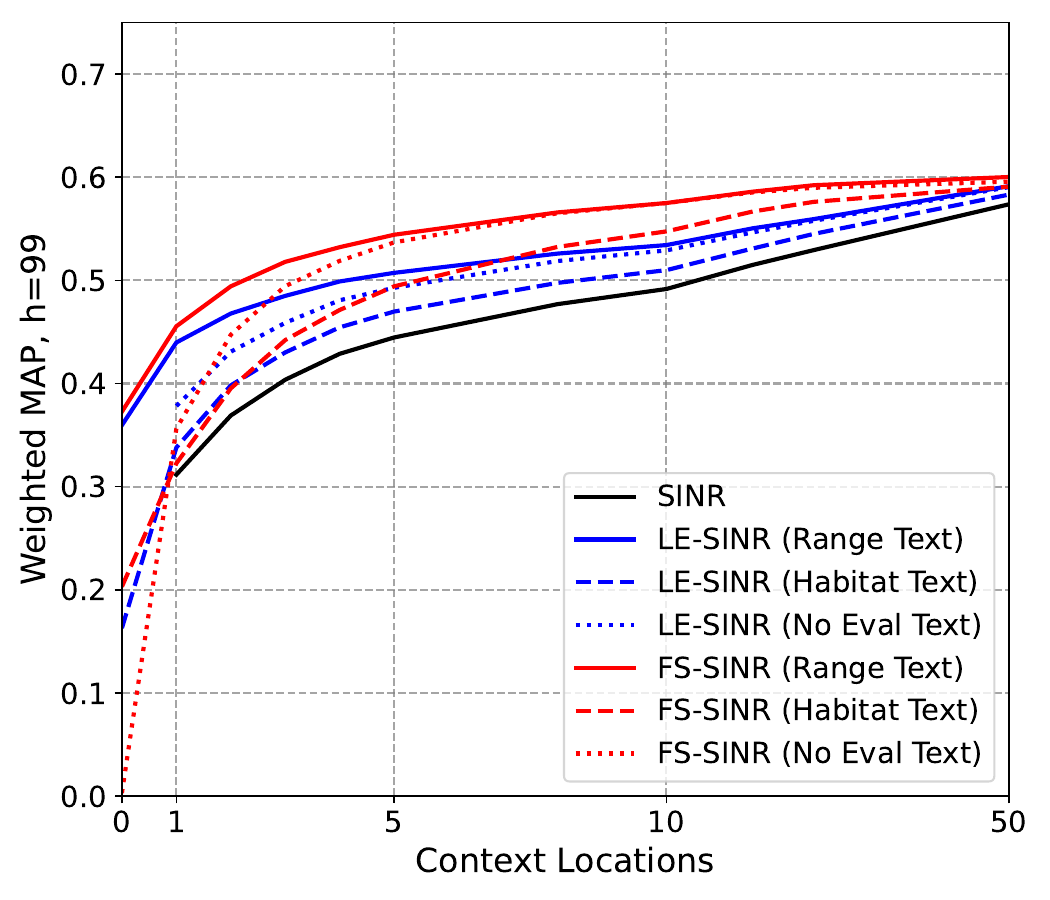}
    \end{minipage}%
    \vspace{-10pt}
    \caption{\textbf{Zero-shot and few-shot performance using our distance weighted MAP metric on the IUCN evaluation dataset.}
    We find that increasing the distance weight hyperparameter, $h$, reduces performance across the board without significantly changing the order of different models \ie FS-SINR continues to outperform LE-SINR and SINR.
    We do see that approaches using habitat text decrease in performance more as $h$ increases, relative to approaches not using text or using range text.}
    \label{weighted_map_IUCN}
\end{figure}

\begin{figure}[h]
    \centering
    \renewcommand{\arraystretch}{1.2} %

    \begin{tabular}{cc}
        {\bf Gravenhorst's Mabuya} & {\bf African Jacana} \\

        \begin{overpic}[trim={0 1cm 0 0},clip,width=0.48\textwidth]{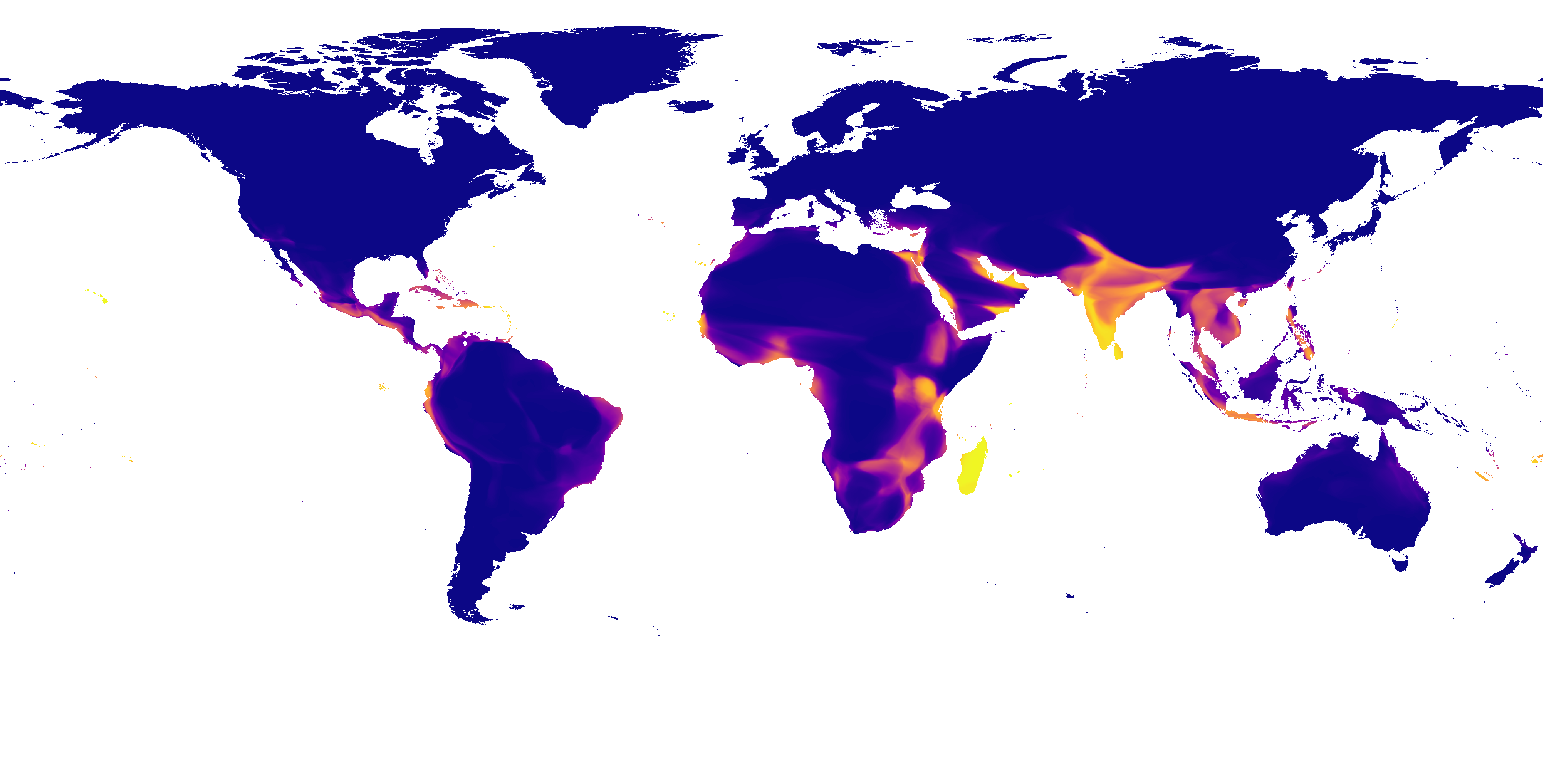}
            \put(0,0){%
              {%
                \setlength\fboxsep{0pt}%
                \setlength\fboxrule{1pt}%
                \fbox{%
                  \includegraphics[width=0.13\textwidth]{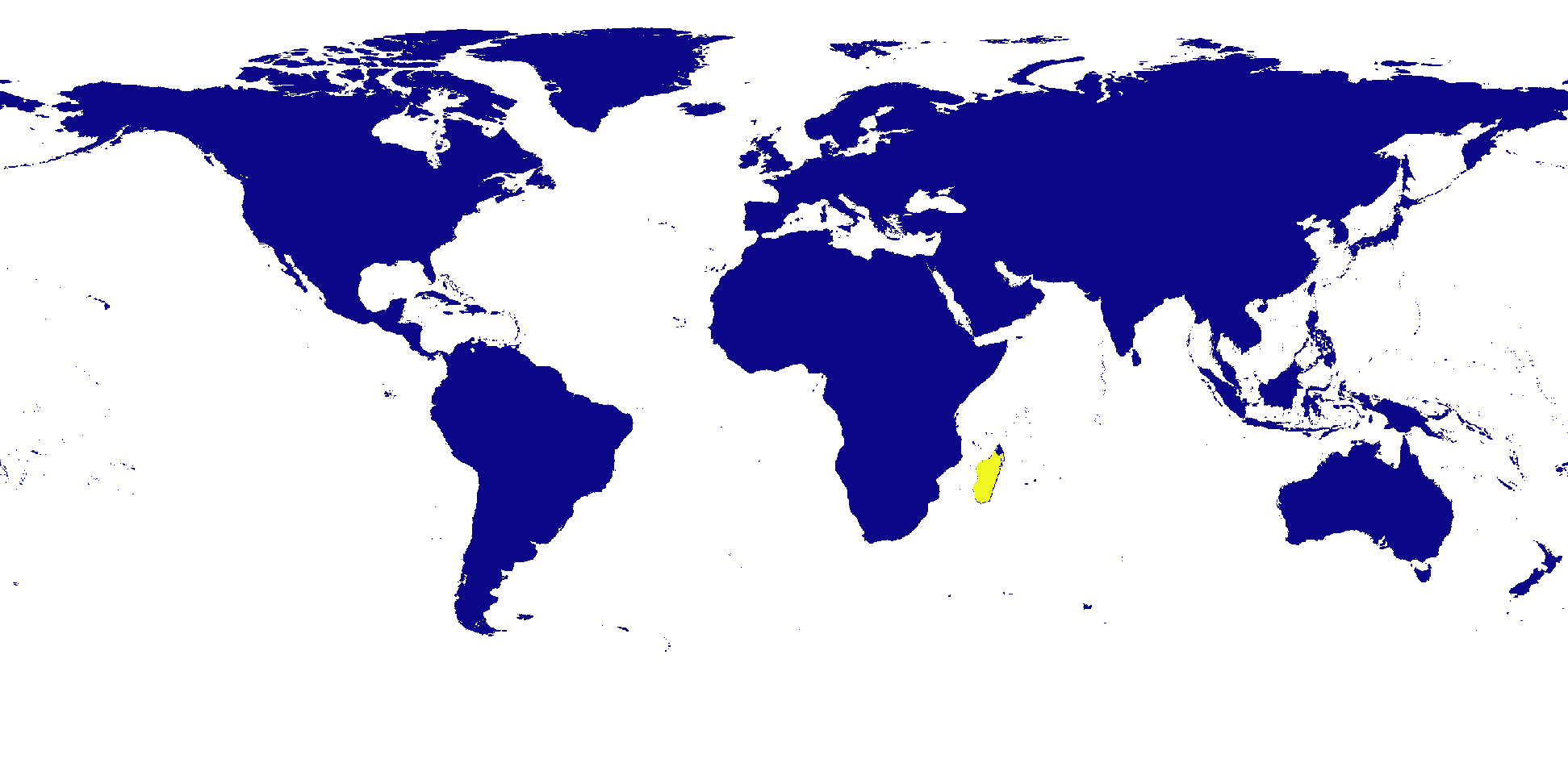}%
                }%
              }%
            }
        \end{overpic}
        &
        \begin{overpic}[trim={0 1cm 0 0},clip,width=0.48\textwidth]{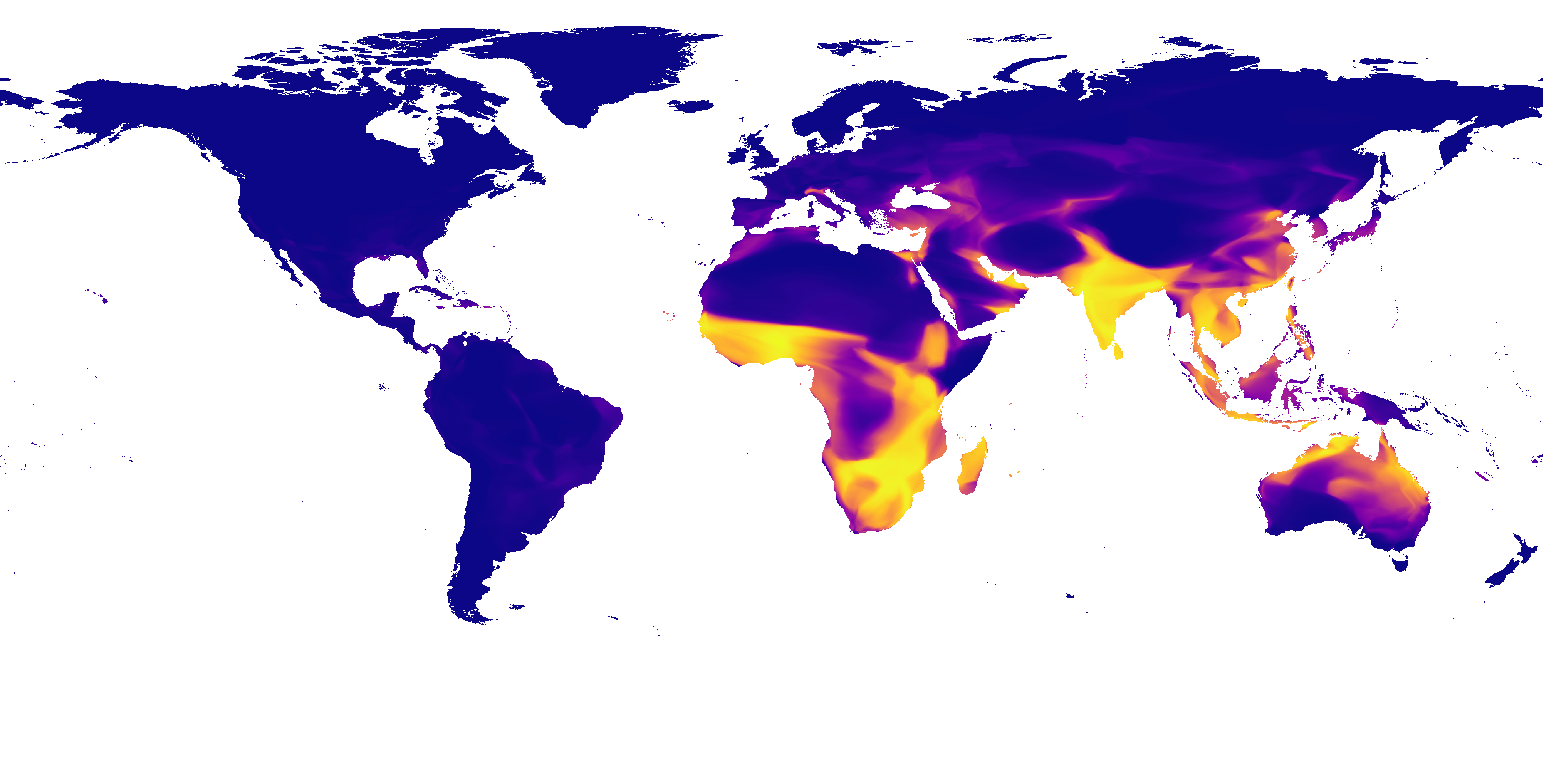}
            \put(0,0){%
              {%
                \setlength\fboxsep{0pt}%
                \setlength\fboxrule{1pt}%
                \fbox{%
                  \includegraphics[width=0.13\textwidth]{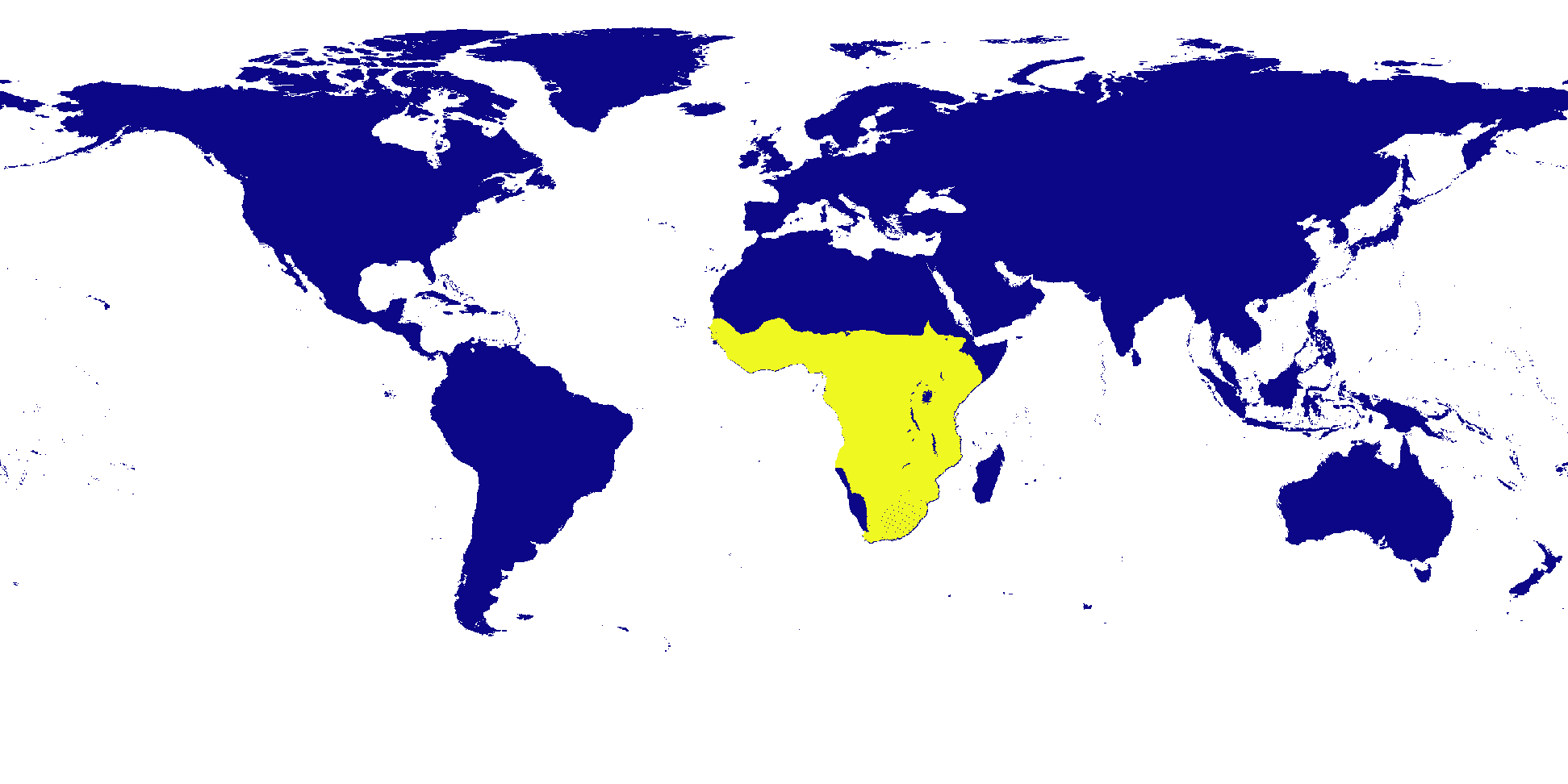}%
                }%
              }%
            }
        \end{overpic}
        \\[6pt]

        \begin{minipage}[t]{0.48\textwidth}
            \scriptsize{\textit{Habitat Text:} ``T. gravenhorstii prefers rocky areas, grassland, shrubland, and forest, but is also found in coffee plantations and ylang-ylang plantations. The species is adapted to a variety of environmental conditions, including different types of ecosystems."}
        \end{minipage}
        &
        \begin{minipage}[t]{0.48\textwidth}
            \scriptsize{\textit{Habitat Text:} ``The African jacana prefers shallow lakes and its preferred habitat is floating vegetation. It has a specific environmental condition, which is not mentioned in the article, but it can be inferred that it requires a certain level of water depth and vegetation cover."}
        \end{minipage}
    \end{tabular}

    \caption{\change{ \textbf{Examples of two species with poor distanced weighted MAP performance.} 
    Here we visualize \modelname's zero-shot predictions using habitat text for two species  where there is a large difference between the evaluation scores using the standard MAP metric compared to the distance weighted one (here using  $h=9$).
    For the \texttt{Gravenhorst's Mabuya} (left), which is endemic to Madagascar, we obtain an MAP of 0.419 but a lower distance weighted MAP of  0.175.  
    For the \texttt{African Jacana} (right), found in most of sub-Saharan Africa, we obtain an MAP of 0.457 and a distance weighted MAP of  0.226. 
    The distance weighted metric more heavily penalizes mistakes for these species that are very far from their true range.}
    }    
    \label{fig:weighted_eval}
\end{figure}

\change{\subsection{Alternative Evaluation Metric}

Here we provide additional results for the main models from \cref{fig:low_shot} using a `distance weighted' MAP evaluation metric. 
This is inspired by the evaluation conducted in LD-SDM~\citep{sastry2023ld}. 
This metric is based on mean average precision (MAP), however we now weight predictions by distance from the true range, \ie predicting the presence of a species far from where it is said to be found is penalized more than predicting the presence of a species in a location that is very close to existing observations, but is still actually outside the range. 
We intend that this metric more closely aligns with a human's judgment on how `correct' a range is, compared to standard MAP.  
By considering both metrics we can be more confident that the improvement in range mapping performance that \modelname provides is not just a consequence of how we are quantifying it.  
We determine the weight for location $\bm{x}$ as:  
\begin{equation}
w_{\bm{x}} = 1 +\frac{d_{range}(\bm{x})}{d_{antipodal}}{h}, 
\end{equation}
where $d_{range}(\bm{x})$ is the distance along the earth's surface from point $\bm{x}$ to the nearest point of the expert-derived range using for evaluation, and $d_{antipodal}$ is the distance along the earth's surface between two points on opposite sides of the earth. 
While this distance does vary very slightly in different locations as the earth is not a perfect sphere, for this experiment we have set $d_{antipodal}$ to 20,037.5 km. 
$h$ is the `distance weight hyperparameter' and determines how much this metric penalizes incorrect predictions far from the range relative to close to the range. 
The metric is implemented equivalent to scikit-learn's \texttt{average\_precision\_score} \texttt{sample\_weight} parameter~\citep{pedregosa2011scikit}. 
We evaluate performance using the standard `unweighted MAP', \ie where $h=0$ and so we are calculating MAP as usual, and `distance weighted MAP' with $h=9$ and $h=99$. We selected these settings so that errors on the opposite side of earth from the true range are penalized 10 and 100 times more than errors close to the true range.

Results on the IUCN evaluation dataset can be found in \cref{weighted_map_IUCN}.
We do not present results for the S\&T dataset as we require knowledge of the range of each species globally to fairly apply the distance-weighted MAP, while the S\&T dataset only provides range estimates for portions of the globe for each species.
As the weight is increased, we observe a general reduction in overall performance.
While there is no change in the relative ordering of different models, and FS-SINR outperforms LE-SINR and SINR across all settings of $h$, we do observe that FS-SINR and LE-SINR models that use habitat text during evaluation seem to decrease in performance more with larger $h$ compared to other approaches. 
They are likely most effected by the larger weight, as habitat text can cause the model to predict presence in locations around the world with similar habitat features such as mountains, forest, or desert, despite these locations being far from the true range. 
This appears to be true of both FS-SINR and LE-SINR. 
For LE-SINR, using habitat text outperforms not using text for the \emph{unweighted} MAP, but using habitat text performs worse than not using text for the weighted MAP.
In \cref{fig:weighted_eval}, we display zero-shot results for two species where there is a large difference in performance based on the two metrics. 
In both cases \modelname using only text incorrectly predicts the species to be present far from the expert-derived range. 

}

\subsection{Expanded Results}

In \cref{tab:main_results_table_iucn,tab:main_results_table_snt} we present and expanded set of results  from \cref{fig:low_shot}, for the IUCN and S\&T datasets.
We also report additional results for the non-few-shot setting using 500 and 1,000 context locations. 
Performance for most approaches including \modelname seems to plateau around 50 context locations, with some approaches gaining a minor boost in performance in the non-few-shot setting (SINR and Prototype-SINR), while others perform much worse (Active SINR).
We also report results for a model trained and evaluated using context locations and features from a visual encoder with a DINOv2 backbone.
This performs worse and has significantly more variable performance across runs compared to the  pre-trained EVA-02 ViT, which was fine-tuned using species images during its original pre-training, used in the main paper.

\begin{table}[h]
    \centering
    \caption{\change{\textbf{IUCN zero-shot and few-shot results.}
    Here we present expanded IUCN evaluation results for the models shown in \cref{fig:low_shot} in tabular form.
    We also show results using a DINOv2 based image encoder}.
    SINR and LE-SINR without text cannot produce a range map without at least one context location.
    Results are presented as MAP, where higher is better.}
    \label{tab:main_results_table_iucn}
    \vspace{2mm}
    \resizebox{0.95\columnwidth}{!}{%
\begin{tabular}{l|ccccc|cc|c|c|c}
           & \multicolumn{5}{c|}{FS-SINR} & \multicolumn{2}{|c|}{LE-SINR} & SINR & Prototype SINR & Active SINR   \\ 
\# Context & Text   & Image & Text + Image & No Text \textbackslash Image & DINOv2 & Text    & No Text & No Text & No Text & No Text \\ \hline
0          & 0.52    & 0.19 & 0.46 & 0.05  & 0.13 & 0.48      & -       & -     & -    & -    \\
1          & 0.57    & 0.45 & 0.55 & 0.48  & 0.40 & 0.55      & 0.47    & 0.42  & 0.48 & 0.48 \\
2          & 0.60    & 0.54 & 0.59 & 0.56  & 0.49 & 0.57      & 0.52    & 0.47  & 0.53 & 0.55 \\
3          & 0.62    & 0.58 & 0.61 & 0.60  & 0.53 & 0.58      & 0.54    & 0.50  & 0.56 & 0.58 \\
4          & 0.63    & 0.60 & 0.62 & 0.62  & 0.55 & 0.59      & 0.56    & 0.52  & 0.57 & 0.59 \\
5          & 0.64    & 0.62 & 0.63 & 0.63  & 0.56 & 0.60      & 0.57    & 0.54  & 0.58 & 0.60 \\
8          & 0.65    & 0.64 & 0.64 & 0.65  & 0.59 & 0.61      & 0.59    & 0.56  & 0.60 & 0.62 \\
10         & 0.66    & 0.65 & 0.65 & 0.66  & 0.60 & 0.62      & 0.60    & 0.57  & 0.61 & 0.62 \\
15         & 0.67    & 0.66 & 0.66 & 0.67  & 0.61 & 0.63      & 0.62    & 0.59  & 0.61 & 0.62 \\
20         & 0.67    & 0.66 & 0.66 & 0.67  & 0.61 & 0.64      & 0.63    & 0.61  & 0.62 & 0.62 \\
50         & 0.68    & 0.67 & 0.67 & 0.67  & 0.61 & 0.66      & 0.66    & 0.64  & 0.62 & 0.60 \\
500        & 0.68    & 0.66 & 0.66 & 0.67  & 0.57 & 0.67      & 0.67    & 0.65  & 0.63 & 0.37 \\
1000       & 0.68    & 0.66 & 0.66 & 0.67  & 0.57 & 0.67      & 0.67    & 0.65  & 0.63 & 0.36 \\
\end{tabular}
    }
\end{table}

\begin{table}[h]
    \centering
    \caption{\change{\textbf{S\&T zero-shot and few-shot results.}
    Here we present expanded S\&T evaluation results for the models shown in \cref{fig:low_shot} in tabular form.
    We also show results using a DINOv2 based image encoder.
    SINR and LE-SINR without text cannot produce a range map without at least one context location.
    Results are presented as MAP, where higher is better.}}
    \vspace{2mm}
    \label{tab:main_results_table_snt}
    \resizebox{0.95\columnwidth}{!}{%
\begin{tabular}{l|ccccc|cc|c|c|c}
           & \multicolumn{5}{c|}{FS-SINR} & \multicolumn{2}{|c|}{LE-SINR} & SINR & Prototype SINR & Active SINR   \\ 
\# Context & Text   & Image & Text + Image & No Text \textbackslash Image & DINOv2 & Text    & No Text & No Text & No Text & No Text \\ \hline
0          & 0.64   & 0.38 & 0.64  & 0.18  & 0.28 & 0.60      & -       & -    & -    & -   \\
1          & 0.66   & 0.49 & 0.66  & 0.50  & 0.44 & 0.64      & 0.52    & 0.49 & 0.54 & 0.53 \\
2          & 0.67   & 0.57 & 0.67  & 0.58  & 0.53 & 0.66      & 0.57    & 0.55 & 0.59 & 0.59 \\
3          & 0.68   & 0.61 & 0.68  & 0.61  & 0.57 & 0.67      & 0.60    & 0.58 & 0.61 & 0.62 \\
4          & 0.69   & 0.64 & 0.69  & 0.64  & 0.61 & 0.67      & 0.61    & 0.59 & 0.62 & 0.65 \\
5          & 0.70   & 0.66 & 0.70  & 0.65  & 0.63 & 0.68      & 0.62    & 0.60 & 0.63 & 0.65 \\
8          & 0.71   & 0.69 & 0.71  & 0.68  & 0.65 & 0.69      & 0.65    & 0.63 & 0.64 & 0.67 \\
10         & 0.72   & 0.70 & 0.71  & 0.69  & 0.67 & 0.69      & 0.66    & 0.64 & 0.65 & 0.67 \\
15         & 0.72   & 0.71 & 0.72  & 0.70  & 0.68 & 0.70      & 0.68    & 0.67 & 0.65 & 0.67 \\
20         & 0.72   & 0.71 & 0.72  & 0.71  & 0.68 & 0.71      & 0.69    & 0.68 & 0.65 & 0.66 \\
50         & 0.73   & 0.72 & 0.73  & 0.71  & 0.69 & 0.73      & 0.72    & 0.72 & 0.66 & 0.64 \\
500        & 0.73   & 0.71 & 0.72  & 0.71  & 0.62 & 0.73      & 0.72    & 0.73  & 0.67 & 0.24 \\
1000       & 0.72   & 0.71 & 0.72  & 0.71  & 0.62 & 0.73      & 0.72    & 0.73  & 0.67 & 0.24 \\
\end{tabular}
    }
\end{table}

\begin{figure}[h]
    \centering
        \includegraphics[width=0.42\textwidth]{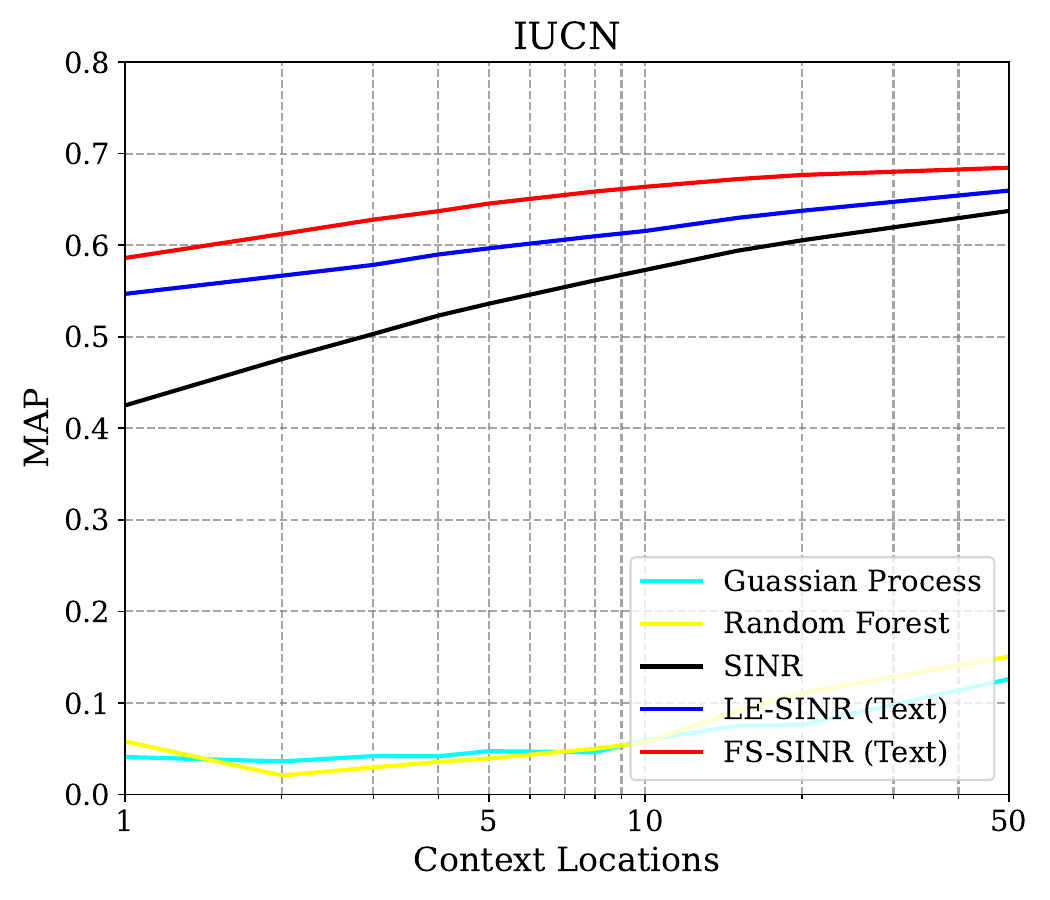}
        \hspace{15pt}
        \includegraphics[width=0.42\textwidth]{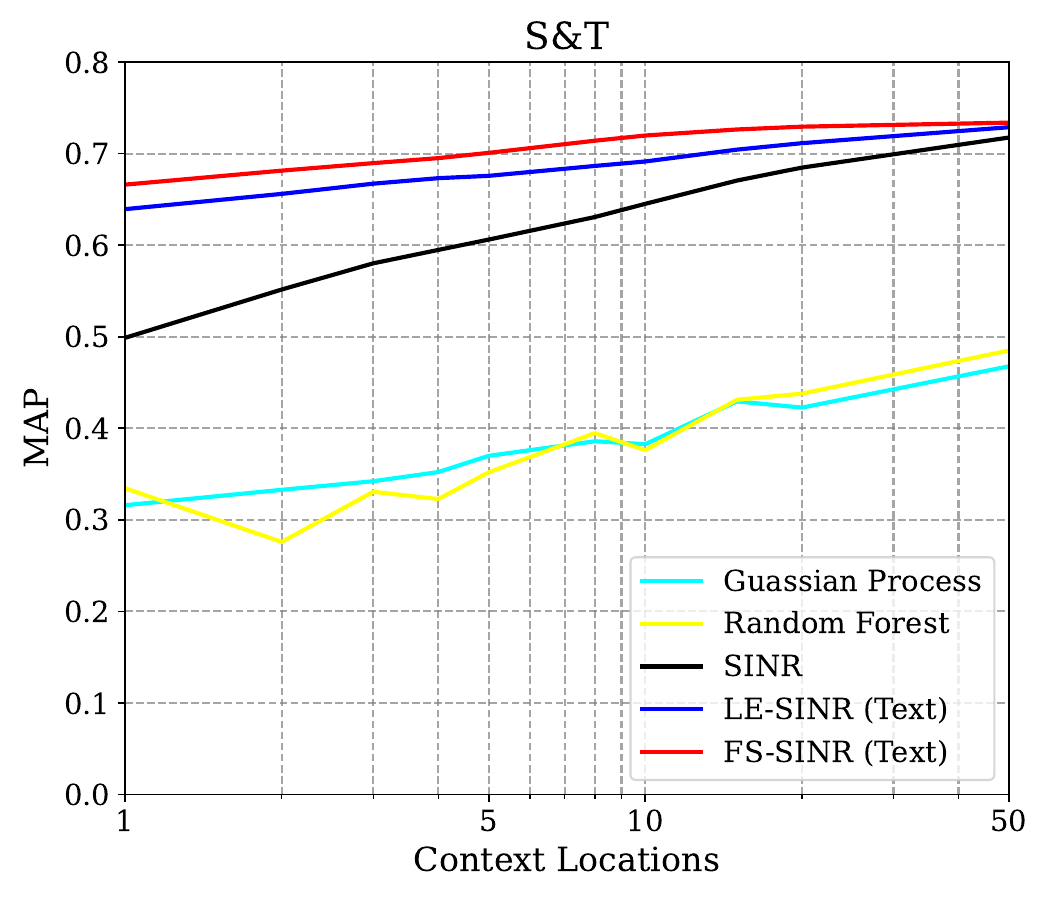} 
        \vspace{-15pt}
        \caption{\change{\textbf{Comparison to traditional machine-learning approaches.} Here we evaluate a Gaussian Process approach and a Random Forest on the IUCN (left) and S\&T (right) datasets. We find that both approaches perform far worse than \modelname, LE-SINR or SINR for species range estimation using presence-only data. `Range' text is used here for \modelname and LE-SINR, while other approaches take only context locations as input.}}
    \label{fig:gp_rf_plot}
    \vspace{-15pt}
\end{figure}

\subsection{Comparison to Traditional Machine Learning Methods}

In \cref{fig:gp_rf_plot} we compare \modelname with two traditional machine learning approaches for classifying whether a species is present or absent at a given location. Both approaches were implemented using scikit-learn \citep{pedregosa2011scikit}.
We compare with a Gaussian Process approach adapted from \citet{gaussianprocessapproach}.
While this method is designed for presence-absence data, we adapt it for the presence-only setting by providing a number of pseudo-negatives equal to the provided number of context locations, which performed best out of several strategies we investigated. 
Using a large number of pseudo-negatives as we do for \modelname both performed poorly and took excessively long to run as Gaussian Process computations scale cubically with the amount of data.
For this approach we use a logit link function and a squared-exponential kernel.
We also compare to a random forest classifier~\citep{breiman2001random}.
We investigated providing the same number of pseudo-negatives as for \modelname with appropriate class weights, however we found better performance by providing the same number of pseudo-negatives as context locations.
In both cases, performance for this task is significantly worse than \modelname, LE-SINR and SINR.

\section{Ecologically Relevant Analysis of Results}
\label{sec:eco_results}

To give a more ecologically relevant analysis of our results we present them here by continent, species range size, and taxonomic class.
In these cases we display results for the IUCN evaluation dataset only, as it provides expert-derived presence-absence information globally and includes a range of taxonomic groups.
In comparison the S\&T dataset only includes \texttt{Aves}, \ie bird species, and evaluates the presence or absence of each species over a portion of the globe, preventing us from calculating global range sizes and performance by continent.

\change{\subsection{Results by Region}}

\begin{figure}[h]
    \centering
    \begin{subfigure}{0.42\textwidth}
        \includegraphics[width=\linewidth]{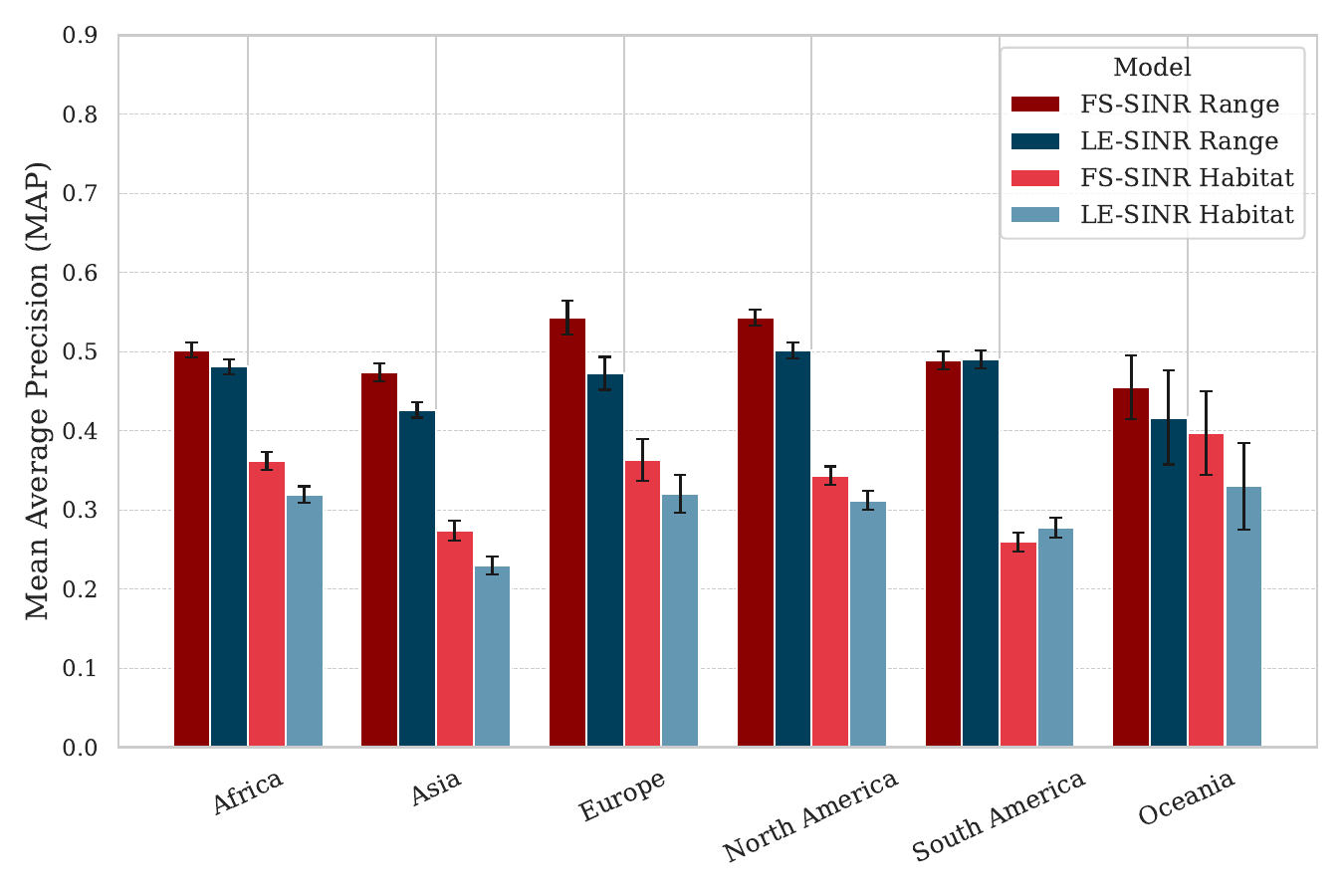}
        \caption{0 Context Locations}
    \end{subfigure}%
    \hspace{1em}
    \begin{subfigure}{0.42\textwidth}
        \includegraphics[width=\linewidth]{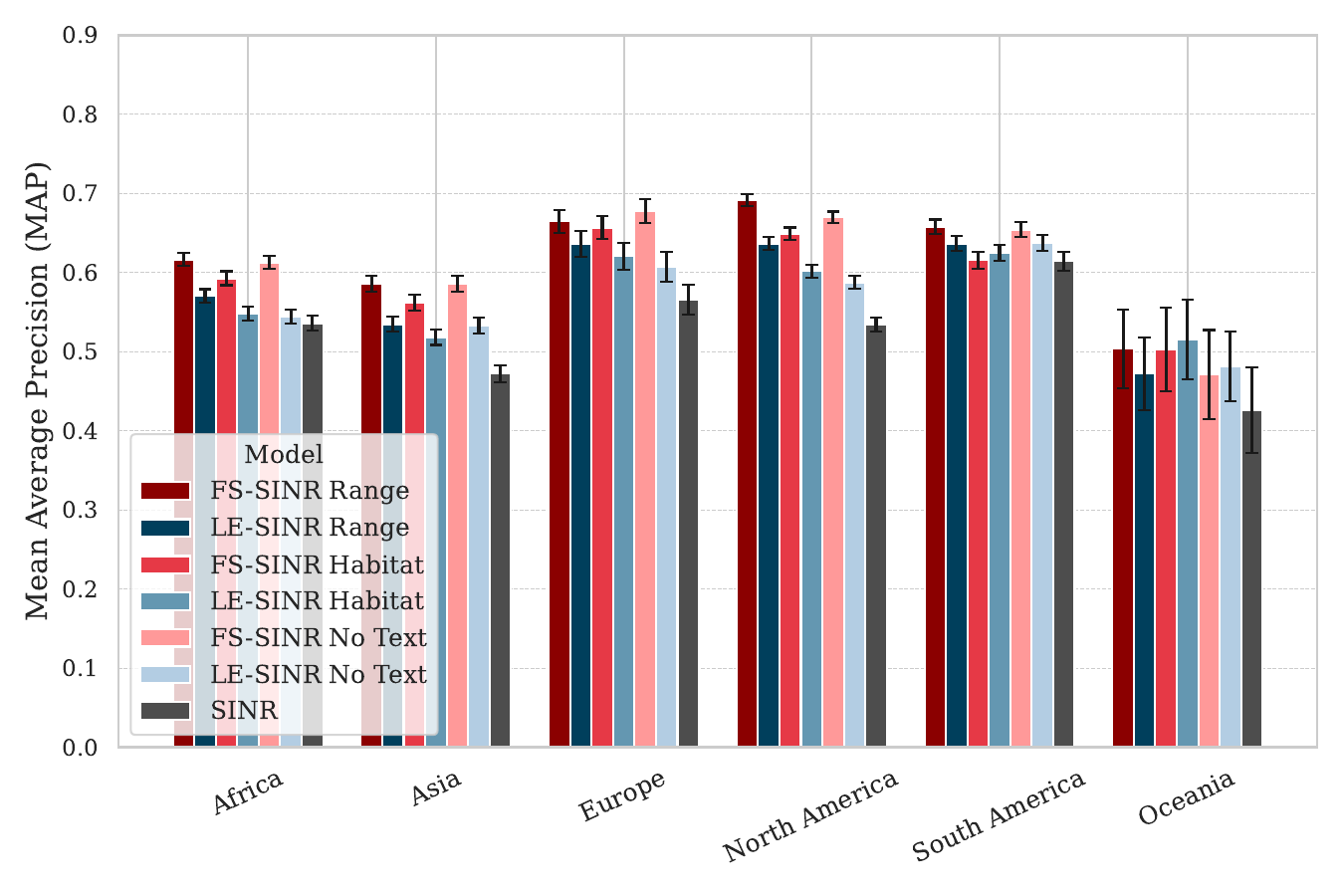}
        \caption{5 Context Locations}
    \end{subfigure}

    \begin{subfigure}{0.42\textwidth}
        \includegraphics[width=\linewidth]{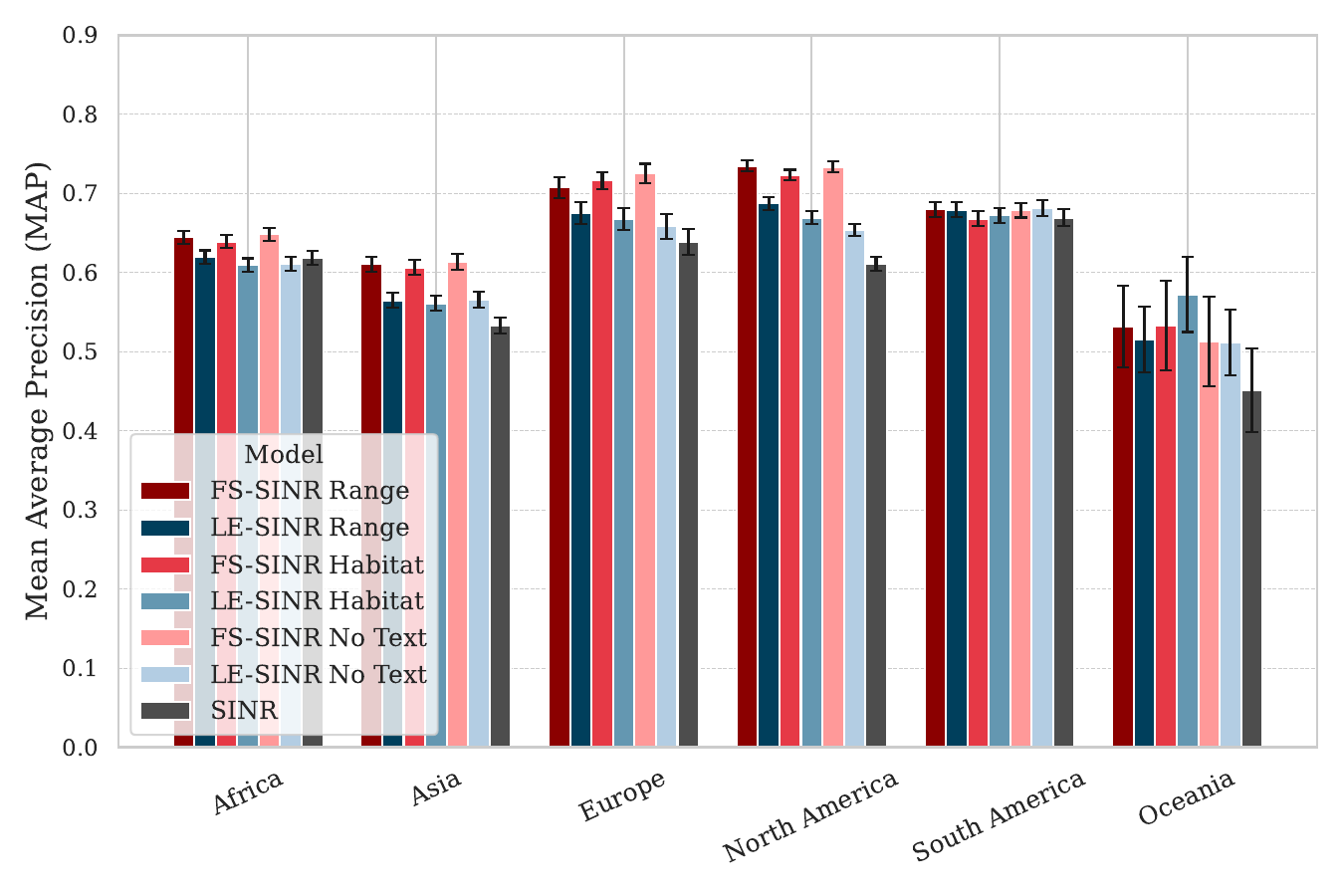}
        \caption{20 Context Locations}
    \end{subfigure}%
    \hspace{1em}
    \begin{subfigure}{0.42\textwidth}
        \includegraphics[width=\linewidth]{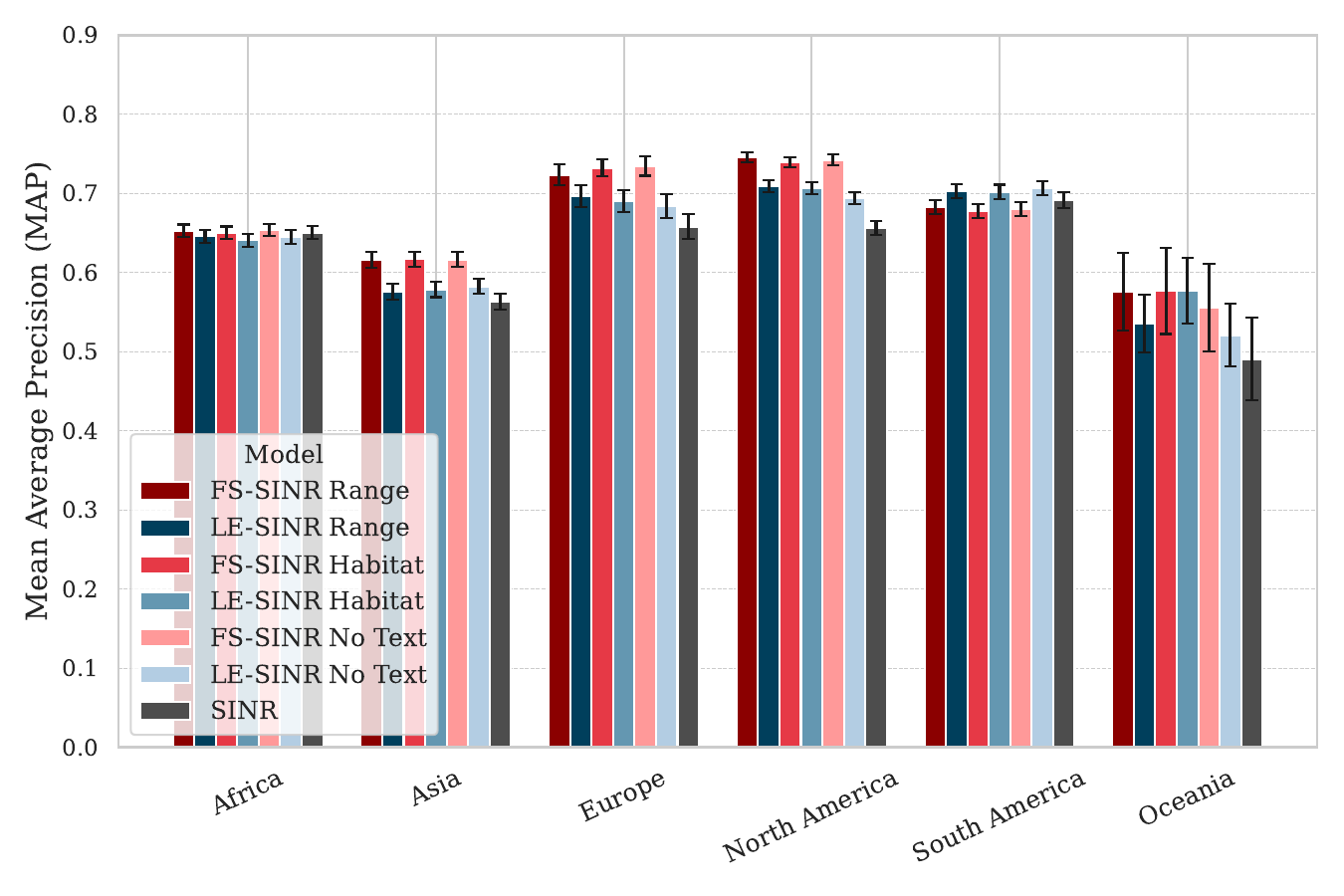}
        \caption{50 Context Locations}
    \end{subfigure}

    \vspace{-10pt}
    \caption{\textbf{IUCN Performance by continent.} Error bars show standard error of the mean.}
    \vspace{-10pt}
    \label{fig:continent_eval}
\end{figure}

In \cref{fig:continent_eval} we show performance of \modelname, LE-SINR, and SINR models by continent for few-shot and zero-shot text-only predictions.
\modelname outperforms other approaches on all continents except South America and Oceania, where at larger numbers of context locations LE-SINR becomes comparable.
Biodiversity data and particularly global-scale citizen science datasets such as the iNaturalist-derived data we use to train \modelname can contain large biases~\citep{geldmann2016determines, hughessamplingbias}, and here we can see the impact of this bias within our training data.

\begin{figure}[h]
    \centering
    \begin{minipage}{0.04\textwidth}
    \end{minipage}%
    \hspace{0.5em}
    \begin{minipage}{0.29\textwidth}
        \centering \textbf{Range Text}
    \end{minipage}%
    \hspace{0.5em}
    \begin{minipage}{0.29\textwidth}
        \centering \textbf{Habitat Text}
    \end{minipage}%
    \hspace{0.5em}
    \begin{minipage}{0.29\textwidth}
        \centering \textbf{No text}
    \end{minipage}
    
    \vspace{1em}

    \begin{minipage}{0.04\textwidth}
        \rotatebox{90}{\tiny\textbf{1 Context Location}}
    \end{minipage}%
    \hspace{0.5em}
    \begin{minipage}{0.29\textwidth}
        \centering
        \includegraphics[width=\linewidth]{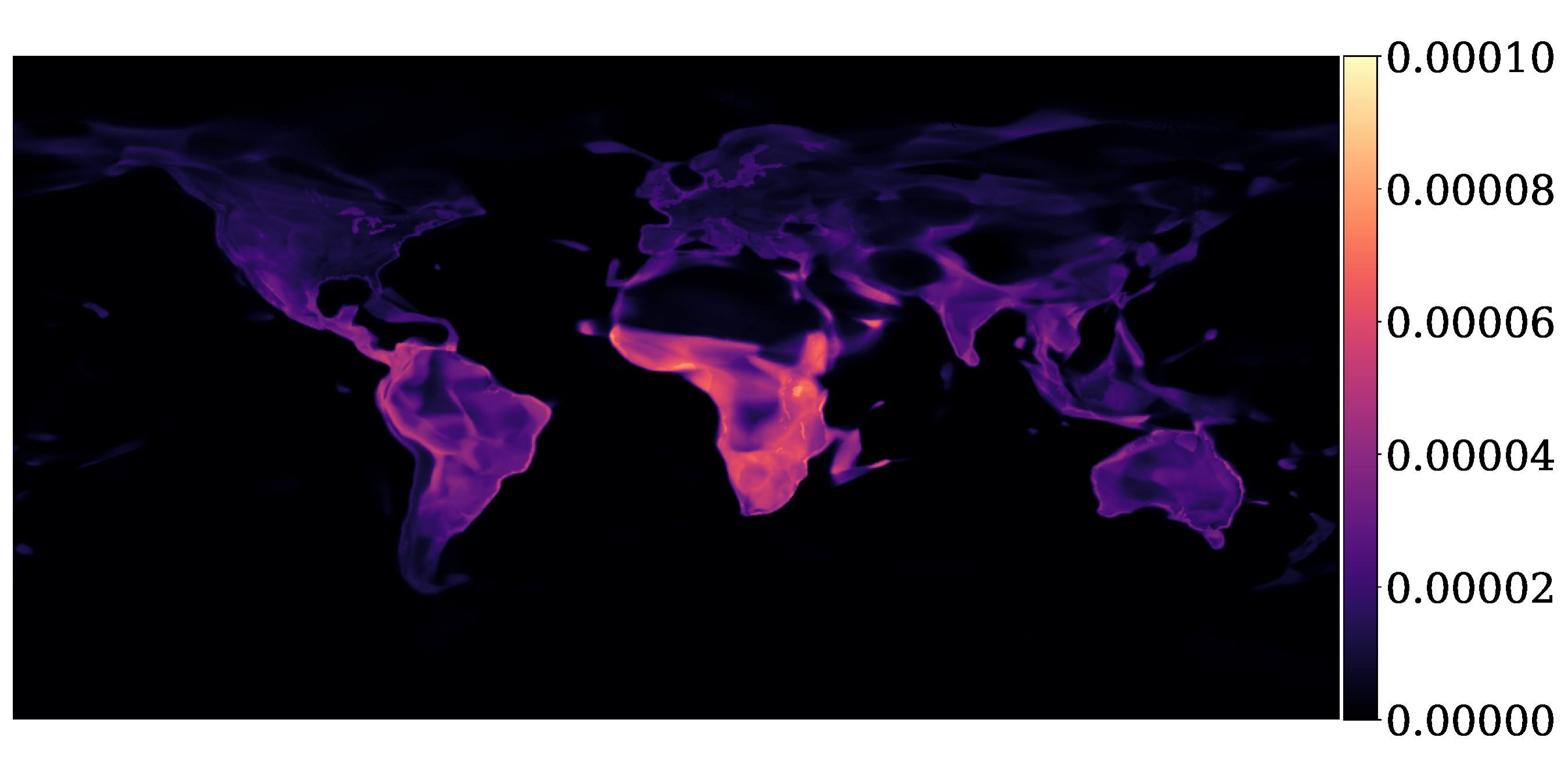}
    \end{minipage}%
    \hspace{0.5em}
    \begin{minipage}{0.29\textwidth}
        \centering
        \includegraphics[width=\linewidth]{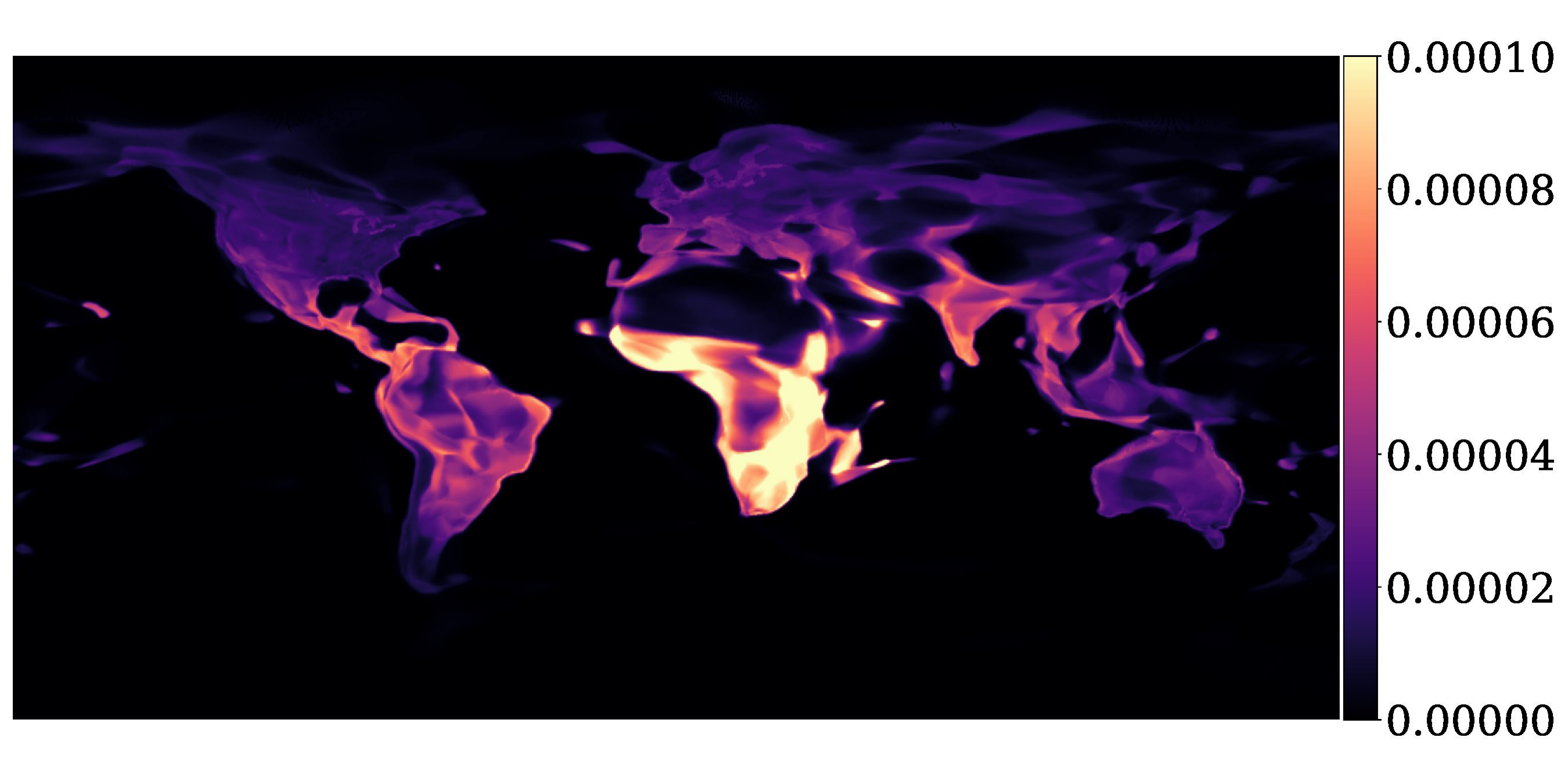}
    \end{minipage}%
    \hspace{0.5em}
    \begin{minipage}{0.29\textwidth}
        \centering
        \includegraphics[width=\linewidth]{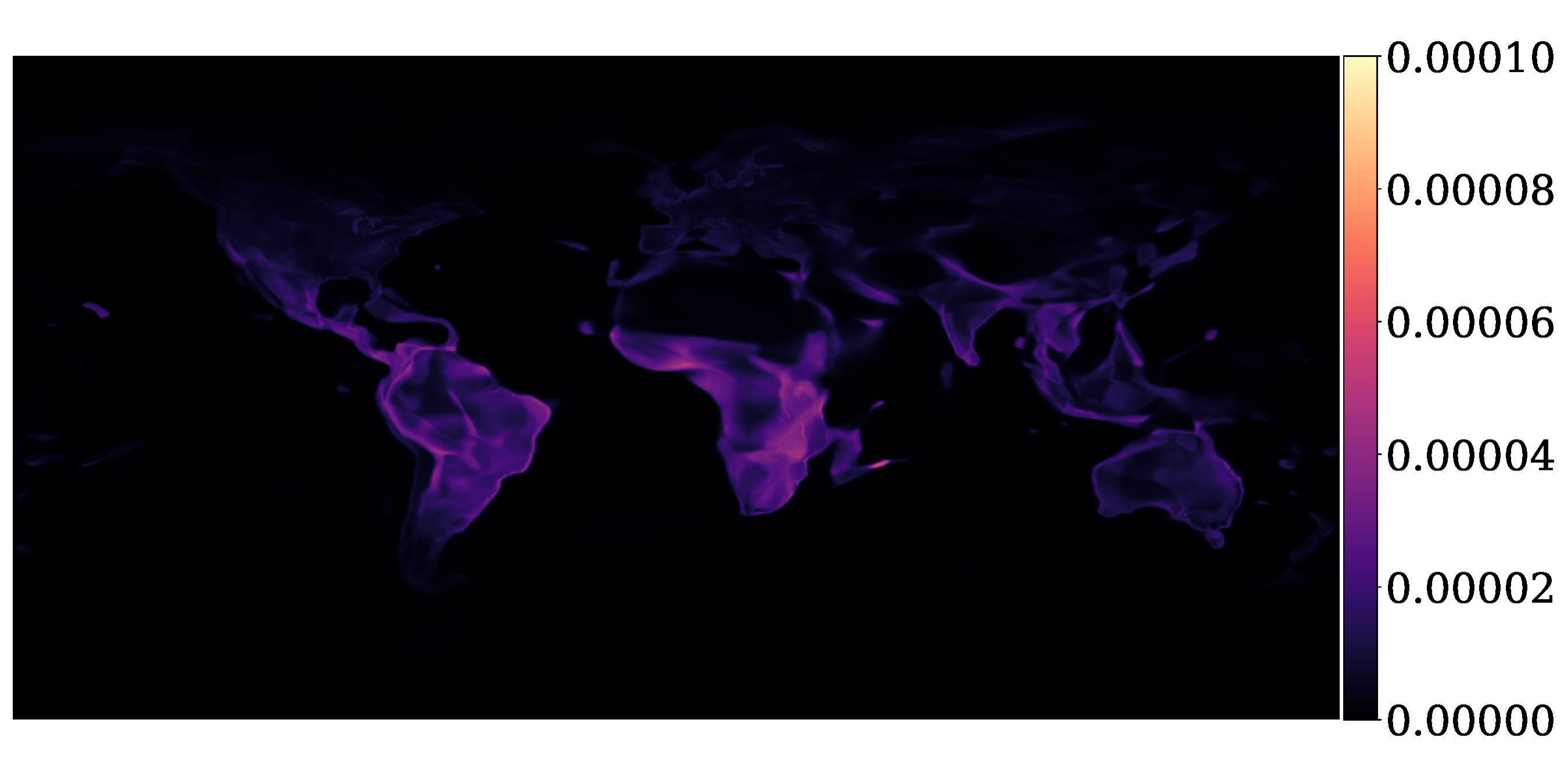}
    \end{minipage}

    \vspace{1em}

    \begin{minipage}{0.04\textwidth}
        \rotatebox{90}{\tiny\textbf{5 Context Locations}}
    \end{minipage}%
    \hspace{0.5em}
    \begin{minipage}{0.29\textwidth}
        \centering
        \includegraphics[width=\linewidth]{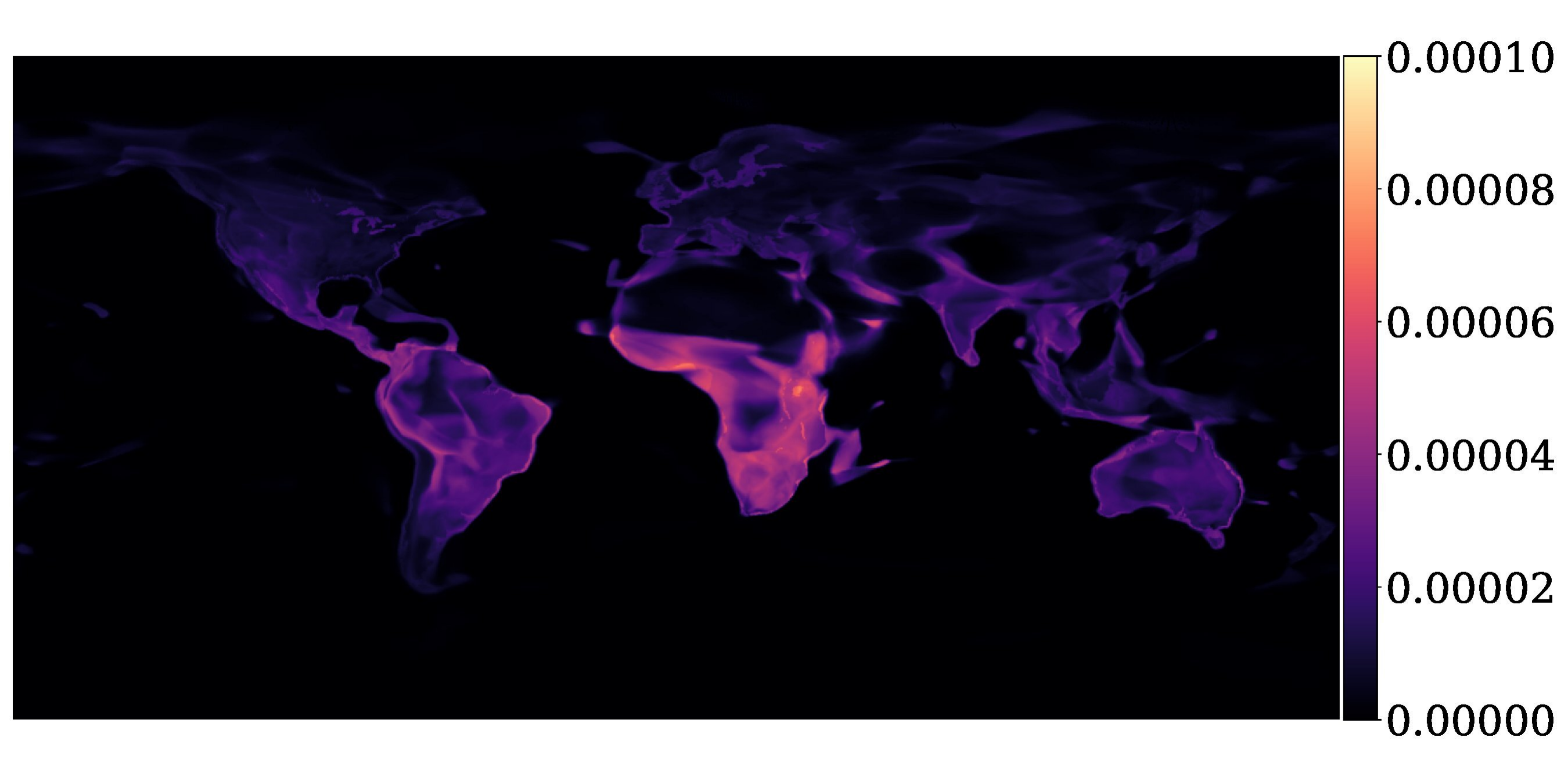}
    \end{minipage}%
    \hspace{0.5em}
    \begin{minipage}{0.29\textwidth}
        \centering
        \includegraphics[width=\linewidth]{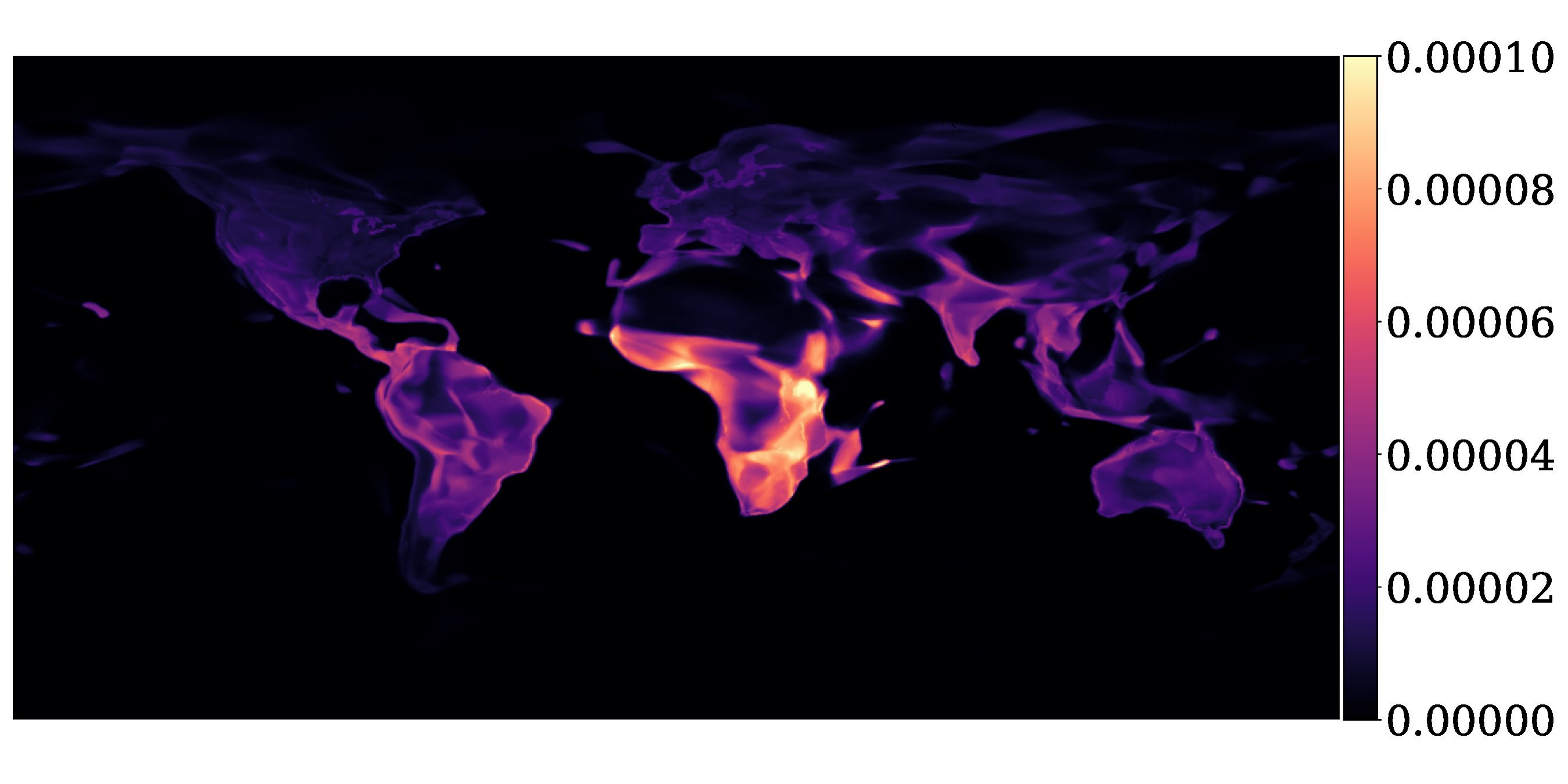}
    \end{minipage}%
    \hspace{0.5em}
    \begin{minipage}{0.29\textwidth}
        \centering
        \includegraphics[width=\linewidth]{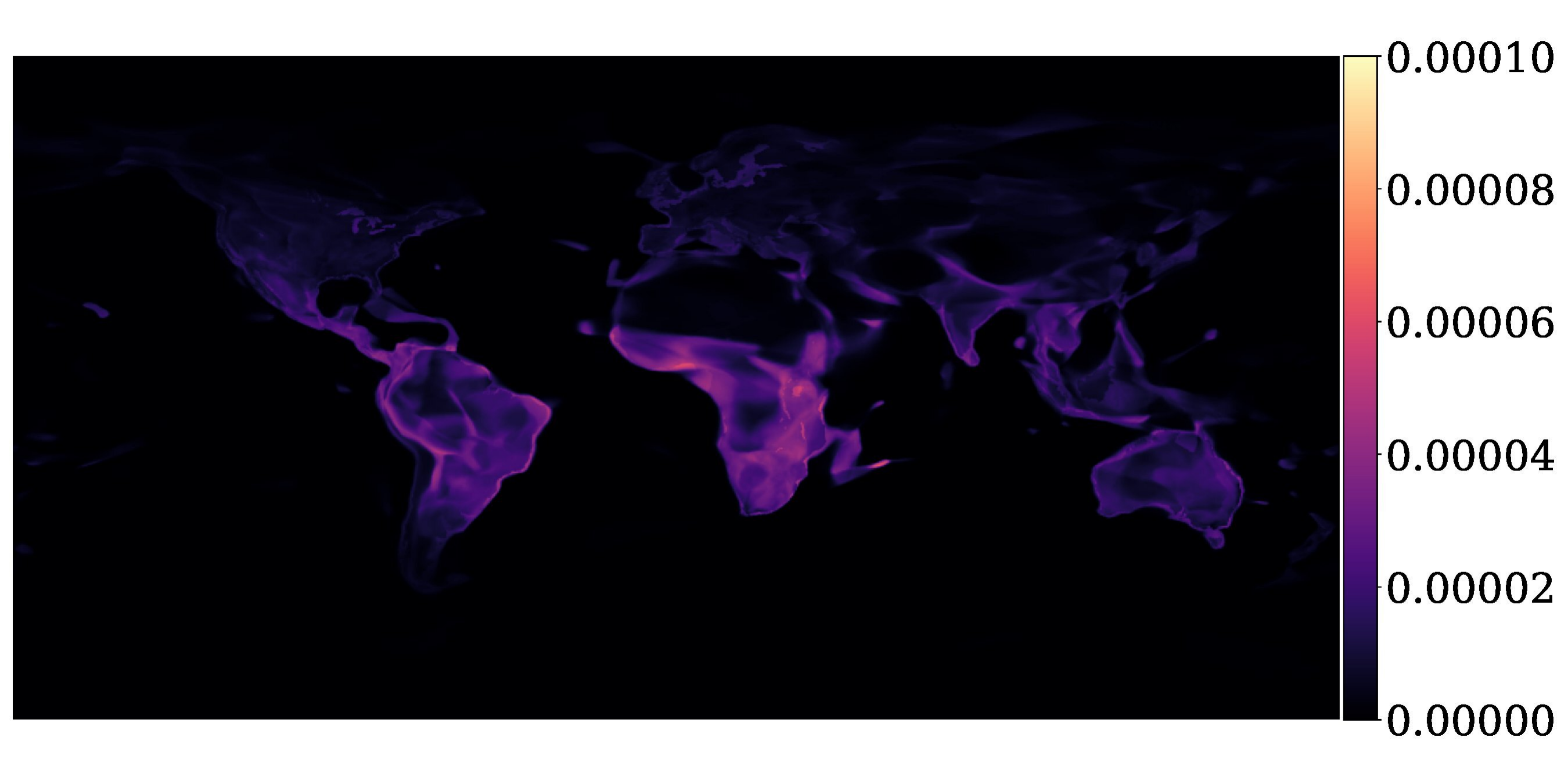}
    \end{minipage}

    \vspace{1em}

    \begin{minipage}{0.04\textwidth}
        \rotatebox{90}{\tiny\textbf{20 Context Locations}}
    \end{minipage}%
    \hspace{0.5em}
    \begin{minipage}{0.29\textwidth}
        \centering
        \includegraphics[width=\linewidth]{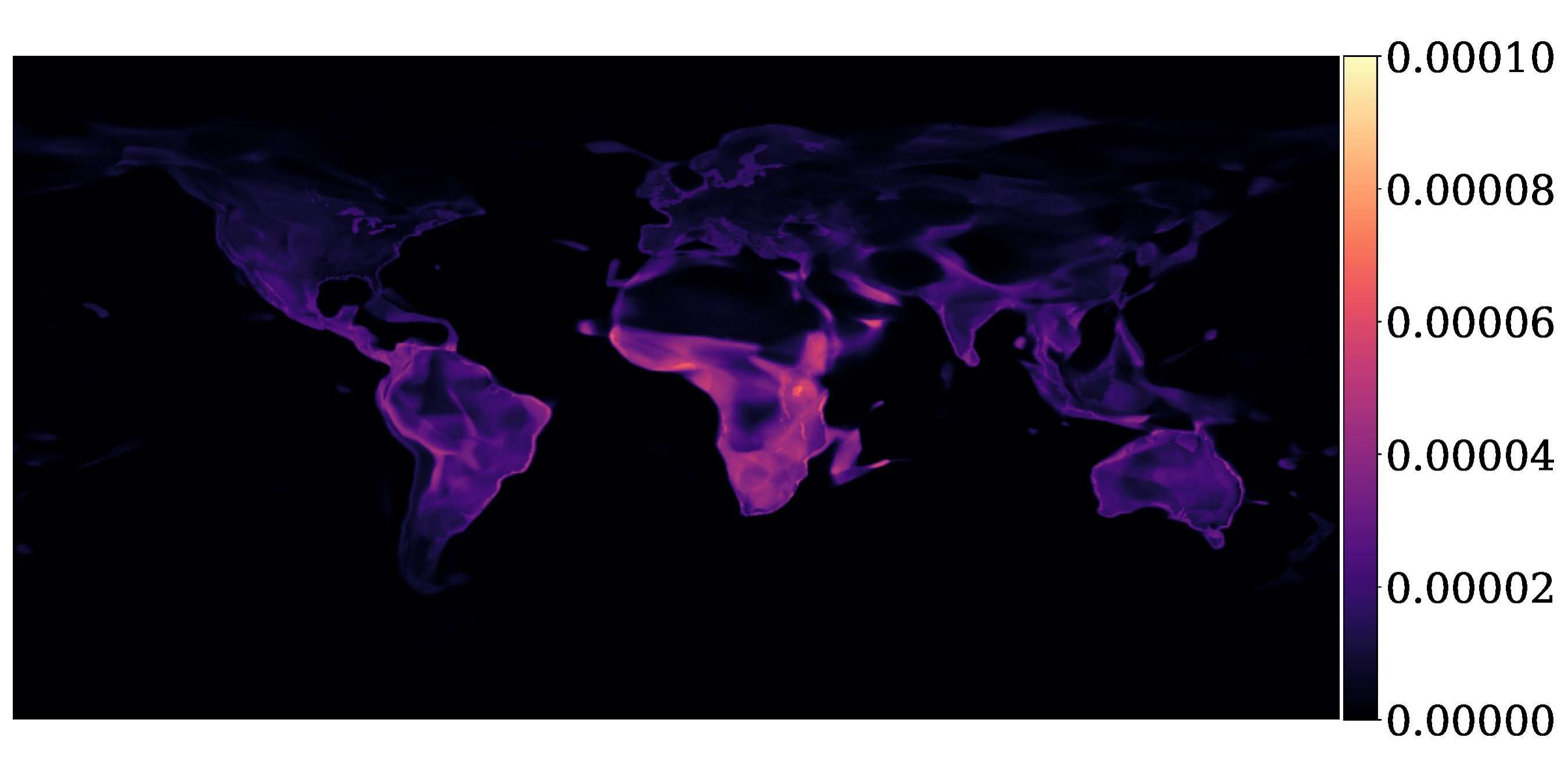}
    \end{minipage}%
    \hspace{0.5em}
    \begin{minipage}{0.29\textwidth}
        \centering
        \includegraphics[width=\linewidth]{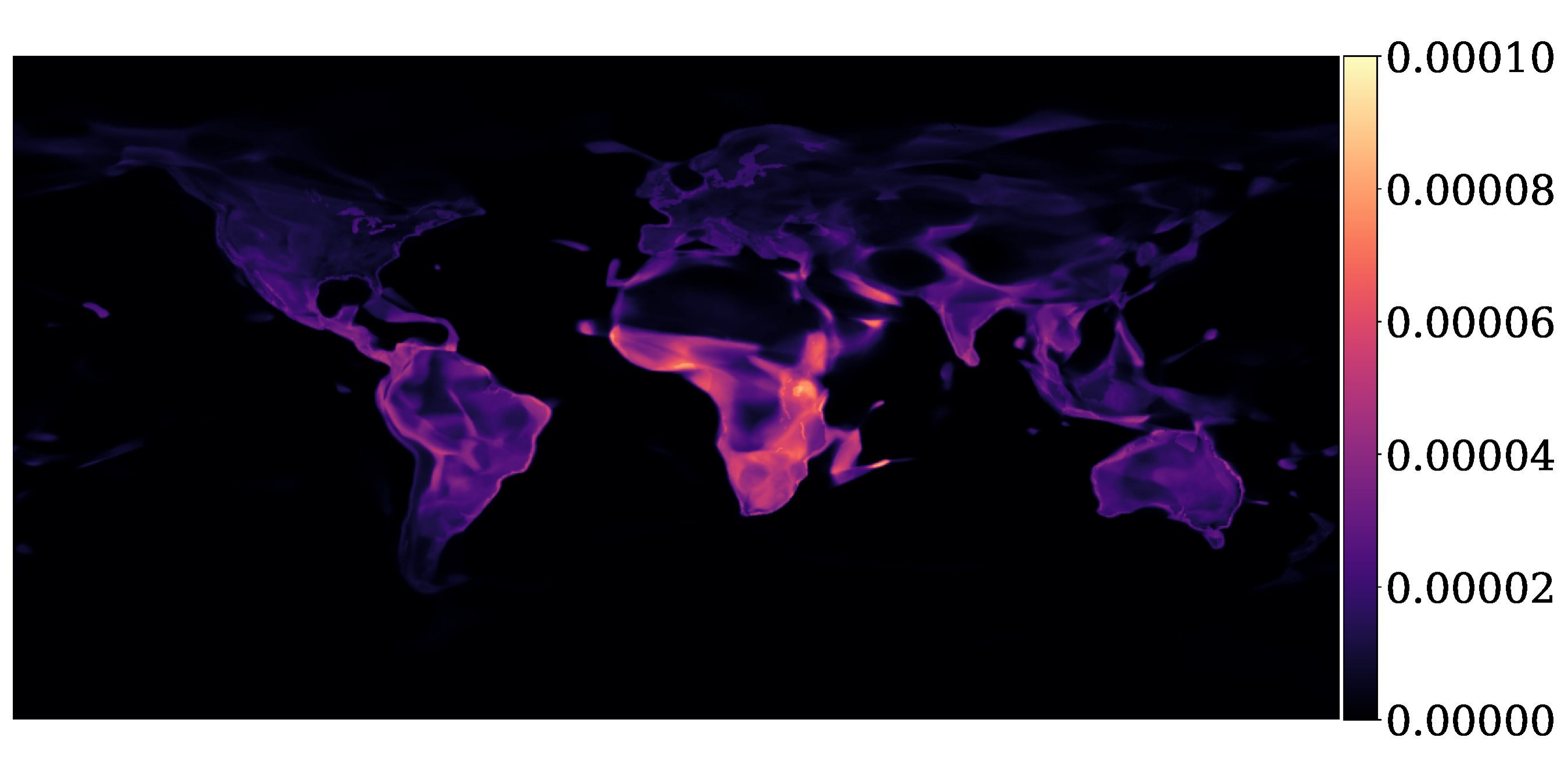}
    \end{minipage}%
    \hspace{0.5em}
    \begin{minipage}{0.29\textwidth}
        \centering
        \includegraphics[width=\linewidth]{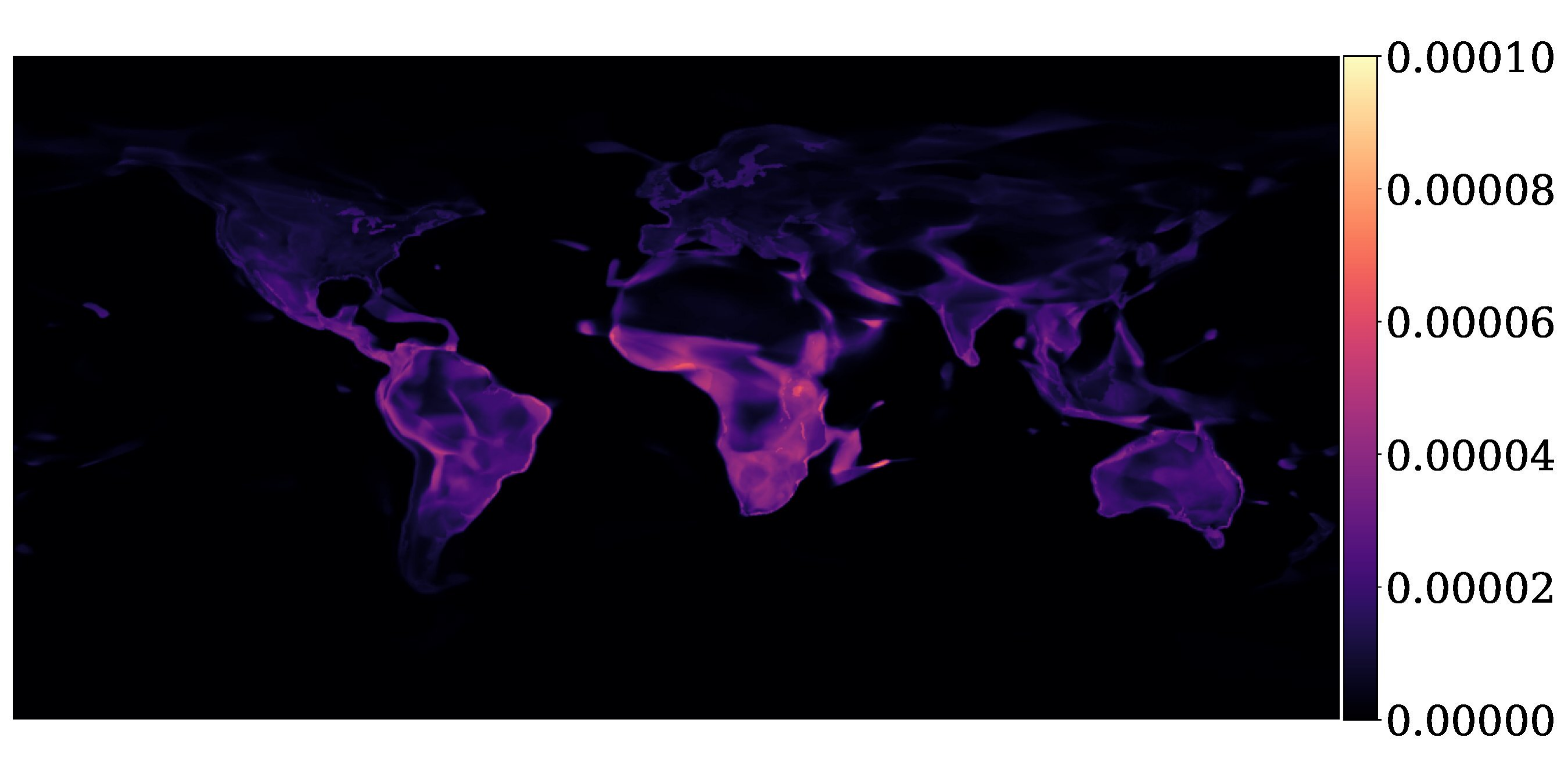}
    \end{minipage}

    \vspace{1em}

    \begin{minipage}{0.04\textwidth}
        \rotatebox{90}{\tiny\textbf{50 Context Locations}}
    \end{minipage}%
    \hspace{0.5em}
    \begin{minipage}{0.29\textwidth}
        \centering
        \includegraphics[width=\linewidth]{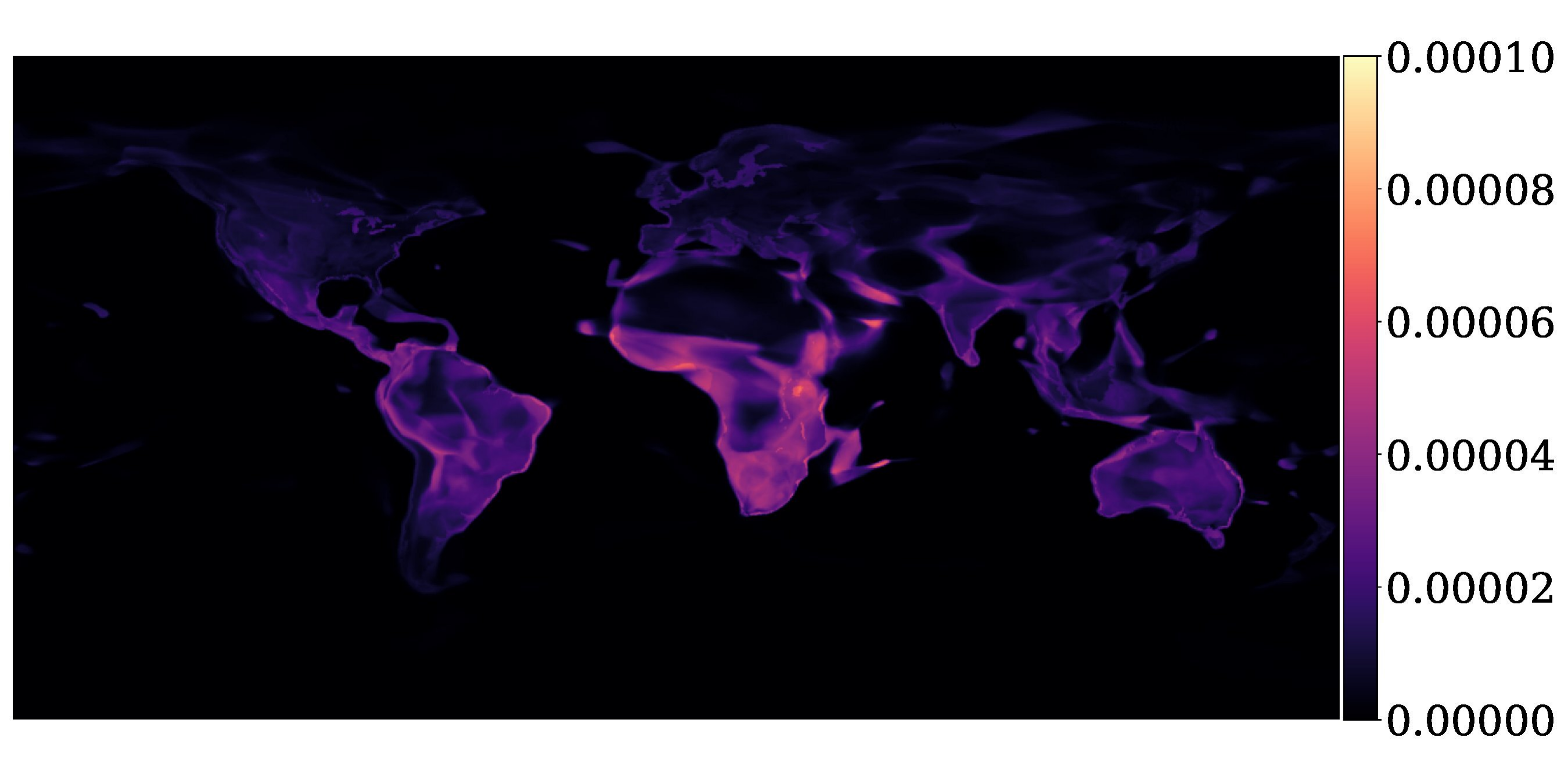}
    \end{minipage}%
    \hspace{0.5em}
    \begin{minipage}{0.29\textwidth}
        \centering
        \includegraphics[width=\linewidth]{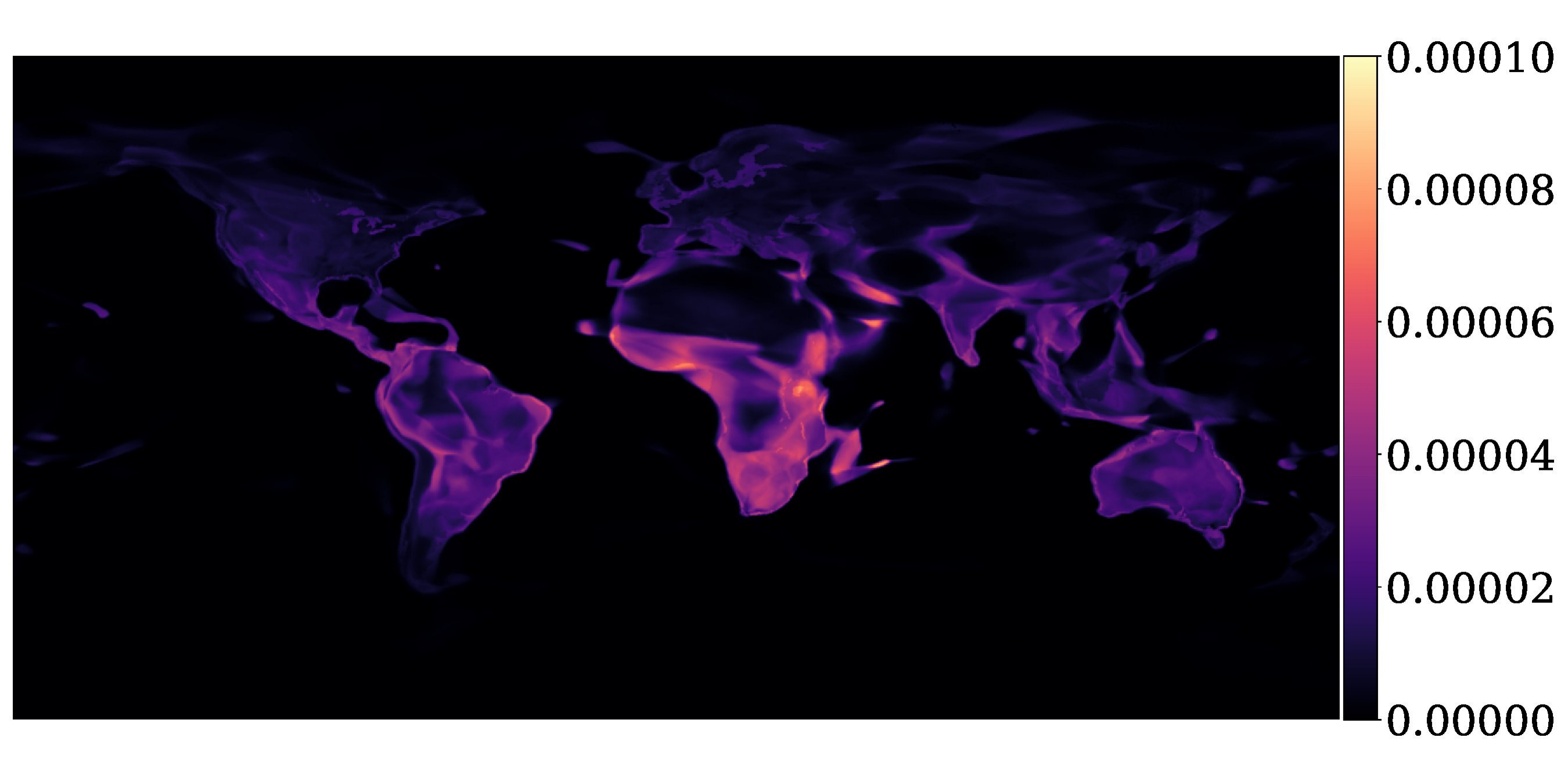}
    \end{minipage}%
    \hspace{0.5em}
    \begin{minipage}{0.29\textwidth}
        \centering
        \includegraphics[width=\linewidth]{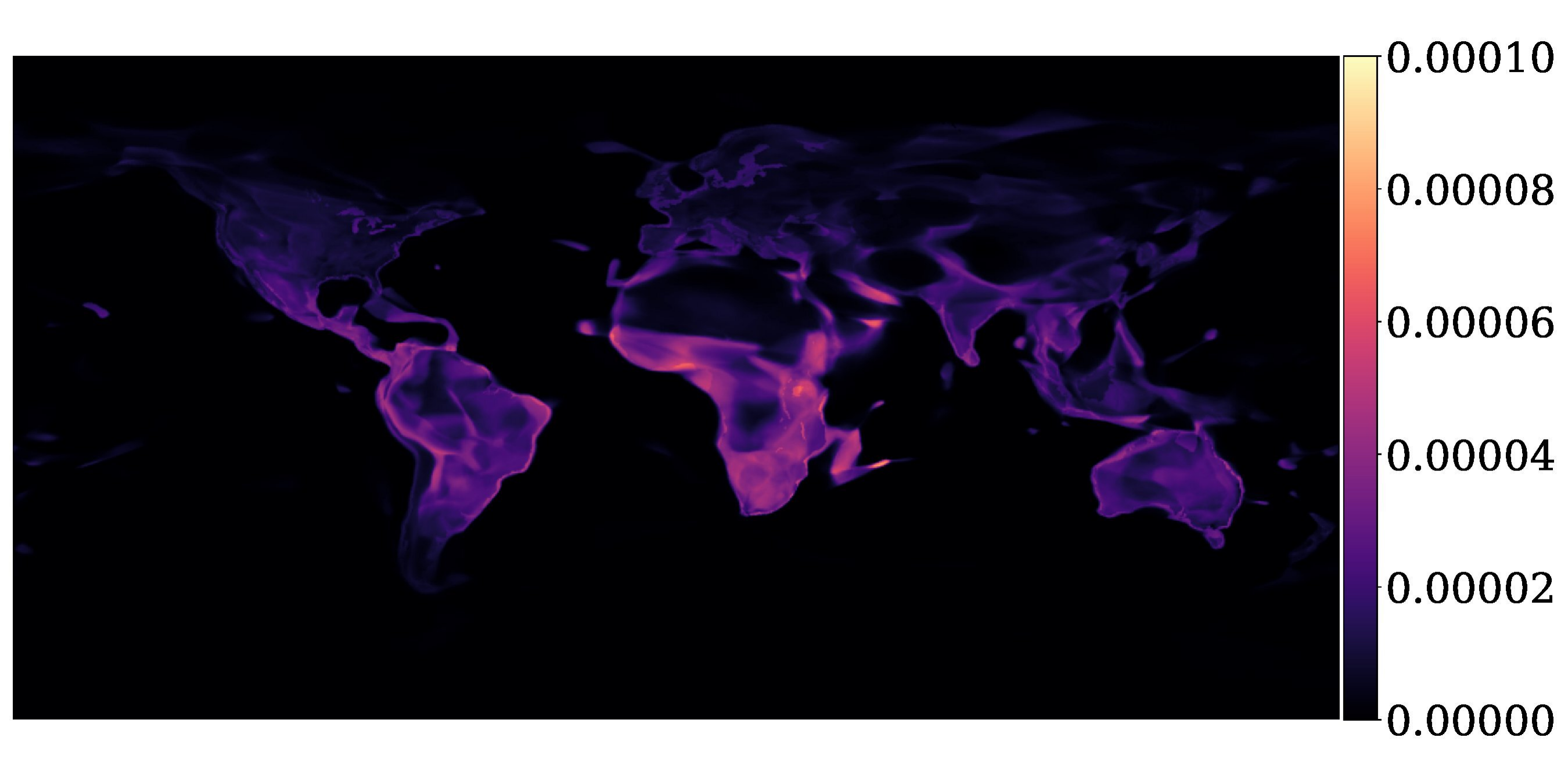}
    \end{minipage}

    \caption{\change{\textbf{Average false positive error by location for few-shot approaches.}
    Here we see average false positive error of FS-SINR on the IUCN dataset.
    Providing any text leads to an increase in the false positive error, although \cref{fig:low_shot} suggests that this text still helps with range estimation performance.
    As the number of provided context locations increases, the impact of the text is reduced and the distribution of errors appear similar.}}
    \label{fig:av_error_low_shot}
\end{figure}

We have more species observation training data (taken from \citet{cole2023spatial}, which also visualizes the data distribution) from Europe and North America than from other areas, and the observations in these regions are relatively well distributed spatially.
In other continents, there are large areas with very few or no training observations for models to learn from, along with small areas that are highly observed.
On top of this, our text descriptions are taken from English language Wikipedia and may be more descriptive for species found in areas where English is widely spoken, and our pre-trained large language model may have more knowledge of North American and European geography and ecology due to biases in the text data used for training. 
Combined, these factors lead to higher performance in North America and Europe compared to other regions. 
In \cref{fig:av_error_low_shot} we show the average false positive error for few-shot range estimation for \modelname on the IUCN evaluation dataset. 
This indicates greater error in regions where we have less training data.

\subsection{Results by Species Range Size} 
\label{range_size_section}
\begin{figure}[h]
    \centering
    \begin{subfigure}{0.42\textwidth}
        \includegraphics[width=\linewidth]{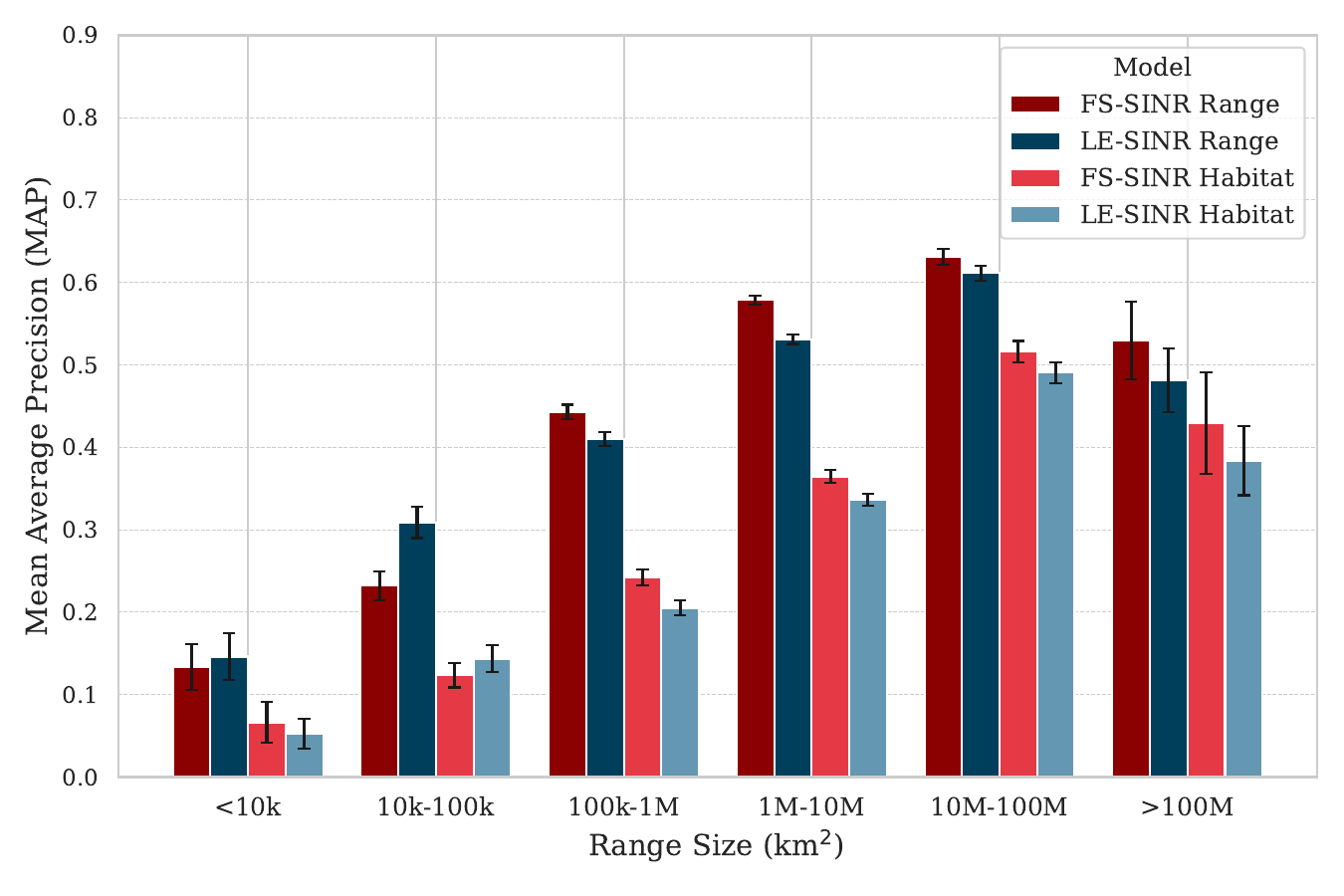}
        \caption{0 Context Locations}
    \end{subfigure}%
    \hspace{1em}
    \begin{subfigure}{0.42\textwidth}
        \includegraphics[width=\linewidth]{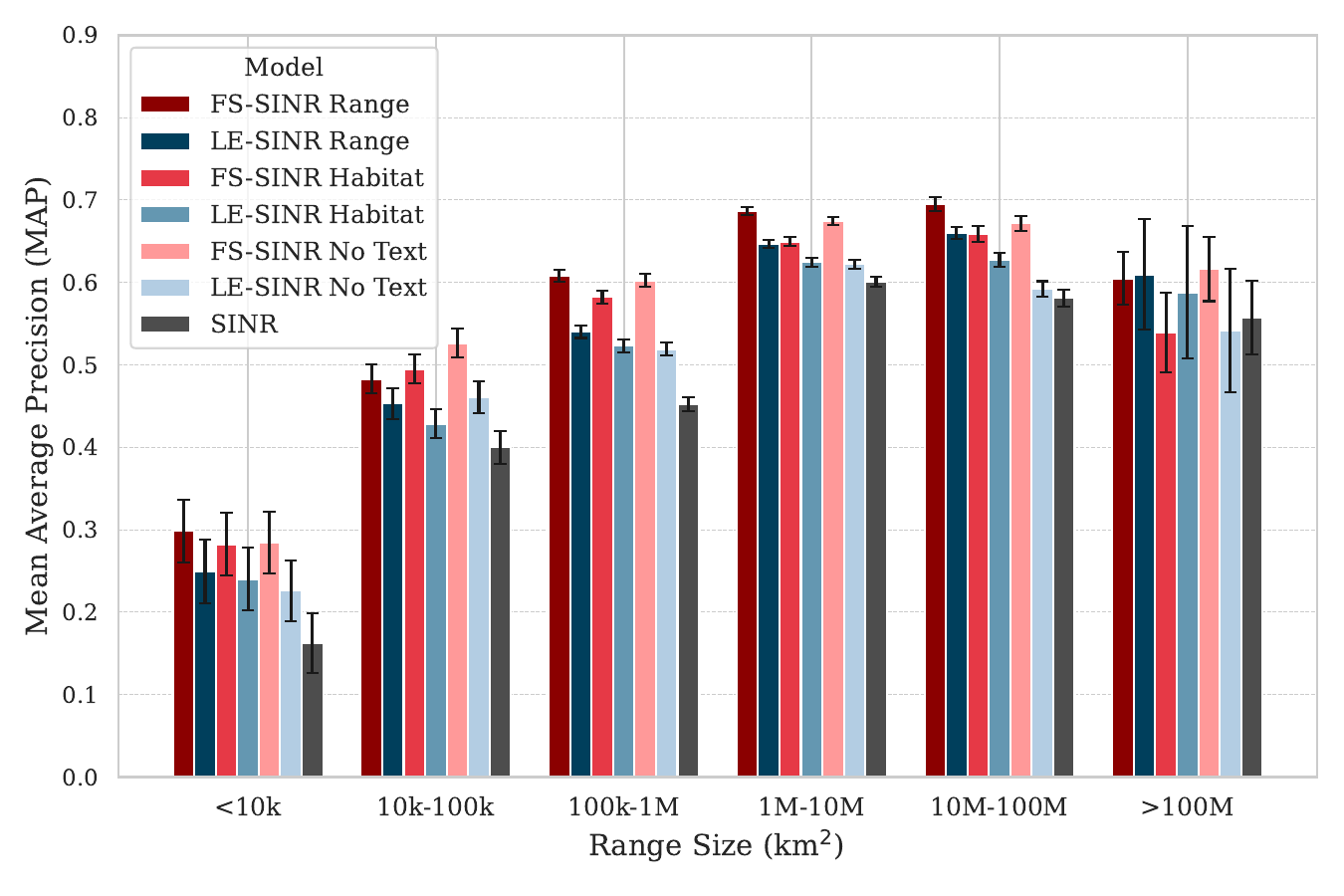}
        \caption{5 Context Locations}
    \end{subfigure}

    \begin{subfigure}{0.42\textwidth}
        \includegraphics[width=\linewidth]{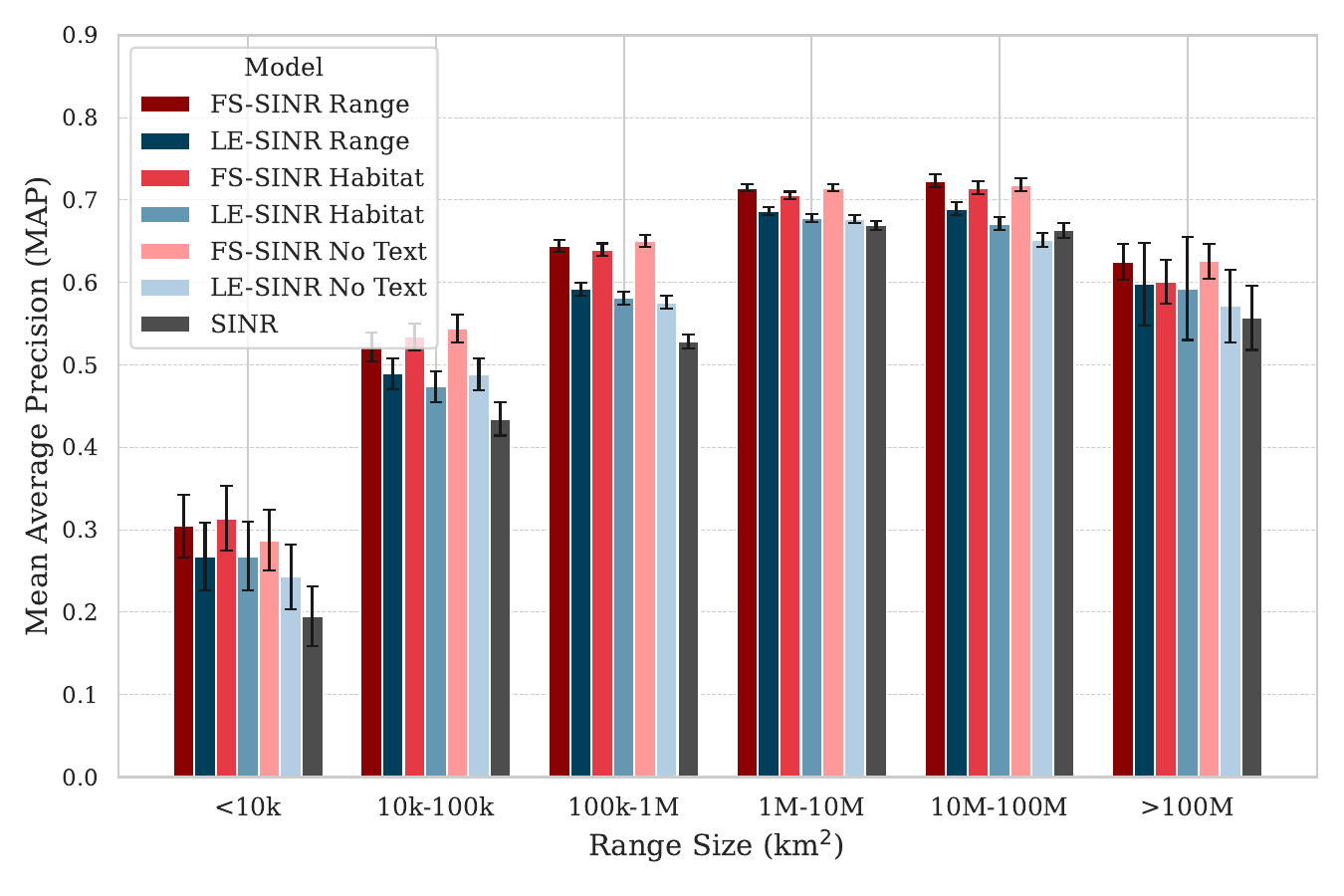}
        \caption{20 Context Locations}
    \end{subfigure}%
    \hspace{1em}
    \begin{subfigure}{0.42\textwidth}
        \includegraphics[width=\linewidth]{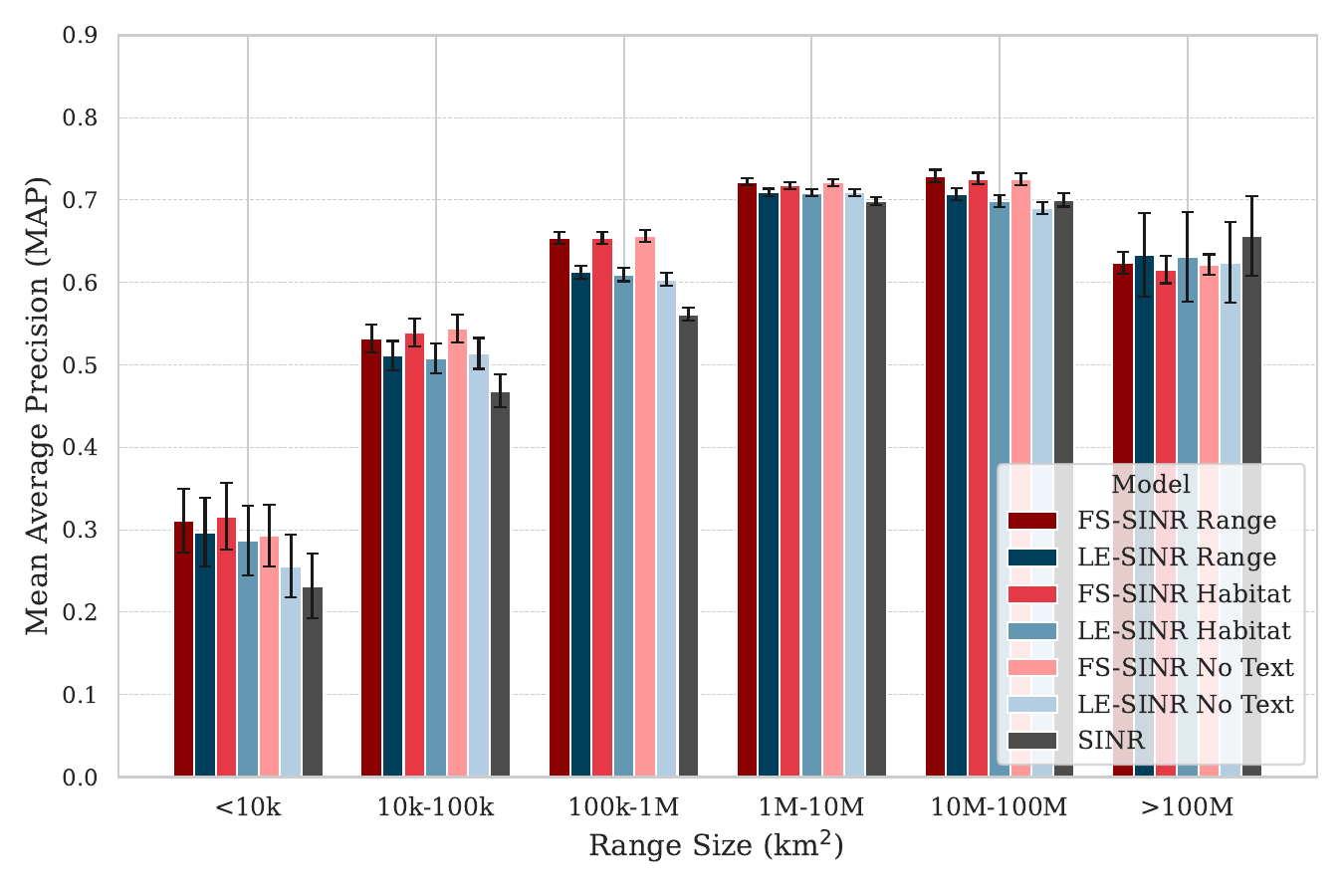}
        \caption{50 Context Locations}
    \end{subfigure}

    \vspace{-10pt}
    \caption{\textbf{IUCN Performance by range size.} Error bars show standard error of the mean. }
    \vspace{-10pt}
    \label{fig:range_size_eval}
\end{figure}

Here, we display results showing the average MAP for species in our IUCN evaluation dataset, grouped by range size, where range size is computed from the expert-derived range maps.
In \cref{fig:range_size_eval} we break down performance of zero-shot approaches by range size for FS-SINR, LE-SINR, and SINR. 
We find that for all models and settings, performance varies very strongly with range size. 
This is most significant in the zero-shot setting.
FS-SINR performs well compared to the baselines, though all models struggle with very small ranges.
As small-ranged species are especially vulnerable to extinction~\citep{CHICHORRO2019220}, current methods performing poorly for these species when evaluated globally may be of concern and improving the modeling of these species may be a priority from a conservation perspective.
We also see that performance worsens for the very largest ranges. 

\subsection{Results by Taxonomic Class}

\begin{figure}[h]
    \centering
    \begin{subfigure}{0.42\textwidth}
        \includegraphics[width=\linewidth]{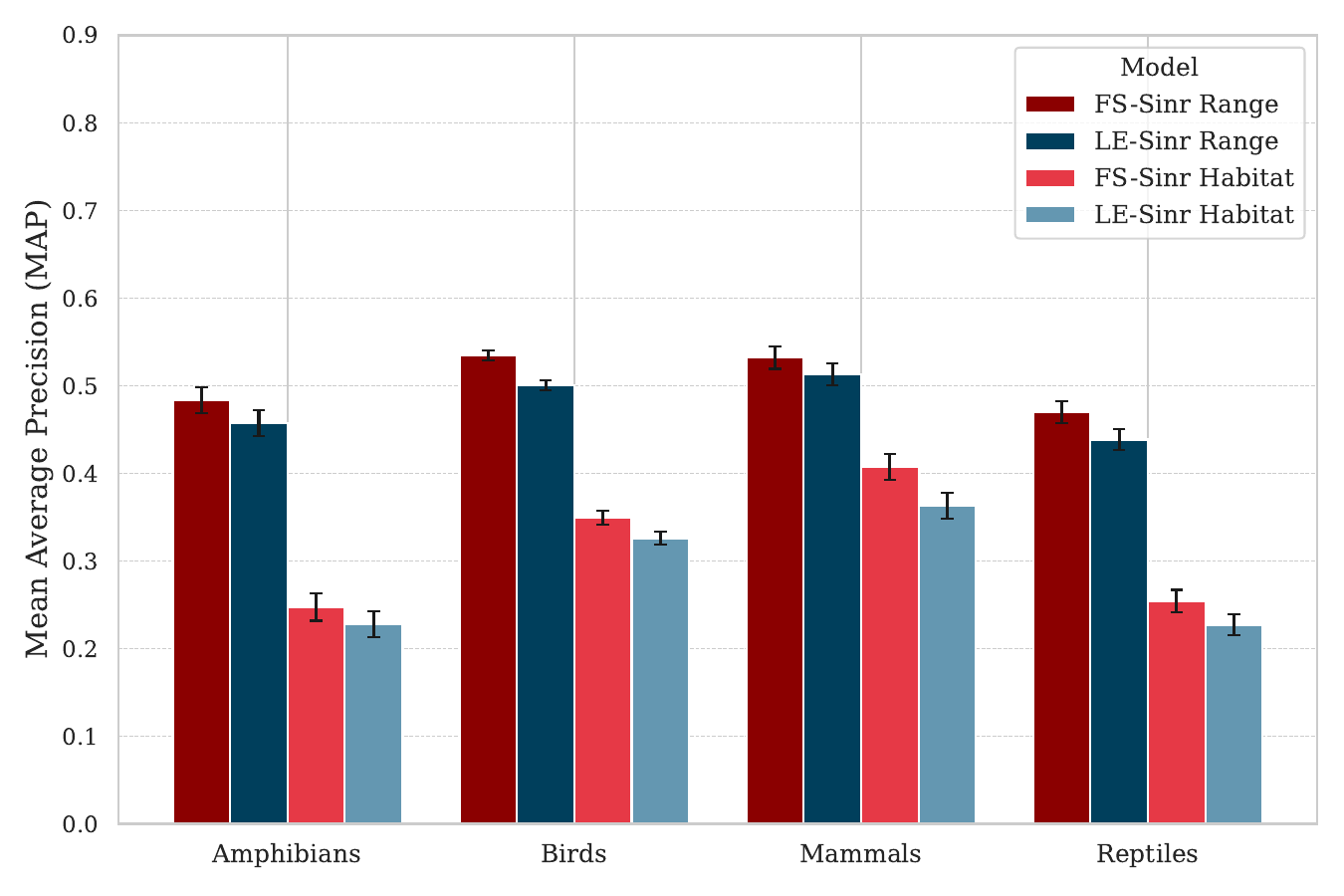}        
        \caption{0 Context Locations}
    \end{subfigure}%
    \hspace{1em}
    \begin{subfigure}{0.42\textwidth}
        \includegraphics[width=\linewidth]{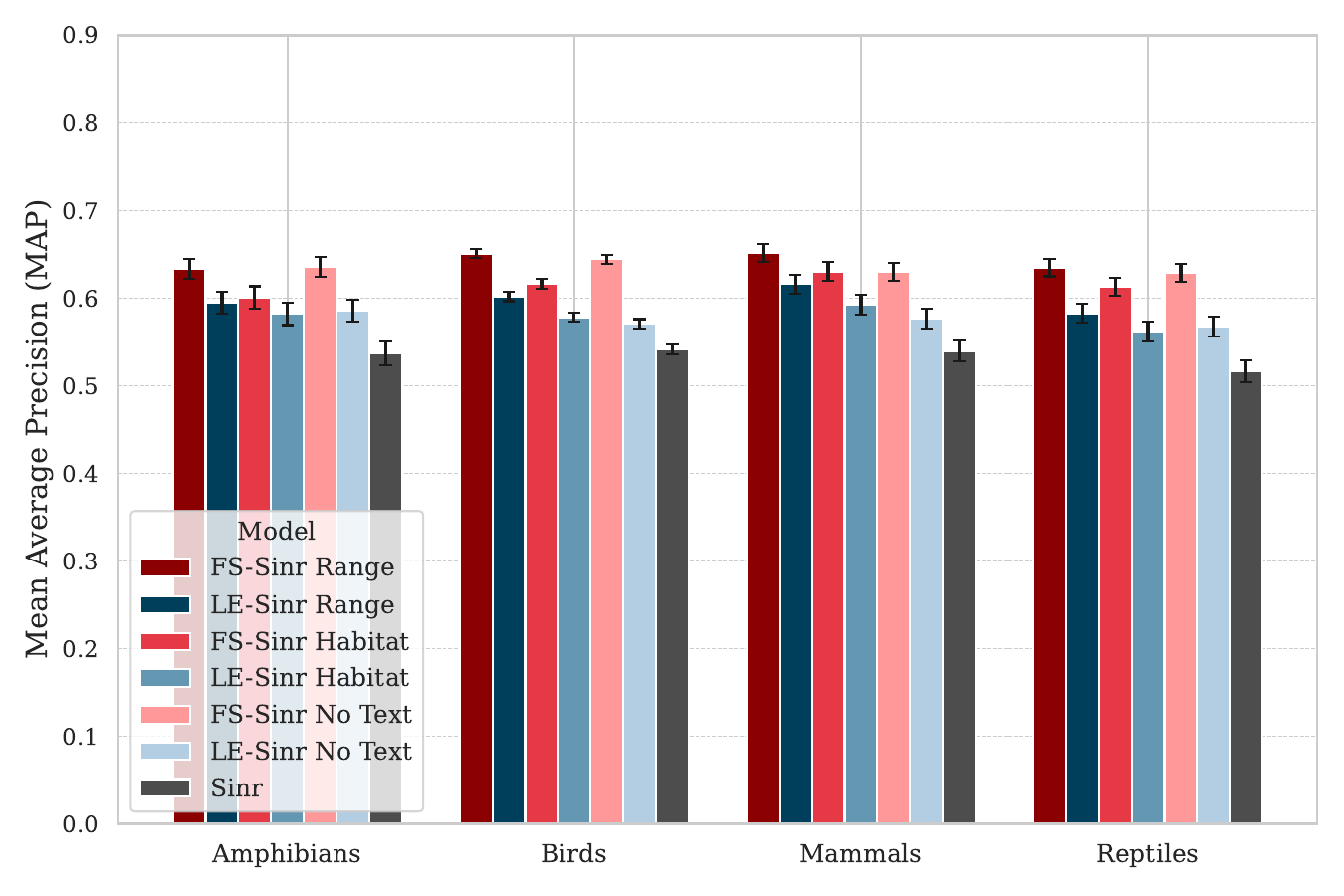}
        \caption{5 Context Locations}
    \end{subfigure}

    \begin{subfigure}{0.42\textwidth}
        \includegraphics[width=\linewidth]{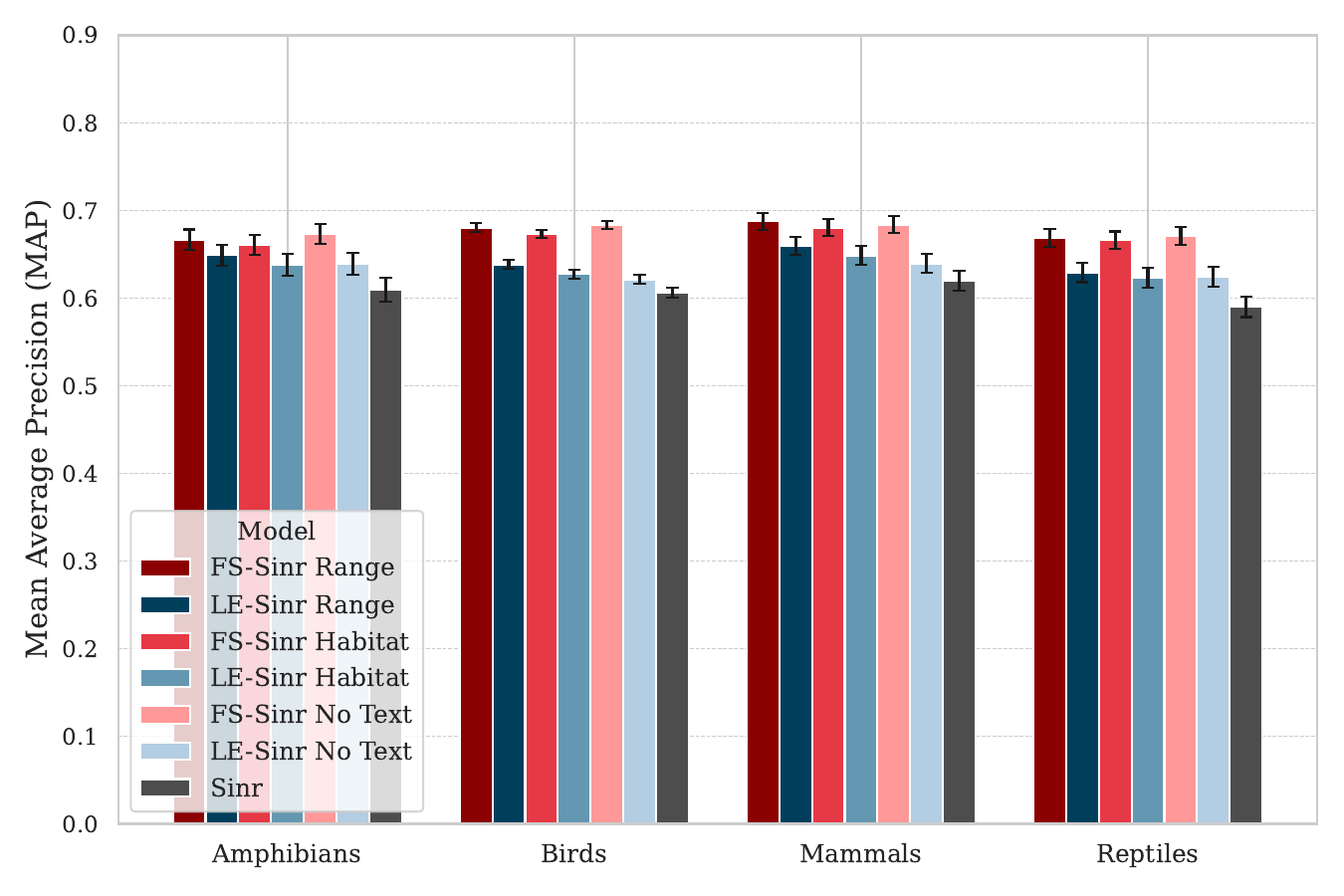}
        \caption{20 Context Locations}
    \end{subfigure}%
    \hspace{1em}
    \begin{subfigure}{0.42\textwidth}
        \includegraphics[width=\linewidth]{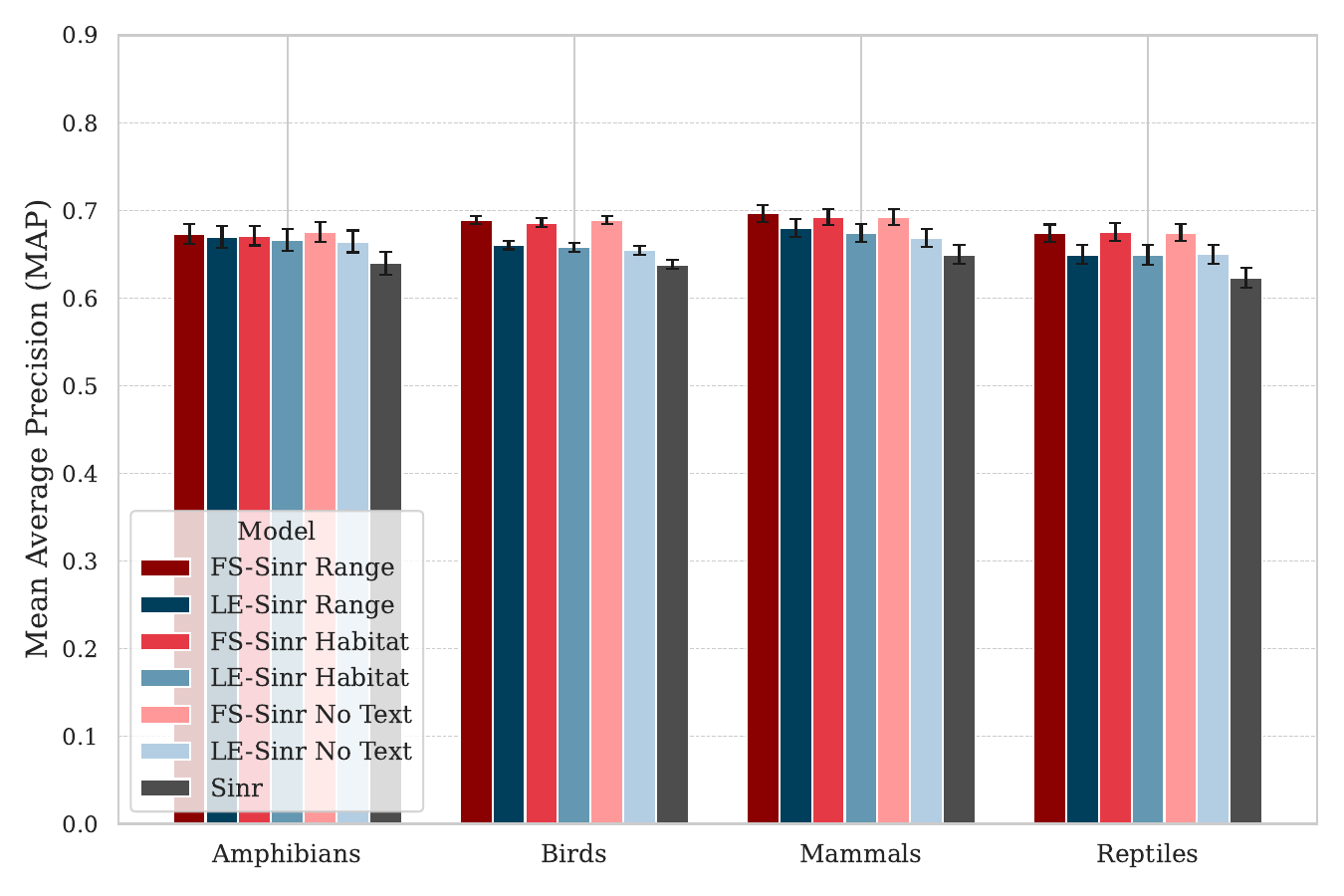}
        \caption{50 Context Locations}
    \end{subfigure}

    \vspace{-10pt}
    \caption{\textbf{IUCN Performance by taxonomic group.} Error bars show standard error of the mean. }
    \vspace{-10pt}
    \label{fig:taxonomic_eval}
\end{figure}

Here we break down performance on the IUCN evaluation dataset by taxonomic class. 
Four taxonomic classes are present in our training data, namely \texttt{Amphibia}, \texttt{Aves}, \texttt{Mammalia}, and \texttt{Reptilia}.  
In \cref{fig:taxonomic_eval} (a) we display zero-shot performance for FS-SINR and LE-SINR using range and habitat text.
We observe that \texttt{Aves} and especially \texttt{Mammalia} outperform the other classes, particularly when habitat text is provided.
\citet{albert2018twenty} suggest that of the 20 most `charismatic' species in the western world, all but the \texttt{Great White Shark} and \texttt{Crocodile} are mammals, and \citet{trimble2010species} show that scientific research is heavily focused on mammals.
We may be seeing the impact of this, where mammals are more likely to have detailed Wikipedia pages which we draw our textual training and evaluation data from.
In \cref{fig:taxonomic_eval} (b), (c), and (d), we investigate how these differences in performance between taxonomic classes change as more location data is provided.
We see that for both FS-SINR and LE-SINR, providing context locations significantly reduces the differences in performance between taxonomic class, though mammals do continue to very slightly outperform other taxonomic groups for a given model and setting.

\section{Implementation Details} 
\label{supp:impl_details}

\subsection{FS-SINR}  
\label{supp:fs_impl_details}
\subsubsection{Model Architecture}
Code for \modelname is available at https://github.com/Chris-lange/fs-sinr. The architecture for \modelname consists of four components: the location encoder, $f$; the text encoder,  $t$; the image encoder, $a$; the transformer encoder, $m$; and the species decoder, $s$. These components comprise 8,154,368 learnable parameters in total.
All non-linearities in \modelname  are ReLUs.

The location encoder, $f$, is identical to the one used in \citet{hamilton2024}, which is taken from ~\citet{cole2023spatial}. It is composed of an initial linear layer and ReLU non-linearity followed by four residual layers, where each is a two-layer fully connected network with residual connections \citep{He2015DeepRL} between the input and output of each residual layer. Each layer contains 256 neurons, and there are 527,616 learnable parameters in total.

The text encoder, $t$, follows the structure of the text-based species encoder from \citet{hamilton2024}. 
In $t$, a pre-trained and frozen large language model, GritLM~\citep{muennighoff2024generative}, is used to produce a fixed 4,096 length embedding from input text. This is then passed through a smaller network to reduce the dimensionality to 256. This smaller network consists of two residual layers with a hidden layer size of 512. In total, the text encoder contains 3,410,432 learnable parameters.

The image encoder, $a$, has a structure similar to $t$. 
In $a$, a pre-trained and frozen vision transformer, EVA-02~\citep{FANG2024105171}, pre-trained on images from 10,000 species from the iNaturalist species classification dataset~\citep{van2021benchmarking}, is used to produce a fixed 1,024 length embedding from an input image, by extracting the \texttt{CLS} token from the final layer of the model.
This is then passed through a smaller network to reduce the dimensionality to 256. This smaller network consists of two residual layers with a hidden layer size of 512. In total, the image encoder contains 1,837,568 learnable parameters. \cref{tab:main_results_table_iucn,tab:main_results_table_snt,tab:zeros_shot_appendix} include results using a DINOv2-large image encoder~\citep{oquab2023dinov2} instead of the EVA-02 ViT where all other architecture choices remain the same.

The transformer encoder, $m$, takes in an arbitrary length set of unordered 256 dimensional tokens produced by $f$,  $t$, and $a$, as well as two learned tokens that are added to each set of inputs. 
The \texttt{CLS}, class, token  produces the species range, and a `Register' token, inspired by~\citet{darcet2023vision}, which acts as an additional repository of global information during encoding.
Element-wise addition between each token and one of five learned 256 dimensional `token type embeddings' is performed to allow the model to differentiate between tokens from different sources.
The transformer itself is composed of four transformer encoder layers, implemented using PyTorch's \texttt{nn.TransformerEncoderLayer}~\citep{pytorch}, based on~\citet{vaswani2017attention}. 
Key-query-value multi-head attention is used with two `heads.' 
The feed forward components contain 512 neurons per layer, while the token dimensionality is 256. 
Layer norm is used in each layer, using a default epsilon value of 1e-5 for enhanced numerical stability. 
In total, $m$ contains 2,176,256 learnable parameters.
Finally the species decoder, $s$, is a simple fully connected network with two hidden layers. 
Each layer contains 256 neurons, and in total the decoder contains 197,376 learnable parameters.

\subsubsection{Training}
For all training we use the Adam optimizer \citep{Kingma2014AdamAM} with a learning rate of 0.0005, and an exponential learning rate scheduler with a learning rate decay of 0.98 per epoch, and we use a batch size of 2048. 
Our training data comes from \citet{cole2023spatial}, comprising 35.5 million species observations with locations, covering 47,375 species observed prior to 2022 on the iNaturalist platform. 
However, we remove all species found in our evaluation datasets, leaving us with 44,181 species in our training set. 

Training comprises of two steps. 
First, the location encoder, $f$, is trained. 
This follows the training procedure of \citet{cole2023spatial} using the $\mathcal{L}_{\text{AN-full}}$ loss function with positive weighting, $\lambda$, set to 2,048, training for 20 epochs with a dropout of 0.5. 
To reduce training time without significantly impacting performance we only train on a maximum of 1,000 examples per-species, as done in~\citet{cole2023spatial}. 
Thus, our training dataset for this step contains 13.8 million location observations.
Secondly, we train all components of \modelname, except the pre-trained large language model and the pretrained vision transformer, using our $\mathcal{L}_{\text{AN-full-b}}$ loss with $\lambda$  set to 2,048.
We train the location encoder, $f$, again as this improves performance compared to freezing it,as seen in \cref{fig:backbone ablation}. 
For this part of training, we use a dropout of 0.2.
We further reduce the training data used to a maximum of 100 examples per-species, leaving 4.0 million training examples, which again increases training speed without a significant impact on performance, as seen in \cref{fig:training_cap_ablation}.
For this step we additionally train with images and text descriptions of the training species. 
Each instance in the training set is used once per epoch as a training example to compute the loss. The training example is not passed through the transformer encoder, $m$, and so does not contribute to the species embedding vector produced by this part of the model.
Instead, additional context information is provided to produce the species embedding. 
By default, this context information consists of 20 context locations, a section of text describing the target species, and an image of the species.
With 0.2 probability the context locations are dropped from the context information, and with 0.5 probability each the text or image is dropped.
These context locations are taken from the training data for the target species.
As such, a single instance from the training set can be used multiple times per epoch, once as a training example, and potentially many times as a context location.
The impact of different distributions of locations and text provided during training is shown in \cref{fig:context_info_ablation}.

For the text inputs required during this stage of training, we use the text dataset from~\citet{hamilton2024} consisting of multiple sections of Wikipedia~\citep{wikipedia} articles for each species in the train set where these are available.  
This dataset contains 127,484 sections from 37,889 species’ articles. 
The evaluation text either describes the habitat or range a species, where habitat text tends to describe the local environment and range text is typically more informative as it often lists specific countries or geographic regions where the species can be found. 
Note that not all 44,181 train species have text data available.
The images used are taken from~\citet{inatWeb}, and this dataset comprises 204,064 images of our train species.
When an image or piece of text is not available for a species during training,  and we are trying to provide these modalities and context locations to the model, 
we simply ignore the additional modality and only provide the context locations. 
When we are attempting to provide just an image or text as context, we instead skip that training instance.

In practice, during training, we pass all text sections through the frozen large language model once and then store the embeddings produced to use in the current training run and all future runs, and similarly extract and store all image embeddings after passing the images through the frozen vision transformer. This prevents us having to repeatedly query these frozen, but resource-intensive, models during training.
Training takes approximately ten hours on a single NVIDIA A6000 GPU, requiring approximately six gigabytes of RAM. 

\subsection{Baselines} 
\label{sec:appendix_baselines}

\subsubsection{LE-SINR}
We compare our approach to the recently introduced species range model LE-SINR~\citep{hamilton2024} that can incorporate text information. We follow the original architecture and training procedure for LE-SINR and SINR, with the exception that we enforce that SINR, like LE-SINR and our approach, is trained on our reduced set of 44,181 species which do not include any of the evaluation species. 

We also follow the original evaluation procedure for LE-SINR. 
For few-shot evaluation without text, logistic regression with L2 regularization is performed with location features as input using the few positive examples provided alongside a set of pseudo-negatives drawn half from a uniform random distribution and half from the training data distribution.
The regularization weight is set to 20. 
For text-based zero-shot evaluation, we directly make use of the output of the text encoder with the dot product between this and location features giving us a probability of species presence. 
For few-shot evaluation, when text is provided, we again perform logistic regression, but the output of the text encoder is used as the `target' that the weights are drawn towards in a modified L2 regularization term,see~\citet{hamilton2024} for more details. 
The regularization weight is again set to 20.
In total, this model comprises 25,715,202 learnable parameters. 

\subsubsection{SINR}
We also compare to SINR~\citep{cole2023spatial}. 
The original SINR implementation requires all evaluation species to be part of the training set. 
We match the adaptations from~\citet{hamilton2024} to allow evaluation on unseen species. After training we remove the learned species heads and keep only the location encoder. During evaluation, we perform logistic regression with L2 regularization using location features as input.
The regularization weight is again set to 20, and the same method of selecting pseudo-negatives as above is used. 
In total, this model comprises 11,941,120 learnable parameters. 

\subsubsection{Prototype SINR} 
\label{sec:prototype}
Here we describe our few-shot baseline based on Prototypical Networks~\citep{prototypes}, which we refer to as Prototype SINR.
Our approach is very similar to~\citet{prototypes} although we use the SINR location encoder of our models as the `embedding function', allowing us to generate few-shot results for a novel species without any retraining.
This SINR location encoder is trained only on species found in the training set and not those used for evaluation.
Using this method, SINR and LE-SINR models can be used to estimate the range of a novel species without requiring training to learn a new species embedding vector. 

In order to do this, we first encode our known `presence' locations using the location encoder of our chosen model and then take an average of these points to generate a `prototype' for the presence class.
We select pseudo-negatives in the same manner as in~\citet{hamilton2024} and similarly encode and average these in order to generate a prototype for the `absent' class.
We represent these prototypes as: 
\begin{equation}
    \bm{r}_k = \frac{1}{\left| S_k \right|}\sum_{\bm{x}_i\in S_k}f_{\bm{\theta}}(\bm{x}_i), 
\end{equation}
where $k \in \{\text{present}, \text{absent}\}$ indicates the class of the prototype, and $S$ is the `support set', \ie the set of locations $\bm{x}$ that we use to create our prototypes.
In our case, $S_{present}$ is the small set of available context locations for our target species, \ie $\mathcal{C}^t$, while $S_{absent}$ is the set of pseudo-negative locations that we have selected according to~\citet{hamilton2024}.
$f_{\bm{\theta}}()$ denotes the location encoder of our model.

To generate a probability of presence or absence for any location $\bm{x}$, we encode $\bm{x}$ using our location encoder and calculate the cosine distance in `location encoder space' between $\bm{x}$ and each prototype. 
We then use these values as the `logits' in a softmax function to generate our probabilities. 
The parameters of the location encoder are not changed. 
Putting this together, we can calculate the probability of presence as:  

\begin{equation}
    p_{\text{present}}(\bm{x}) = \frac{e^{d({f_{\bm{\theta}}(\bm{x}), \bm{r}_{\text{present}})}}}{e^{d({f_{\bm{\theta}}(\bm{x}), \bm{r}_{\text{present}})}} + e^{d({f_{\bm{\theta}}(\bm{x}), \bm{r}_{\text{absent}})}}},
\end{equation}

where $d(\bm{a},\bm{b})$ represents a distance metric between $\bm{a}$ and $\bm{b}$, in this case, cosine distance.

While the original implementation in~\citet{prototypes} uses the squared Euclidean distance instead of cosine distance we find that this performs significantly worse and actually results in decreased MAP as the number of context locations increases.
We suggest the SINR location encoder is more suited to using cosine distance, as during training presence predictions are generated by taking the dot product of the location and species embeddings. 
However, when the location encoder is trained from scratch for Prototype SINR as in~\citet{prototypes} we find that using the squared Euclidean distance performs better than cosine distance, although performance is still lower than cosine distance with a SINR location encoder.

\begin{figure}[h]
    \centering
    \includegraphics[width=0.9\textwidth, trim={0 0 0 1cm},clip]{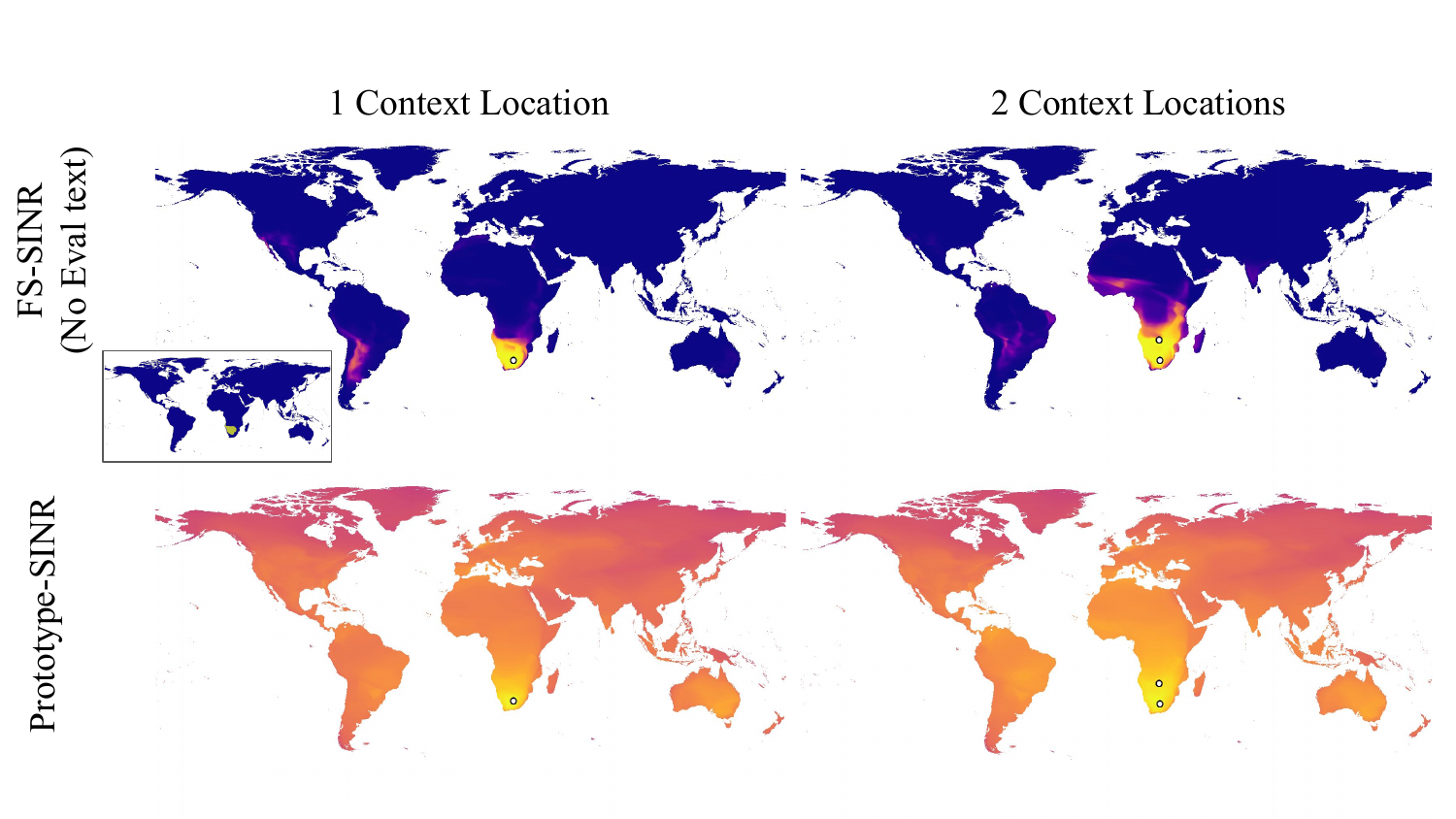}
    \vspace{-20pt}
    \caption{\change{ \textbf{Qualitative comparison of the Prototype SINR baseline.}  
    Here we compare the predictions of our \modelname approach (without any text) and our Prototype SINR baseline on the \texttt{Kalahari Scrub-Robin} species that is found in Southern Africa.
    The Prototype SINR approach obtains an MAP of 0.54 and 0.79, for one and two context locations respectively, while \modelname obtains 0.79 and 0.85.
    As MAP is tied to the ranking of predicted probabilities rather than their absolute values, it can remain high even if the model is somewhat overconfident across the board.
    As long as the highest probabilities consistently align with areas where the species is truly present, the model will achieve a strong MAP, which we can see with the predictions from Prototype SINR.
    }
    }
    \label{fig:prototypes_qualitative}
    \vspace{-10pt}
\end{figure}

In \cref{fig:low_shot} we see that the performance of Prototype SINR is worse than \modelname and the SINR and LE-SINR baselines.
In \cref{tab:zeros_shot_appendix} we provide zero-shot results where the positive prototype is a species embedding produced from text, in the same manner as LE-SINR zero-shot predictions.
In both cases, Prototype SINR underperforms compared to our approach.
In \cref{fig:prototypes_qualitative} we present qualitative results visualizing the few-shot estimated range for the \texttt{Kalahari Scrub-Robin} produced by \modelname and by the Prototype SINR baseline.

\subsubsection{Active SINR}
We also compare to the model introduced for active learning in~\citet{lange2024active}, which we call Active SINR, although in our setting there is no active learning component. 
This approach begins with a SINR model trained on our reduced 44,181 species which do not include the evaluation species.
The weights $\bm{W}$ of the multi-label classifier of this model can be viewed as a set of species embeddings where each column vector $\bm{w_j}$ of $\bm{W}$ represents an individual species $j$. 
We can combine these species embeddings with a location embedding $f_{\bm{\theta}}(\bm{x})$ via an inner product to compute the probability that the species $j$ is present at $\bm{x}$. 
At inference time, we compute the presence probabilities for all species in the training set, for all locations $c$ in the set of available context locations $\mathcal{C}^t$ for our target species $t$. We then produce a new species embedding $\bm{w}_t$ by taking a weighted average of the existing $\bm{w}_j$'s where the weight for each is the product of the probabilities of presence for that species:
\begin{equation}
\bm{w}_t = \sum_{j=1}^{s} P(\bm{w}_j|\mathcal{C}^t)\bm{w}_j. 
\label{eqn:param_est}
\end{equation}

We can then use this new species embedding for our target species to produce a probability of presence for any location $x$ as in SINR~\citep{cole2023spatial}.
We present few-shot results using this method in \cref{fig:low_shot}.
We see that the performance of the Active SINR approach is competitive with \modelname when provided with no text, though worse than \modelname when provided with this additional context.
However, increasing the number of provided context locations beyond a small number actually hurts performance, as it is unable to accurately represent the range of a previously unseen species via the weighted combination of those from the training set.  

\subsection{Evaluation}
We perform three runs for each experiment using different initial random seeds and report the mean. 
We display the standard deviation as error bars in our figures.
For all evaluations across SINR, LE-SINR, Prototype SINR, Active SINR, and FS-SINR, the same set of context locations are used for a given species, and these context locations are accessed in the same order, so all evaluations using five context locations are performed with the same five points, and four of those points are those used for evaluations using four context locations, \etc 
In our few-shot setting, we use at most 50 context locations during both training and evaluation.

\section{Additional Ablations}
\label{sec:app_ablations}
Here we present additional results to investigate the impact of a variety of design choices and training procedures for \modelname. 
We present plots on a ``Symlog'' scale, where a linear scale is used between 0 and 10, in order to allow us to show zero-shot results alongside few-shot results. 
We display the mean of three runs with standard deviations shown as error bars and also show just the mean values alongside for easier interpretation.

\subsection{Ablating Training Context Locations} 

\begin{figure}[h]
    \centering
        \includegraphics[width=0.42\textwidth]{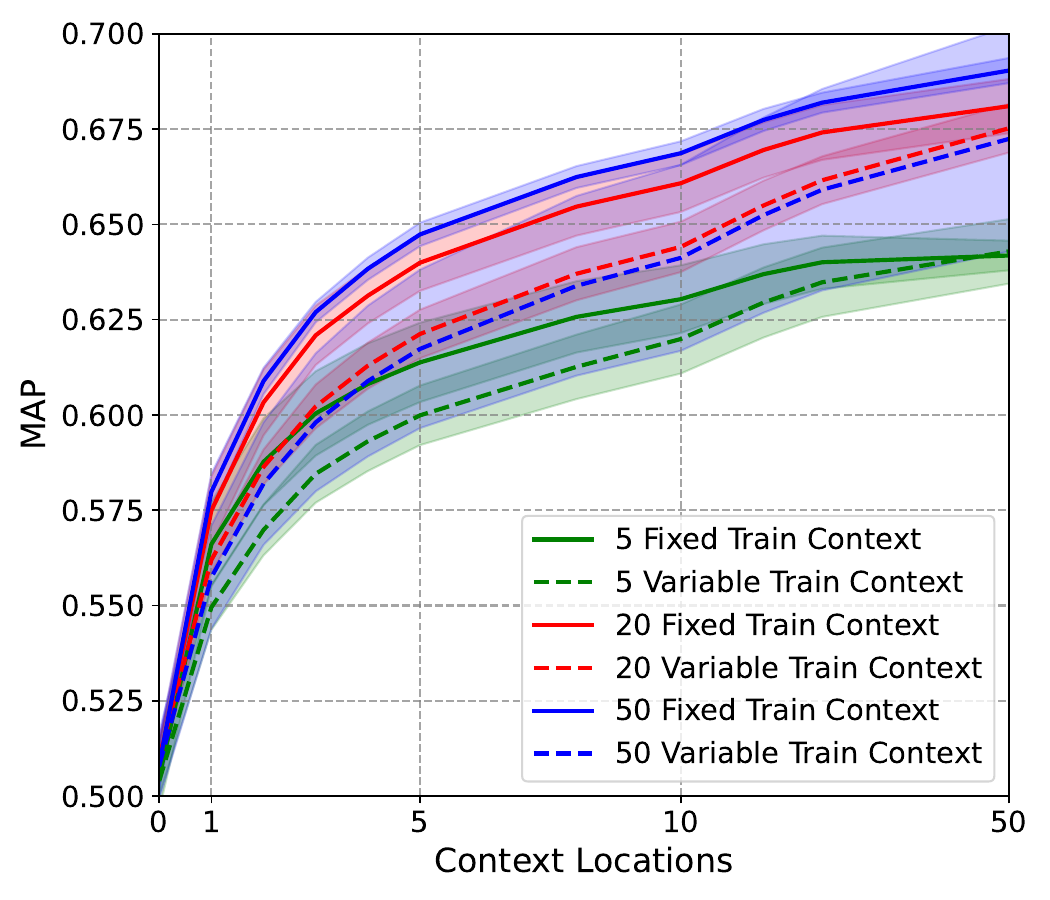}
        \hspace{15pt}
        \includegraphics[width=0.42\textwidth]{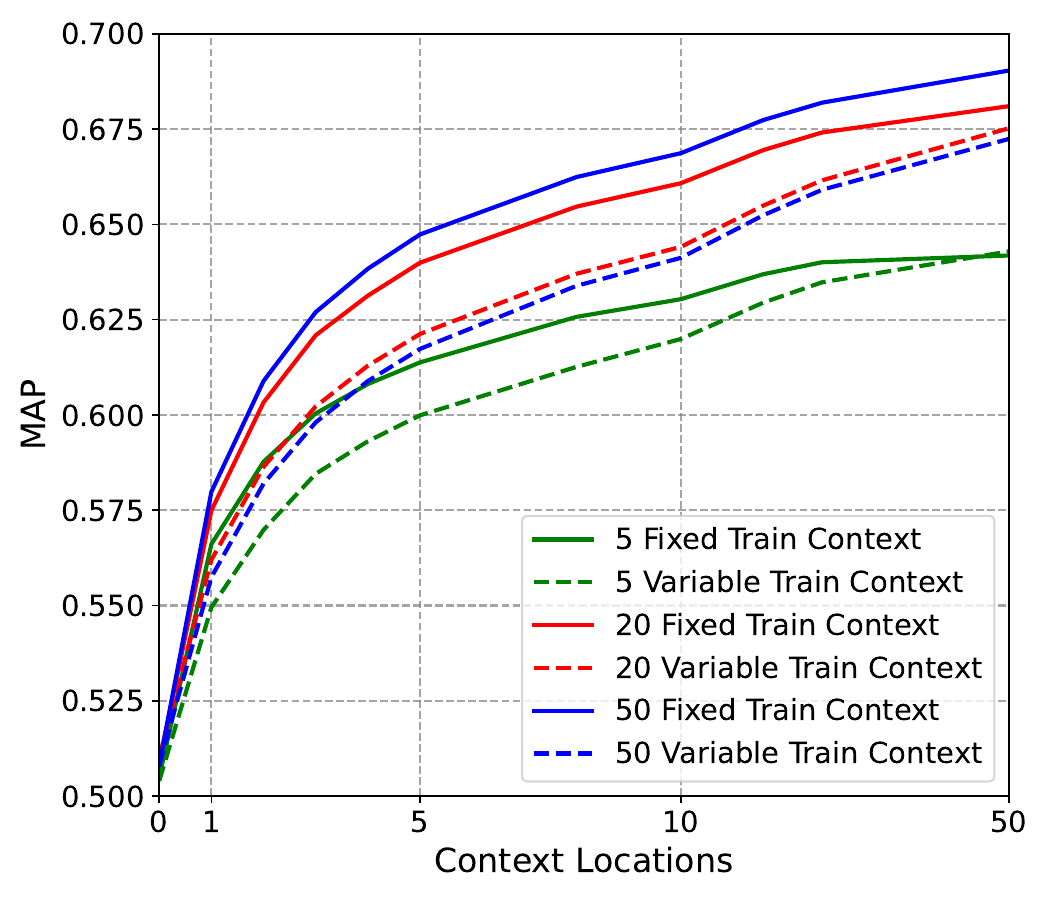}
    \vspace{-15pt}
    \caption{\change{\textbf{Impact of number of training context locations.} Here we evaluate FS-SINR models trained using different numbers of context locations. Results are shown with standard deviations from three runs (left), and without (right) for clarity.
    Evaluation is performed with `Range Text' on the IUCN dataset.
    `Fixed' indicates the same number of context locations were provided for every training example.
    `Variable' indicates that a uniform random distribution of context locations up to the specified number were provided with each training example.
    We see that `Variable' generally under-performs compared to `Fixed' and that increasing the train context length tends to increase evaluation performance.}}
    \label{fig:train_context_ablation}
\end{figure}

In \cref{fig:train_context_ablation} we show `Range Text' evaluation performance on the IUCN dataset for \modelname models trained using different amounts of context information at training time. 
We see that generally increasing the context used during training improves performance, and that having a fixed number of context locations is also beneficial.

\subsection{Ablating Context Information}
\begin{figure}[h!]
    \centering
        \includegraphics[width=0.42\textwidth]{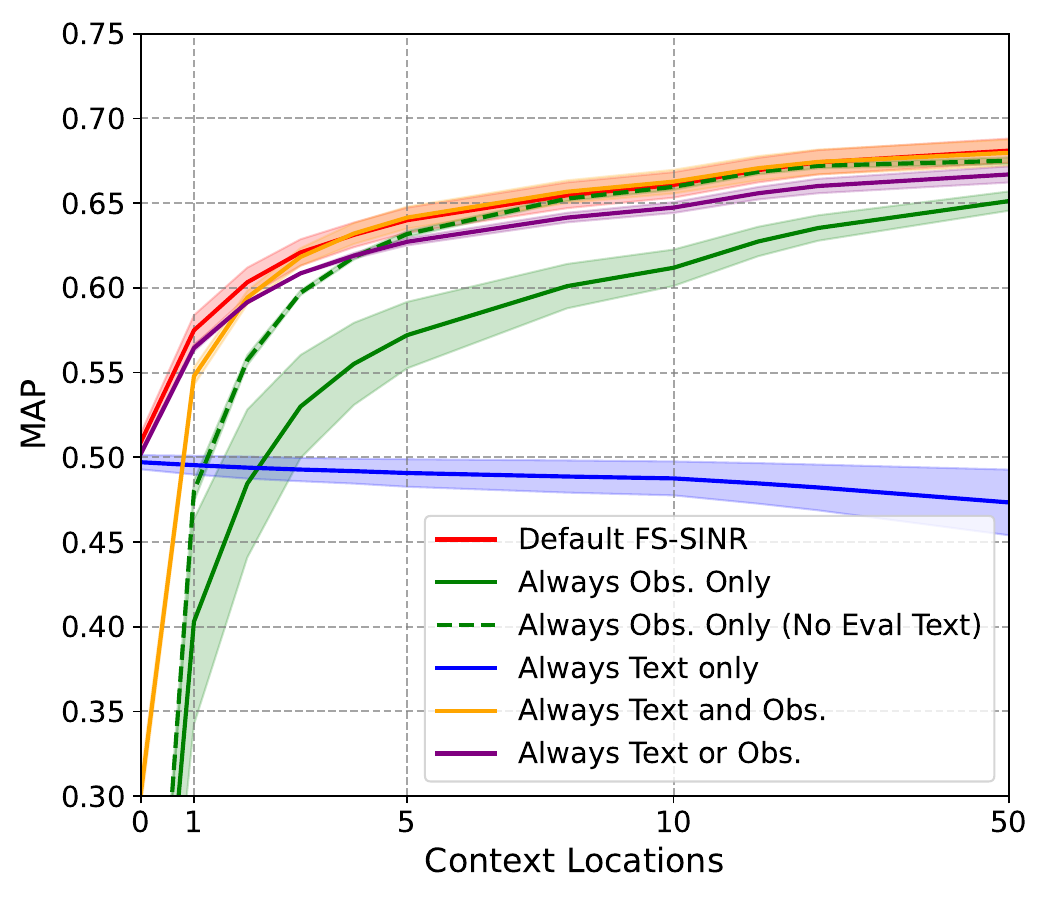}
        \hspace{15pt}
        \includegraphics[width=0.42\textwidth]{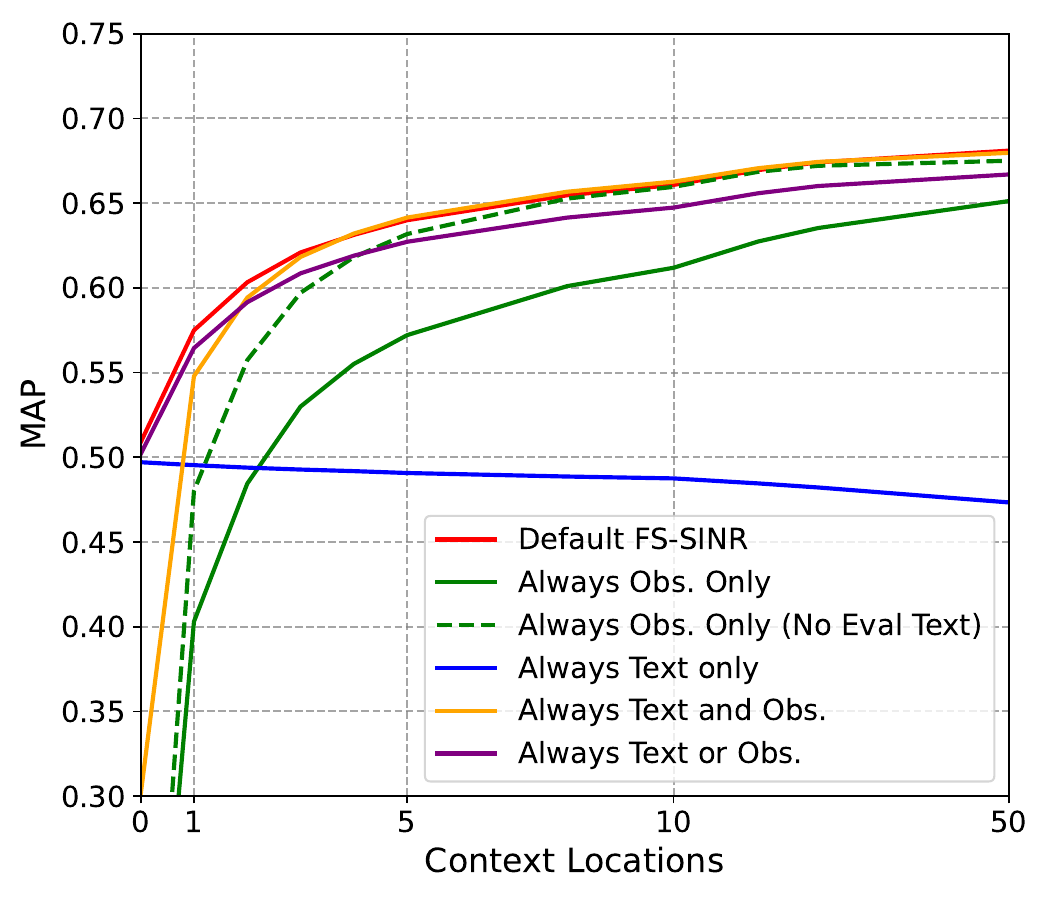}
    \vspace{-15pt}
    \caption{\change{\textbf{Impact of train context information.} Here we evaluate \modelname models trained using different context information on the IUCN dataset.
    Results are shown with standard deviations from three runs (left), and without (right) for clarity.
    Evaluation is performed with `Range Text' unless `No Eval Text' is specified, in which case just locations are provided during eval.
    70\% of training examples for `Default FS-SINR' provide both location and text context, 20\% provide just locations 10\% and provide just text.
    `Always Obs. Only' has only seen locations during training. `Always Text Only' has only seen Text during training.
    `Always Text and Obs' is always provided with both locations and text during training.
    `Always Text or Obs.' is provided with just locations for 90\% of training examples, and just text for the remaining 10\%.}}
    \label{fig:context_info_ablation}
\end{figure}

In \cref{fig:context_info_ablation} we display `Range Text' evaluation performance on the IUCN dataset for \modelname models trained using different combinations of text and context locations during training. 
We observe that good text-only zero-shot performance requires sometimes providing just text as context information during training. 
This forces the model to learn to produce ranges from only text information.
Models that are sometimes provided with both text and locations for the same training examples perform best as the number of provided context locations increases.
We also see that models trained without text can perform on par with those that see text during training when enough context locations are provided (5 - 10).
As we might expect, models that are provided with token types they have not seen during training perform poorly.

\subsection{Ablating Input Features}
In~\cref{tab:zeros_shot_appendix} we provide additional zero-shot results expanding on those in~\cref{tab:zeros_shot} from the main paper. 
Specifically, we add comparisons to using a different location encoder (\ie SATCLIP~\citep{klemmer2024satclipglobalgeneralpurposelocation} instead of SINR), comparisons to using a DINOv2 pre-trained image encoder (DINOv2-large)~\citep{oquab2023dinov2}, comparisons to using the environmental covariates as in SINR~\citep{cole2023spatial} that contain information about a locations' climate in addition to the spatial coordinates.

\begin{table*}[t]
\centering
\caption{\textbf{Additional zero-shot results}. 
We report zero-shot performance where no location information is provided to each model, comparing SINR~\citep{cole2023spatial}, LE-SINR~\citep{hamilton2024}, and variants. 
We denote additional metadata as: 
\textbf{EN} for additional environmental covariates~\citep{fick2017worldclim} used in~\citet{cole2023spatial},
\textbf{HT} for `Habitat Text', 
\textbf{RT} for `Range Text',
\textbf{I} for `Image' using our default EVA-02 image encoder,
\textbf{I (DINOV2)} for Image using a DINOV2 based image encoder~\citep{oquab2023dinov2},
\textbf{TST} for `Test Species in Train',
\textbf{TRT} for using full taxonomic rank text,
\textbf{SATCLIP} for where the SINR encoders are replaced with the image derived location encoders from~\citet{klemmer2024satclipglobalgeneralpurposelocation}, and
\textbf{P-LE-SINR} for `Prototype LE-SINR'. 
Results are reported as MAP, where higher is better.
}
\label{tab:zeros_shot_appendix}

\begin{subtable}[t]{0.48\linewidth}
\centering
\caption{Methods without additional environmental covariates}
\begin{tabular}{l|l|cc}
\toprule
\textbf{Method} & \textbf{Variant} & \textbf{IUCN} & \textbf{S\&T} \\
\multicolumn{4}{l}{\textit{TST (test species in train)}} \\
\midrule
SINR       & TST         & 0.67 & 0.77 \\
FS-SINR    & HT, TST     & 0.38 & 0.59 \\
FS-SINR    & RT, TST     & 0.55 & 0.67 \\
\midrule
\multicolumn{4}{l}{\textit{With SATCLIP encoder}} \\
\midrule
FS-SINR    & HT, SATCLIP & 0.20 & 0.43 \\
FS-SINR    & RT, SATCLIP & 0.33 & 0.55 \\
\midrule
\multicolumn{4}{l}{\textit{Prototype SINR}} \\
\midrule
P-LE-SINR  & HT          & 0.23 & 0.48 \\
P-LE-SINR  & RT          & 0.40 & 0.55 \\
\midrule
\multicolumn{4}{l}{\textit{LE-SINR}} \\
\midrule
LE-SINR    & HT          & 0.28 & 0.52 \\
LE-SINR    & RT          & 0.48 & 0.60 \\
\midrule
\multicolumn{4}{l}{\textit{FS-SINR}} \\
\midrule
FS-SINR    &             & 0.05 & 0.18 \\
FS-SINR    & TRT         & 0.21 & 0.34 \\
FS-SINR    & HT          & 0.33 & 0.53 \\
FS-SINR    & RT          & 0.52 & 0.64 \\
FS-SINR    & I           & 0.19 & 0.38 \\
FS-SINR    & I (DINOv2)      & 0.13 & 0.28 \\
FS-SINR    & I + RT      & 0.46 & 0.64 \\
FS-SINR    & I (DINOv2) + RT & 0.46 & 0.62 \\
\bottomrule
\end{tabular}
\end{subtable}
\hfill
\begin{subtable}[t]{0.48\linewidth}
\centering
\caption{Methods with additional environmental covariates}
\begin{tabular}{l|l|cc}
\toprule
\textbf{Method} & \textbf{Variant} & \textbf{IUCN} & \textbf{S\&T} \\
\midrule
\multicolumn{4}{l}{\textit{TST (test species in train)}} \\
\midrule
SINR       & EN, TST         & 0.76 & 0.81 \\
FS-SINR    & HT, EN, TST     & 0.38 & 0.61 \\
FS-SINR    & RT, EN, TST     & 0.57 & 0.67 \\
\midrule
\multicolumn{4}{l}{\textit{LE-SINR}} \\
\midrule
LE-SINR    & HT, EN          & 0.31 & 0.52 \\
LE-SINR    & RT, EN          & 0.51 & 0.61 \\
\midrule
\multicolumn{4}{l}{\textit{FS-SINR}} \\
\midrule
FS-SINR    & EN              & 0.07 & 0.64 \\
FS-SINR    & HT, EN          & 0.32 & 0.53 \\
FS-SINR    & RT, EN          & 0.51 & 0.65 \\
\bottomrule
\end{tabular}
\end{subtable}
\end{table*}

\subsection{Ablating Location Encoder} 
\begin{figure}[h]
    \centering
        \includegraphics[width=0.42\textwidth]{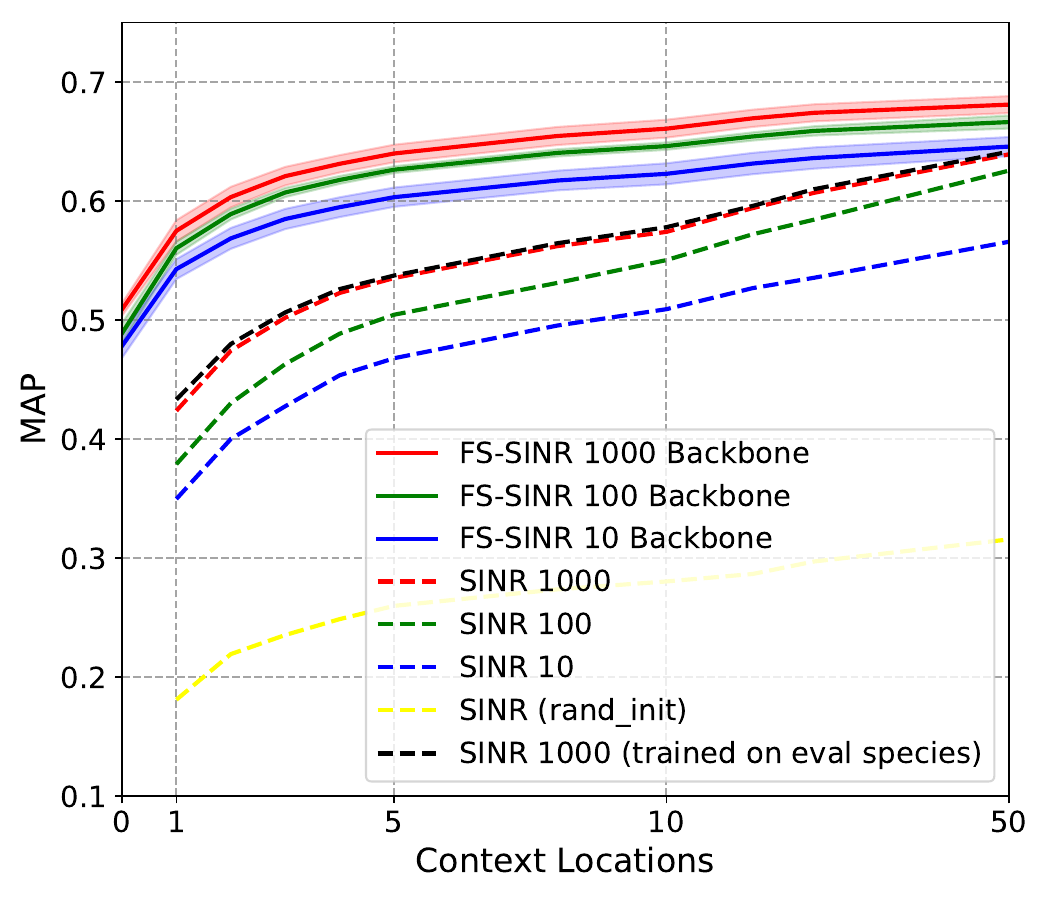}
        \hspace{15pt}
        \includegraphics[width=0.42\textwidth]{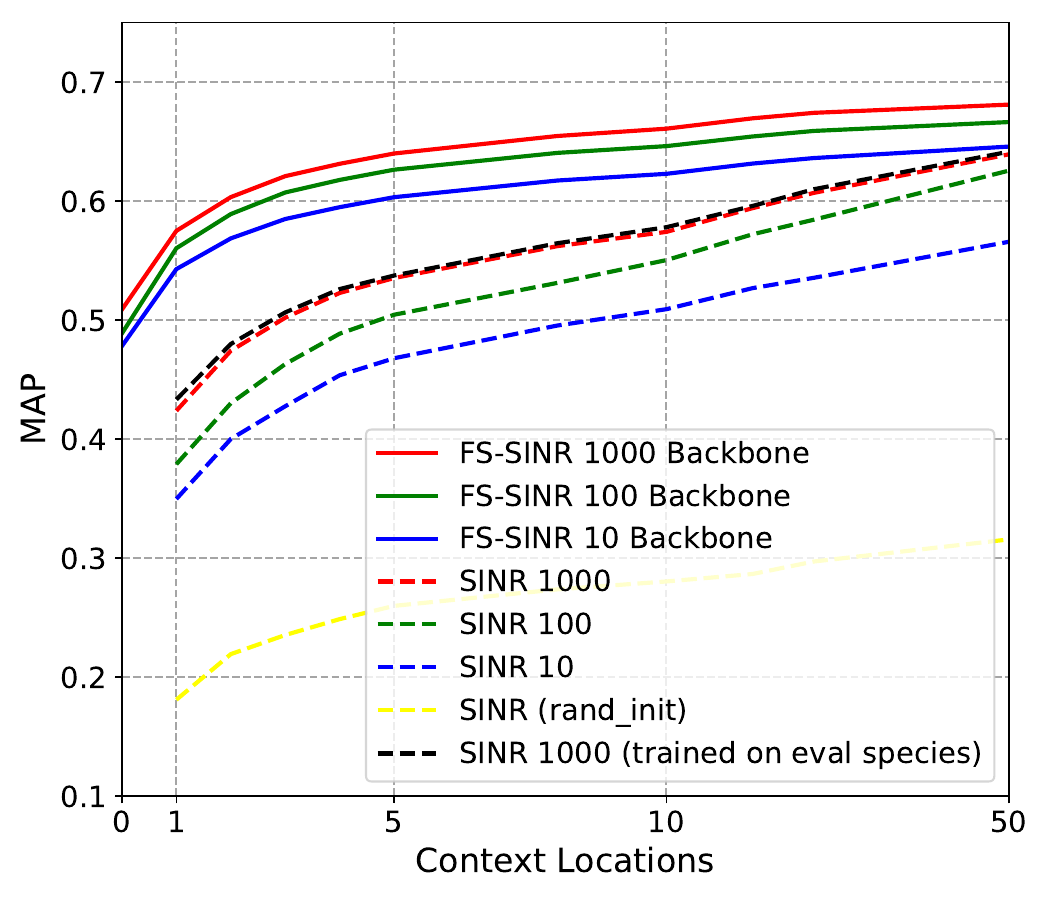}
    \vspace{-15pt}
    \caption{\change{\textbf{Impact of Location Encoder Training.} Here we evaluate the performance of SINR and FS-SINR models when the size of the training dataset for the SINR backbone is varied.
    Results for FS-SINR models are shown with standard deviations from three runs (left), and without (right) for clarity.
    Evaluation on FS-SINR is performed with `Range Text', while SINR can only make use of location data.
    `1000', `100', and `10' represent the maximum number of examples per class the SINR backbone was trained on. `SINR (rand\_init)' is initialized with random weights and is not trained.
    `(trained on eval species)' indicates training on all training and evaluation species.}}
    \label{fig:backbone ablation baselines}
\end{figure}

In~\cref{fig:backbone ablation baselines}, we vary the number of datapoints used to pre-train the SINR encoder used in \modelname. 
For both \modelname and the SINR baseline, we generally observe that more data is better, and for SINR approaches we see that pretraining the encoder is much better than randomly initializing it. 
We also show results for a SINR model trained on evaluation species in addition to train species. As we saw in \cref{tab:zeros_shot} for \modelname, the impact of training the location encoder with evaluation species is small.

\begin{figure}[h]
    \centering
        \includegraphics[width=0.42\textwidth]{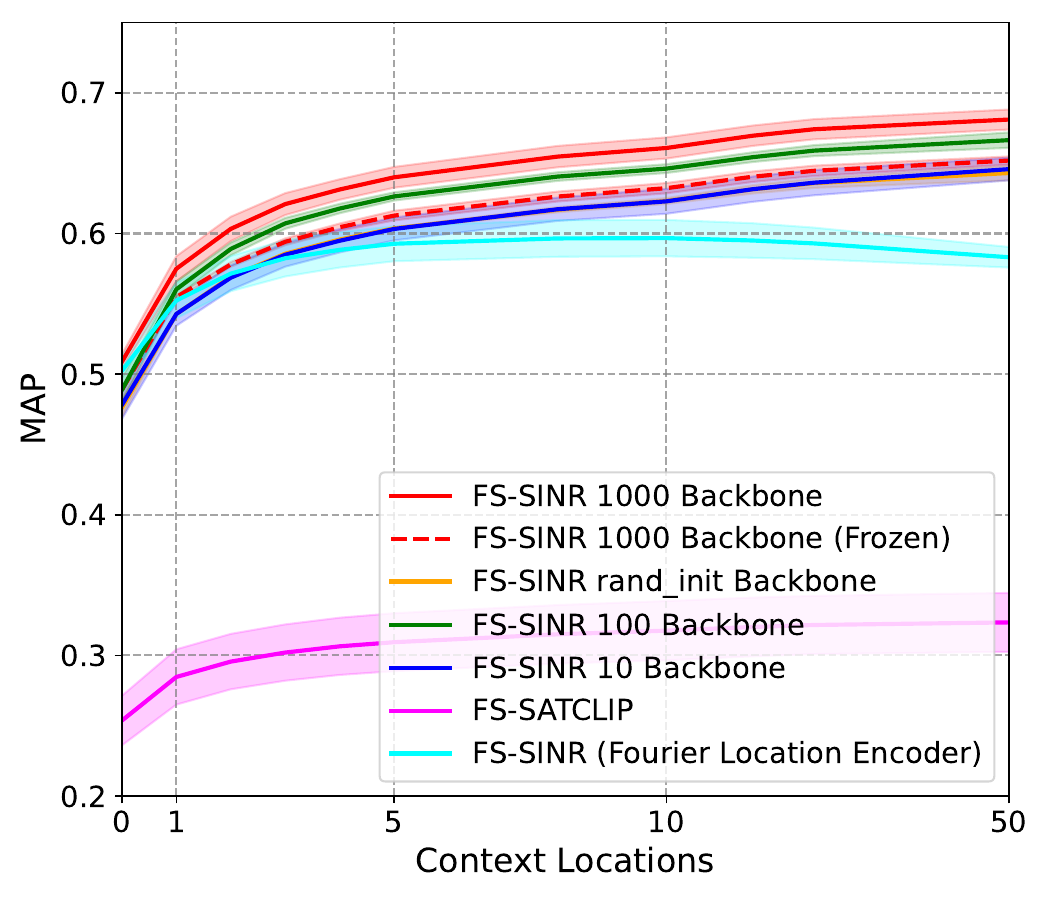}
        \hspace{15pt}
        \includegraphics[width=0.42\textwidth]{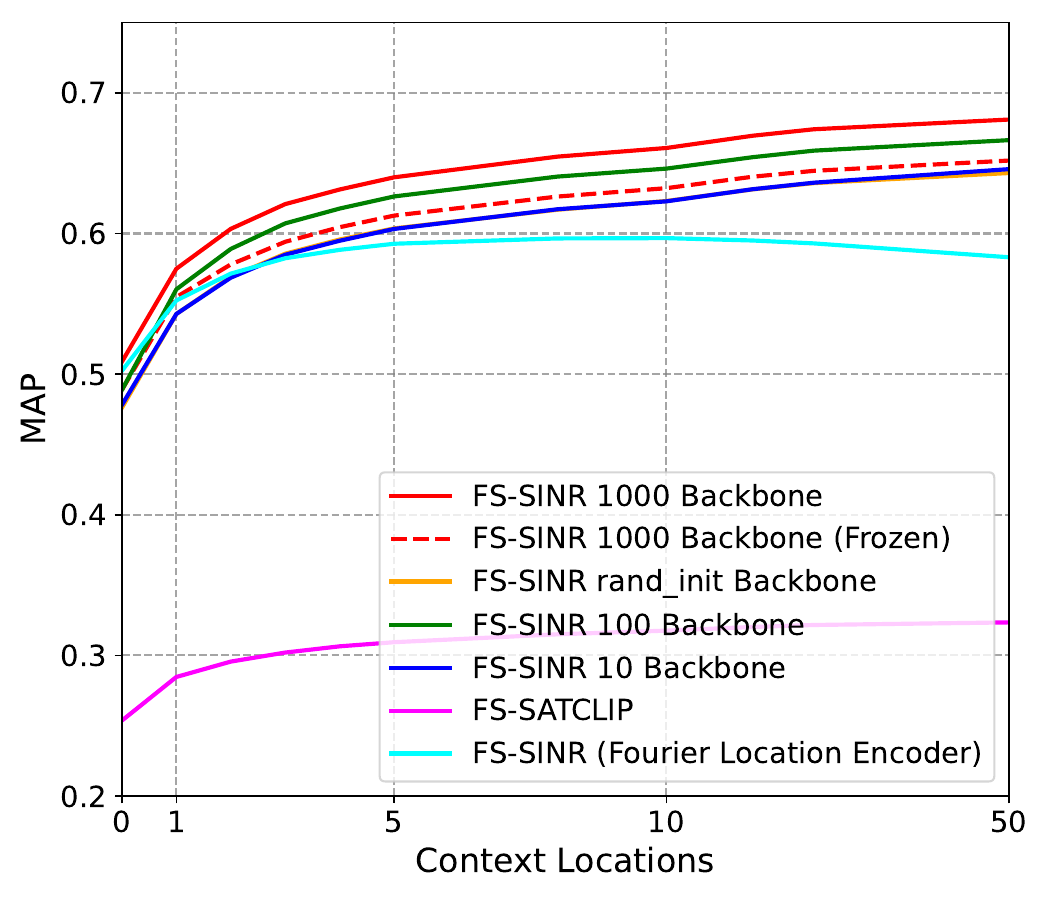}
        \vspace{-15pt}
    \caption{\change{\textbf{Impact of location encoder.} Here we evaluate the performance of FS-SINR style models with different location encoders.
    Results are shown with standard deviations from three runs (left), and without (right) for clarity. Evaluation is performed with `Range Text' on the IUCN dataset.
    `1000', `100', `10' represent the maximum number of examples per class the SINR backbone was trained on.
    `(Frozen)' indicates that the location encoder parameters were not updated during FS-SINR training.
    `FS-SATCLIP' replaces the SINR location encoder with a pretrained, frozen location encoder from \citet{klemmer2024satclipglobalgeneralpurposelocation}.
    `FS-SINR (Fourier Location Encoder)' uses the simple Fourier feature encoding \citep{tancik2020fourierfeaturesletnetworks} used in \citet{mildenhall2020nerf} to match the 256 dimensional outputs of the SINR location encoders.
    These outputs are used directly as inputs to the transformer encoder.
    After a species token is produced in this way, it is attached to a pre-trained and fine-tuned SINR backbone to produce a range.}}
    \label{fig:backbone ablation}
\end{figure}

In \cref{fig:backbone ablation}, we also investigate the impact of changing the location encoder entirely. We see that replacing our SINR location encoder with a pre-trained and frozen `SATCLIP' location encoder \citep{klemmer2024satclipglobalgeneralpurposelocation} significantly harms performance. 
This may be due to this model being frozen and trained on tasks that do not completely match ours. 
In comparison, a randomly initialized and untrained SINR backbone performs almost identically well as one that has seen a small amount of training data (10 examples per-species in the train set).
We also investigate replacing the learned location encoder $f()$ with a simple form of Fourier feature encoding~\citep{tancik2020fourierfeaturesletnetworks} to encode location inputs to the transformer $m_{\bm{\psi}}()$. 
In this setting, a pre-trained and fine-tuned SINR type location encoder $f()$ is still used to encode evaluation locations $x$ to determine the probability of presence of species $j$ via the inner product between the species embedding vector $\bm{w}_j$ and $f(x)$. 
However, $f()$ is not used to encode the context locations $\mathcal{C}^t$ before they are passed to $m_{\bm{\psi}}()$. 
Using these two different encoders performs increasingly poorly as the amount of context information increases.

\begin{figure}[h]
    \centering
        \includegraphics[width=0.42\textwidth]{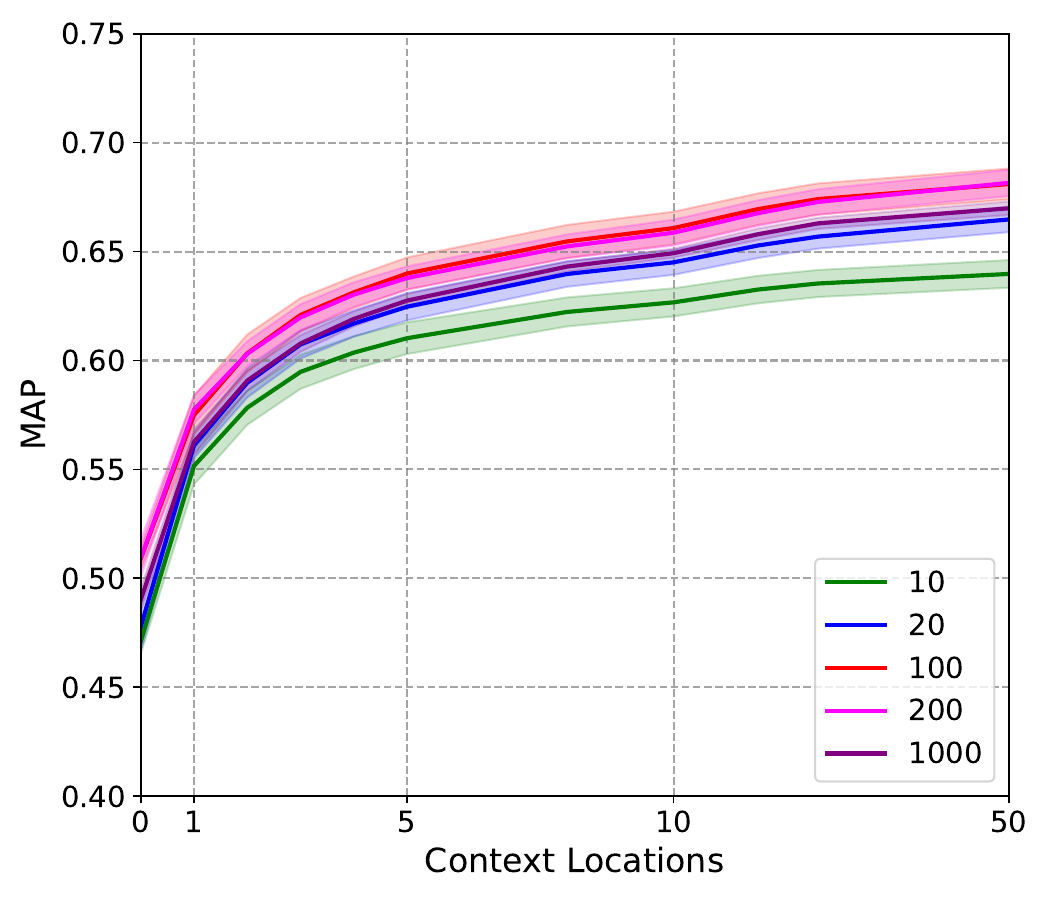}
        \hspace{15pt}
        \includegraphics[width=0.42\textwidth]{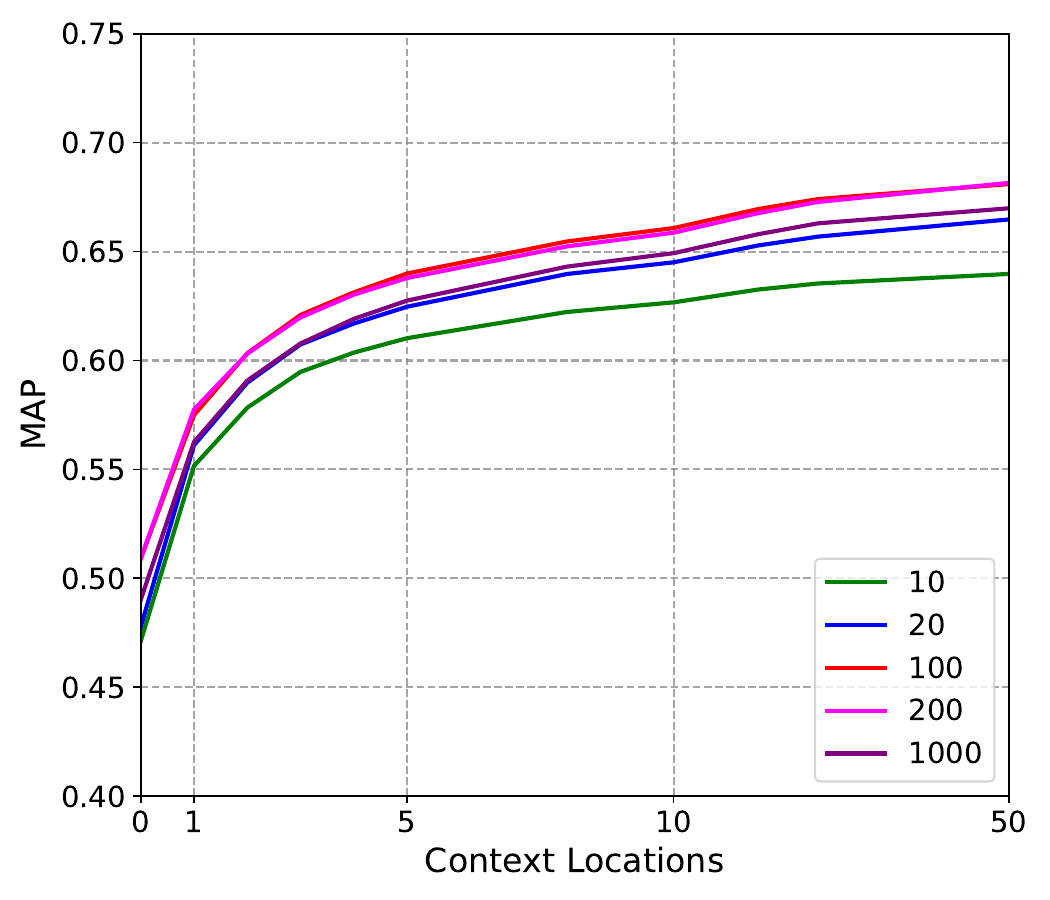}
        \vspace{-15pt}
    \caption{\change{\textbf{Impact of training data.} Here we evaluate FS-SINR models trained with different amounts of data.
    Results are shown with standard deviations from three runs (left), and without (right) for clarity.
    Evaluation is performed with `Range Text' on the IUCN dataset.
    The labels show the maximum number of examples per-species that FS-SINR is trained on.
    We see that training on an intermediate amount of training data leads to best performance.}}
    \label{fig:training_cap_ablation}
\end{figure}

\subsection{Ablating Training Data}

In \cref{fig:training_cap_ablation} we vary the number of examples per-species that are provided during training.
The impact of this is fairly small, with models trained on an intermediate amount of data performing best. 
It is worth noting, that not all species in the training dataset have as many as 1000 observations. 
We find that a model trained on only 10 examples per-species performs significantly worse.

\begin{figure}[h]
    \centering
        \includegraphics[width=0.42\textwidth]{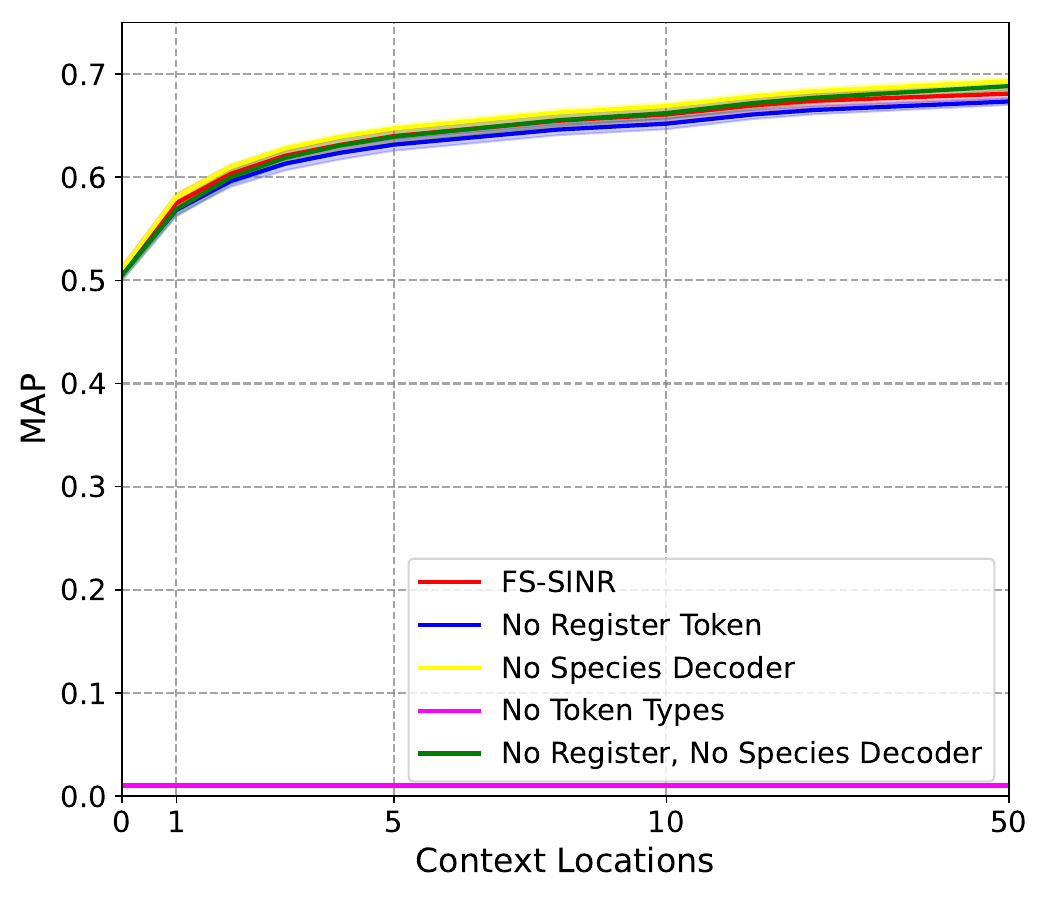}
        \hspace{15pt}
        \includegraphics[width=0.42\textwidth]{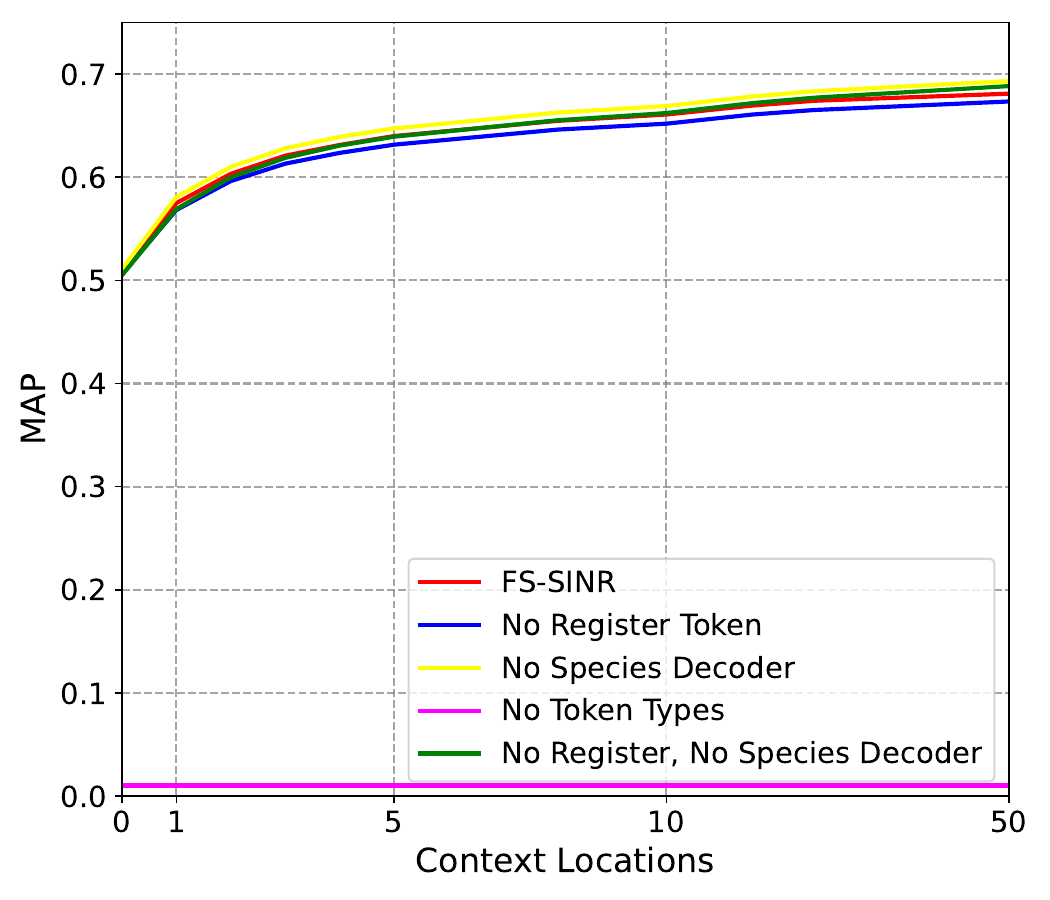} 
    \vspace{-15pt}
    \caption{\change{\textbf{Ablating model architecture components.} Here we evaluate the performance of FS-SINR style models as we ablate various design choices.
    Results are shown with standard deviations from three runs (left), and without (right) for clarity.
    Evaluation is performed with `Range Text' on the IUCN dataset.
    We see  small changes in performance when removing the register token and the species decoder.
    However removing the learned token type embeddings has a large impact.}}
    \label{fig:model ablation}
\end{figure}

\subsection{Ablating \modelname Architecture} 
\begin{figure}[h]
    \centering
        \includegraphics[width=0.42\textwidth]{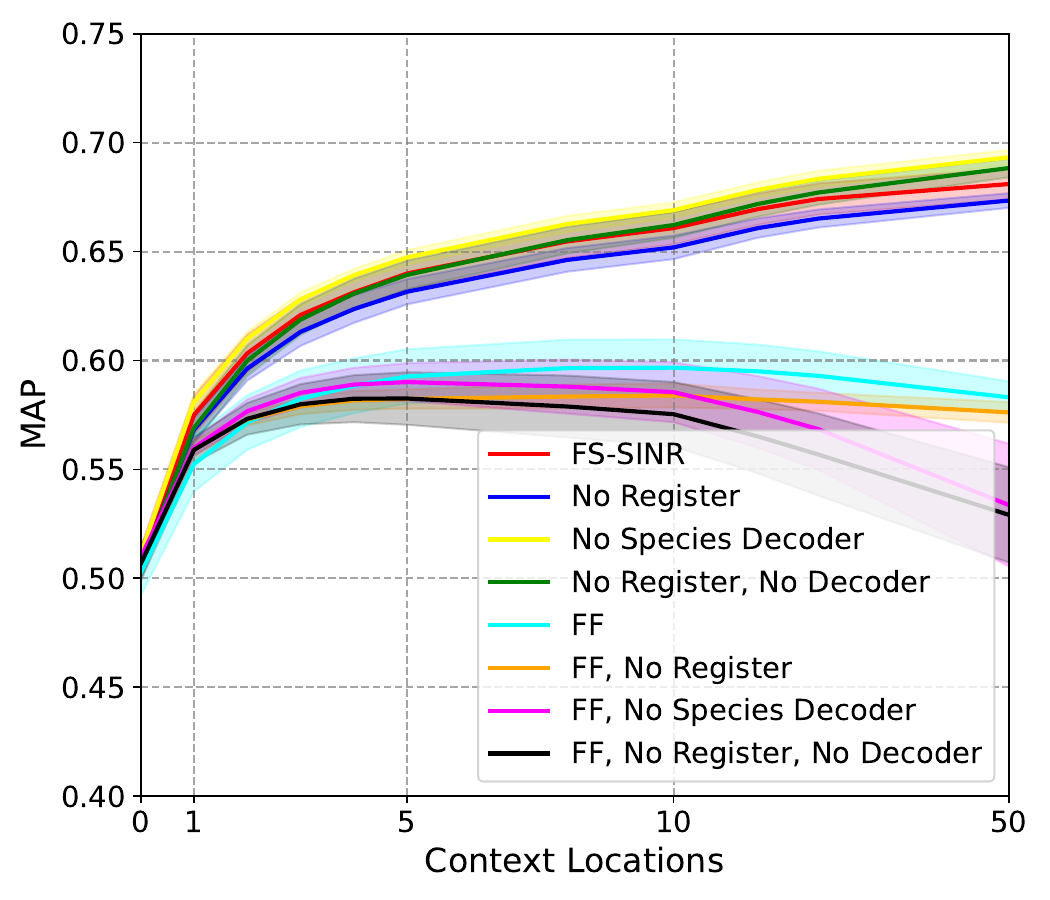}
        \hspace{15pt}
        \includegraphics[width=0.42\textwidth]{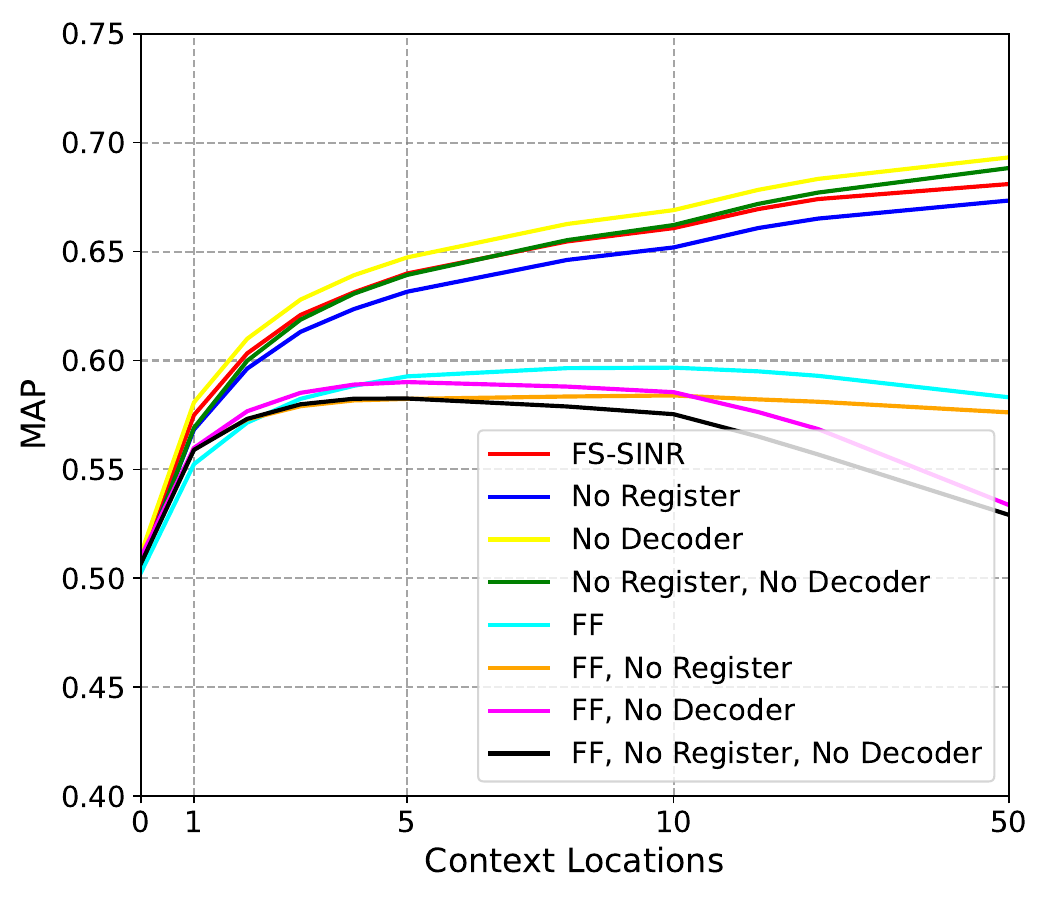} 
        \vspace{-15pt}
    \caption{\change{\textbf{Further ablating model components.} Here we evaluate the performance of FS-SINR style models as we ablate more components.
    Results are shown with standard deviations from three runs (left), and without (right) for clarity.
    Evaluation is performed with `Range Text' on the IUCN dataset. `FF' indicates that the model does not use a SINR backbone to encode location inputs to the transformer encoder. 
    Instead, a simple Fourier feature encoding \citep{tancik2020fourierfeaturesletnetworks} used in \citet{mildenhall2020nerf} is used to increase the dimensionality of location data to match the token dimension of the transformer encoder.
    These are used directly as inputs to the transformer encoder.
    After a species token is produced in this way, it is attached to a standard SINR backbone to produce a range.
    Removing the SINR backbone for encoding inputs to the transformer has a large impact on performance, especially when more context locations are supplied, and makes the model more sensitive to the impact of other ablations.}}
    \label{fig:model ablation 3}
\end{figure}

In~\cref{fig:model ablation} we vary the underlying \modelname architecture. 
Removing different components has a small effect on model performance, with the removal of the species decoder actually improving results when range text is provided.
However, as several ablations perform very similarly, it is difficult to tease out the how much of this effect is due to variance.
It is clear however that removing the learnable token type embeddings causes the model to completely fail to learn during training.
In \cref{fig:model ablation 3} we show further ablations based around removing the learned location encoder for inputs to the transformer and replacing it with the simple Fourier feature encoding also seen in \cref{fig:backbone ablation}.
When this is removed, other ablations seem to further harm performance, although results for these ablations vary significantly between runs.

\section{Additional Qualitative Results}
\label{sec:app_qual_results}

\subsection{Visualizing Non-Species Concepts} 
\begin{figure}[h]
\centering
\begin{overpic}[trim=1100 450 350 100,clip,width=0.24\textwidth]{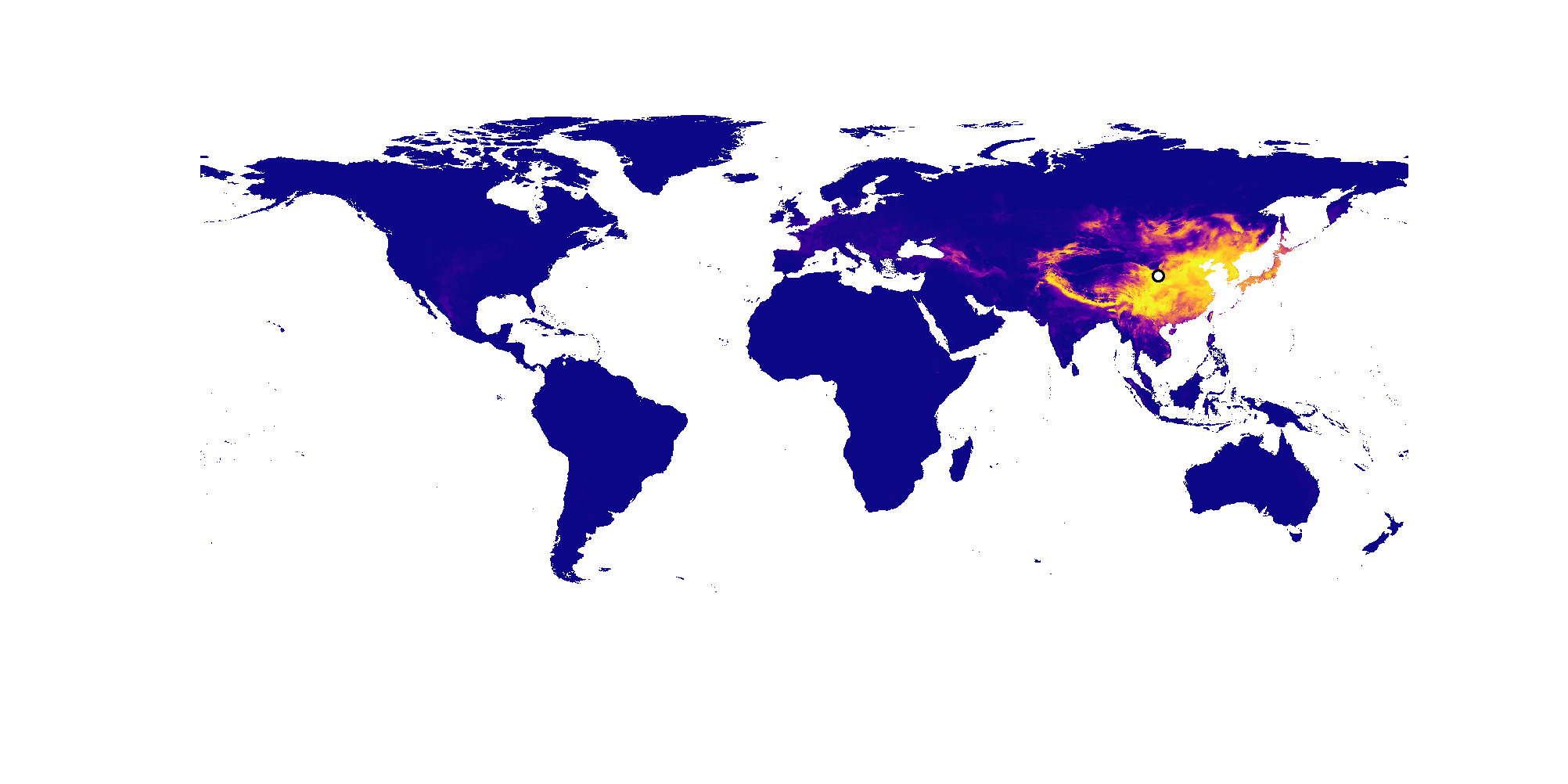}
 \put(0,4){
    \fcolorbox{black}{gray!30}{\strut\normalsize  [No Text]}
  }
\end{overpic}%
\begin{overpic}[trim=1100 450 350 100,clip,width=0.24\textwidth]{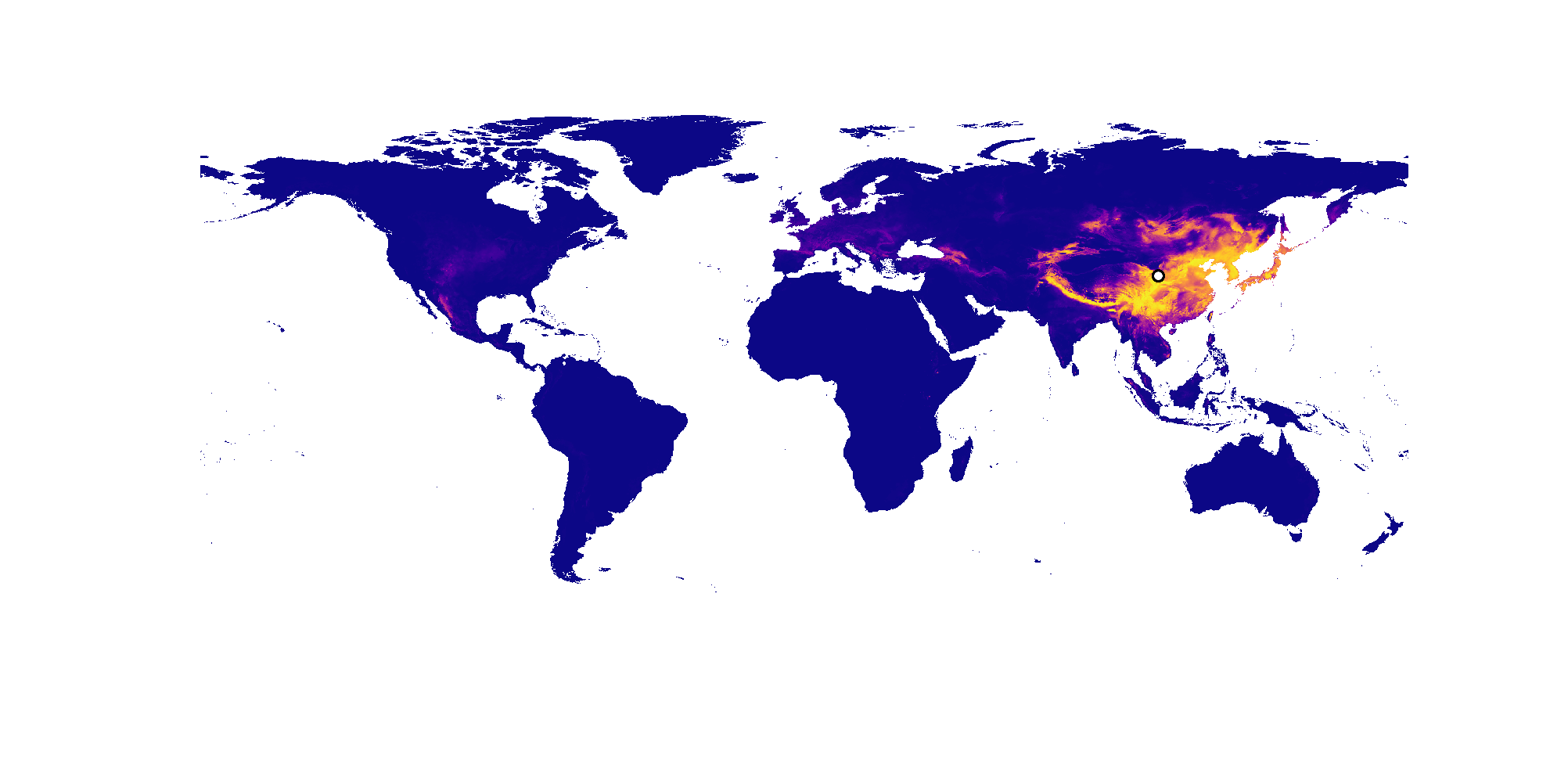}
 \put(0,4){
    \fcolorbox{black}{gray!30}{\strut\normalsize  \search ``mountain range''}
  }
\end{overpic} %
\begin{overpic}[trim=1100 450 350 100,clip,width=0.24\textwidth]{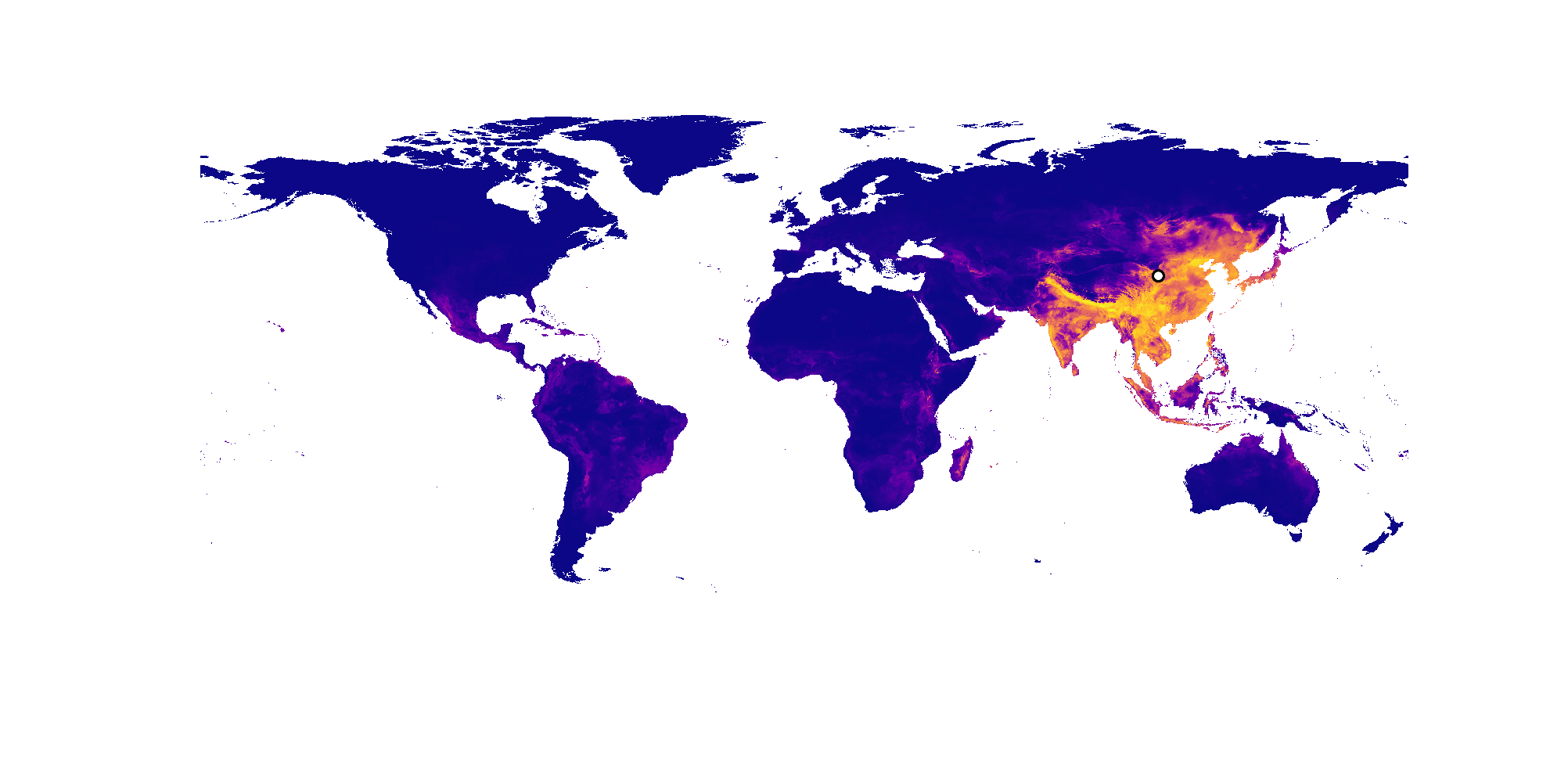}
 \put(0,4){
    \fcolorbox{black}{gray!30}{\strut\normalsize  \search ``tropical climate''}
  }
\end{overpic} %
\begin{overpic}[trim=1100 450 350 100,clip,width=0.24\textwidth]{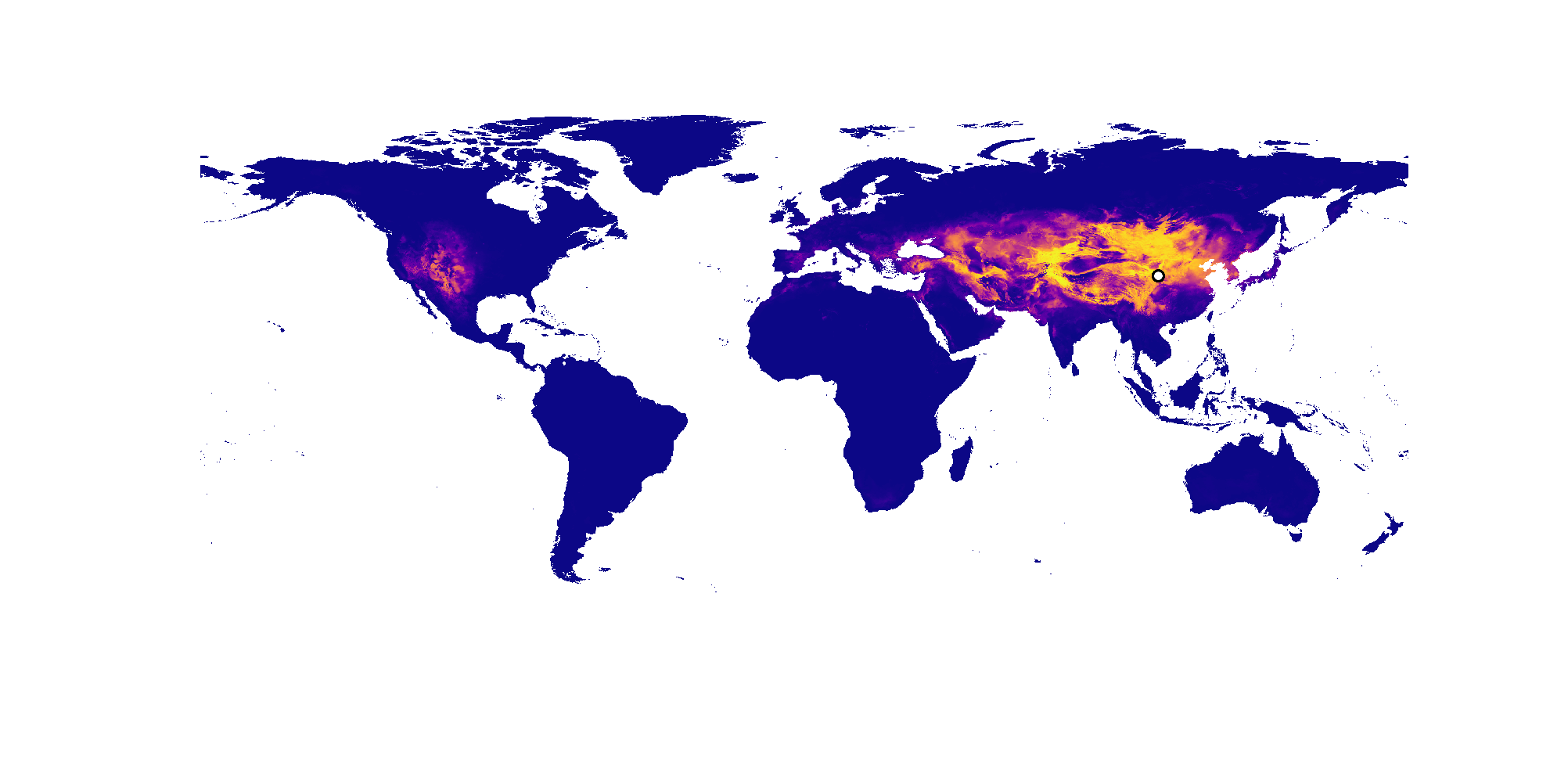}
 \put(0,4){
    \fcolorbox{black}{gray!30}{\strut\normalsize  \search ``arid desert''}
  }
\end{overpic} 
\vspace{-5pt}
\caption{ \textbf{Controlling range predictions using a single context location and text}.  
Here we show another example similar to~\cref{fig:model_vis_south_america} in the main paper. 
Given the same context location, denoted as `$\circ$', \modelname can produce significantly different range predictions depending on the text provided. 
This example illustrates a use case where a user may have limited observations but some additional knowledge regarding what type of habitat a species of interest could be found in. 
}

\vspace{-10pt}
\label{fig:model_vis_asia}
\end{figure}

\begin{figure}[h]
\centering

\begin{overpic}[trim=250 90 200 150,clip,width=0.49\textwidth]{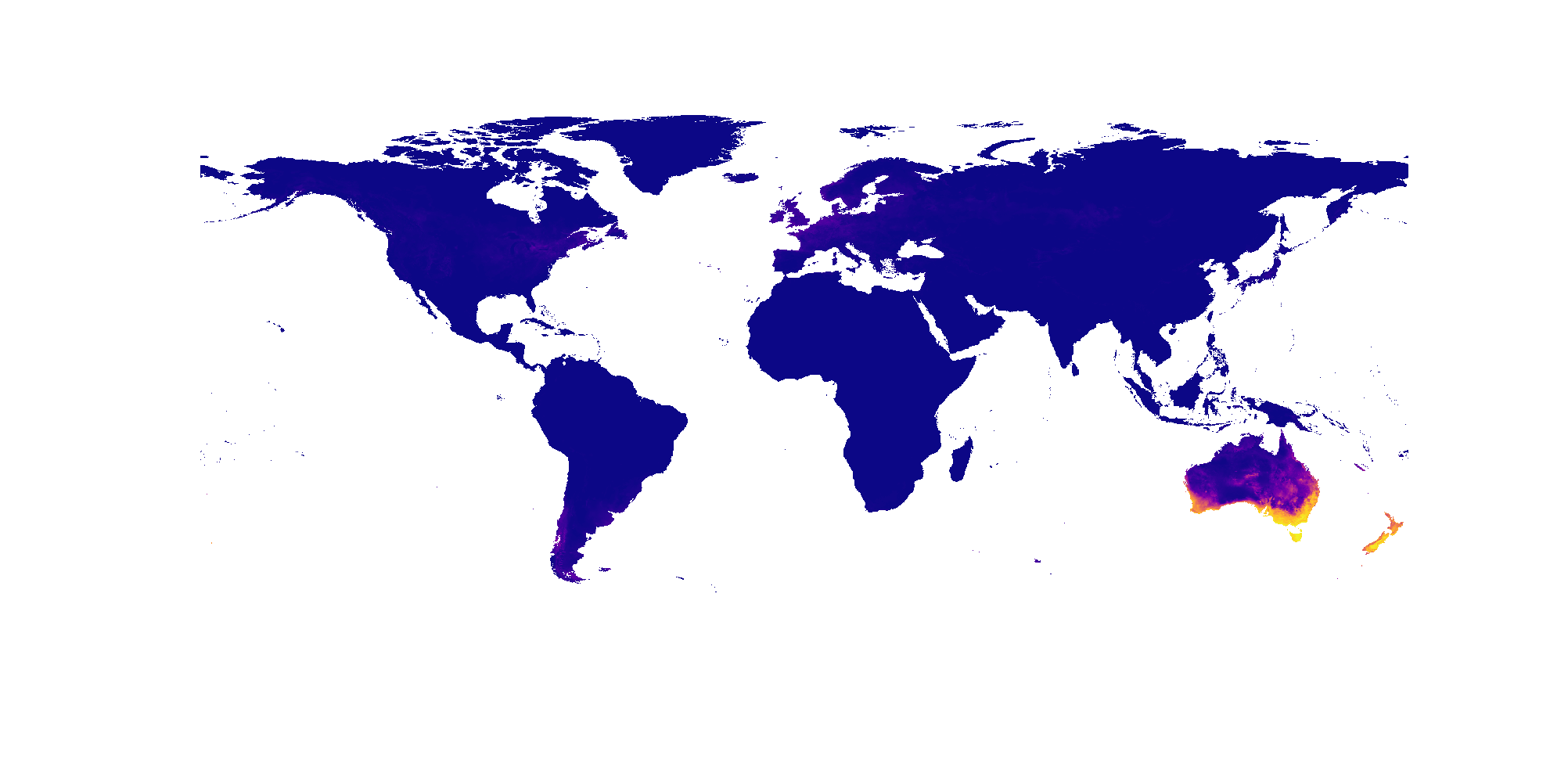}
 \put(0,4){
    \fcolorbox{black}{gray!30}{\strut\normalsize  \search ``Tasmania''}
  }
\end{overpic}\hspace{5pt}
\begin{overpic}[trim=250 90 200 150,clip,width=0.49\textwidth]{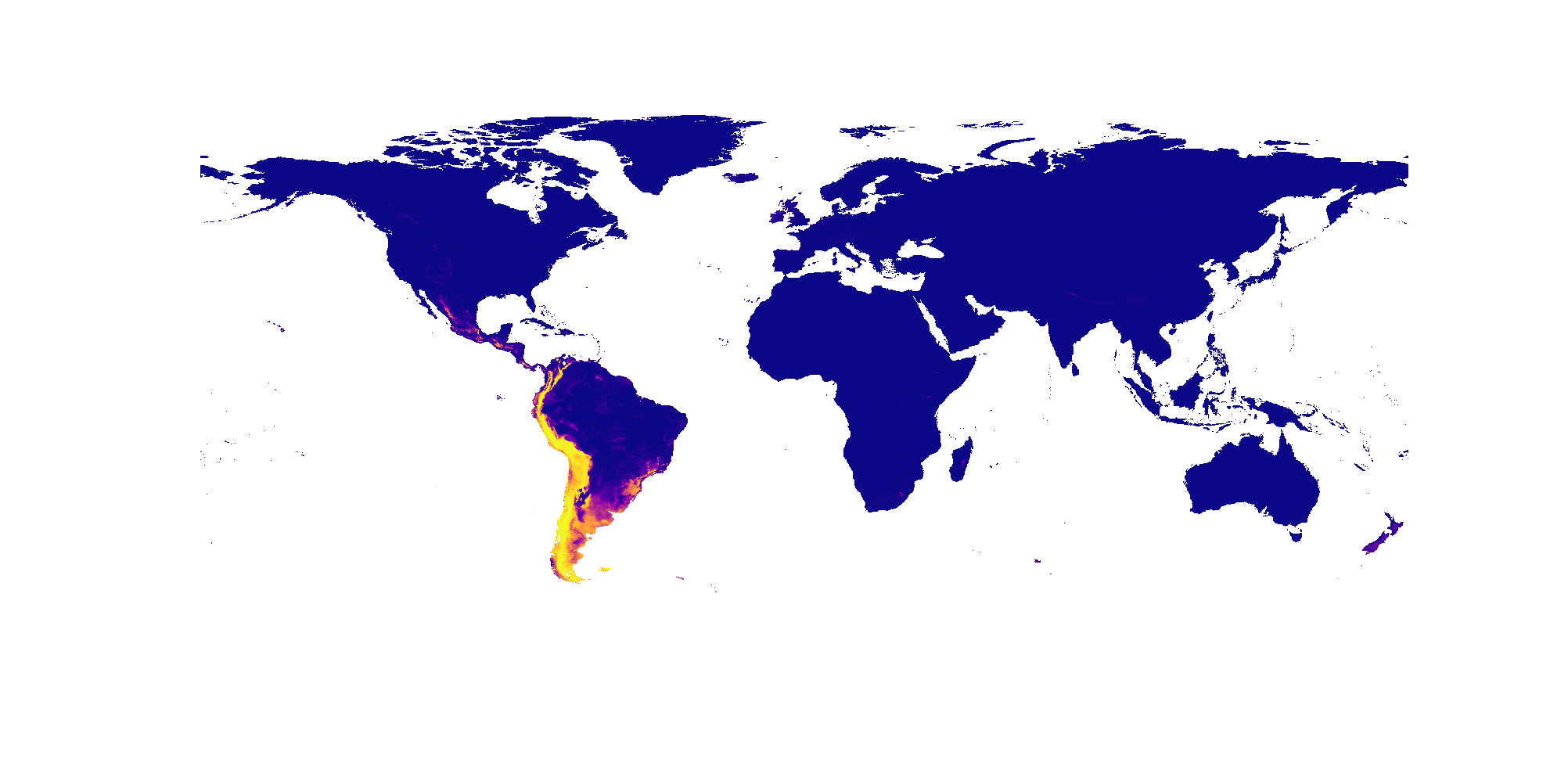}
 \put(0,4){
    \fcolorbox{black}{gray!30}{\strut\normalsize  \search ``The Andes''}
  }
\end{overpic} \\
\begin{overpic}[trim=250 90 200 150,clip,width=0.49\textwidth]{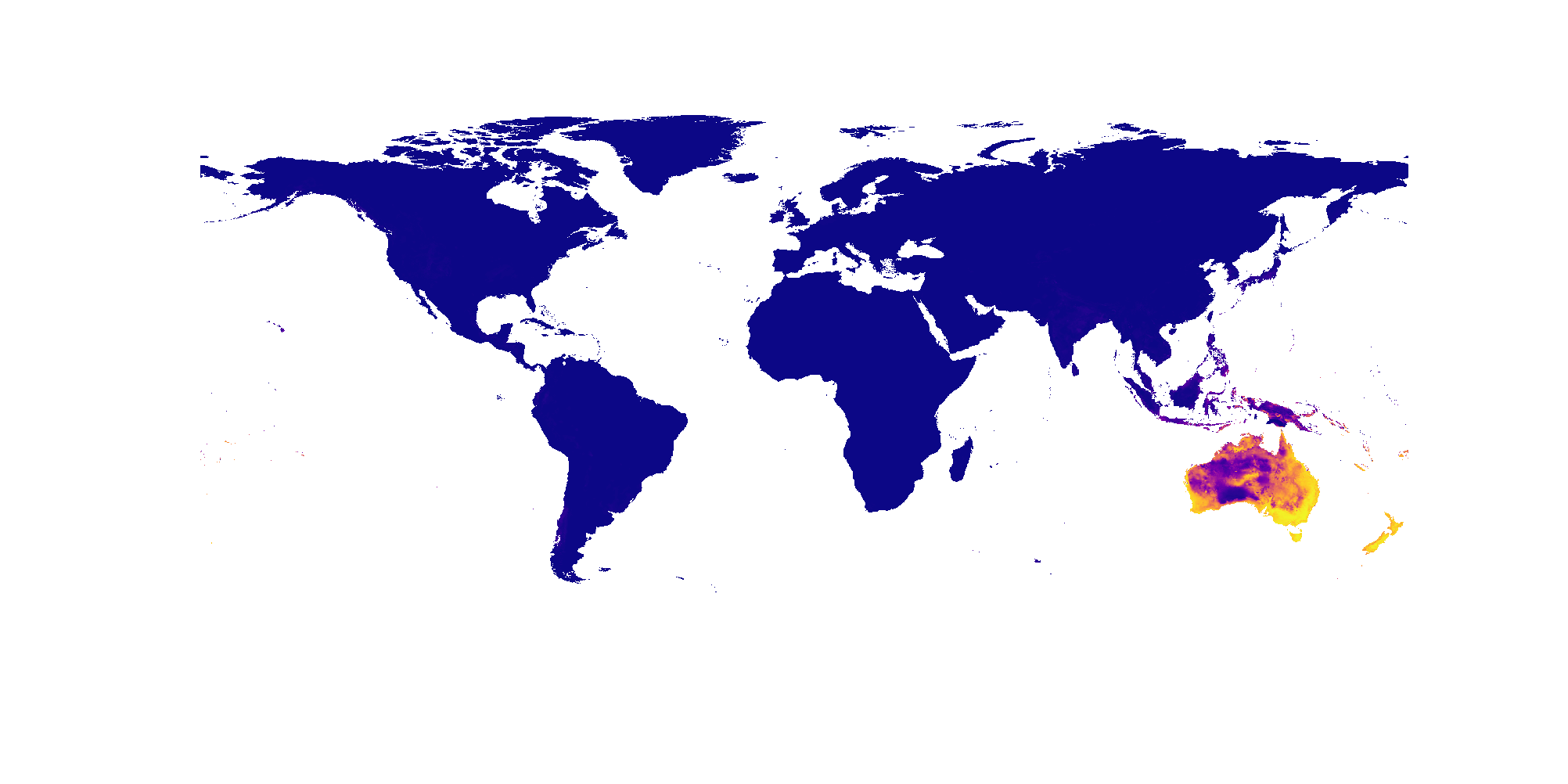}
 \put(0,4){
    \fcolorbox{black}{gray!30}{\strut\normalsize  \search ``Lord of The Rings''}
  }
\end{overpic}\hspace{5pt}
\begin{overpic}[trim=250 90 200 150,clip,width=0.49\textwidth]{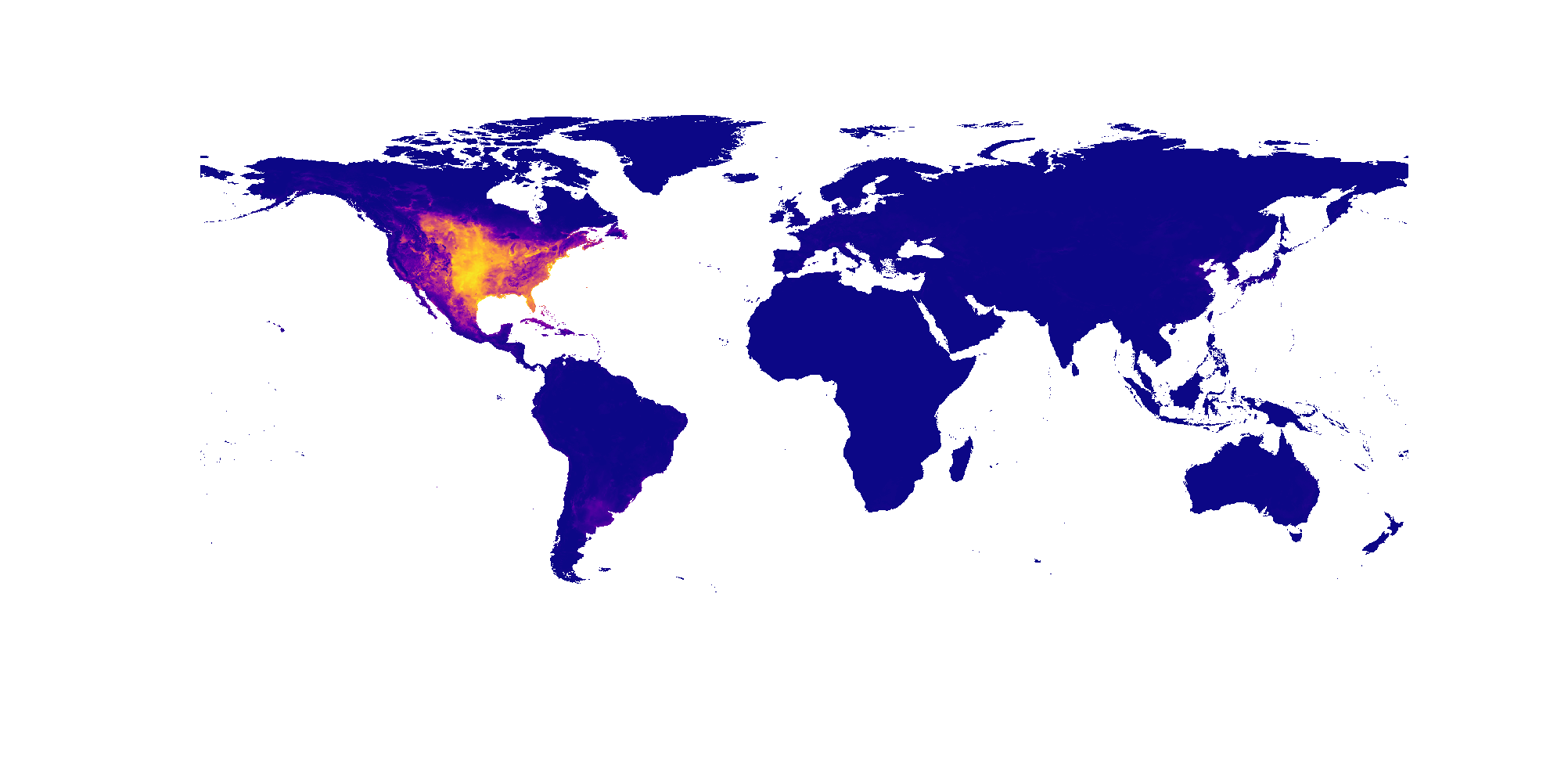}
 \put(0,4){
    \fcolorbox{black}{gray!30}{\strut\normalsize  \search ``Baseball''}
  }
\end{overpic} \\
\begin{overpic}[trim=250 90 200 150,clip,width=0.49\textwidth]{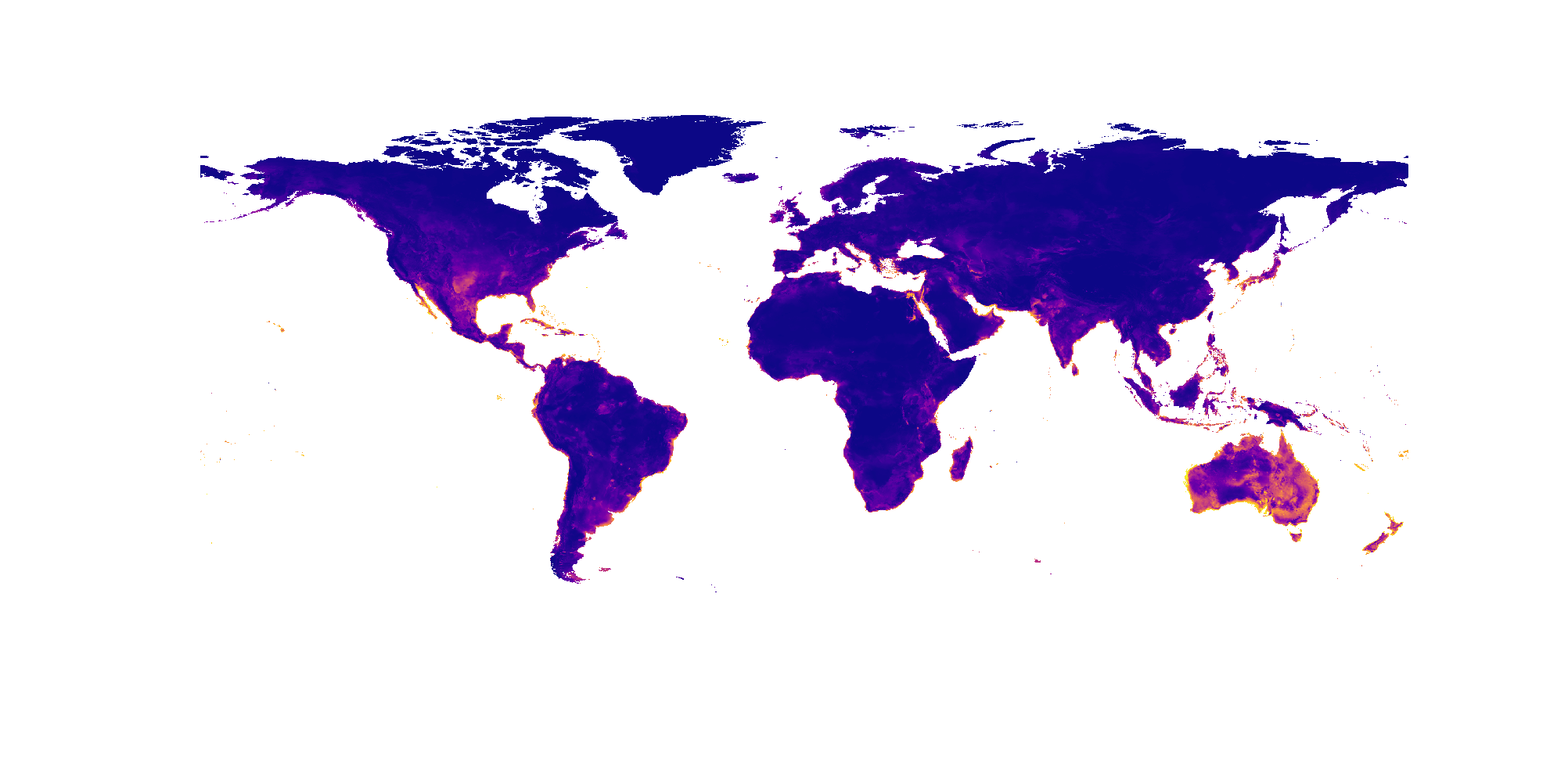}
 \put(0,4){
    \fcolorbox{black}{gray!30}{\strut\normalsize  \search ``Pirates''}
  }
\end{overpic}\hspace{5pt}
\begin{overpic}[trim=250 90 200 150,clip,width=0.49\textwidth]{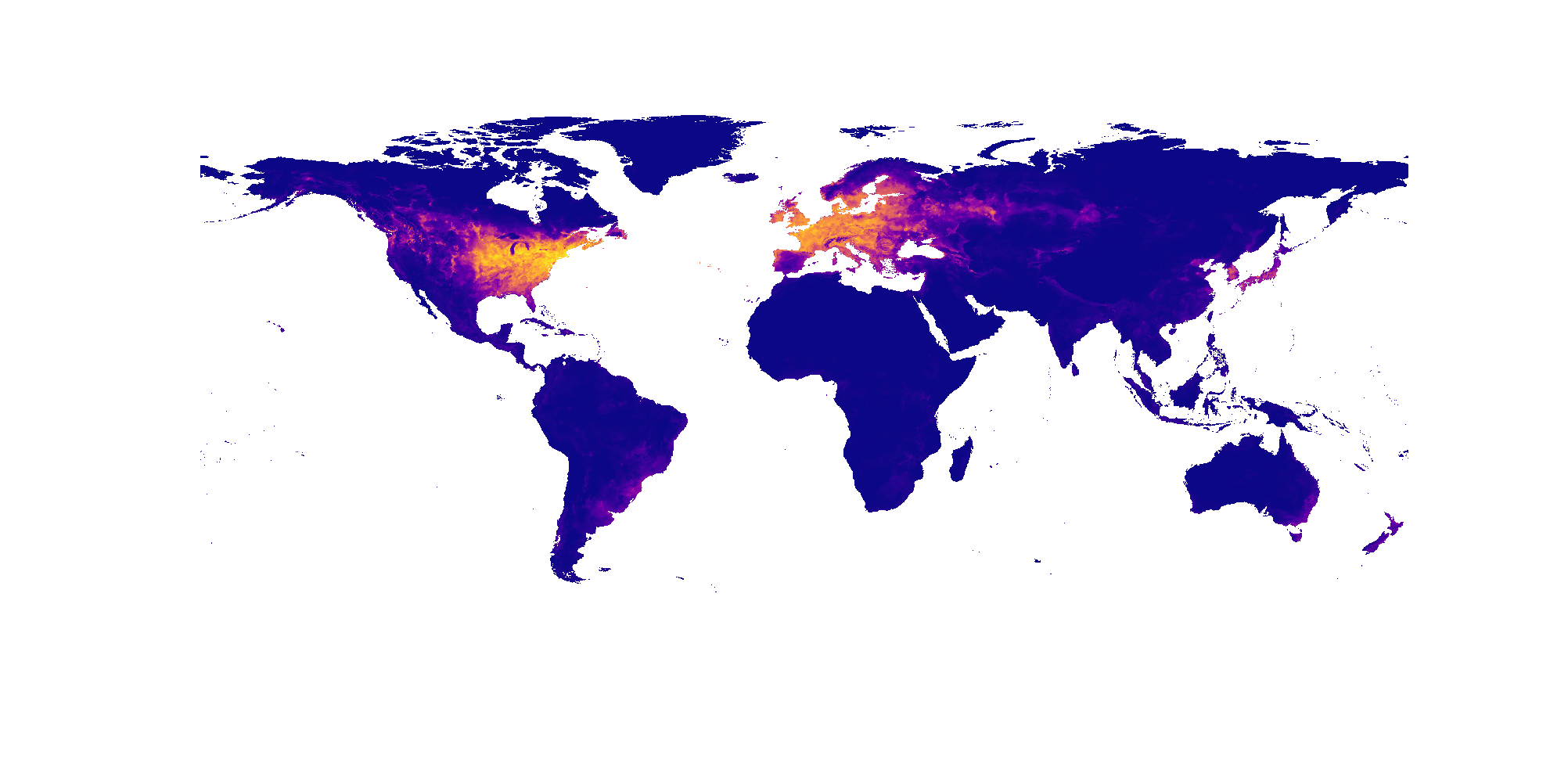}
 \put(0,4){
    \fcolorbox{black}{gray!30}{\strut\normalsize  \search ``Oktoberfest''}
  }
\end{overpic} 
\vspace{-5pt}
\caption{ \textbf{Zero-shot non-species concepts}. 
We can evaluate \modelname in a zero-shot manner using only text information, \ie without any locations.  
Here, we observe that \modelname, like LE-SINR~\citep{hamilton2024}, can localize abstract (\ie non-species) concepts in geographic space, despite never being trained to explicitly do so. 
The model achieves this as it learns to make connections between species text and information already contained in the pretrained language encoder we use. 
However, we do note failure/ambiguous cases such as the ``Pirate'' example in the bottom row.  
}
\vspace{-10pt}
\label{fig:model_vis_zeroshot}
\end{figure}

\begin{figure}[h]
\centering
  \centering
    \rotatebox{90}{\hspace{-15pt}\parbox{20mm}{\small\centering 0 context}}
    \begin{minipage}{0.45\textwidth}
        \includegraphics[trim=250 90 200 150,clip,width=\linewidth]{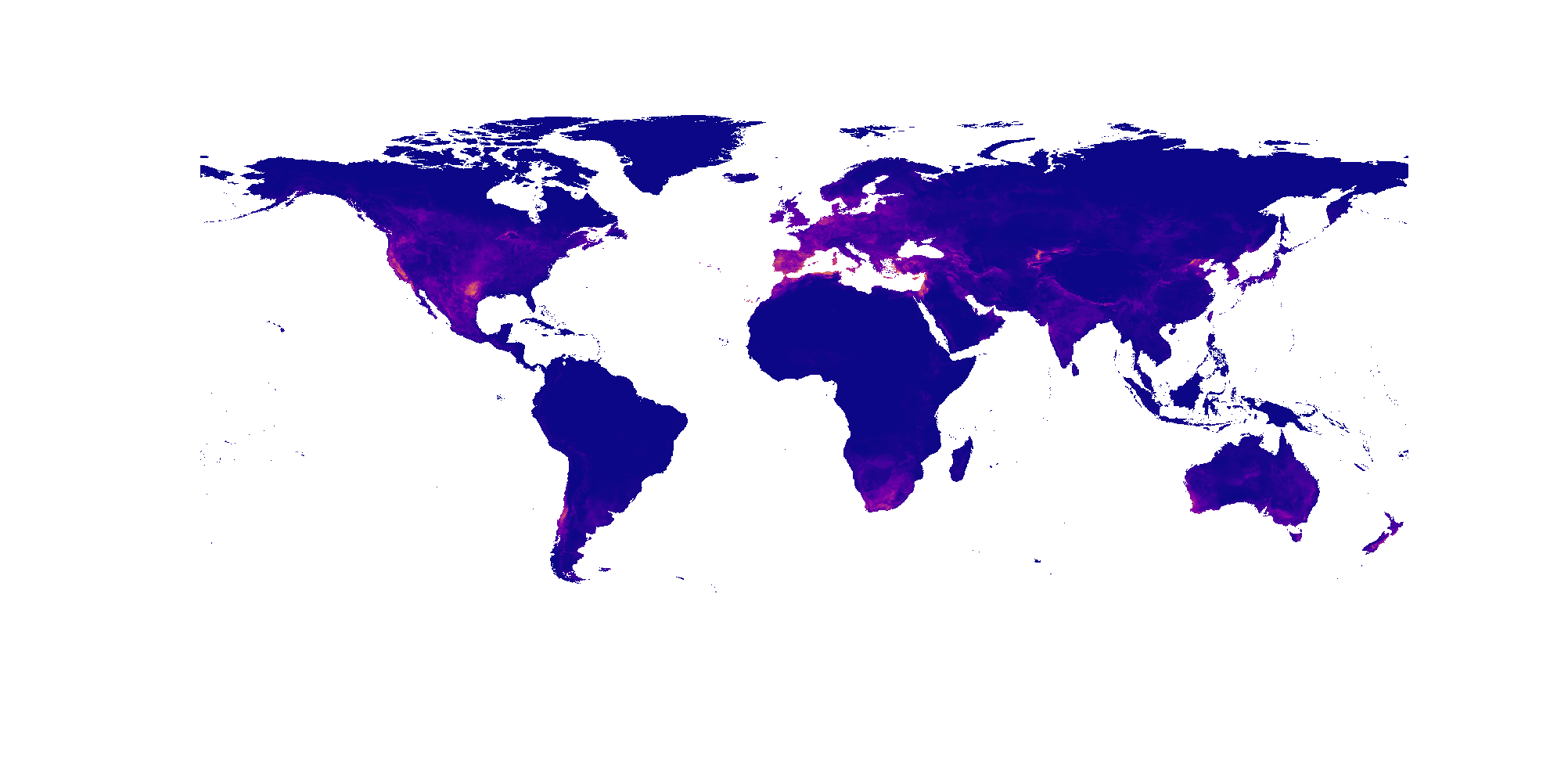} 
    \end{minipage}
    \begin{minipage}{0.45\textwidth}
        \includegraphics[trim=250 90 200 150,clip,width=\linewidth]{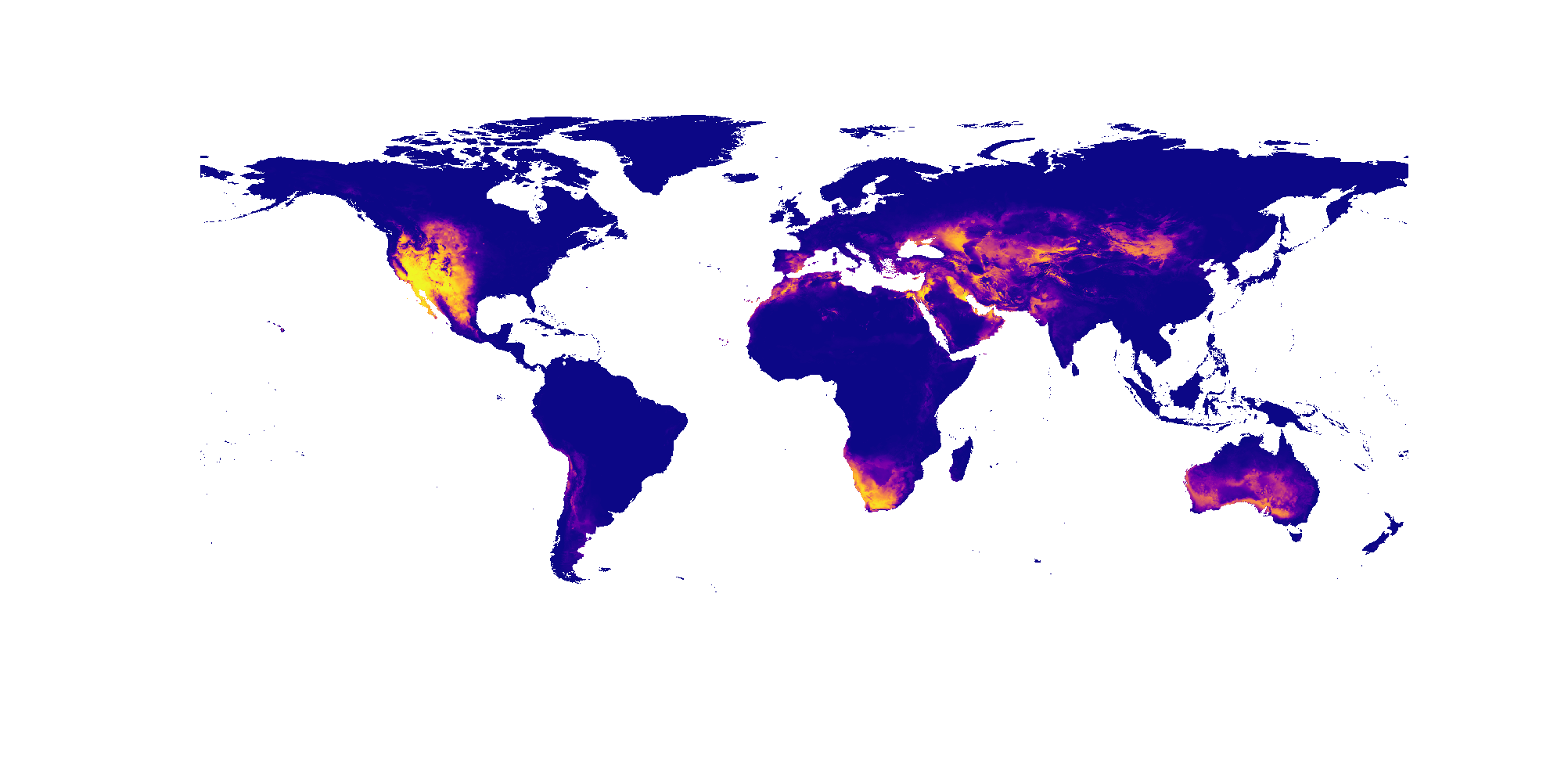} 
    \end{minipage}
    \vspace{-10pt}

    \rotatebox{90}{\hspace{-15pt}\parbox{20mm}{\small\centering 1 context}}
    \begin{minipage}{0.45\textwidth}
        \includegraphics[trim=250 90 200 150,clip,width=\linewidth]{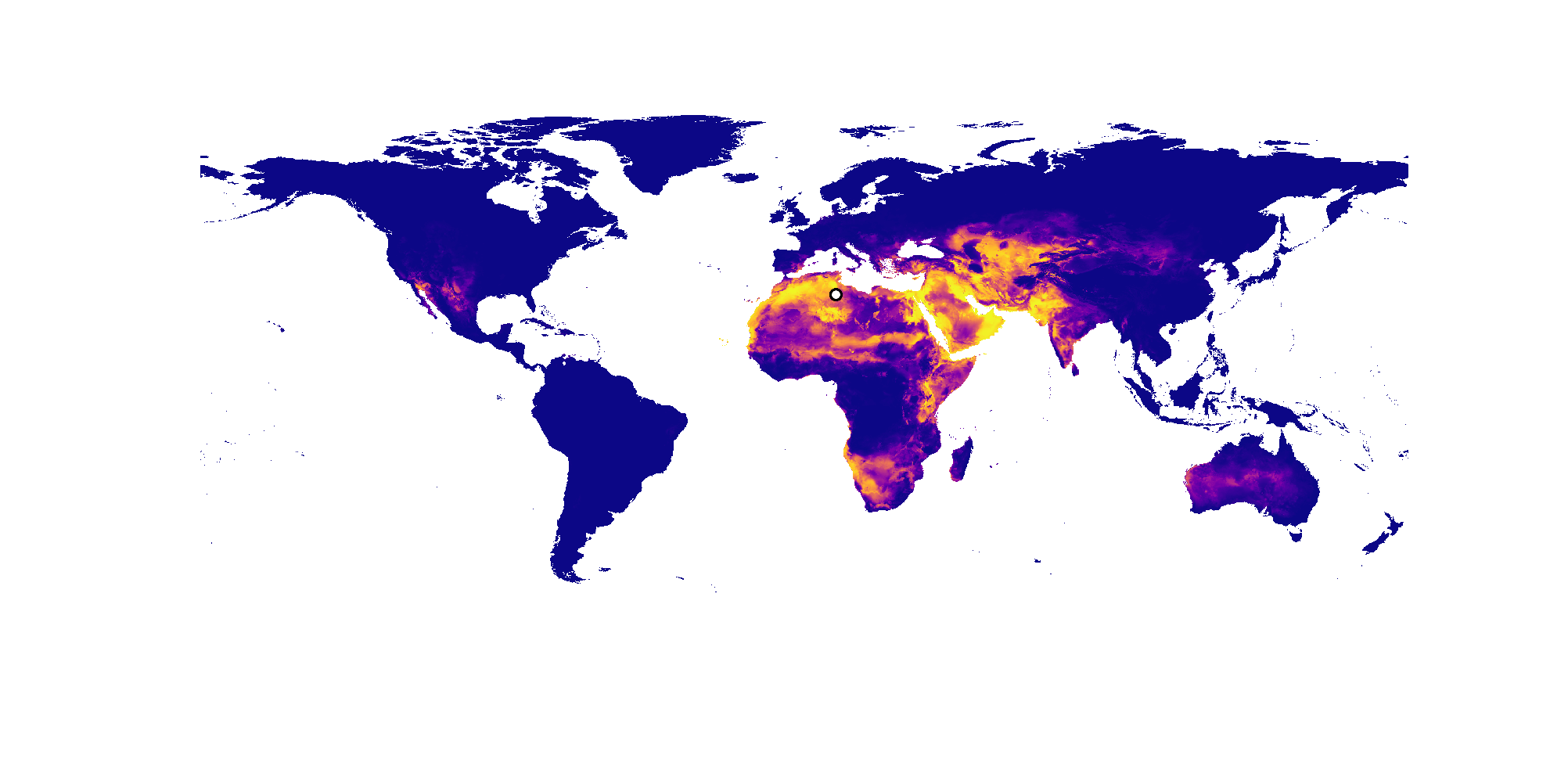} 
    \end{minipage}
    \begin{minipage}{0.45\textwidth}
        \includegraphics[trim=250 90 200 150,clip,width=\linewidth]{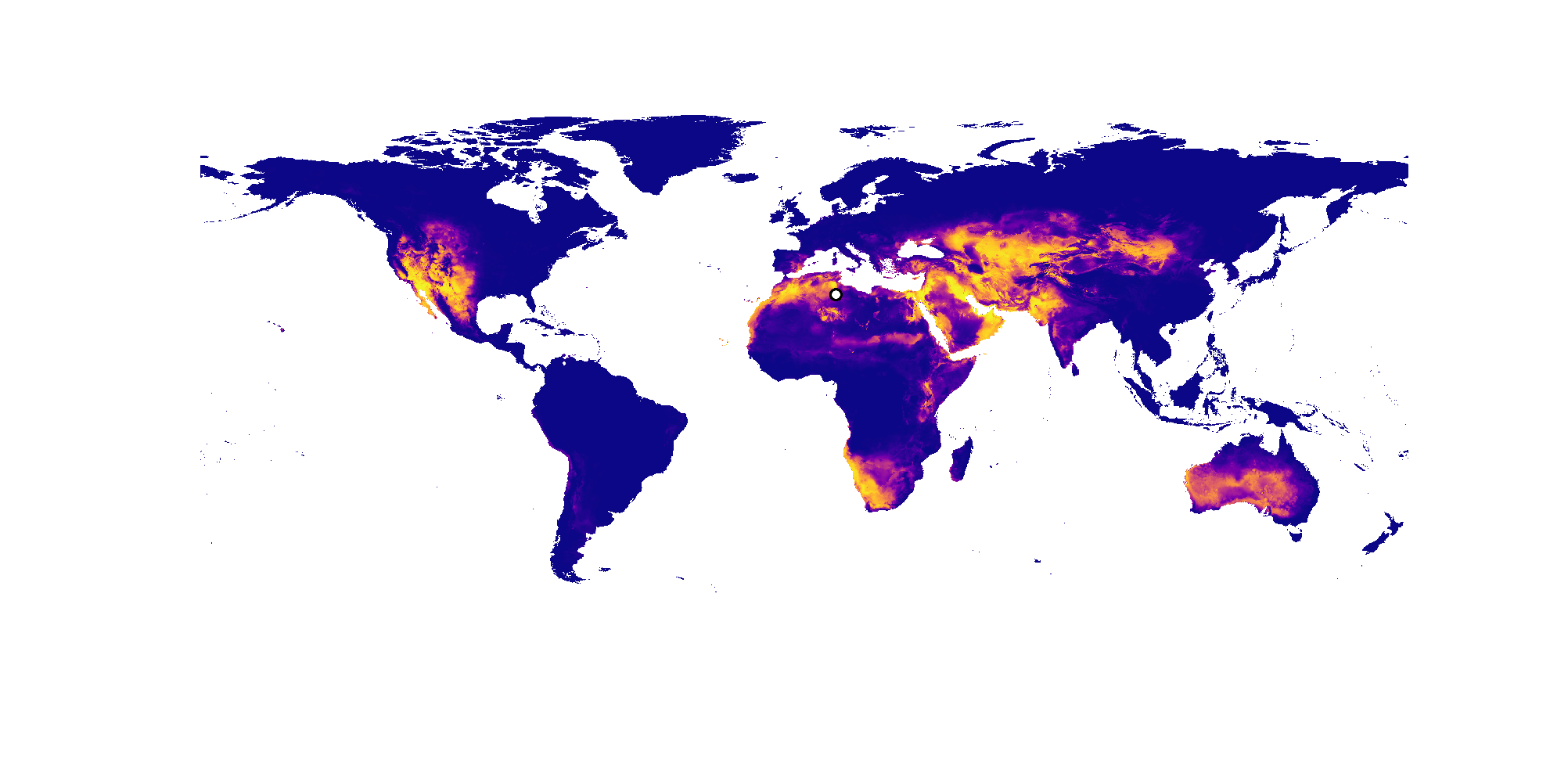}
    \end{minipage}
    \vspace{-10pt}
    
    \rotatebox{90}{\hspace{-15pt}\parbox{20mm}{\small\centering 2 context}}
    \begin{minipage}{0.45\textwidth}
        \includegraphics[trim=250 90 200 150,clip,width=\linewidth]{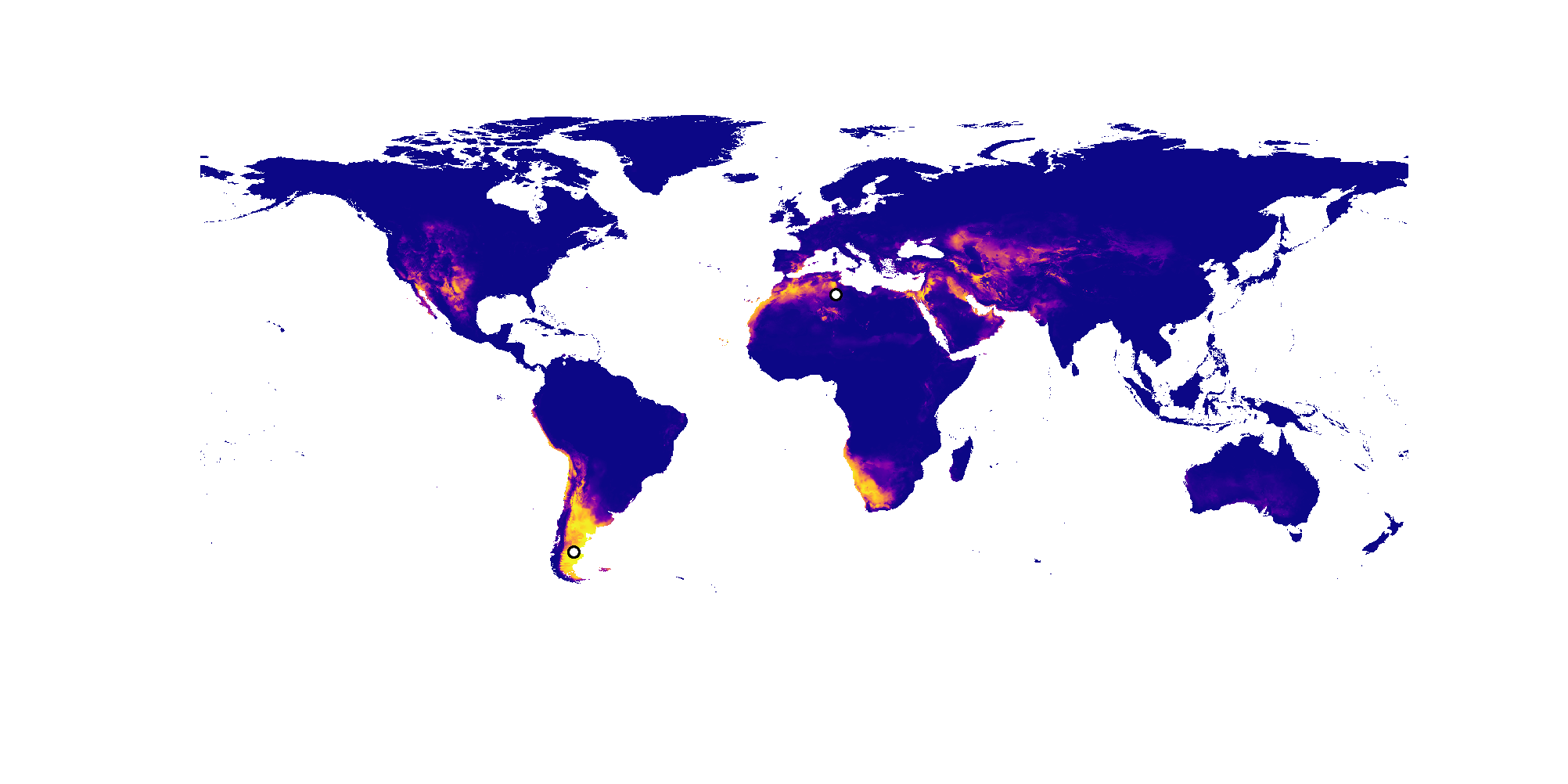} 
    \end{minipage}
    \begin{minipage}{0.45\textwidth}
        \includegraphics[trim=250 90 200 150,clip,width=\linewidth]{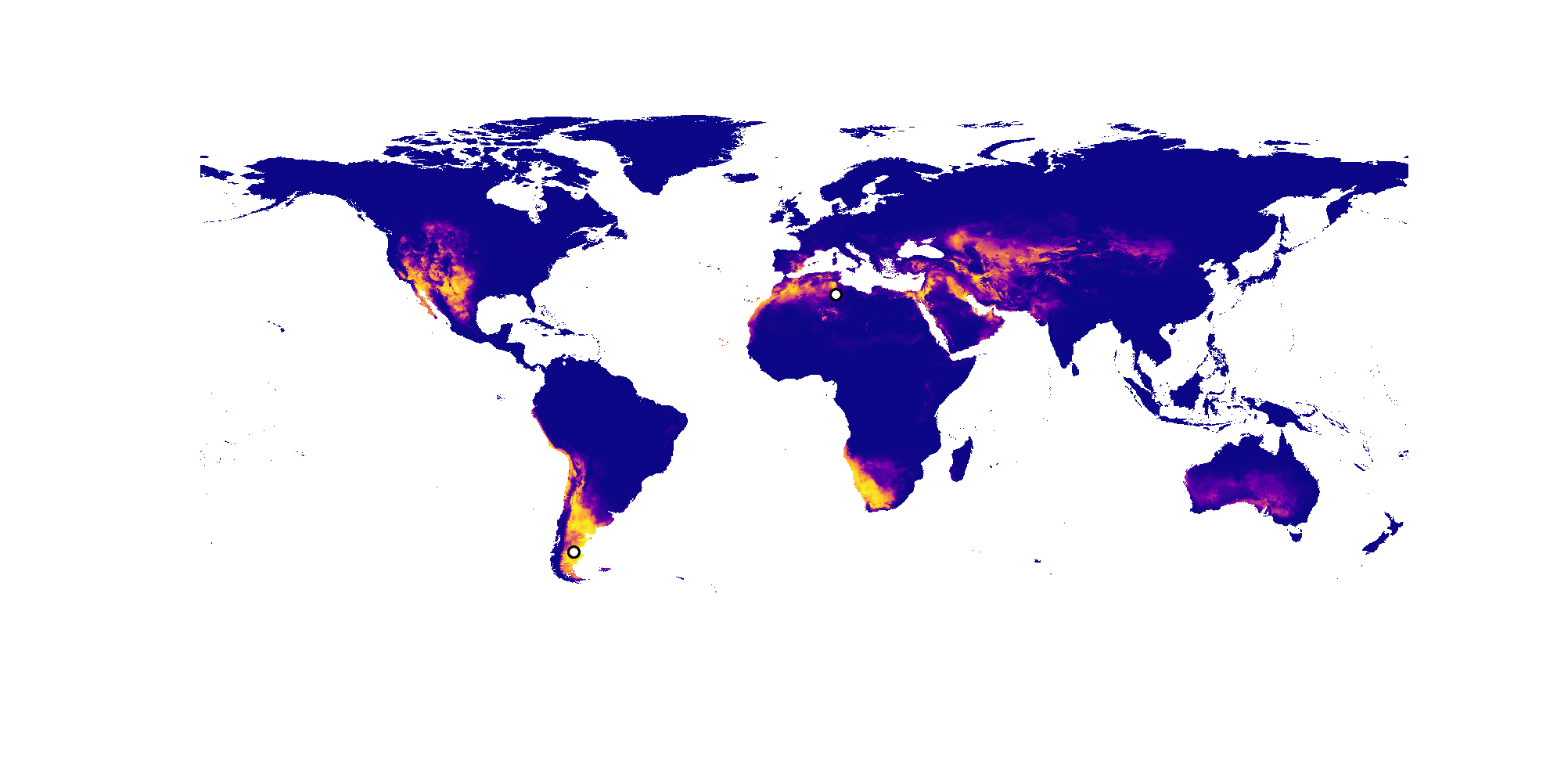} 
    \end{minipage}
    \vspace{-10pt}

    \rotatebox{90}{\hspace{-15pt}\parbox{20mm}{\small\centering 3 context}}
    \begin{minipage}{0.45\textwidth}
        \includegraphics[trim=250 200 200 150,clip,width=\linewidth]{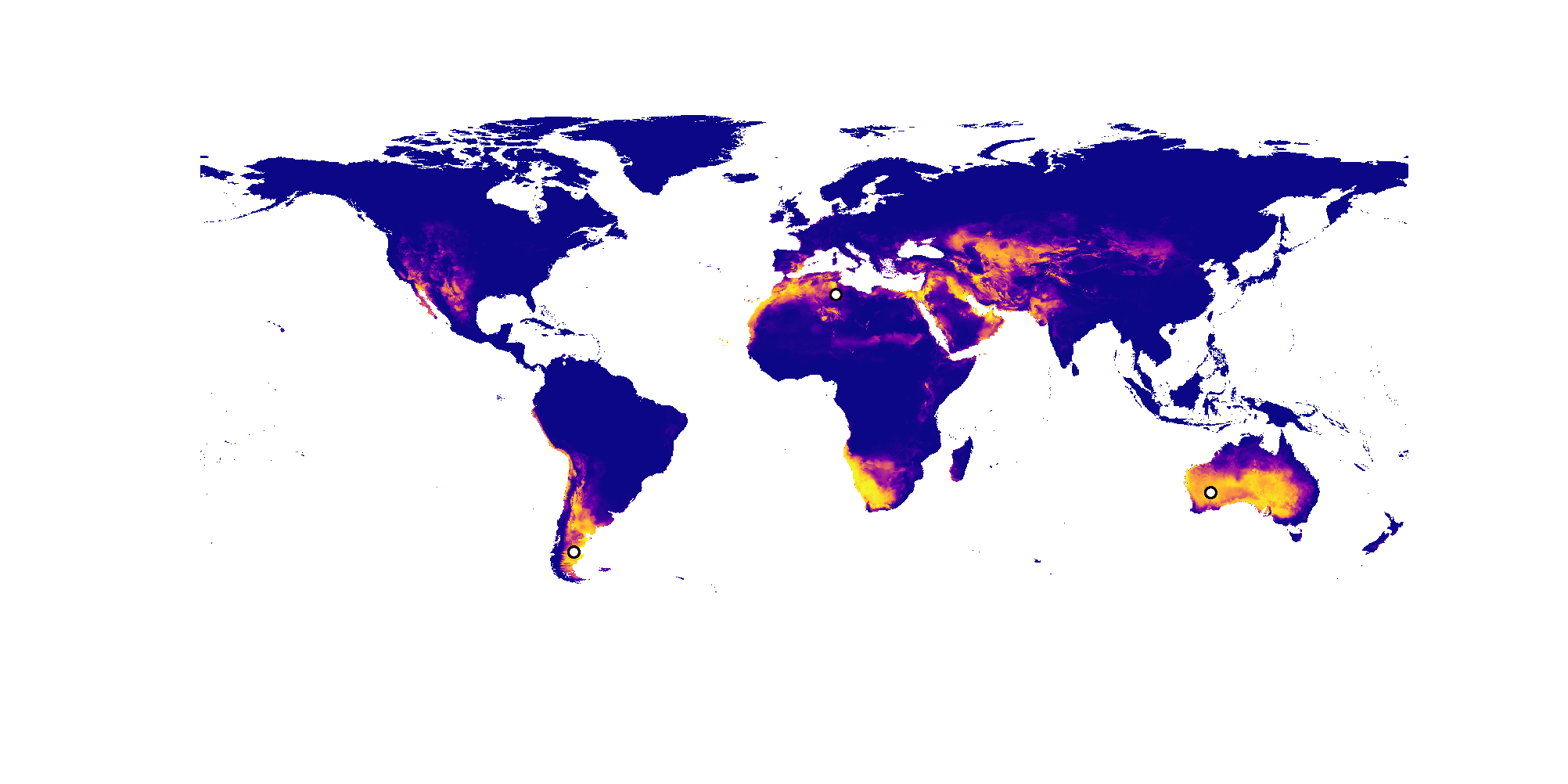} 
        \centering{\fcolorbox{black}{gray!30}{\strut\normalsize [No Text]}}
    \end{minipage}
    \vspace{-10pt}
    \begin{minipage}{0.45\textwidth}
        \includegraphics[trim=250 200 200 150,clip,width=\linewidth]{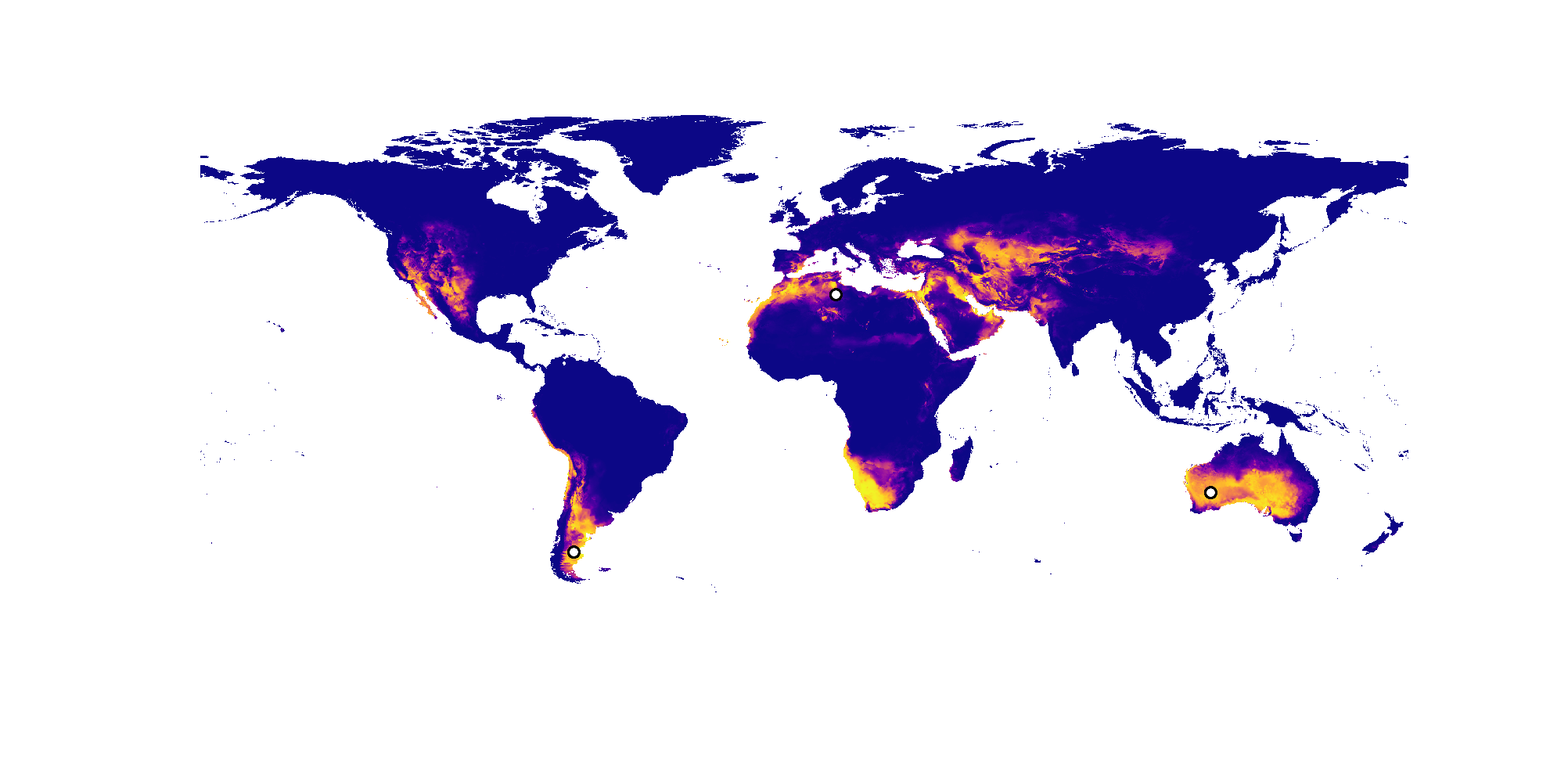} 
        \centering{\fcolorbox{black}{gray!30}{\strut\normalsize  \search ``Desert''}}
    \end{minipage}
    
    \vspace{5pt}
    \caption{ \textbf{Varying the context information provided}. 
    Here we change the context information provided to \modelname.
    The model on the left column receives no text input, but the one on the right gets the text ``Desert''. 
    Additionally, in each row we increase the number of context locations provided, from zero to three, denoted as `$\circ$'. 
    We observe that the model on the right that uses text already has a strong prior about the species being present at desert-like locations, \eg see first row where no context locations are provided. 
    As soon as one context location is added in North Africa (second row), the model generates a new prediction with an increased  probability that the species is present there.  
}
\label{fig:text_model_vis_vary_context}
\end{figure}

By jointly training on text and locations, \modelname is able to spatially ground abstract non-species concepts in a zero-shot manner, as is done with LE-SINR in~\citet{hamilton2024}.
In~\cref{fig:model_vis_asia} we provide another example similar to~\cref{fig:model_vis_south_america} in the main paper. 
Here, we again fix the context location and show the impact of changing the text. 
We can see that different text prompts can result in quite different predicted ranges.
In~\cref{fig:model_vis_zeroshot} we see examples where different text concepts, which are very different from the species-based text provided during training, are grounded in sensible locations on the map. 
In~\cref{fig:text_model_vis_vary_context} we compare predictions made with increasing numbers of context locations in desert regions, with or without the accompanying text prompt ``Desert''.
As we increase the number of context locations, the two different models converge to more similar range predictions.

\subsection{Visualizing Estimated Species Ranges} 

\begin{figure}[h]
    \centering
    \renewcommand{\arraystretch}{1.2} %

    \begin{tabular}{cc}
        {\bf European Robin - Range Text} & {\bf European Robin - Habitat Text} \\

        \begin{overpic}[trim={0 1cm 0 0},clip,width=0.48\textwidth]{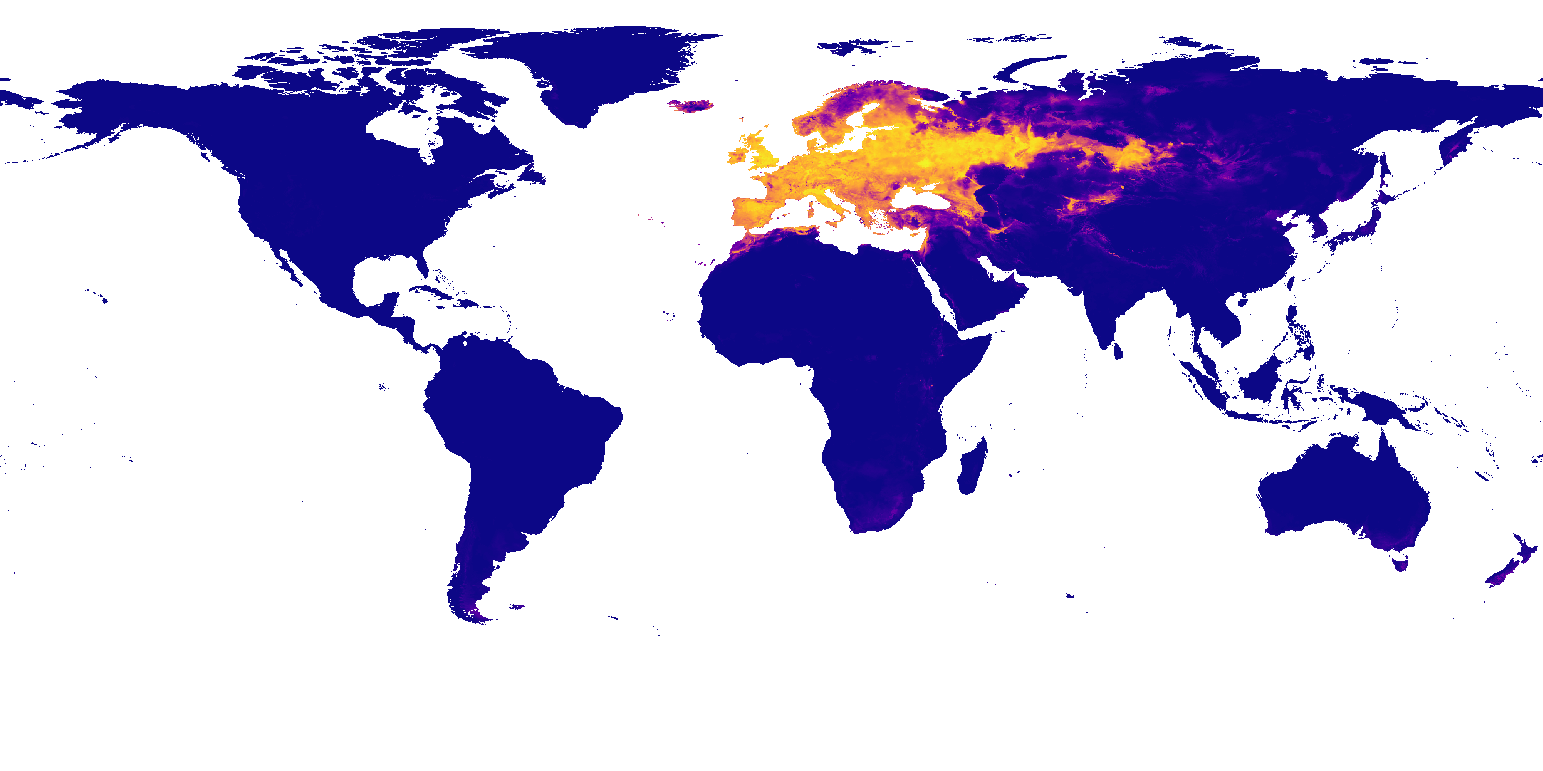}
            \put(0,0){%
              {%
                \setlength\fboxsep{0pt}%
                \setlength\fboxrule{1pt}%
                \fbox{%
                  \includegraphics[width=0.125\textwidth]{figs/range_est/european_robin_crop.png}%
                }%
              }%
            }
        \end{overpic}
        &
        \begin{overpic}[trim={0 1cm 0 0},clip,width=0.48\textwidth]{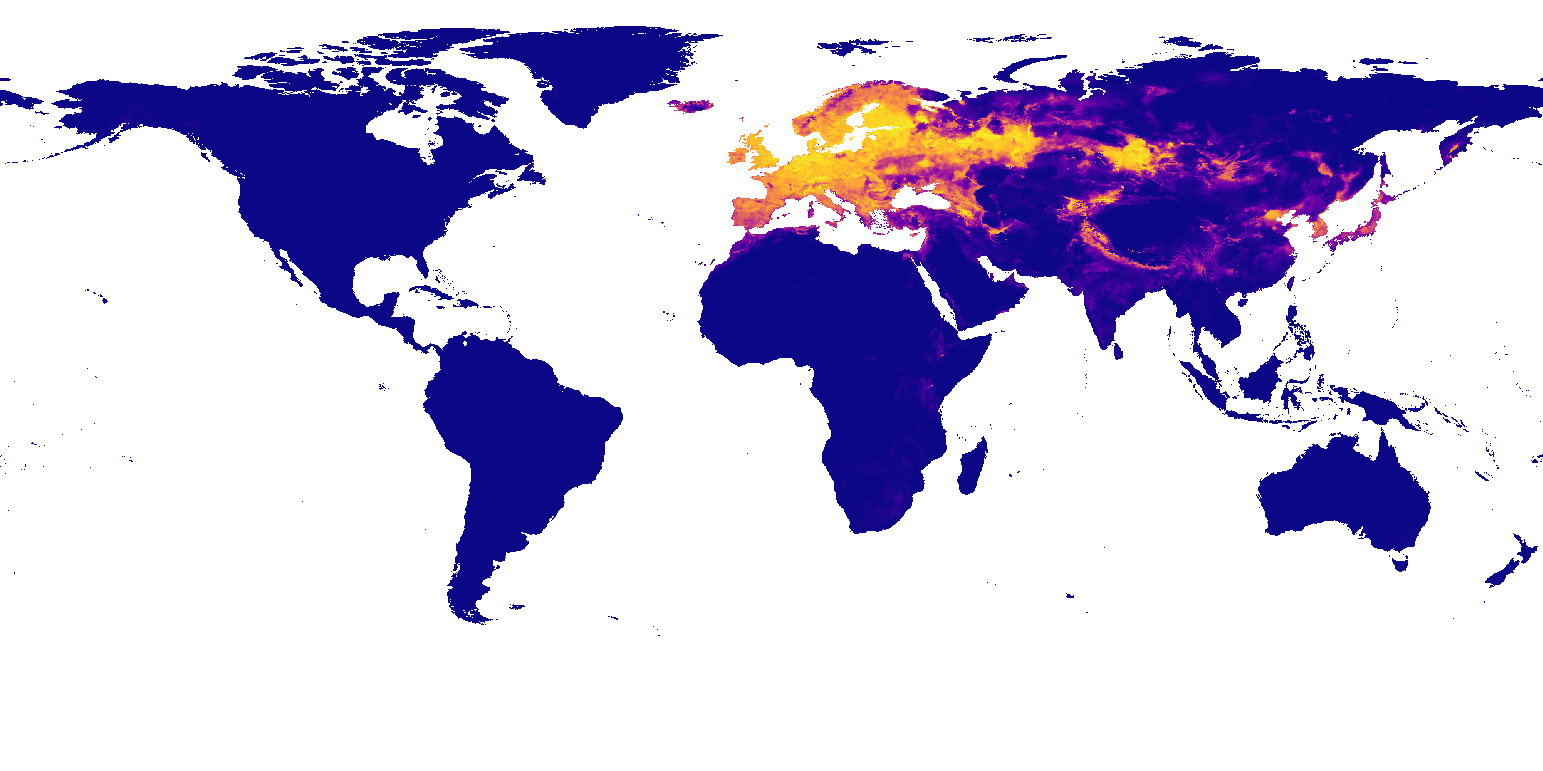}
            \put(0,0){%
              {%
                \setlength\fboxsep{0pt}%
                \setlength\fboxrule{1pt}%
                \fbox{%
                  \includegraphics[width=0.125\textwidth]{figs/range_est/european_robin_crop.png}%
                }%
              }%
            }
        \end{overpic}
        \\[6pt]

        \includegraphics[trim={0 1cm 0 0},clip,width=0.48\textwidth]{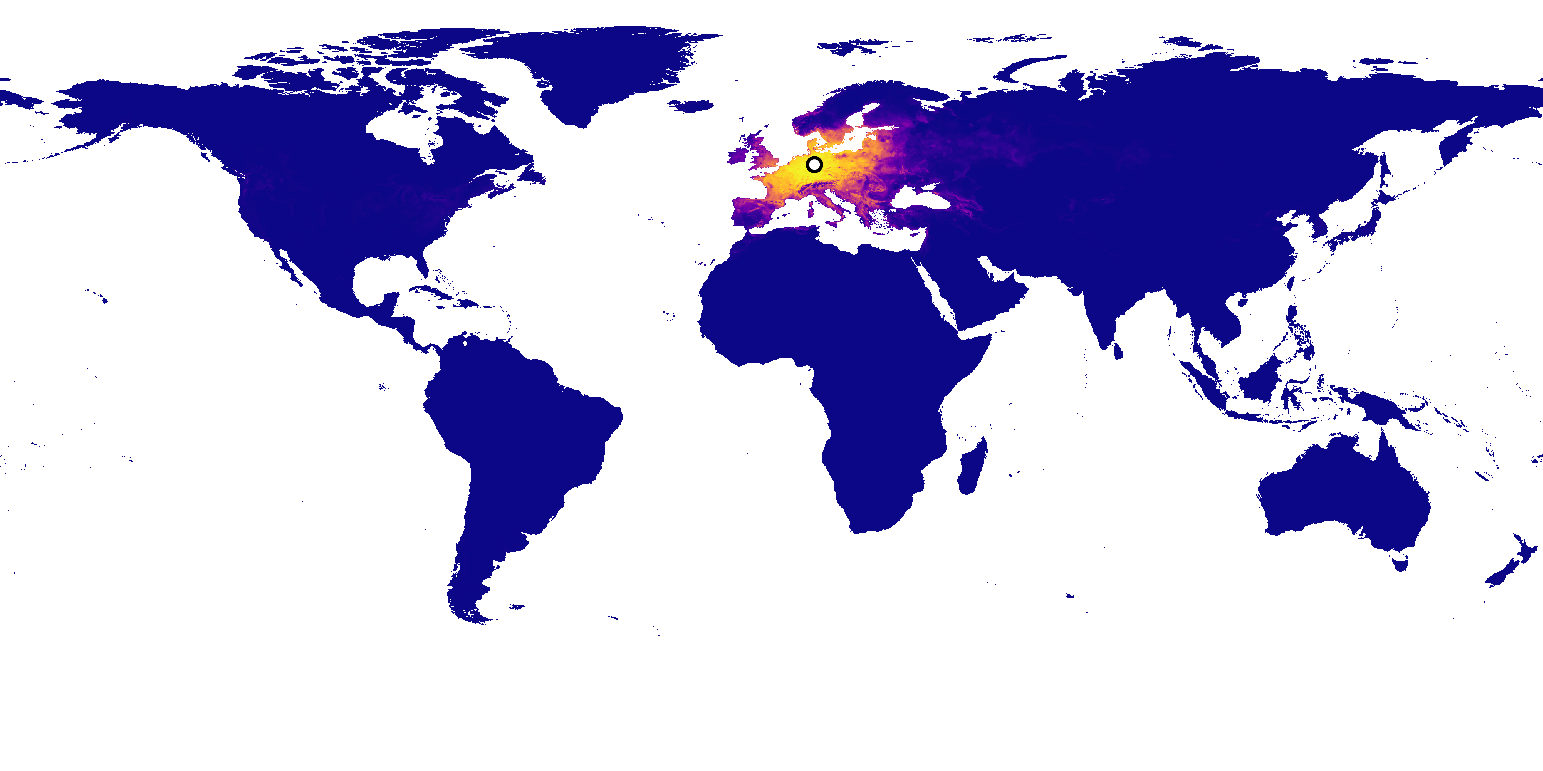}
        &
        \includegraphics[trim={0 1cm 0 0},clip,width=0.48\textwidth]{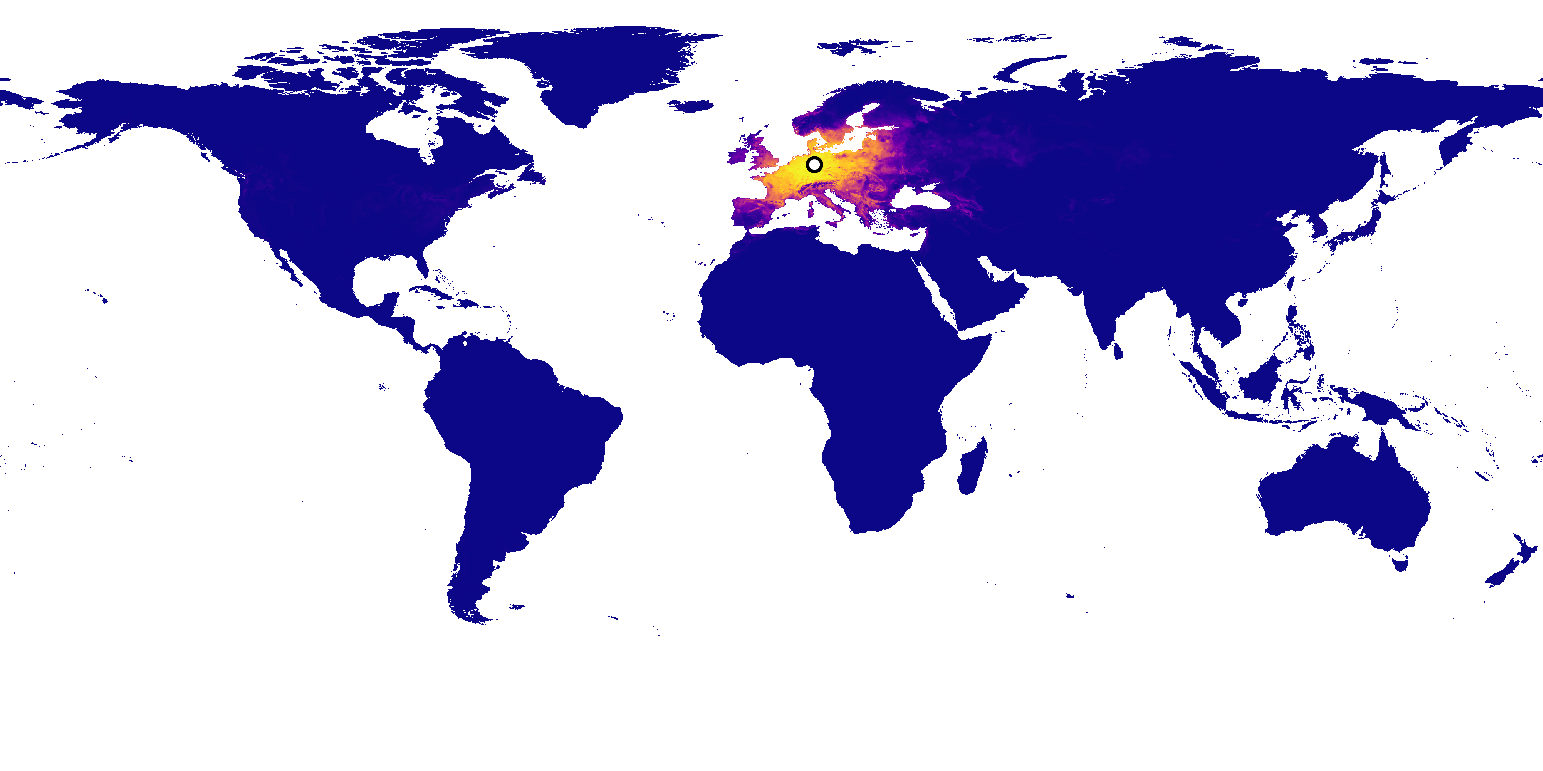}
        \\[6pt]

        \begin{minipage}[t]{0.48\textwidth}
            \scriptsize{\textit{Range Text:} 
            The European robin is found across Europe, east to Western Siberia and south to North Africa; it is sedentary in most of its range except the far north. It also occurs in the Atlantic islands as far west as the Central Group of the Azores and Madeira. It is a vagrant in Iceland and has been introduced to other regions, including North America and Australia, but these introductions were unsuccessful.}
        \end{minipage}
        &
        \begin{minipage}[t]{0.48\textwidth}
            \scriptsize{\textit{Habitat Text:} The European robin inhabits a variety of habitats, including gardens, parks, woodlands, and forests. It prefers areas with dense vegetation and is often found near human settlements. It is also found in mountainous regions and can be seen in urban areas, such as cities and towns.}
        \end{minipage}
    \end{tabular}

    \caption{\textbf{Using text descriptions.} Here we illustrate the zero-shot (top row) and one-shot (bottom row) \modelname range estimations based on text descriptions for the \texttt{European Robin}, using `Range' (left), and `Habitat' (right) text, shown below the range estimates.
    Expert-derived range maps are shown inset in the top row.}
    \label{fig:richer_text_1}
\end{figure}

\begin{figure}[h]
    \centering
    \renewcommand{\arraystretch}{1.2} %

    \begin{tabular}{cc}
        {\bf American Pika - Range Text} & {\bf American Pika - Habitat Text} \\

        \begin{overpic}[trim={0 1cm 0 0},clip,width=0.48\textwidth]{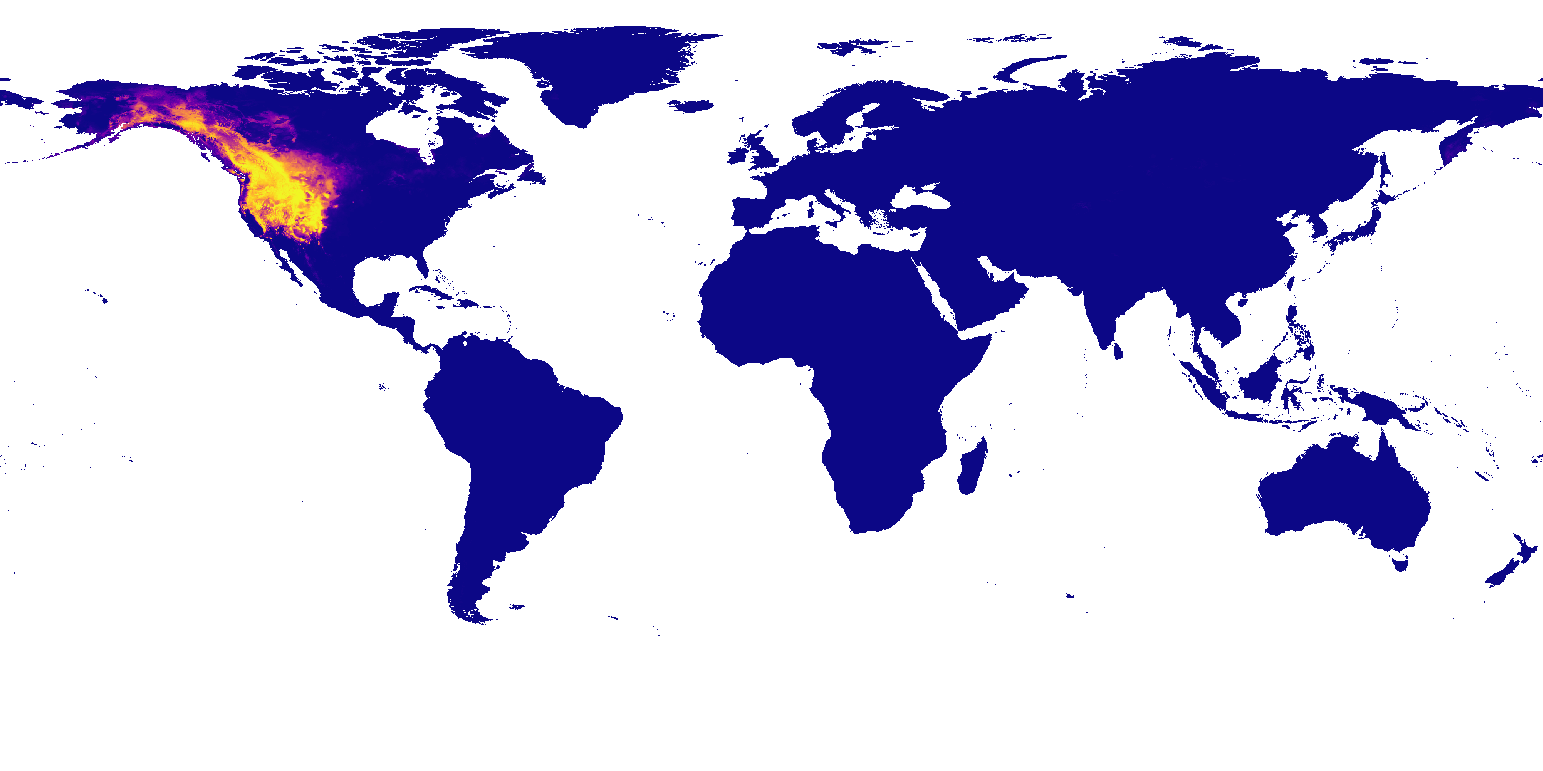}
            \put(0,0){%
              {%
                \setlength\fboxsep{0pt}%
                \setlength\fboxrule{1pt}%
                \fbox{%
                  \includegraphics[width=0.125\textwidth]{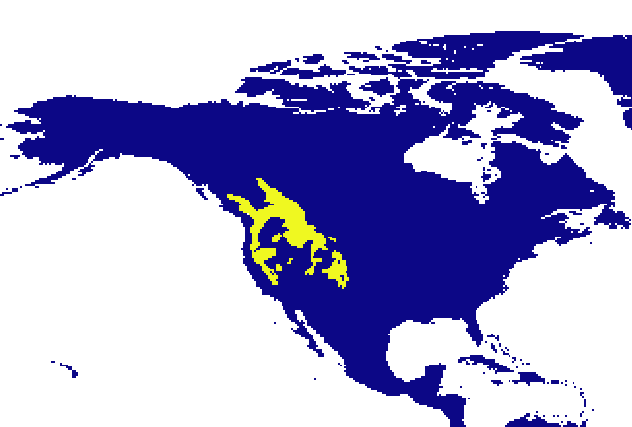}%
                }%
              }%
            }
        \end{overpic}
        &
        \begin{overpic}[trim={0 1cm 0 0},clip,width=0.48\textwidth]{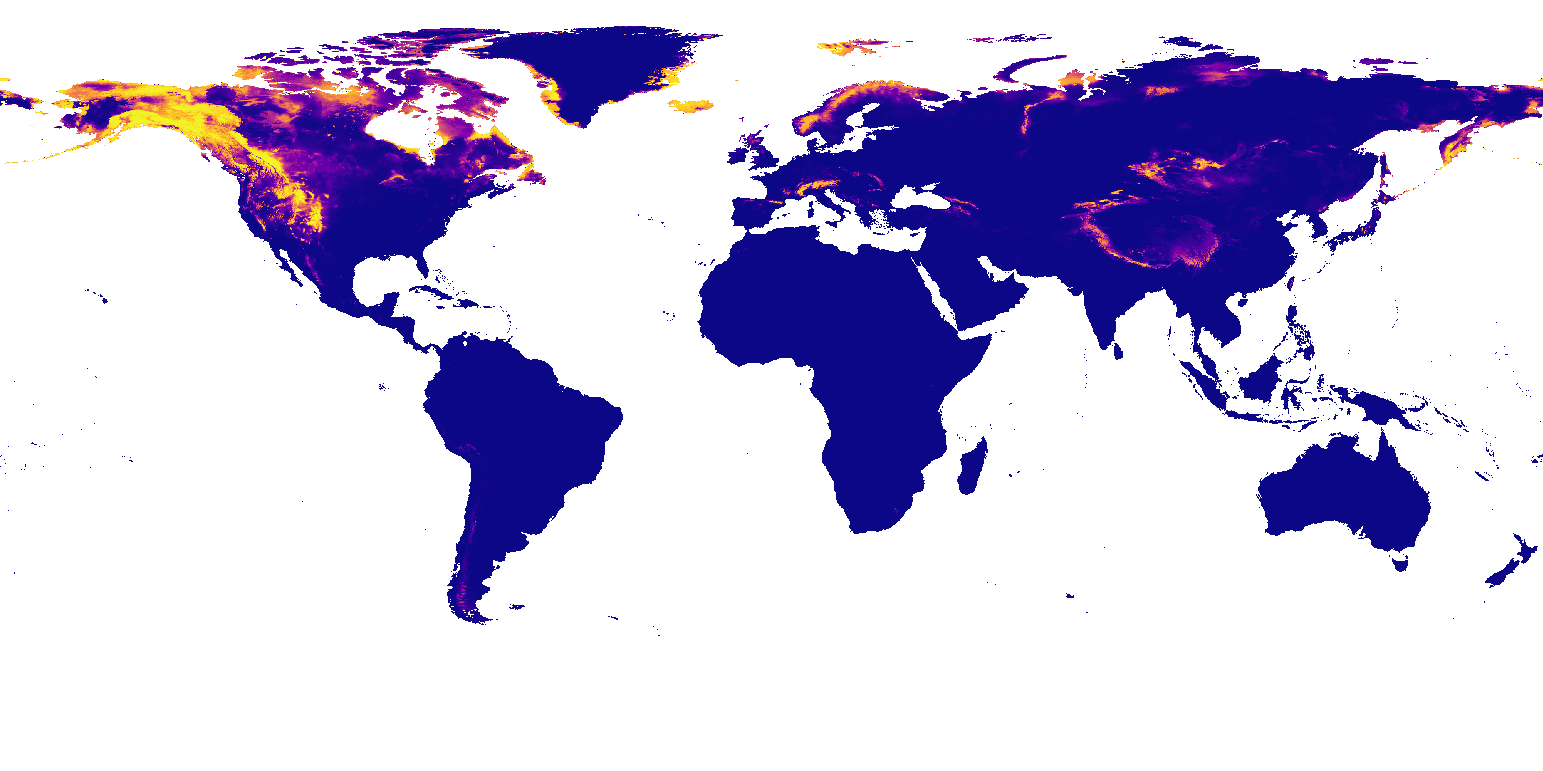}
            \put(0,0){%
              {%
                \setlength\fboxsep{0pt}%
                \setlength\fboxrule{1pt}%
                \fbox{%
                  \includegraphics[width=0.125\textwidth]{figs/range_est/pika_crop.png}%
                }%
              }%
            }
        \end{overpic}
        \\[6pt]

        \includegraphics[trim={0 1cm 0 0},clip,width=0.48\textwidth]{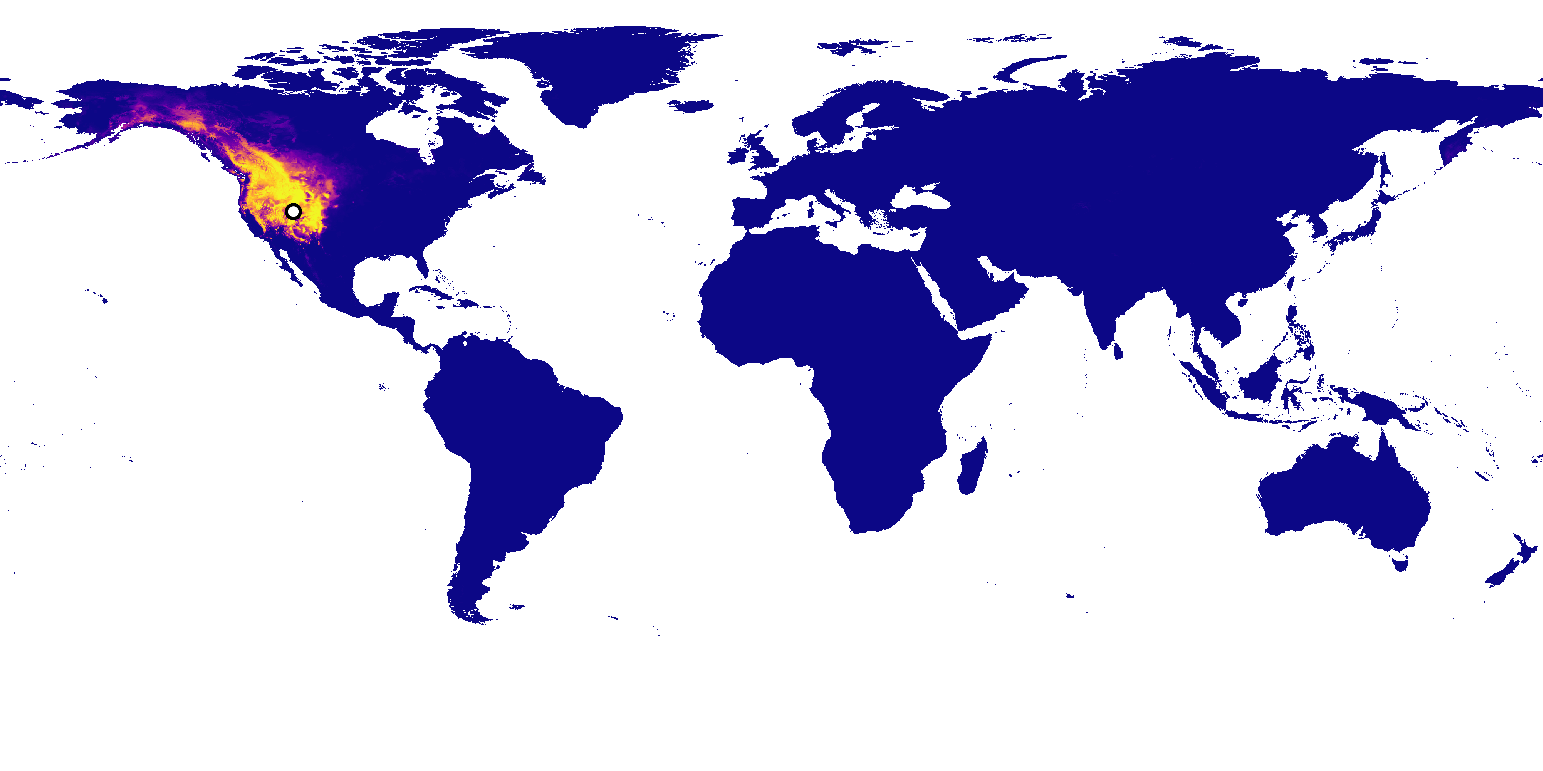}
        &
        \includegraphics[trim={0 1cm 0 0},clip,width=0.48\textwidth]{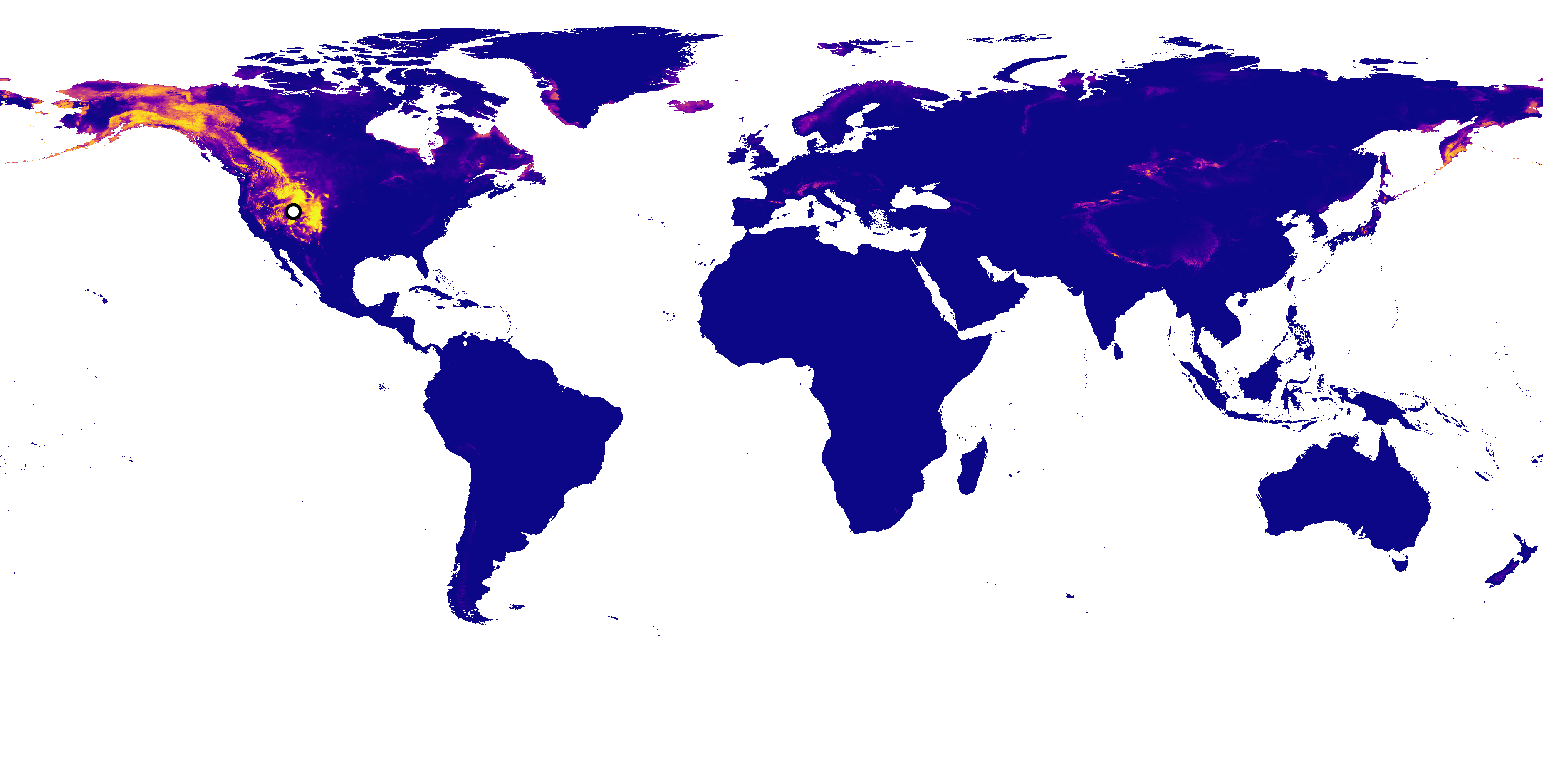}
        \\[6pt]

        \begin{minipage}[t]{0.48\textwidth}
            \scriptsize{\textit{Range Text:} 
            The American pika (Ochotona princeps) is found in the mountains of western North America, usually in boulder fields at or above the tree line, from central British Columbia and Alberta in Canada to the US states of Oregon, Washington, Idaho, Montana, Wyoming, Colorado, Utah, Nevada, California, and New Mexico.}
        \end{minipage}
        &
        \begin{minipage}[t]{0.48\textwidth}
            \scriptsize{\textit{Habitat Text:} Pikas inhabit talus fields that are fringed by suitable vegetation in alpine areas. They also live in piles of broken rock. Sometimes, they live in man-made substrate such as mine tailings and piles of scrap lumber. Pikas usually have their den and nest sites below rock, around 20-100 cm (8-39 in) in diameter, but often sit on larger and more prominent rocks.}
        \end{minipage}
    \end{tabular}

    \caption{\textbf{Using text descriptions.} Here we illustrate the zero-shot (top row) and one-shot (bottom row) \modelname range estimations based on text descriptions for the \texttt{American Pika}, using `Range' (left), and `Habitat' (right) text, shown below the range estimates. 
    Expert-derived range maps are shown inset.}
    \vspace{-10pt}
    \label{fig:richer_text_2}
\end{figure}

\begin{figure*}[t]
\centering
\includegraphics[width=\textwidth]{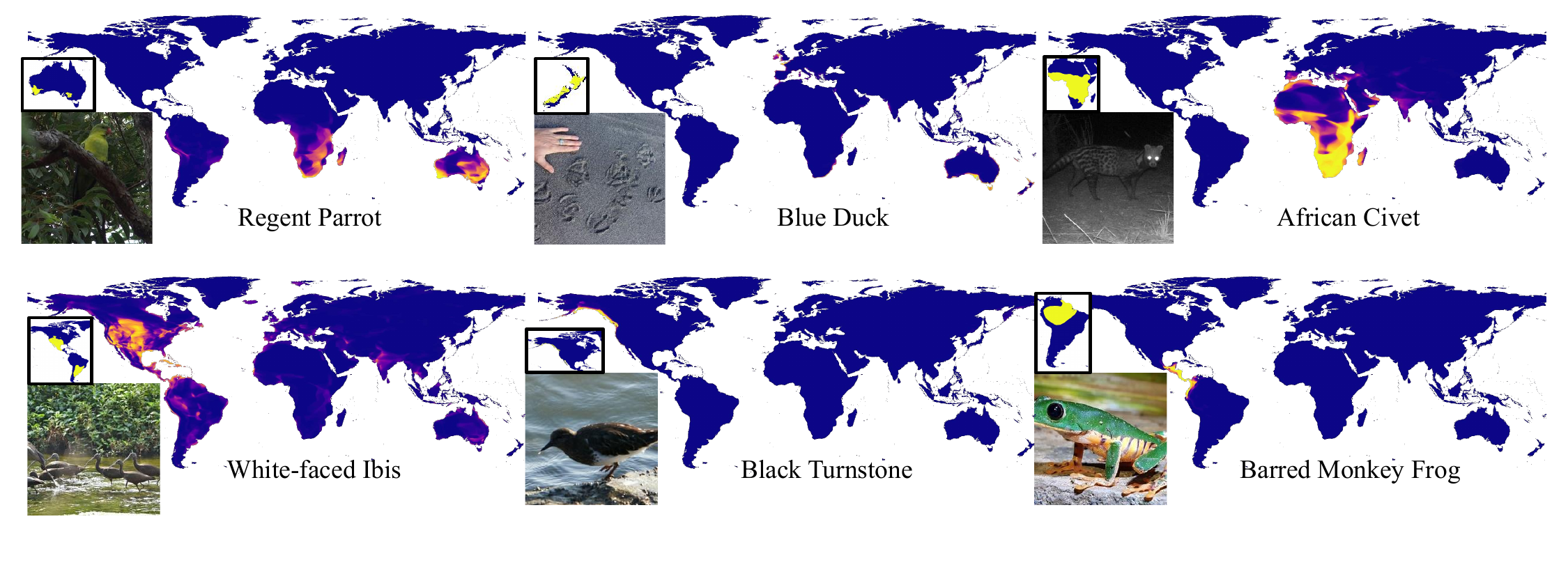} 
\vspace{-35pt}
\caption{{\bf Image zero-shot range estimation.} Here we see zero-shot range estimates for six species in the IUCN evaluation dataset, with expert-derived range and image inset. 
The \texttt{blue duck} image taken from~\citet{inatWeb} only shows evidence of the species from footprints in wet sand.
We see that this image generates predictions in coastal areas in various locations around the globe.
The coastal background for the \texttt{Black Turnstone} could have helped the model to generate a relatively accurate prediction on the northwest coast of North America.
}
\vspace{-10pt}
\label{fig:viz_zero_shot}
\end{figure*}

Here, we provide additional examples of the ranges produced by \modelname using context locations, text, and images.
In~\cref{fig:richer_text_1,fig:richer_text_2} we visualize \modelname range estimates for two different species when habitat or range text is provided. 
We observe that the combination of text and context locations seem to result in better estimates of the range. 
In \cref{fig:viz_zero_shot} we show additional zero-shot image-only examples, where \modelname is provided a single image from a held-out test species at inference time. 
Again, we observe some plausible range predictions even with such limited input data. 

\begin{figure}[h]
    \centering
    \begin{minipage}{0.04\textwidth}
    \end{minipage}%
    \hspace{0.5em}
    \begin{minipage}{0.29\textwidth}
        \centering \textbf{FS-SINR}
    \end{minipage}%
    \hspace{0.5em}
    \begin{minipage}{0.29\textwidth}
        \centering \textbf{LE-SINR}
    \end{minipage}%
    \hspace{0.5em}
    \begin{minipage}{0.29\textwidth}
        \centering \textbf{SINR}
    \end{minipage}
    
    \vspace{1em}

    \begin{minipage}{0.04\textwidth}
        \rotatebox{90}{\tiny\textbf{0 Context}}
    \end{minipage}%
    \hspace{0.5em}
    \begin{minipage}{0.29\textwidth}
        \centering
            \begin{overpic}[trim={0 0 0 0},clip,width=\linewidth]{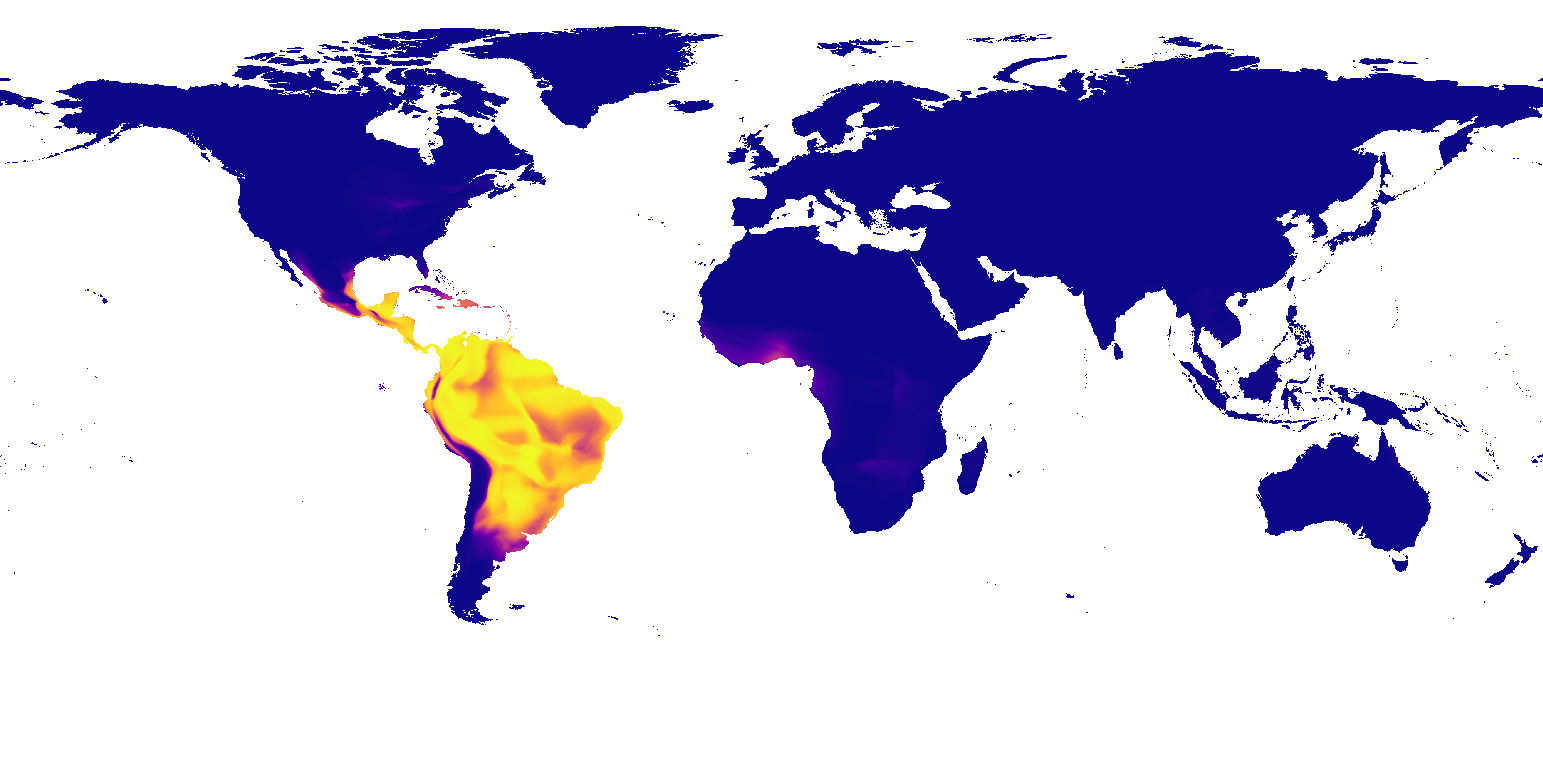}
        \put(0,0){%
          {%
            \setlength\fboxsep{0pt}%
            \setlength\fboxrule{1pt}%
            \fbox{%
              \includegraphics[width=0.25\linewidth]{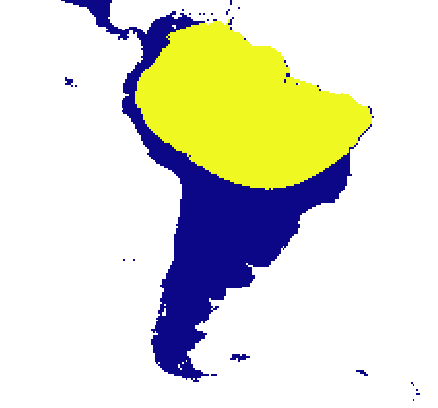}%
            }%
          }%
        }
    \end{overpic}
    \end{minipage}%
    \hspace{0.5em}
    \begin{minipage}{0.29\textwidth}
        \centering
        \includegraphics[width=\linewidth]{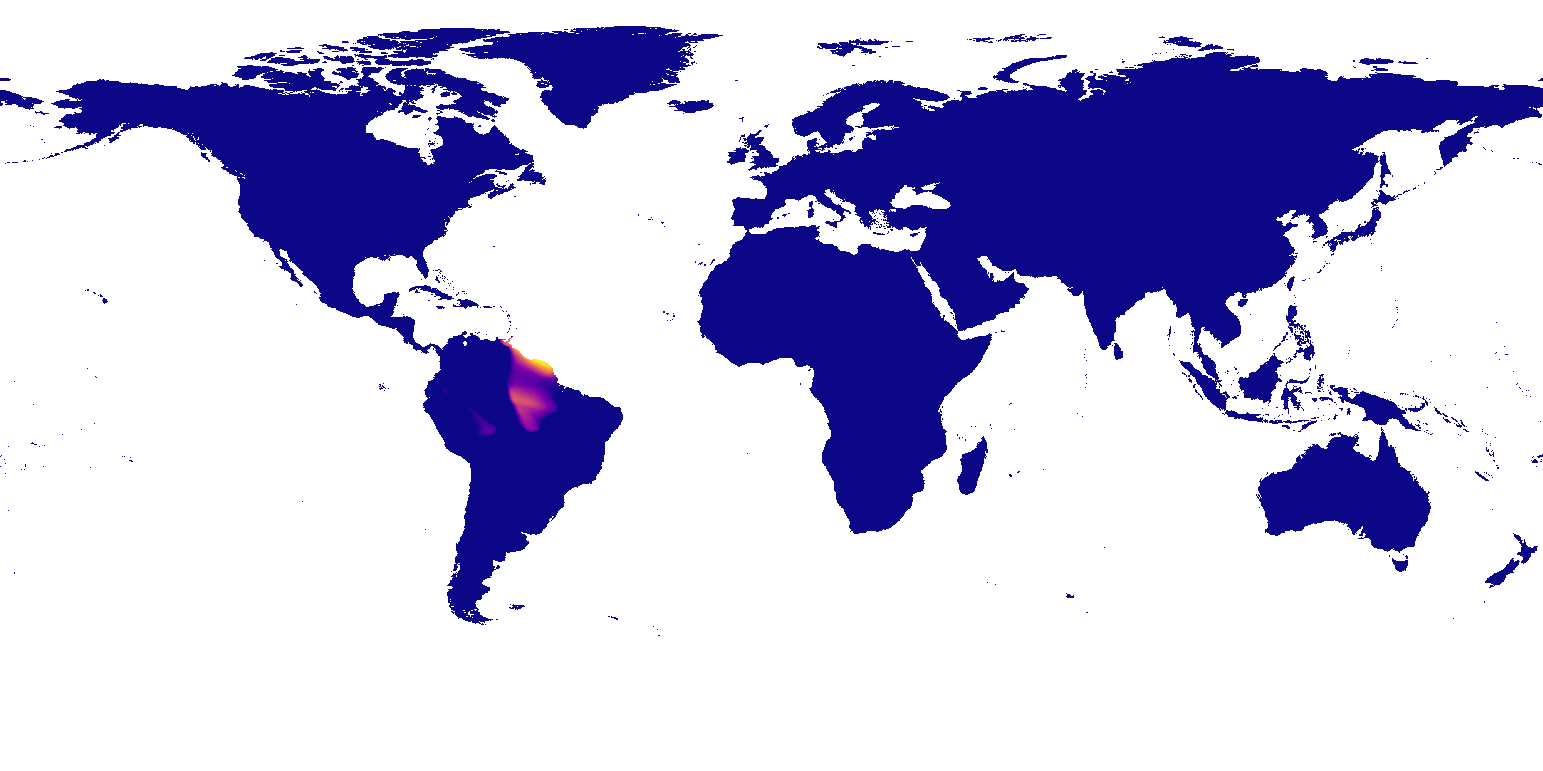}
    \end{minipage}%
    \hspace{0.5em}
    \begin{minipage}{0.29\textwidth}
        \centering
        \includegraphics[width=\linewidth]{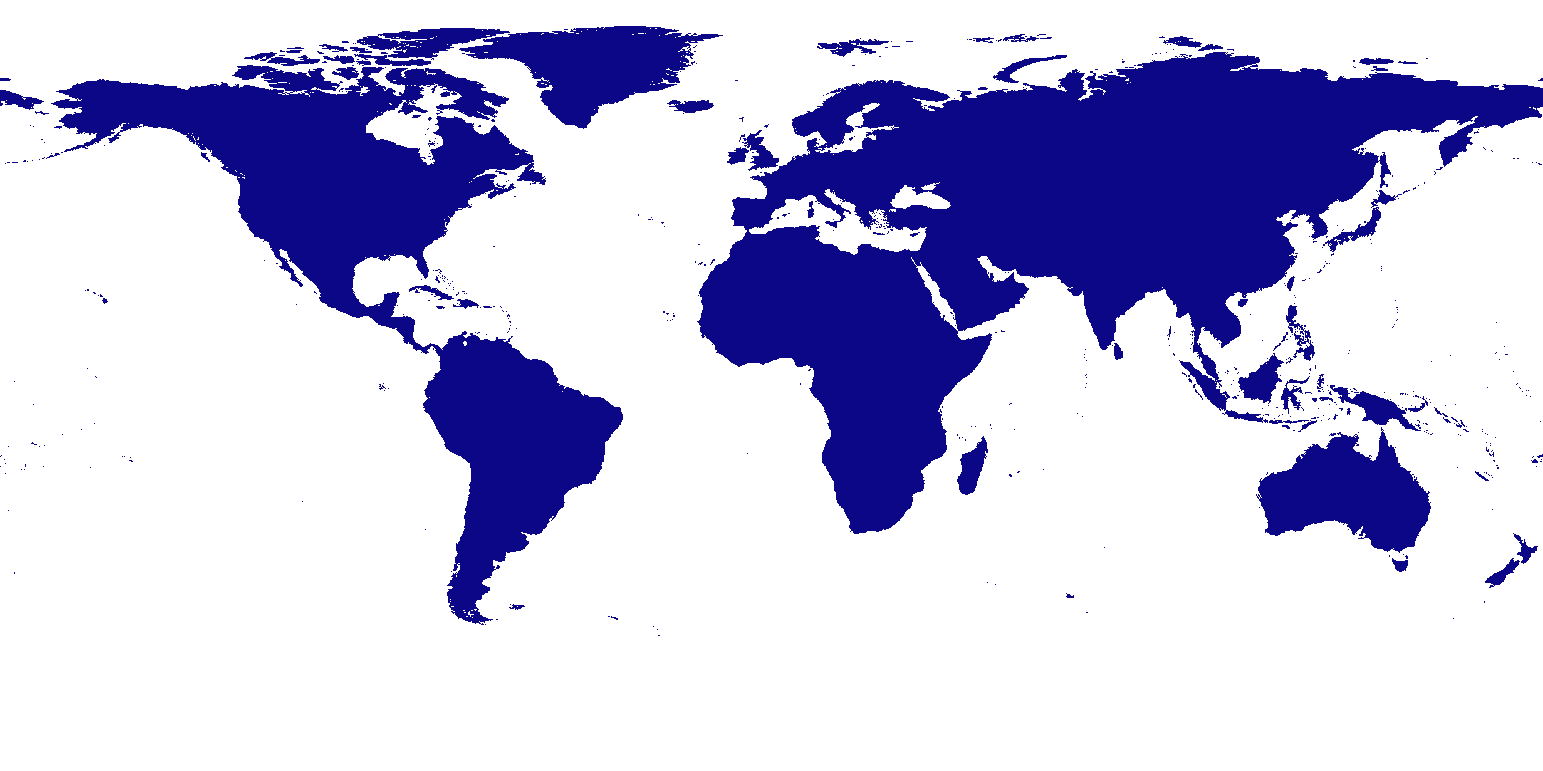}
    \end{minipage}

    \vspace{1em}

    \begin{minipage}{0.04\textwidth}
        \rotatebox{90}{\tiny\textbf{1 Context}}
    \end{minipage}%
    \hspace{0.5em}
    \begin{minipage}{0.29\textwidth}
        \centering
        \includegraphics[width=\linewidth]{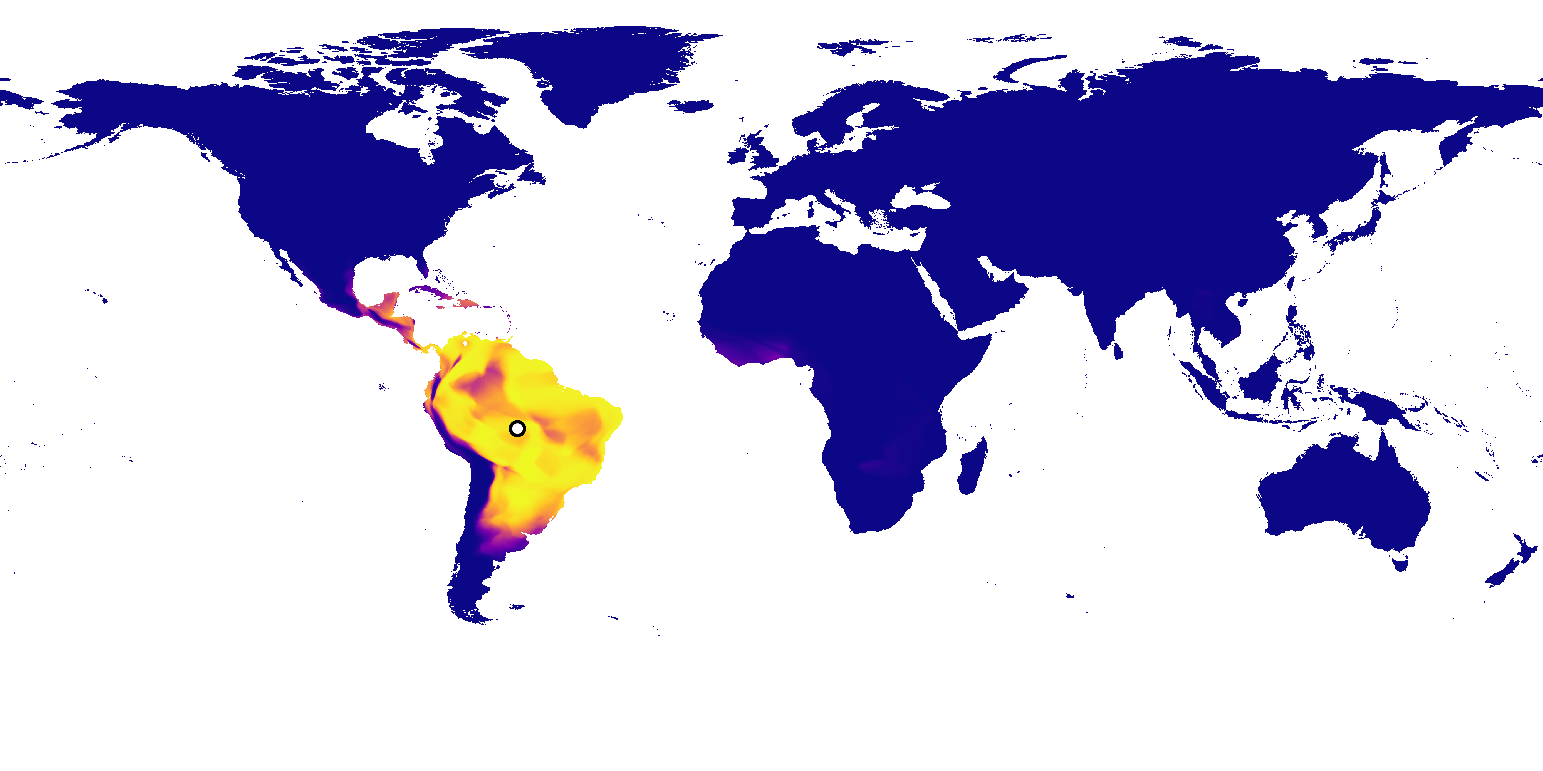}
    \end{minipage}%
    \hspace{0.5em}
    \begin{minipage}{0.29\textwidth}
        \centering
        \includegraphics[width=\linewidth]{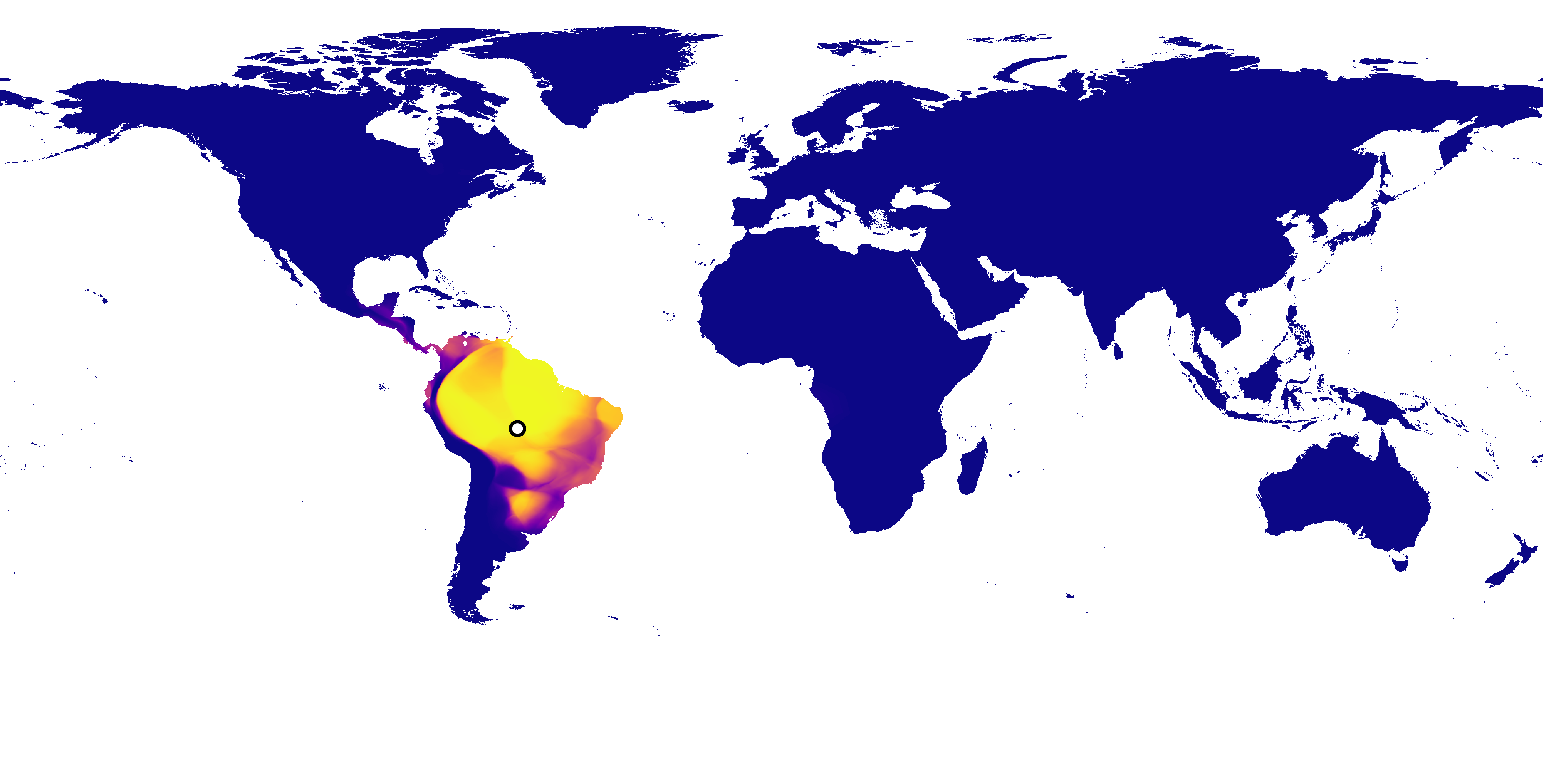}
    \end{minipage}%
    \hspace{0.5em}
    \begin{minipage}{0.29\textwidth}
        \centering
        \includegraphics[width=\linewidth]{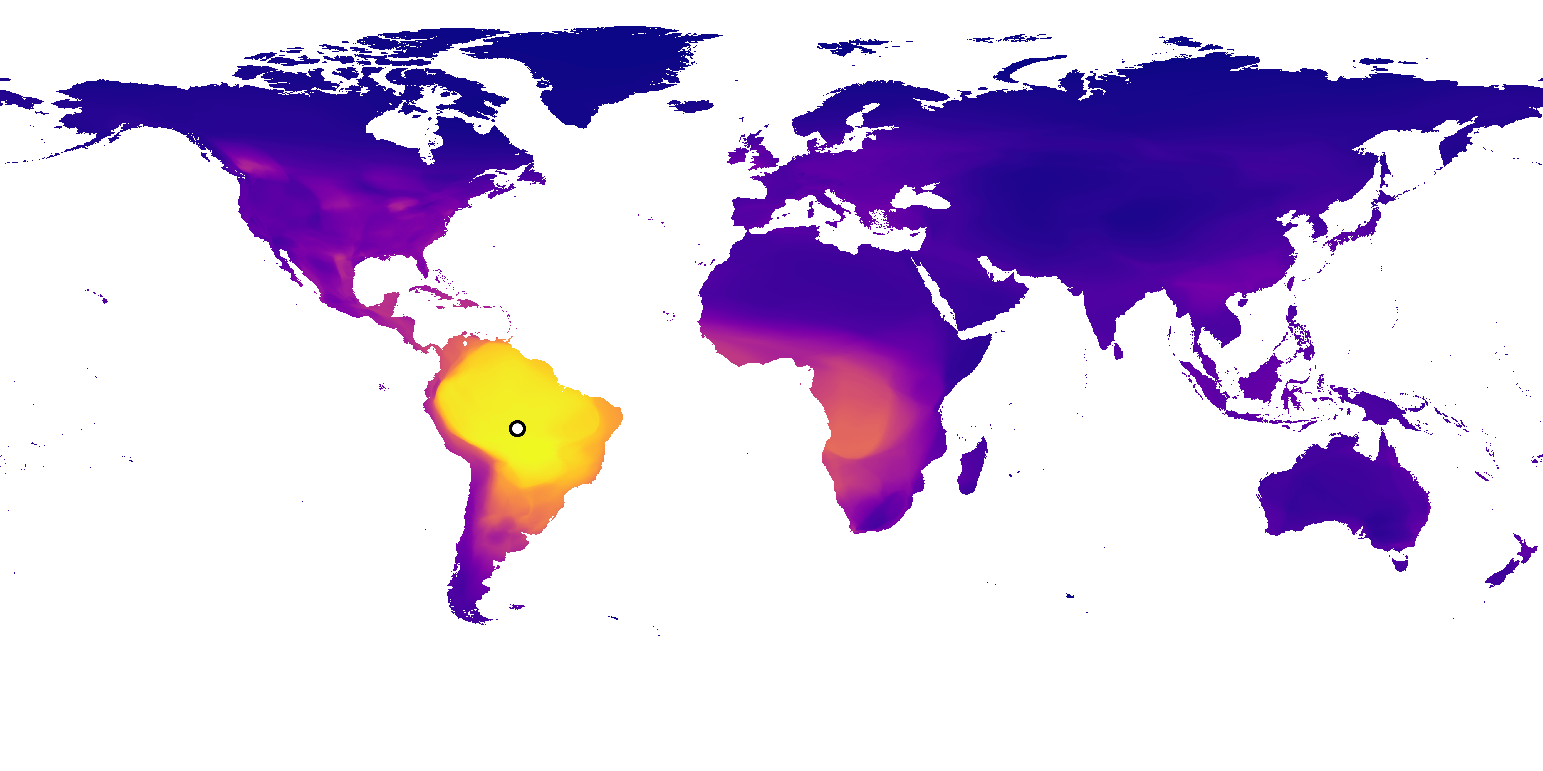}
    \end{minipage}

    \vspace{1em}

    \begin{minipage}{0.04\textwidth}
        \rotatebox{90}{\tiny\textbf{2 Context}}
    \end{minipage}%
    \hspace{0.5em}
    \begin{minipage}{0.29\textwidth}
        \centering
        \includegraphics[width=\linewidth]{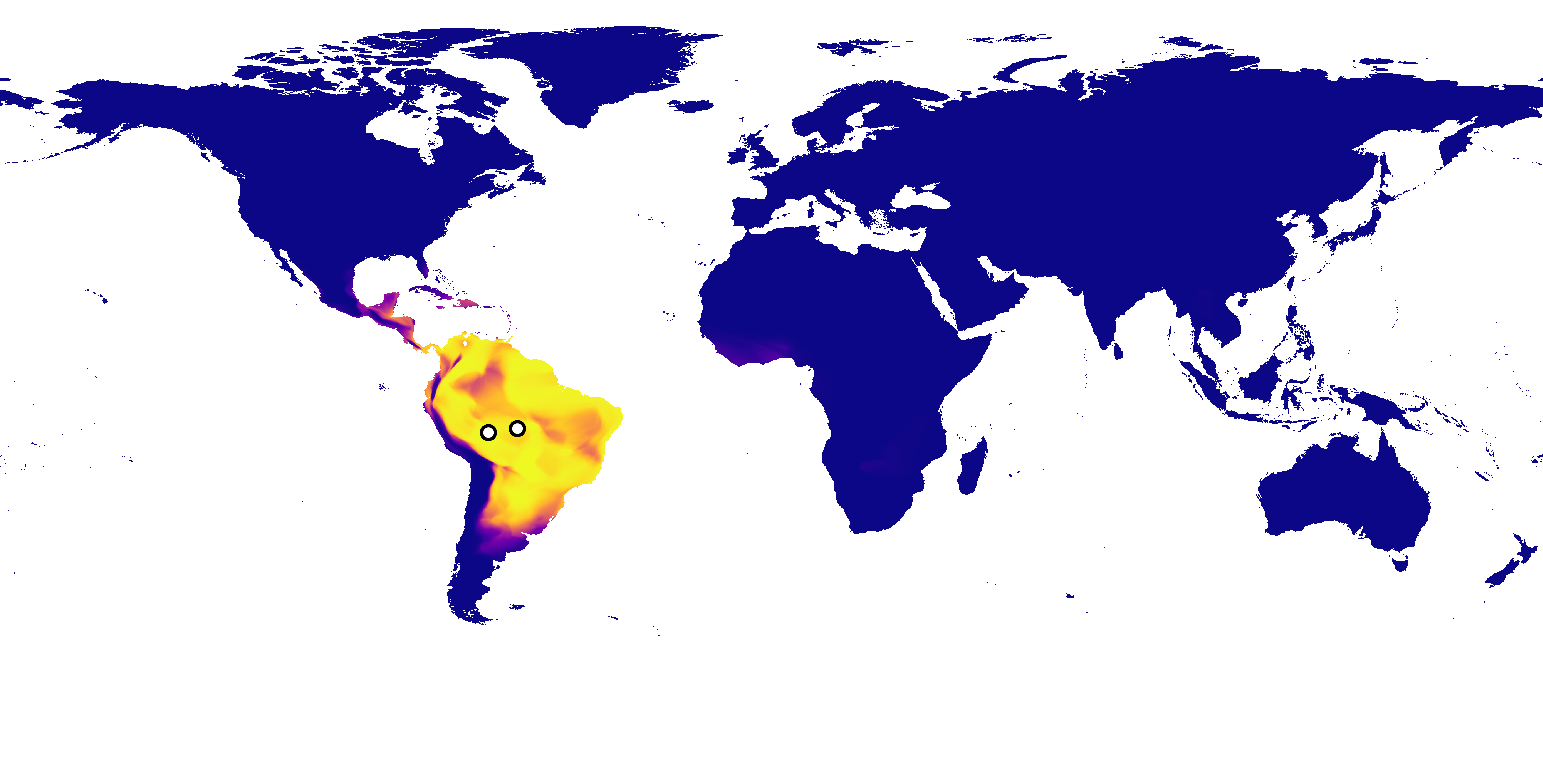}
    \end{minipage}%
    \hspace{0.5em}
    \begin{minipage}{0.29\textwidth}
        \centering
        \includegraphics[width=\linewidth]{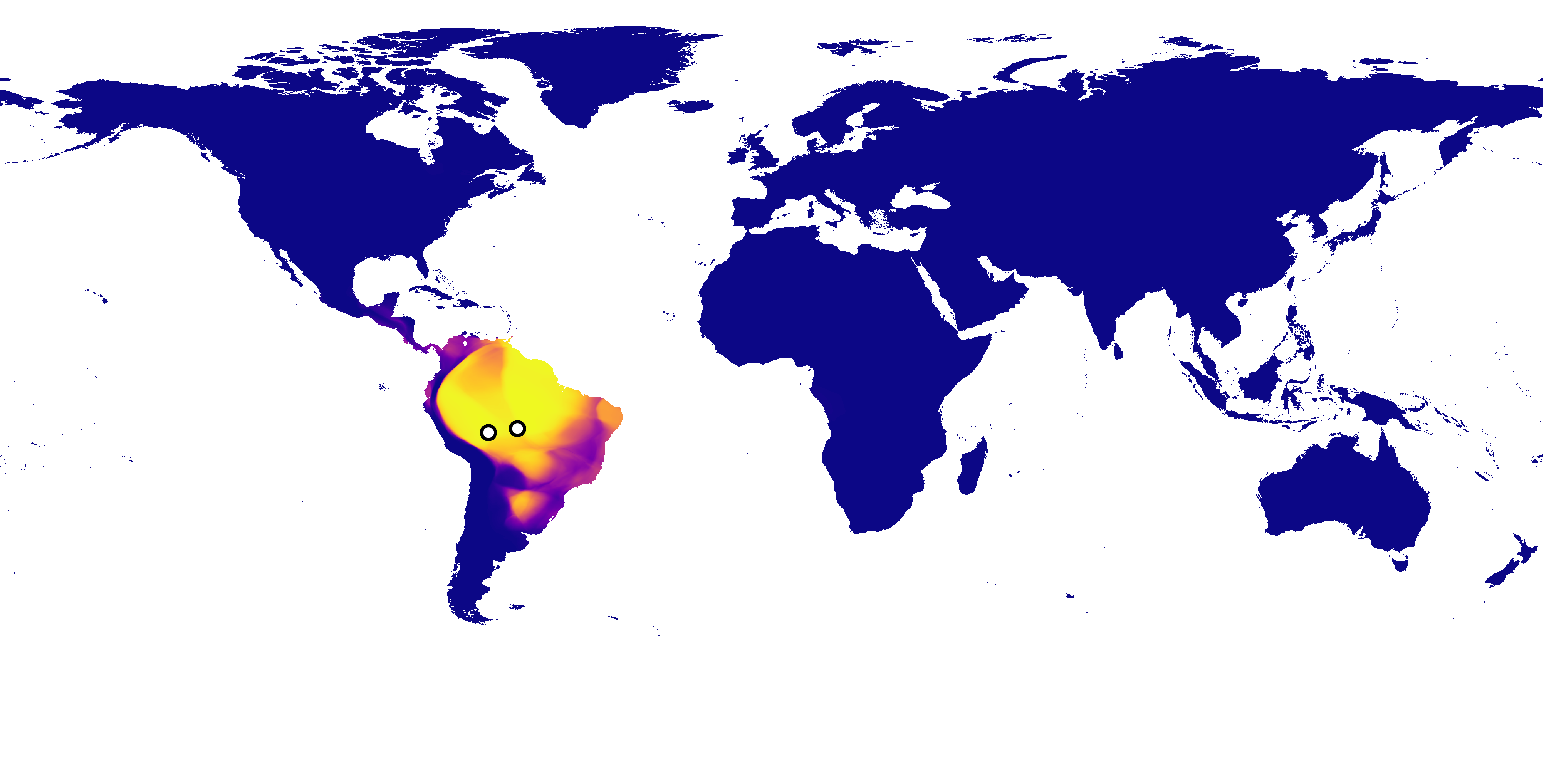}
    \end{minipage}%
    \hspace{0.5em}
    \begin{minipage}{0.29\textwidth}
        \centering
        \includegraphics[width=\linewidth]{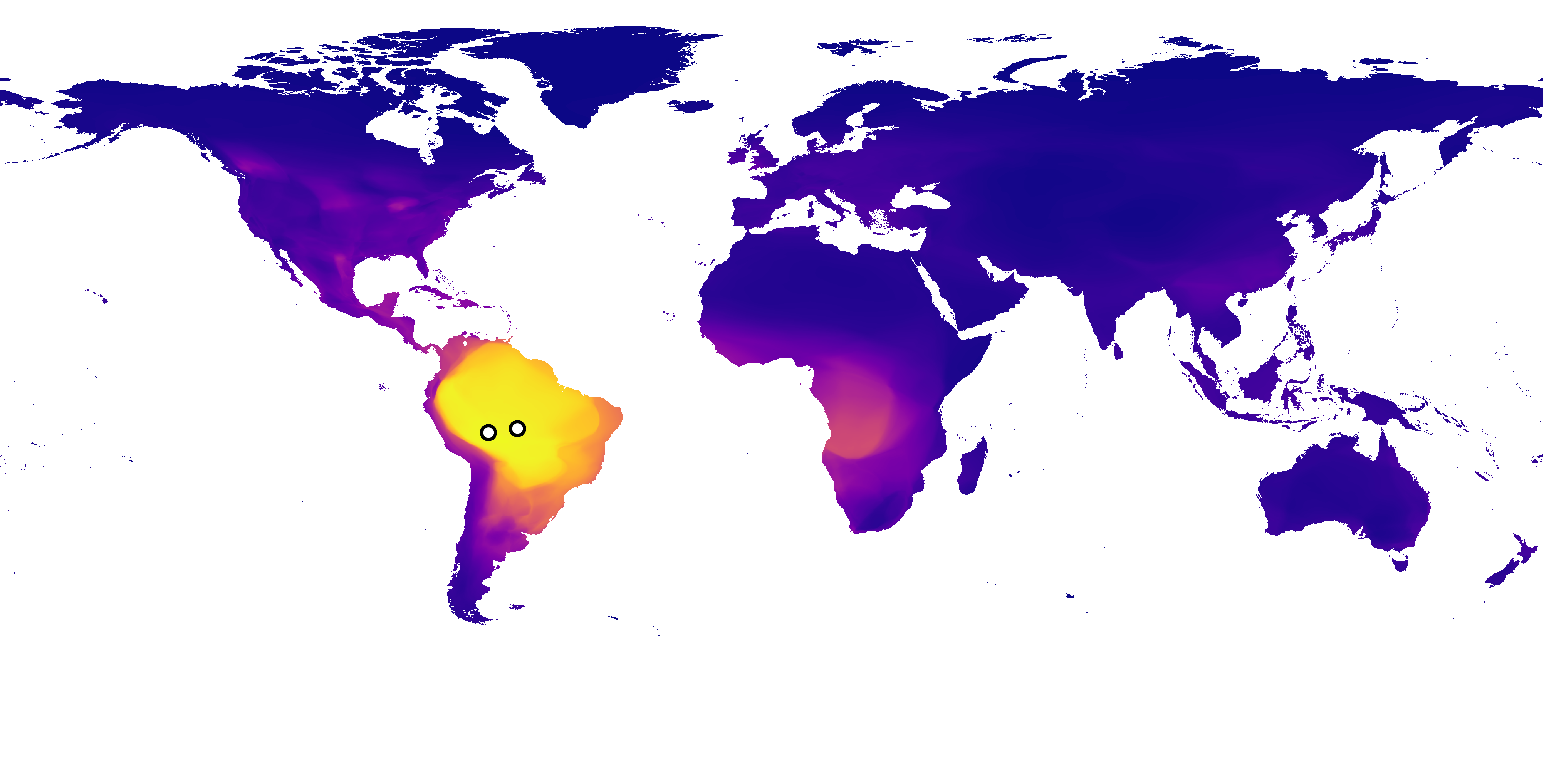}
    \end{minipage}

    \vspace{1em}

    \begin{minipage}{0.04\textwidth}
        \rotatebox{90}{\tiny\textbf{5 Context}}
    \end{minipage}%
    \hspace{0.5em}
    \begin{minipage}{0.29\textwidth}
        \centering
        \includegraphics[width=\linewidth]{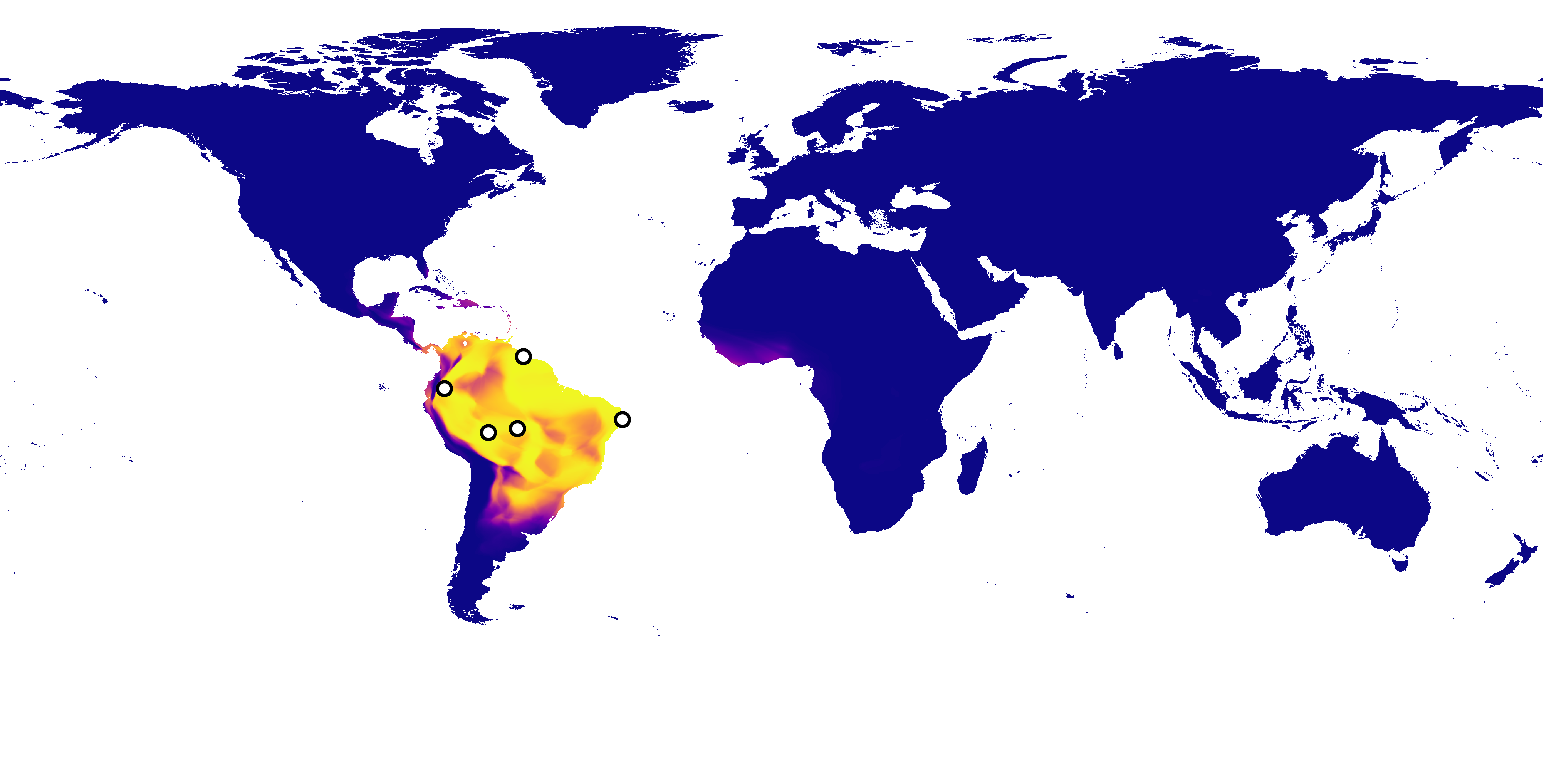}
    \end{minipage}%
    \hspace{0.5em}
    \begin{minipage}{0.29\textwidth}
        \centering
        \includegraphics[width=\linewidth]{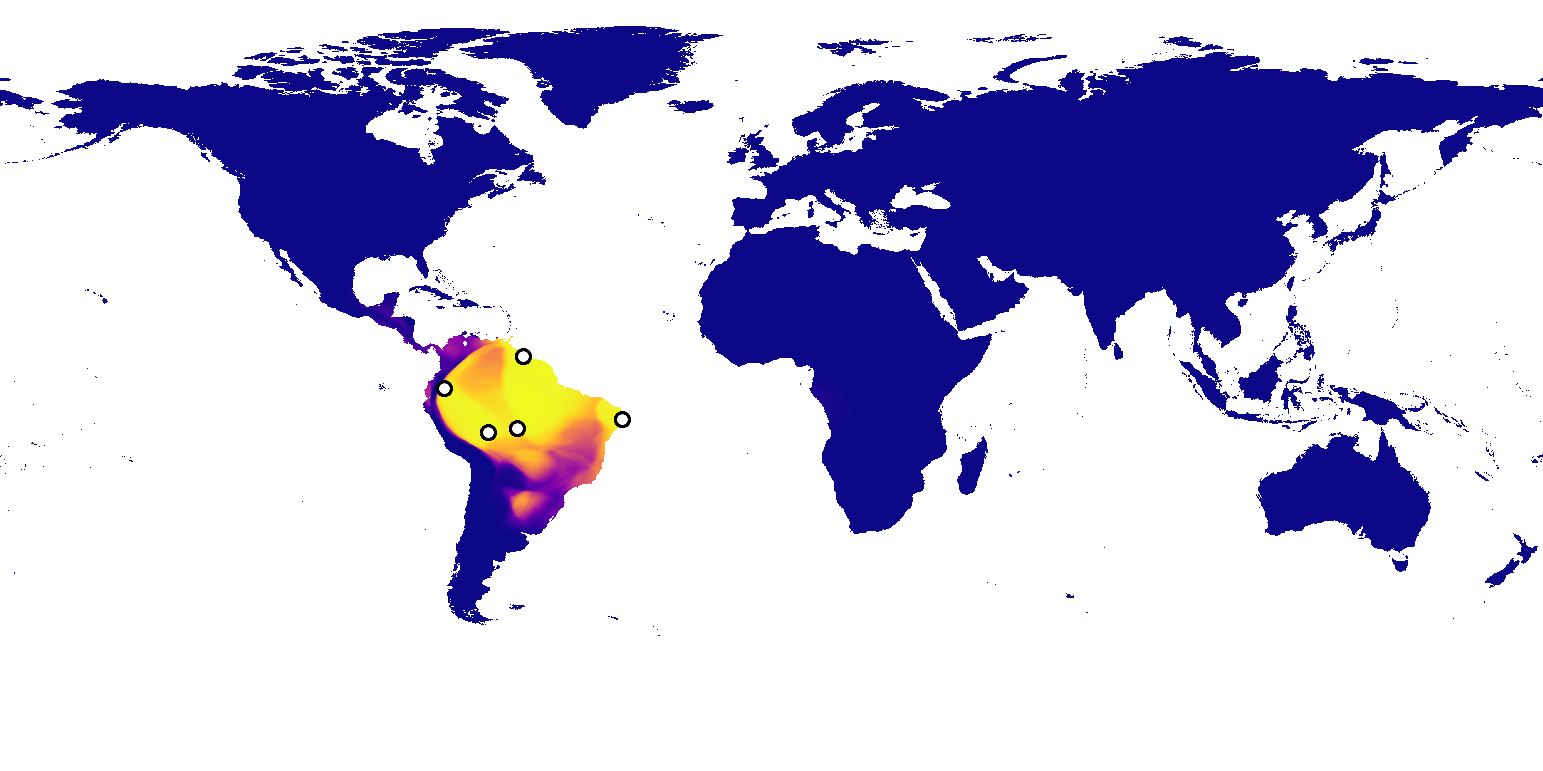}
    \end{minipage}%
    \hspace{0.5em}
    \begin{minipage}{0.29\textwidth}
        \centering
        \includegraphics[width=\linewidth]{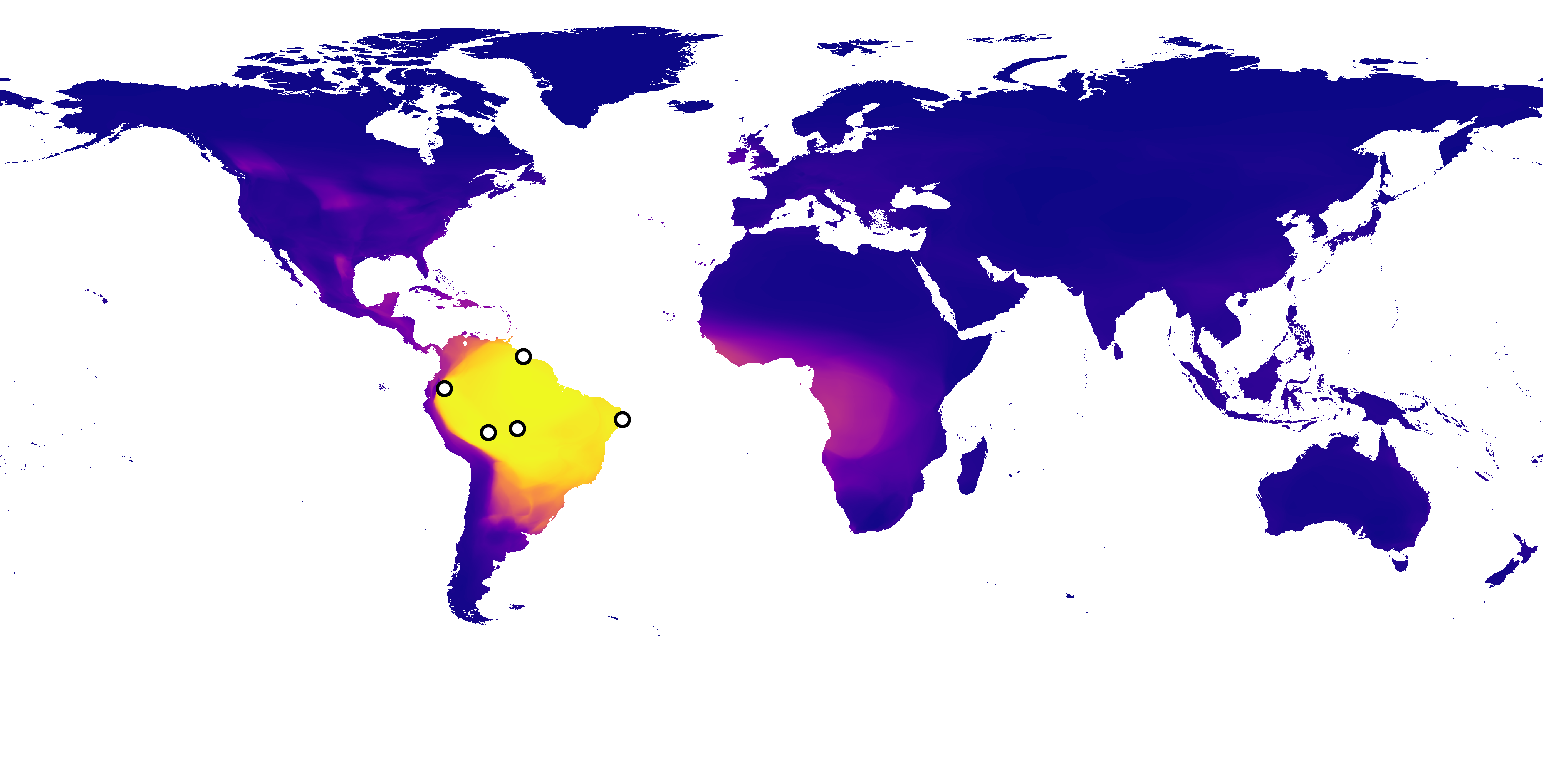}
    \end{minipage}

    \vspace{1em}

    \begin{minipage}{0.04\textwidth}
        \rotatebox{90}{\tiny\textbf{20 Context}}
    \end{minipage}%
    \hspace{0.5em}
    \begin{minipage}{0.29\textwidth}
        \centering
        \includegraphics[width=\linewidth]{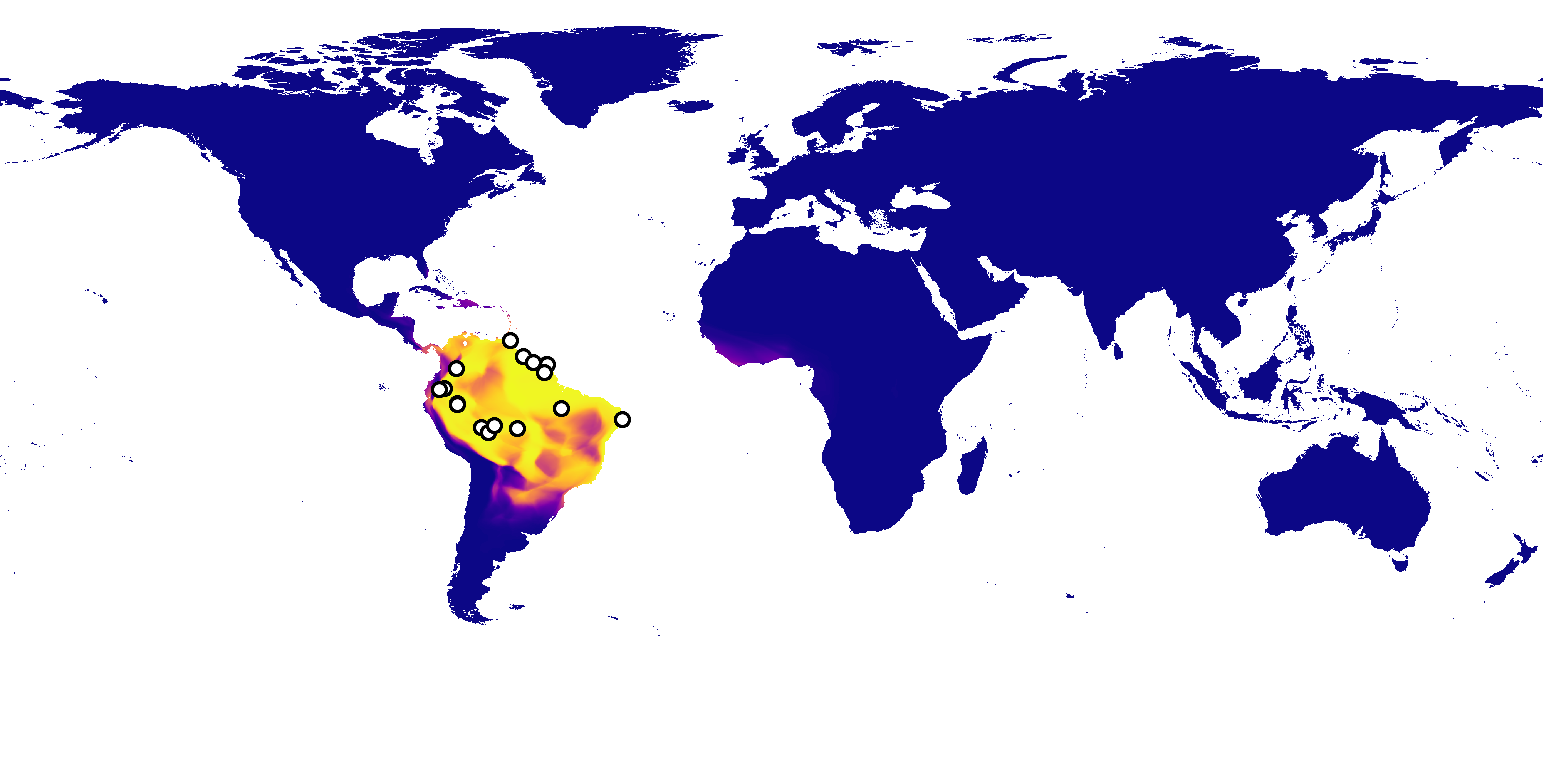}
    \end{minipage}%
    \hspace{0.5em}
    \begin{minipage}{0.29\textwidth}
        \centering
        \includegraphics[width=\linewidth]{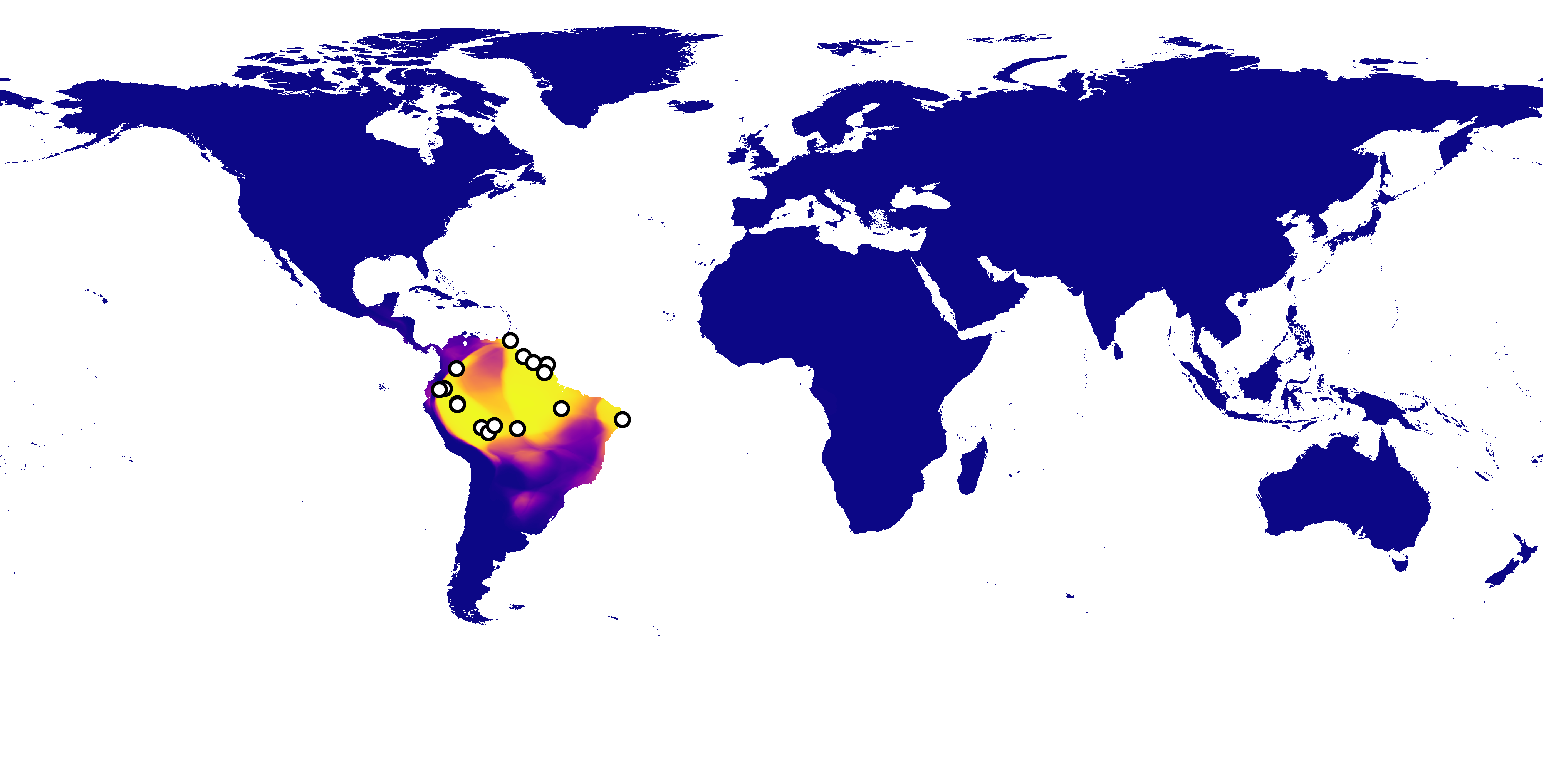}
    \end{minipage}%
    \hspace{0.5em}
    \begin{minipage}{0.29\textwidth}
        \centering
        \includegraphics[width=\linewidth]{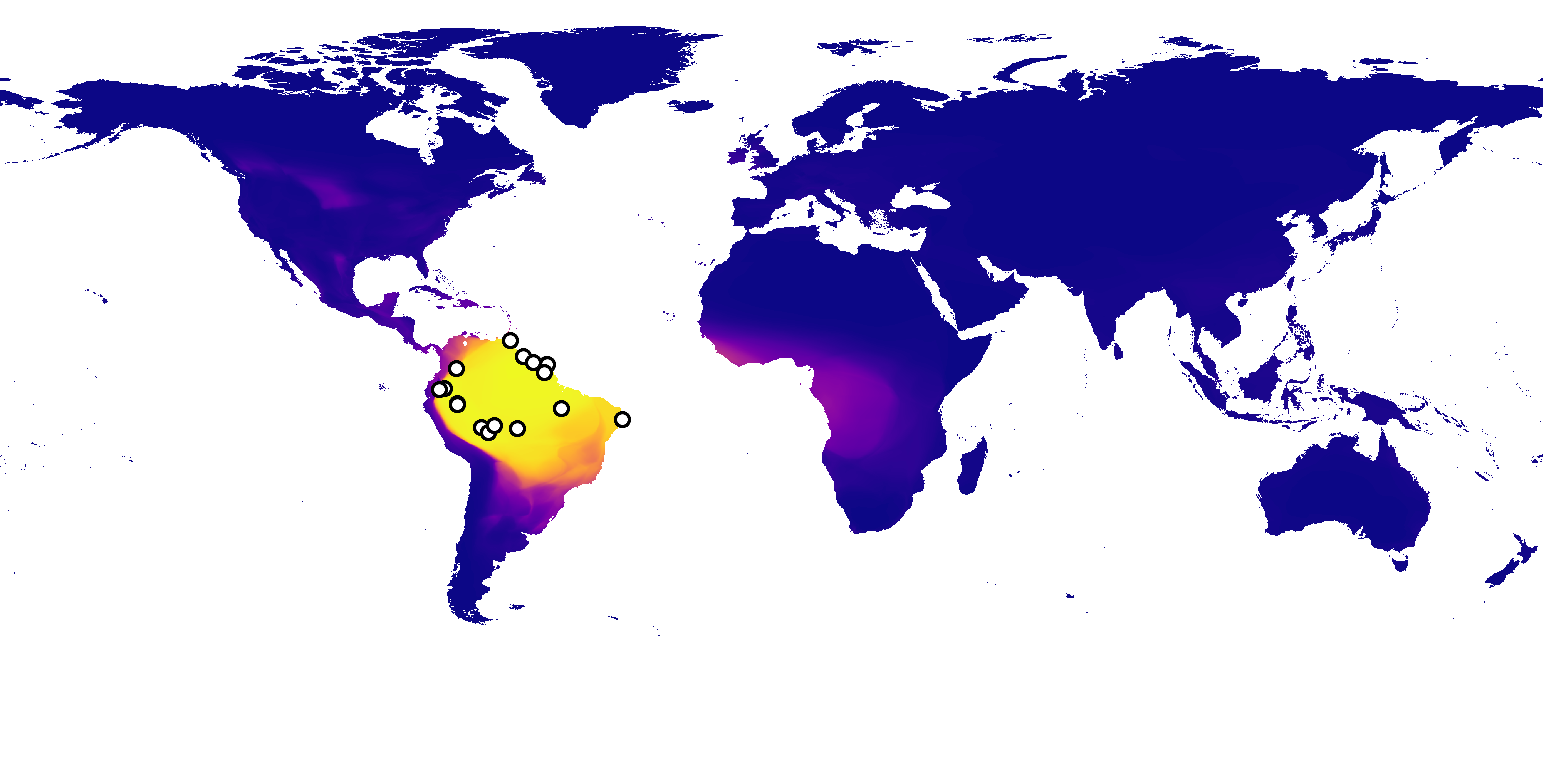}
    \end{minipage}
    \vspace{-15pt}
    \caption{\change{\textbf{Comparing estimated ranges across models with context lcoations and range text.}
    Here we see zero-shot and few-shot range estimates produced by FS-SINR, LE-SINR, and SINR for the \texttt{Brown-banded Watersnake}, with expert-derived range inset.
    We provide range text to FS-SINR and LE-SINR as well as context locations, but SINR is not capable of accepting text and so we show a blank map for the zero-shot range estimate.
    We see that LE-SINR underestimates the range using only text, while FS-SINR overestimates it.
    SINR requires more location data than the other approaches to localize the range to South America.\\
\textit{Range Text:} ``The Brown-banded water snake (Helicops angulatus) is found in tropical South America and Trinidad and Tobago.''}}
    \label{fig:qualitative_different_models_range}
\end{figure}

\begin{figure}[h]
    \centering
    \begin{minipage}{0.04\textwidth}
    \end{minipage}%
    \hspace{0.5em}
    \begin{minipage}{0.29\textwidth}
        \centering \textbf{FS-SINR}
    \end{minipage}%
    \hspace{0.5em}
    \begin{minipage}{0.29\textwidth}
        \centering \textbf{LE-SINR}
    \end{minipage}%
    \hspace{0.5em}
    \begin{minipage}{0.29\textwidth}
        \centering \textbf{SINR}
    \end{minipage}
    
    \vspace{1em}

    \begin{minipage}{0.04\textwidth}
        \rotatebox{90}{\tiny\textbf{0 Context}}
    \end{minipage}%
    \hspace{0.5em}
    \begin{minipage}{0.29\textwidth}
        \centering
                            \begin{overpic}[trim={0 0 0 0},clip,width=\linewidth]{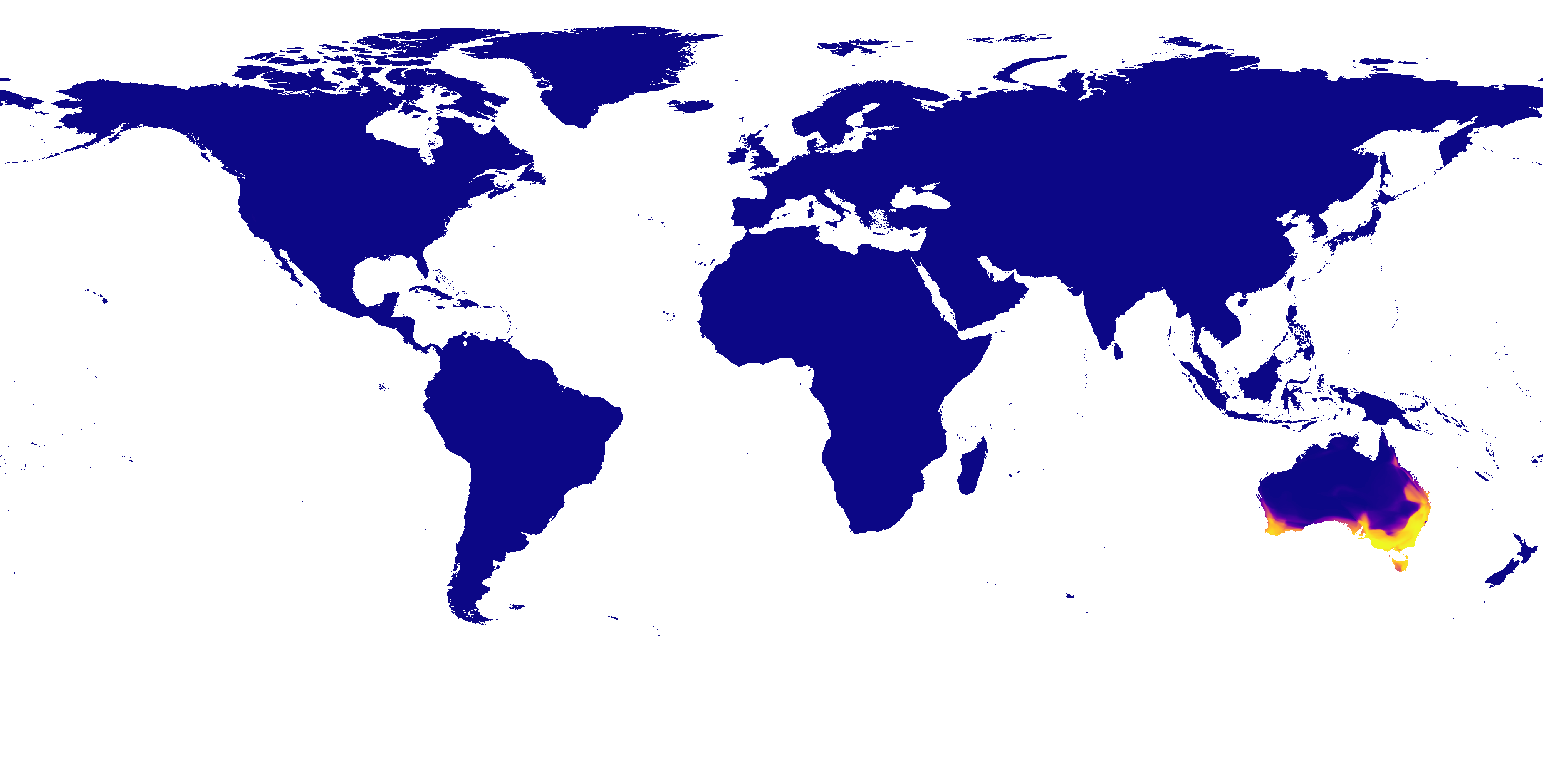}
        \put(0,0){%
          {%
            \setlength\fboxsep{0pt}%
            \setlength\fboxrule{1pt}%
            \fbox{%
              \includegraphics[width=0.25\linewidth]{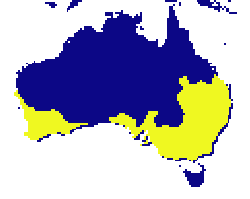}%
            }%
          }%
        }
    \end{overpic}
    \end{minipage}%
    \hspace{0.5em}
    \begin{minipage}{0.29\textwidth}
        \centering
        \includegraphics[width=\linewidth]{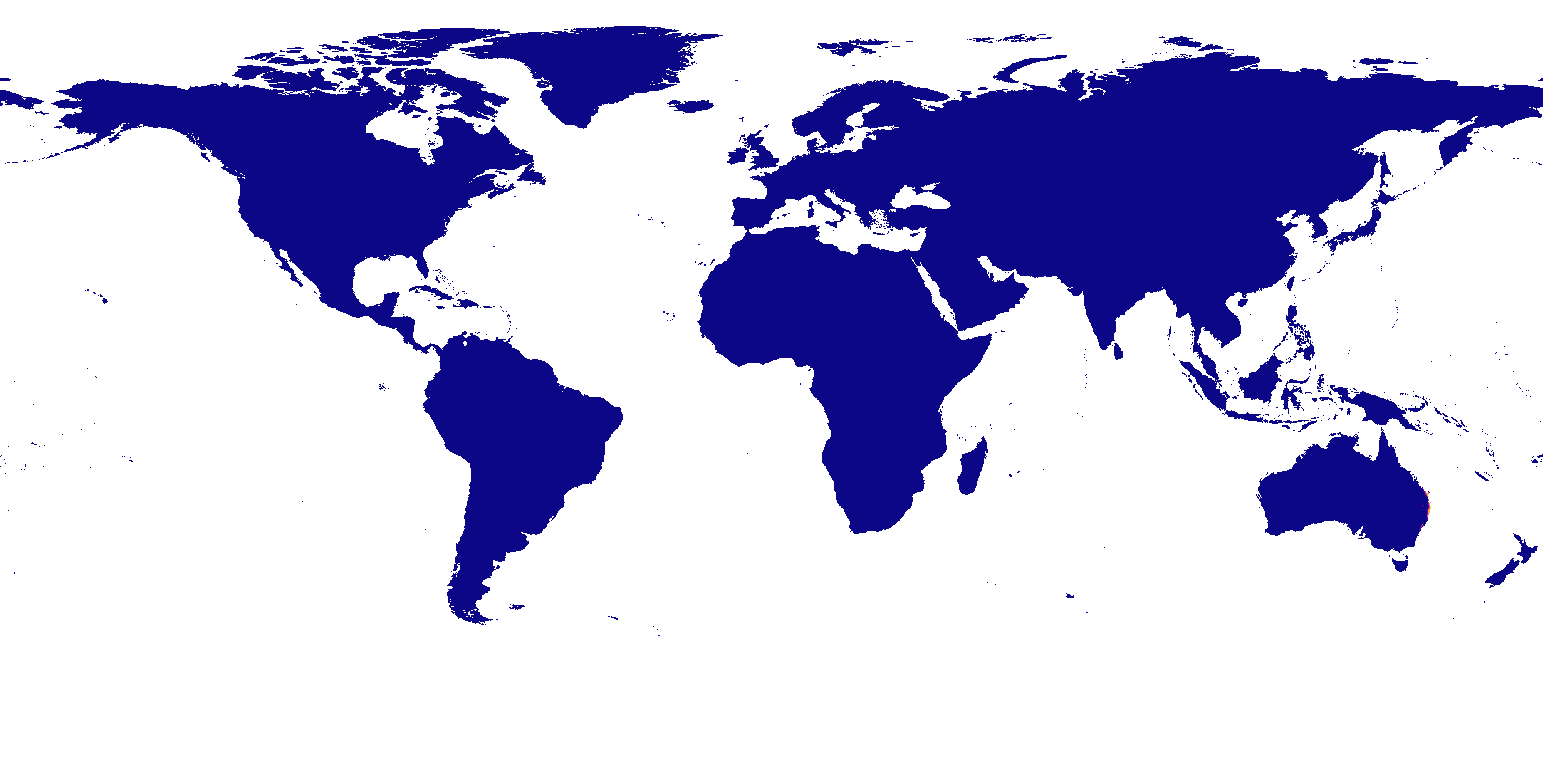}
    \end{minipage}%
    \hspace{0.5em}
    \begin{minipage}{0.29\textwidth}
        \centering
        \includegraphics[width=\linewidth]{figs/raw_map_sm.png}
    \end{minipage}

    \vspace{1em}

    \begin{minipage}{0.04\textwidth}
        \rotatebox{90}{\tiny\textbf{1 Context}}
    \end{minipage}%
    \hspace{0.5em}
    \begin{minipage}{0.29\textwidth}
        \centering
        \includegraphics[width=\linewidth]{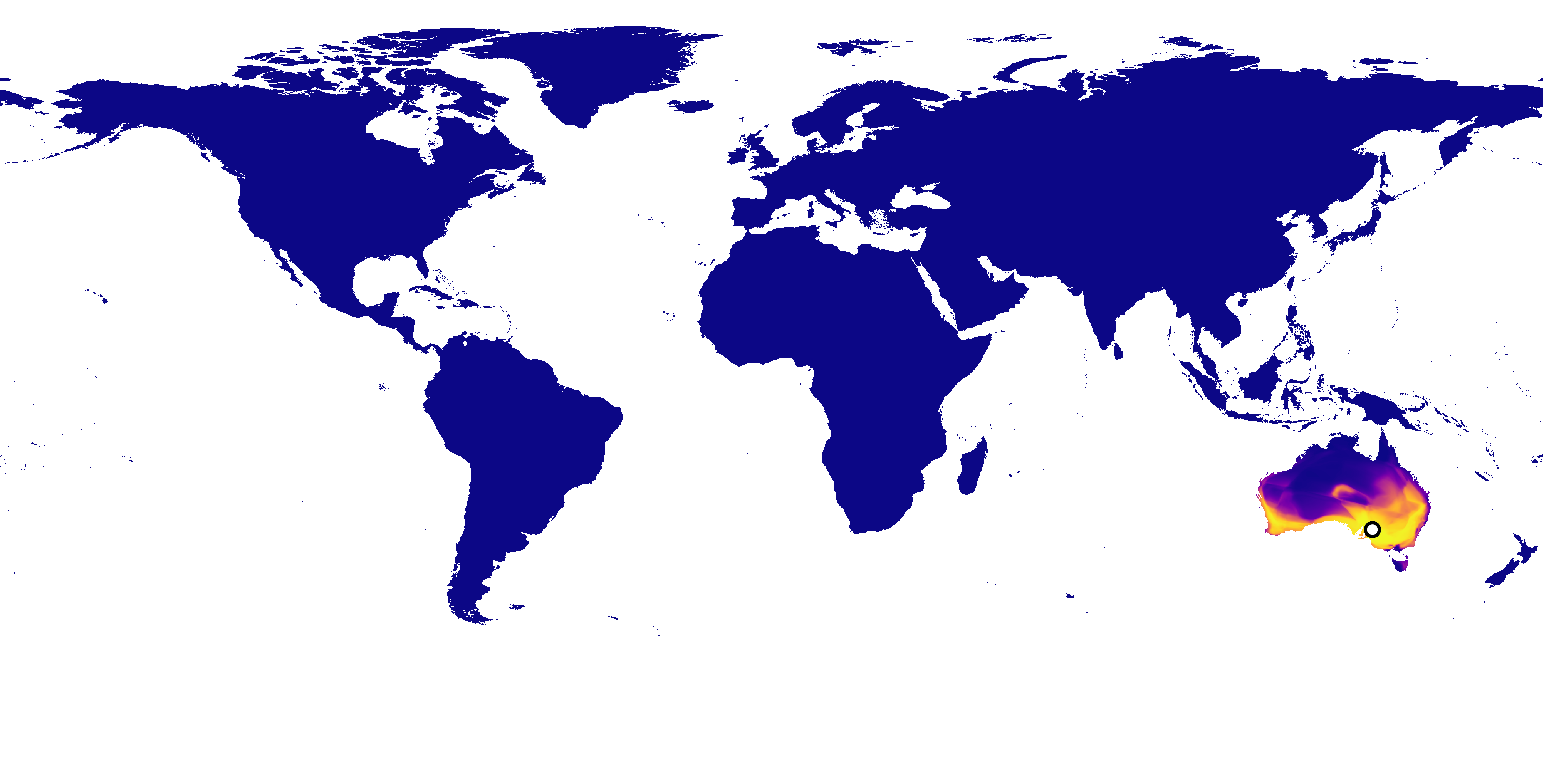}
    \end{minipage}%
    \hspace{0.5em}
    \begin{minipage}{0.29\textwidth}
        \centering
        \includegraphics[width=\linewidth]{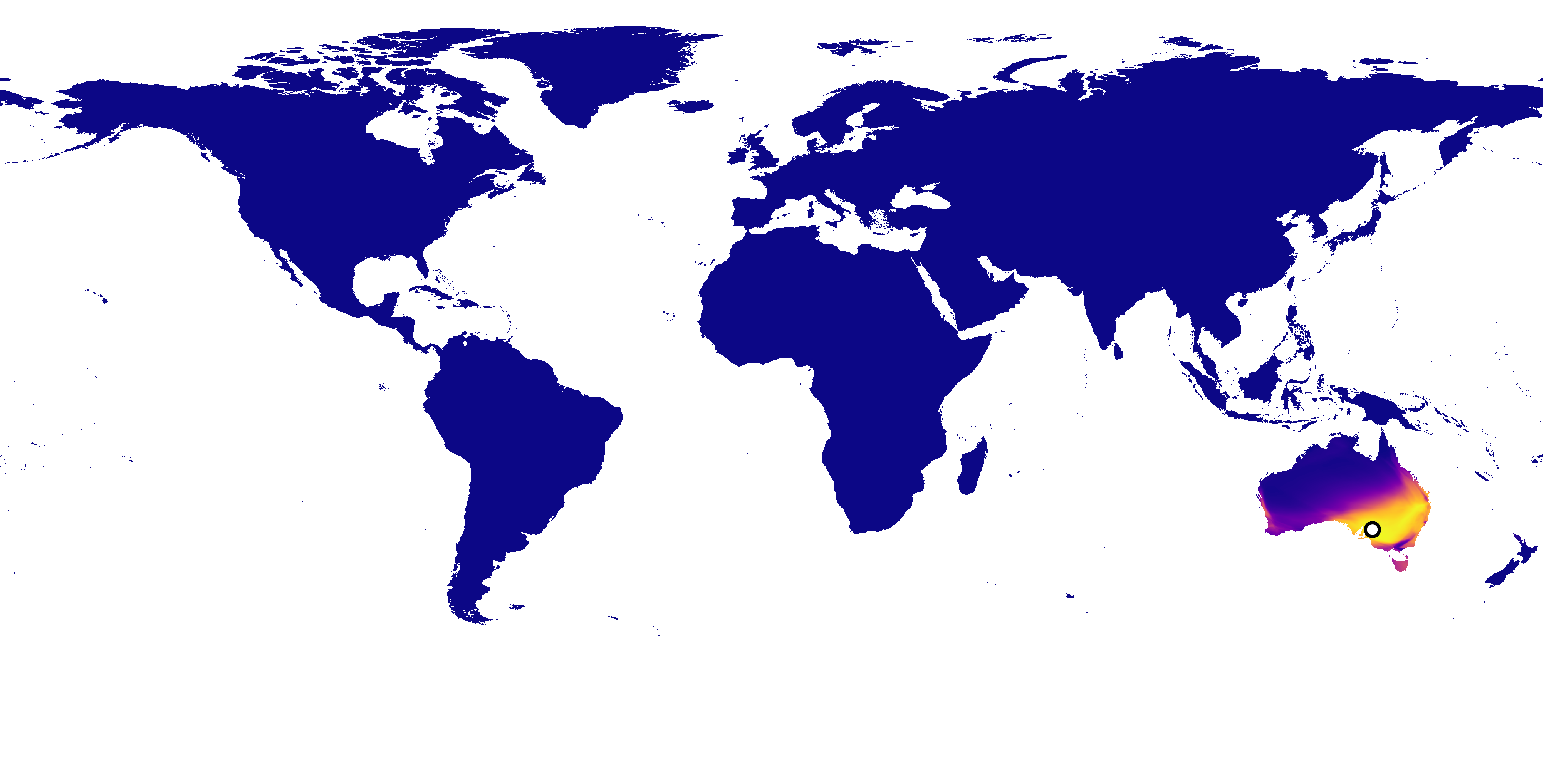}
    \end{minipage}%
    \hspace{0.5em}
    \begin{minipage}{0.29\textwidth}
        \centering
        \includegraphics[width=\linewidth]{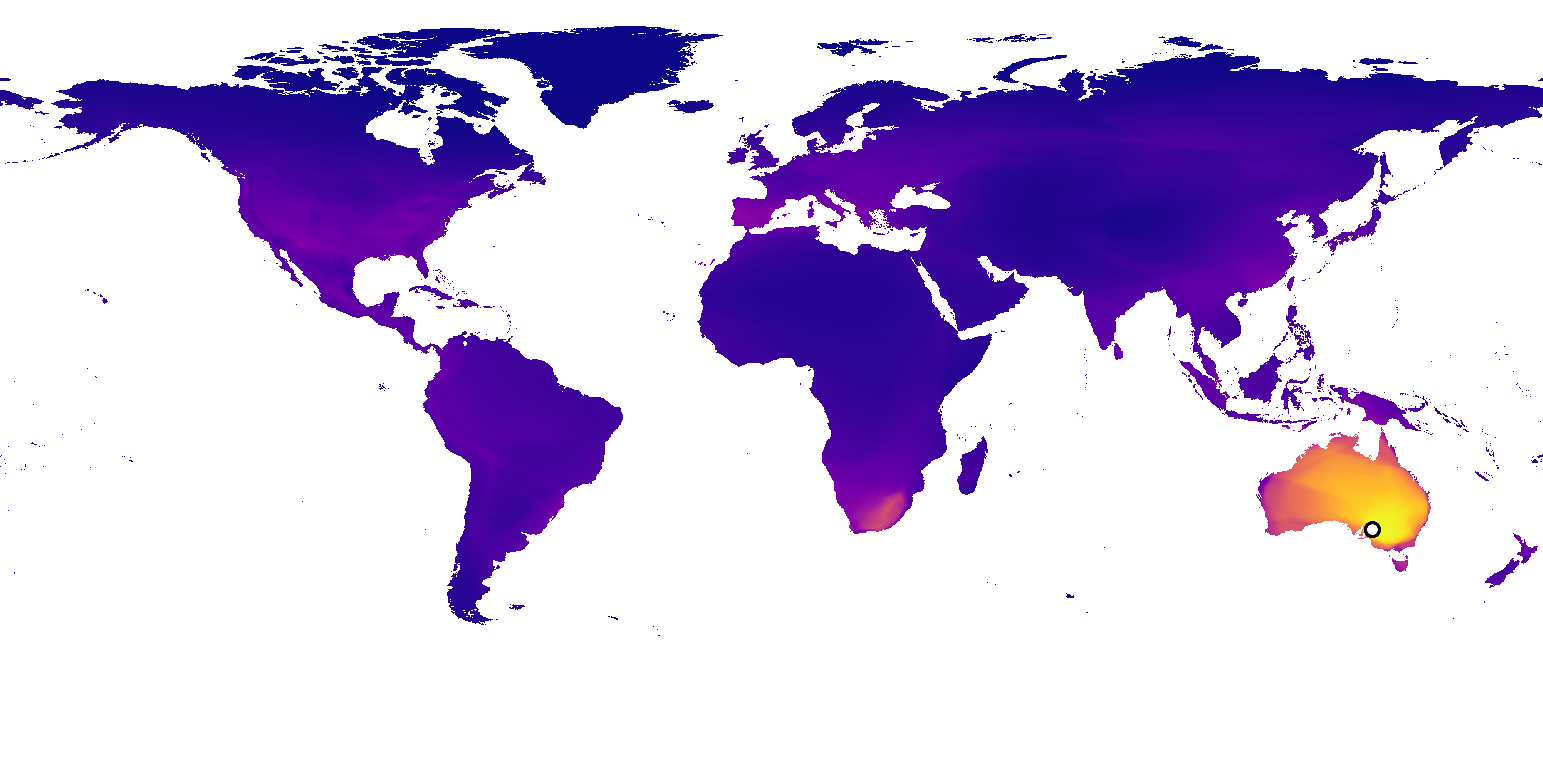}
    \end{minipage}

    \vspace{1em}

    \begin{minipage}{0.04\textwidth}
        \rotatebox{90}{\tiny\textbf{2 Context}}
    \end{minipage}%
    \hspace{0.5em}
    \begin{minipage}{0.29\textwidth}
        \centering
        \includegraphics[width=\linewidth]{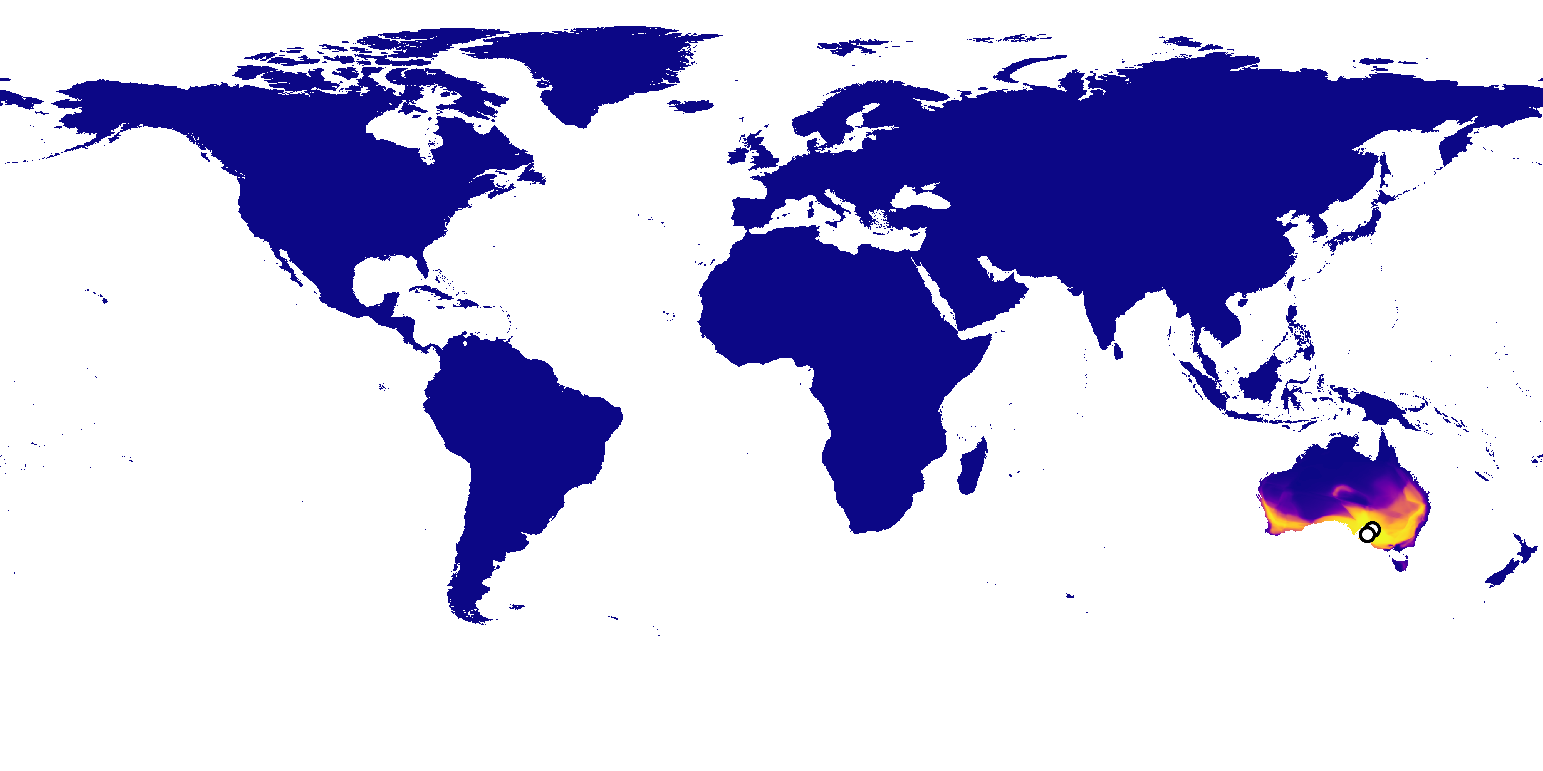}
    \end{minipage}%
    \hspace{0.5em}
    \begin{minipage}{0.29\textwidth}
        \centering
        \includegraphics[width=\linewidth]{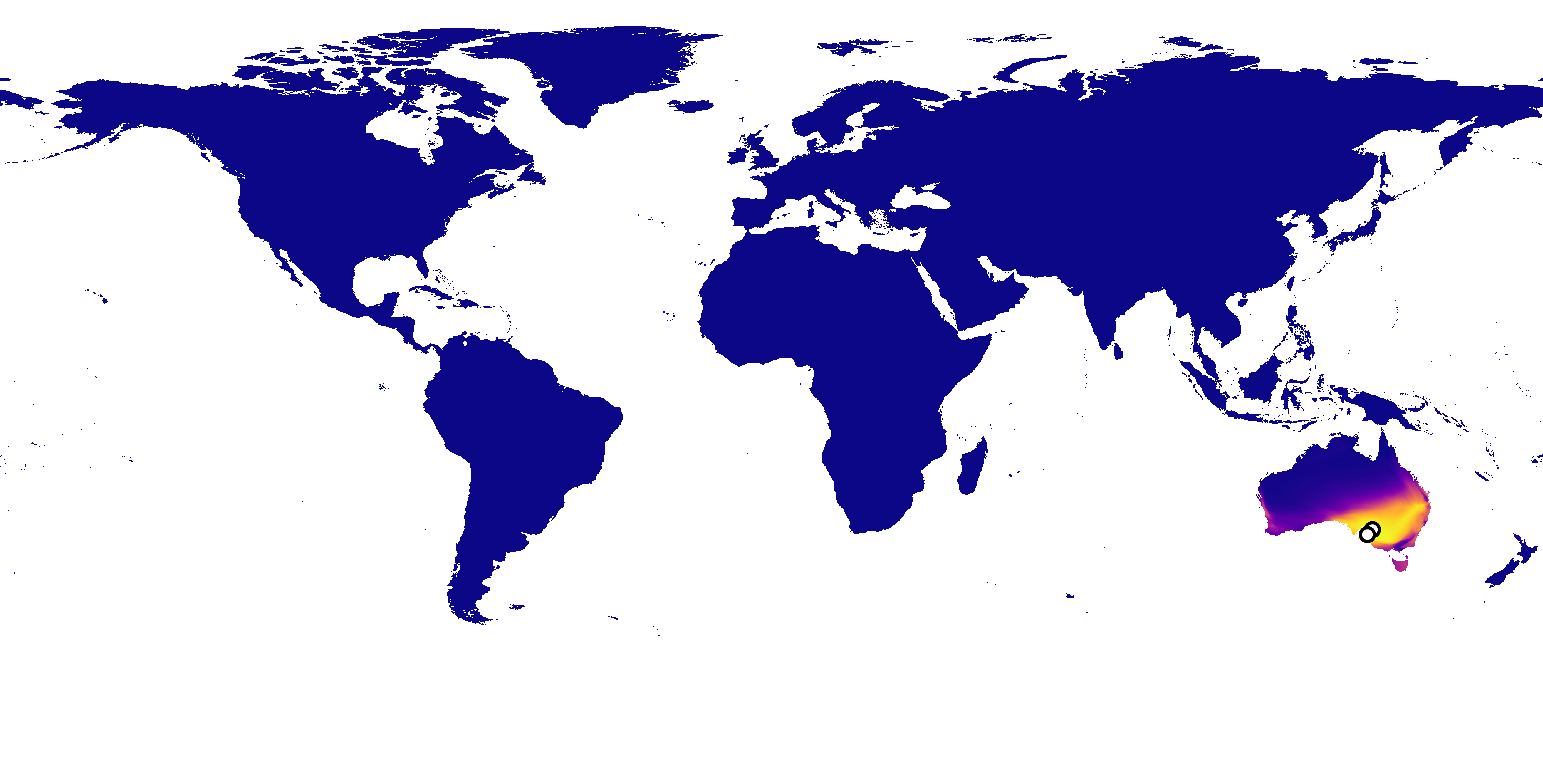}
    \end{minipage}%
    \hspace{0.5em}
    \begin{minipage}{0.29\textwidth}
        \centering
        \includegraphics[width=\linewidth]{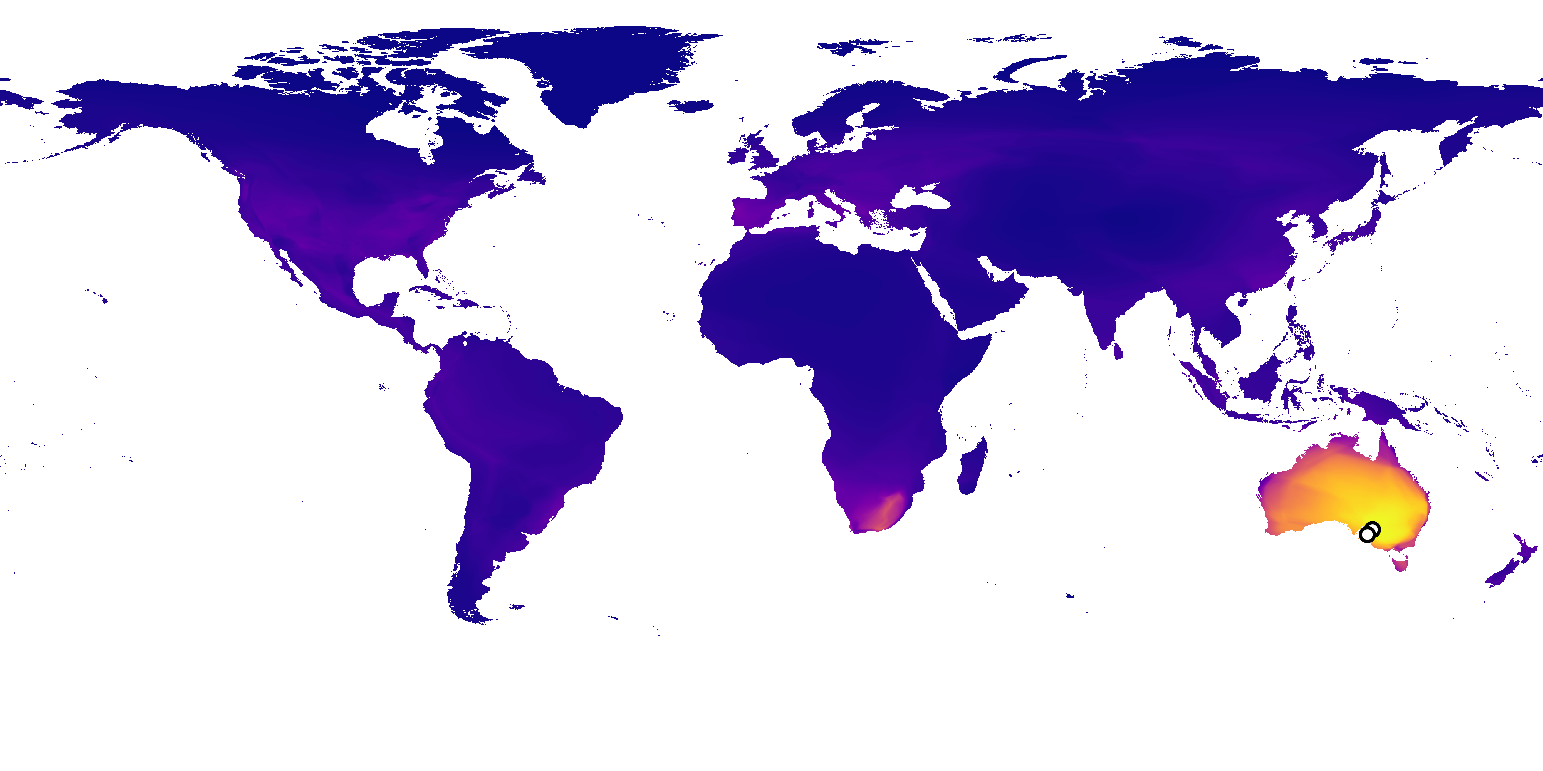}
    \end{minipage}

    \vspace{1em}

    \begin{minipage}{0.04\textwidth}
        \rotatebox{90}{\tiny\textbf{5 Context}}
    \end{minipage}%
    \hspace{0.5em}
    \begin{minipage}{0.29\textwidth}
        \centering
        \includegraphics[width=\linewidth]{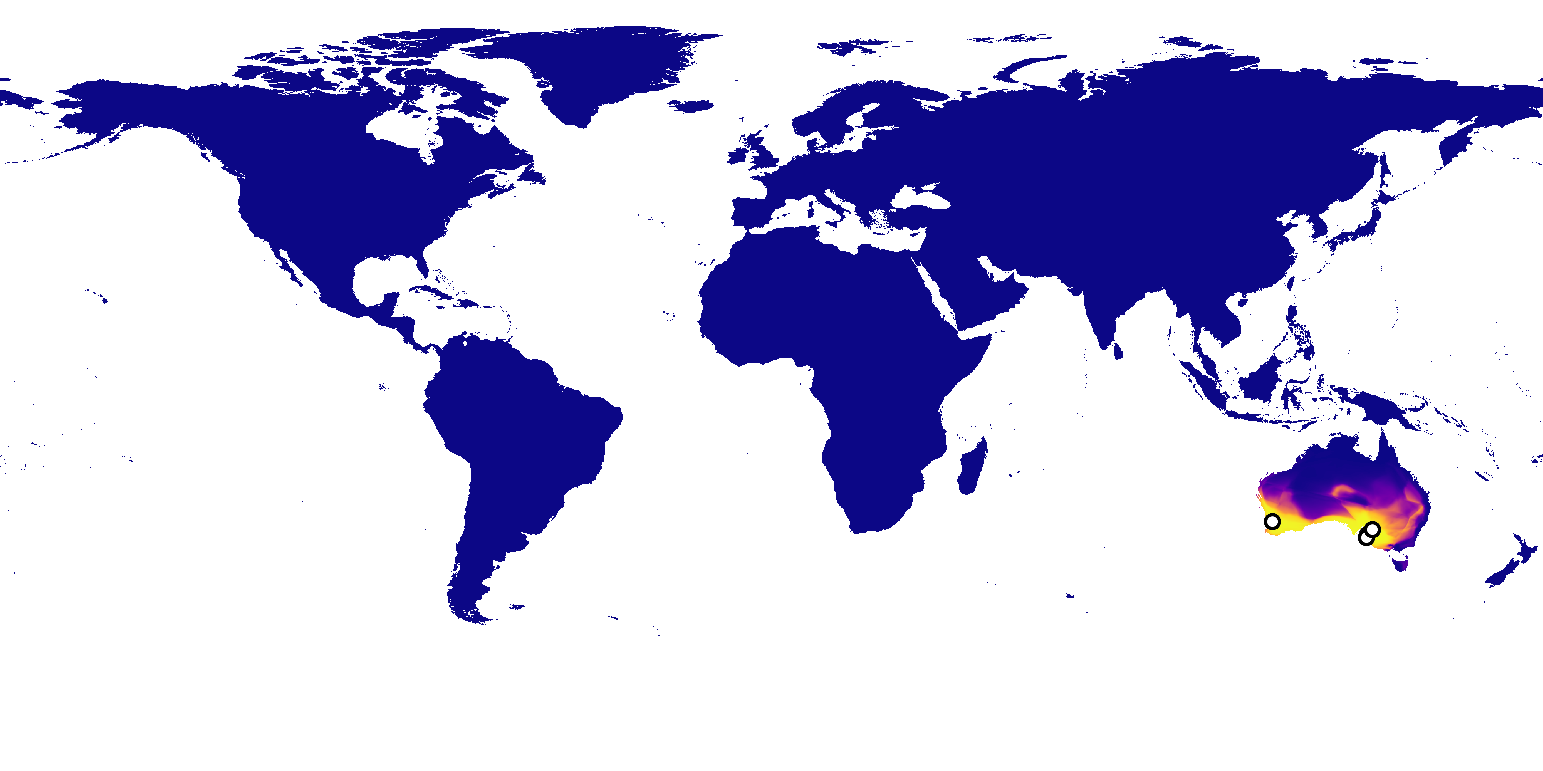}
    \end{minipage}%
    \hspace{0.5em}
    \begin{minipage}{0.29\textwidth}
        \centering
        \includegraphics[width=\linewidth]{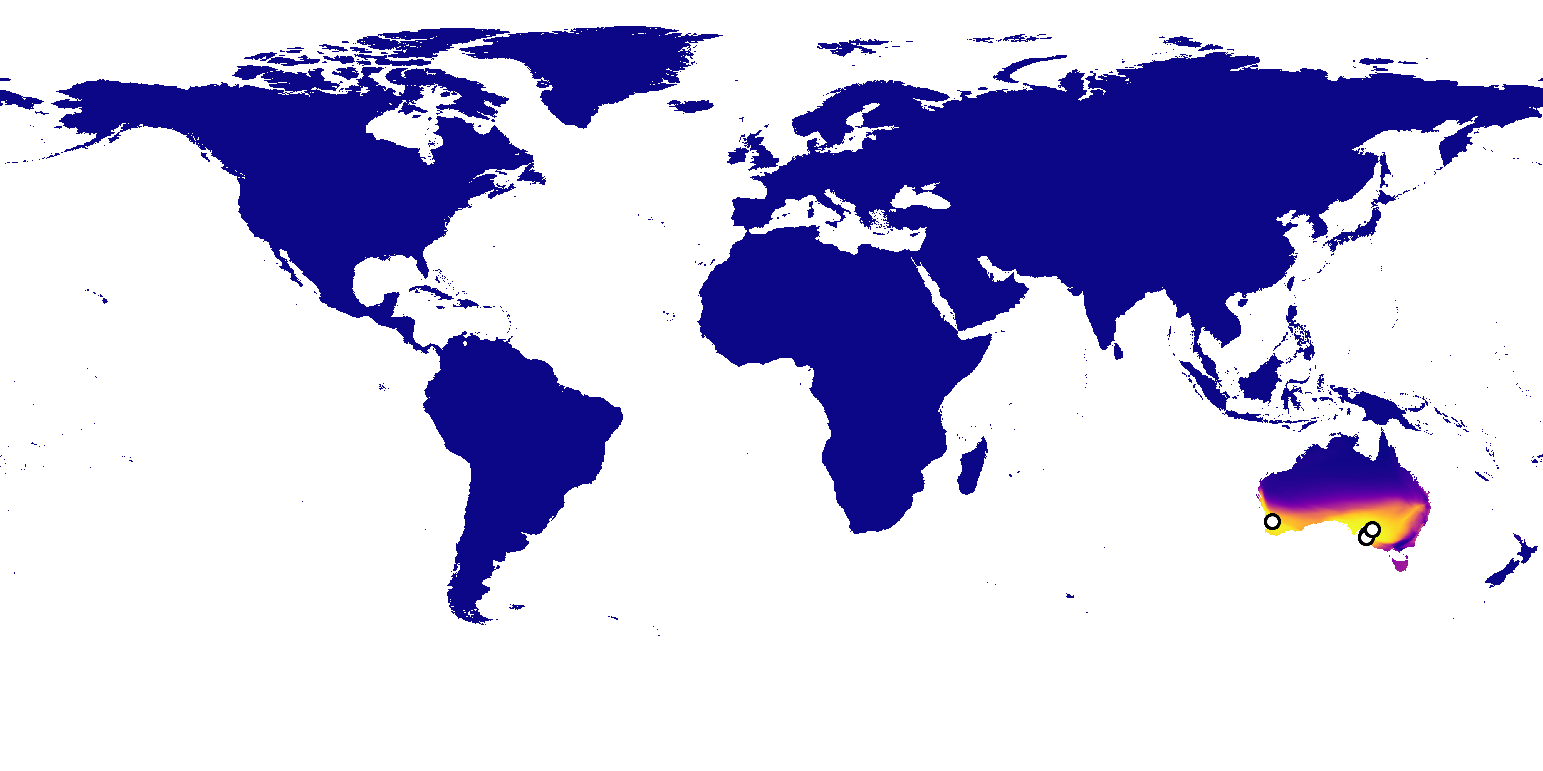}
    \end{minipage}%
    \hspace{0.5em}
    \begin{minipage}{0.29\textwidth}
        \centering
        \includegraphics[width=\linewidth]{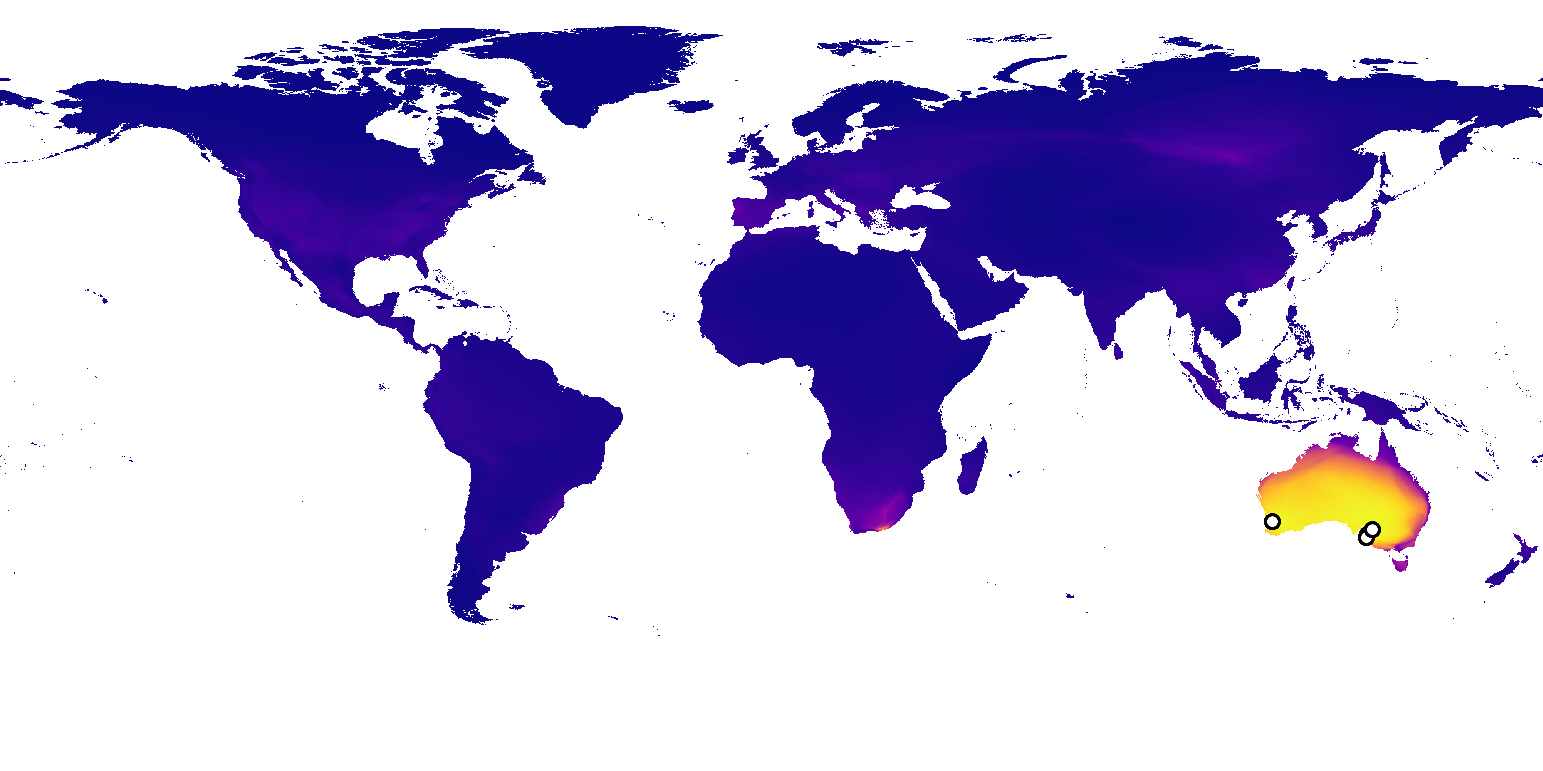}
    \end{minipage}

    \vspace{1em}

    \begin{minipage}{0.04\textwidth}
        \rotatebox{90}{\tiny\textbf{20 Context}}
    \end{minipage}%
    \hspace{0.5em}
    \begin{minipage}{0.29\textwidth}
        \centering
        \includegraphics[width=\linewidth]{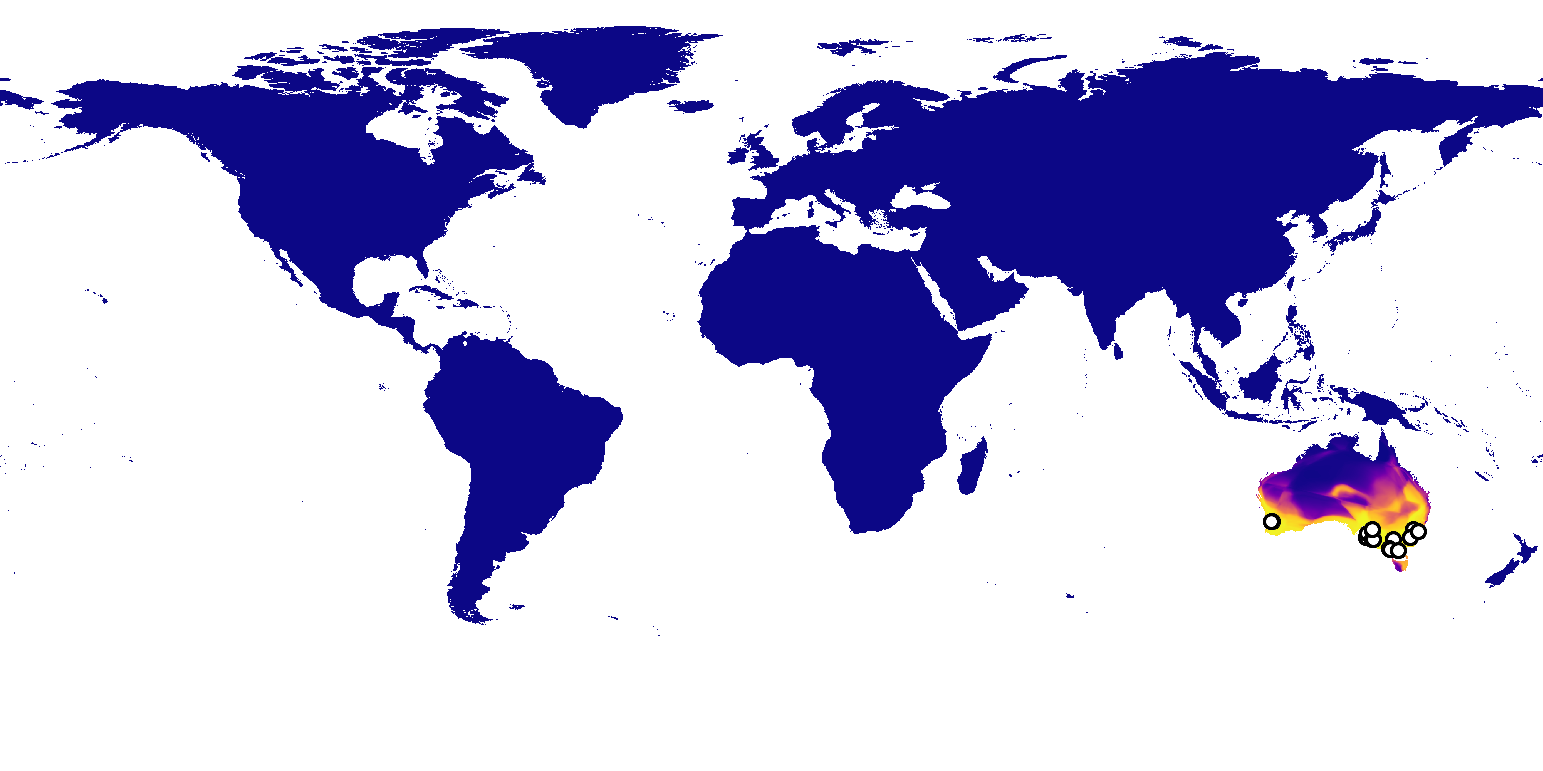}
    \end{minipage}%
    \hspace{0.5em}
    \begin{minipage}{0.29\textwidth}
        \centering
        \includegraphics[width=\linewidth]{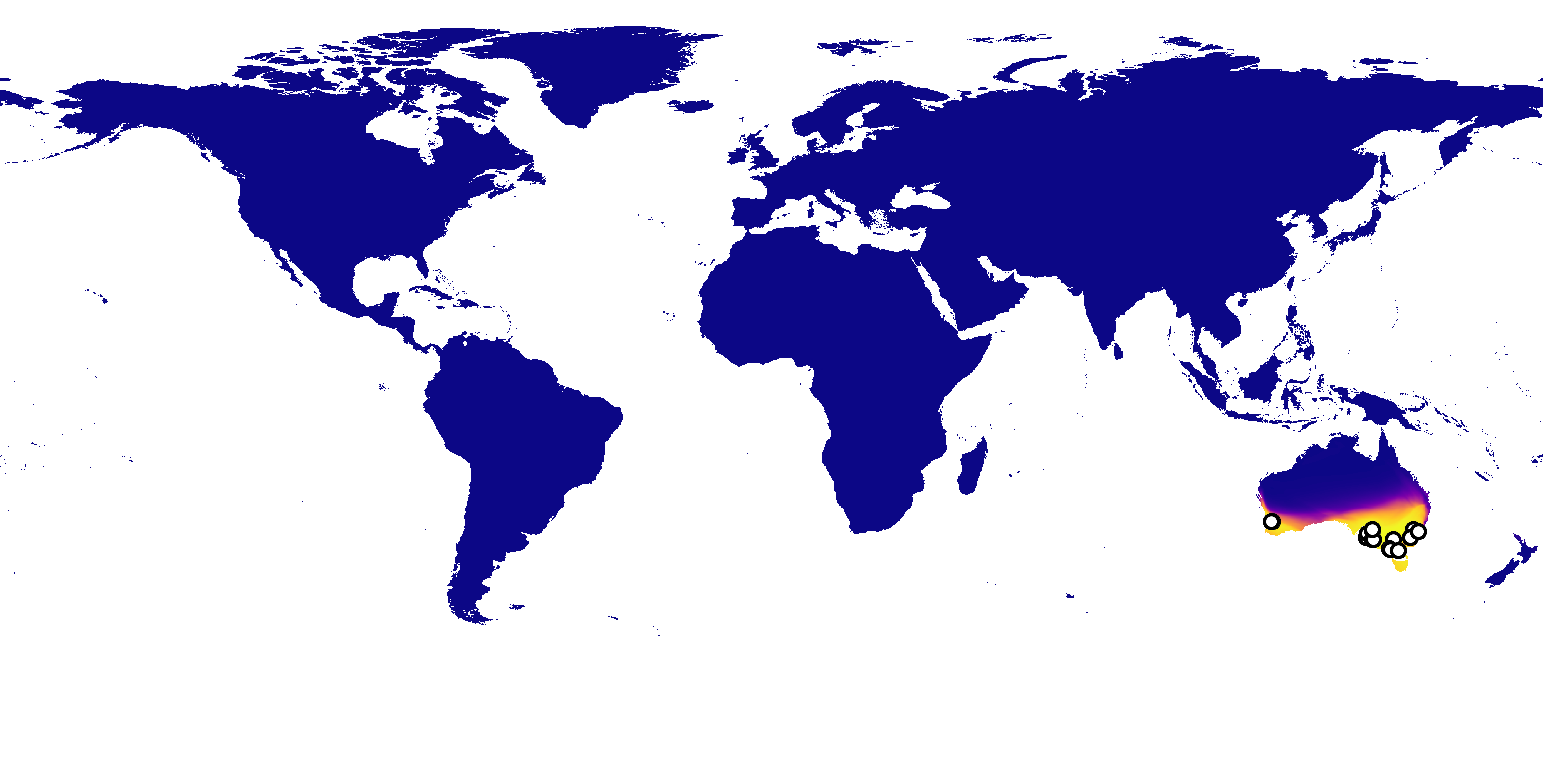}
    \end{minipage}%
    \hspace{0.5em}
    \begin{minipage}{0.29\textwidth}
        \centering
        \includegraphics[width=\linewidth]{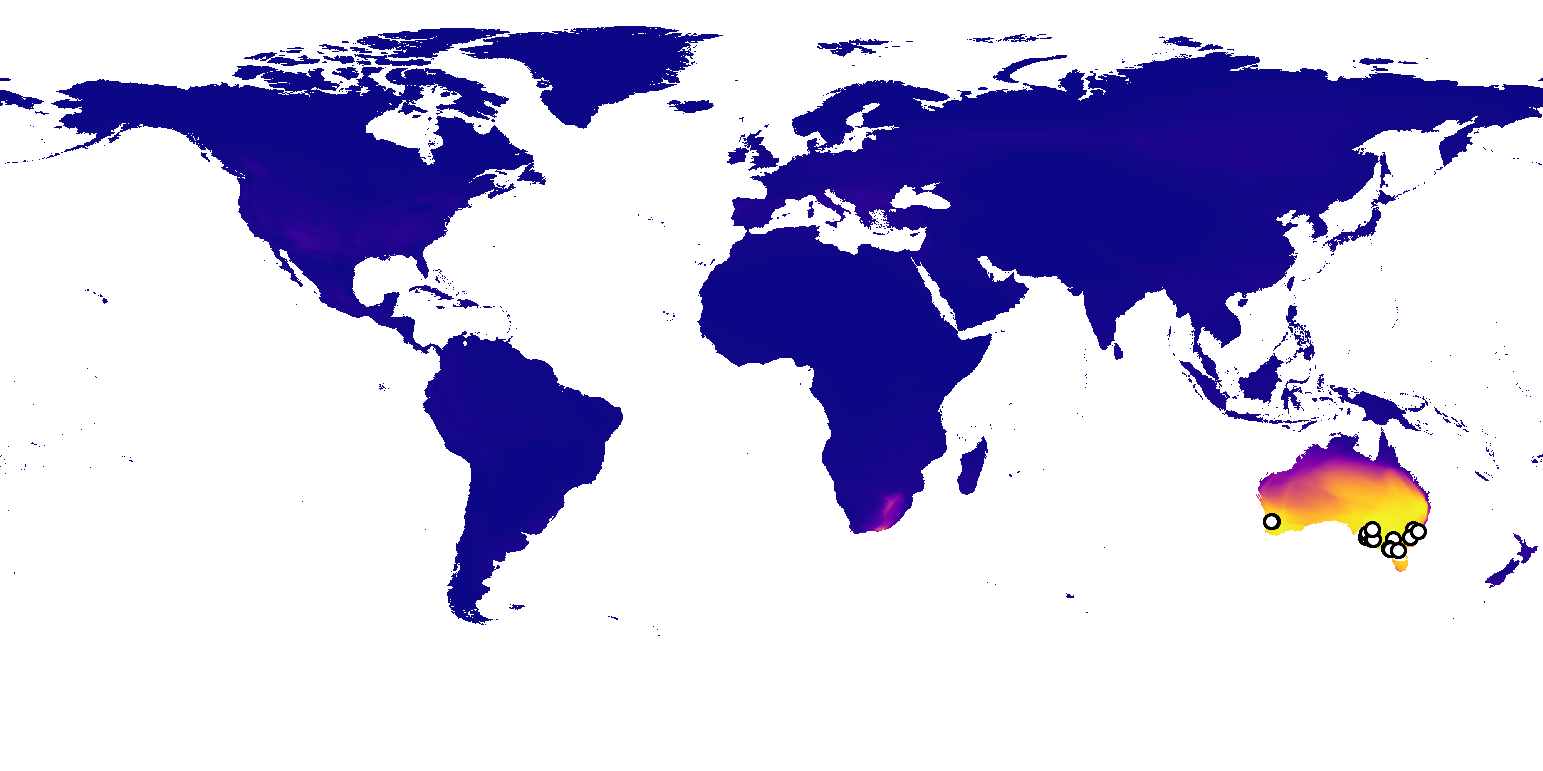}
    \end{minipage}
    \vspace{-15pt}
    \caption{\change{\textbf{Comparing estimated ranges across models with context locations and habitat text.}
    Here we see zero-shot and few-shot range estimates produced by \modelname, LE-SINR, and SINR for the \texttt{Brown-headed Honeyeater}, with expert-derived range inset.
    We provide habitat text to \modelname and LE-SINR as well as context locations, but SINR is not capable of accepting text and so we show a blank map for the zero-shot range estimate.
    We again see LE-SINR underestimate the range using only text, while \modelname has very good zero-shot performance for this species.
    We see that SINR again requires more location data to narrow down the range and even after 20 locations the range is still significantly larger than the other models, extending into South Africa.\\ 
    \textit{Habitat Text:} ``The brown-headed honeyeater inhabits temperate forests and Mediterranean-type shrubby vegetation. It is typically found in tall trees, where it forages by probing in the bark of trunks and branches.''}}
    \label{fig:qualitative_different_models_habitat}
\end{figure}

\begin{figure}[h]
    \centering
    \begin{minipage}{0.04\textwidth}
    \end{minipage}%
    \hspace{0.5em}
    \begin{minipage}{0.29\textwidth}
        \centering \textbf{FS-SINR}
    \end{minipage}%
    \hspace{0.5em}
    \begin{minipage}{0.29\textwidth}
        \centering \textbf{LE-SINR}
    \end{minipage}%
    \hspace{0.5em}
    \begin{minipage}{0.29\textwidth}
        \centering \textbf{SINR}
    \end{minipage}
    
    \vspace{1em}

    \begin{minipage}{0.04\textwidth}
        \rotatebox{90}{\tiny\textbf{0 Context}}
    \end{minipage}%
    \hspace{0.5em}
    \begin{minipage}{0.29\textwidth}
        \centering
                    \begin{overpic}[trim={0 0 0 0},clip,width=\linewidth]{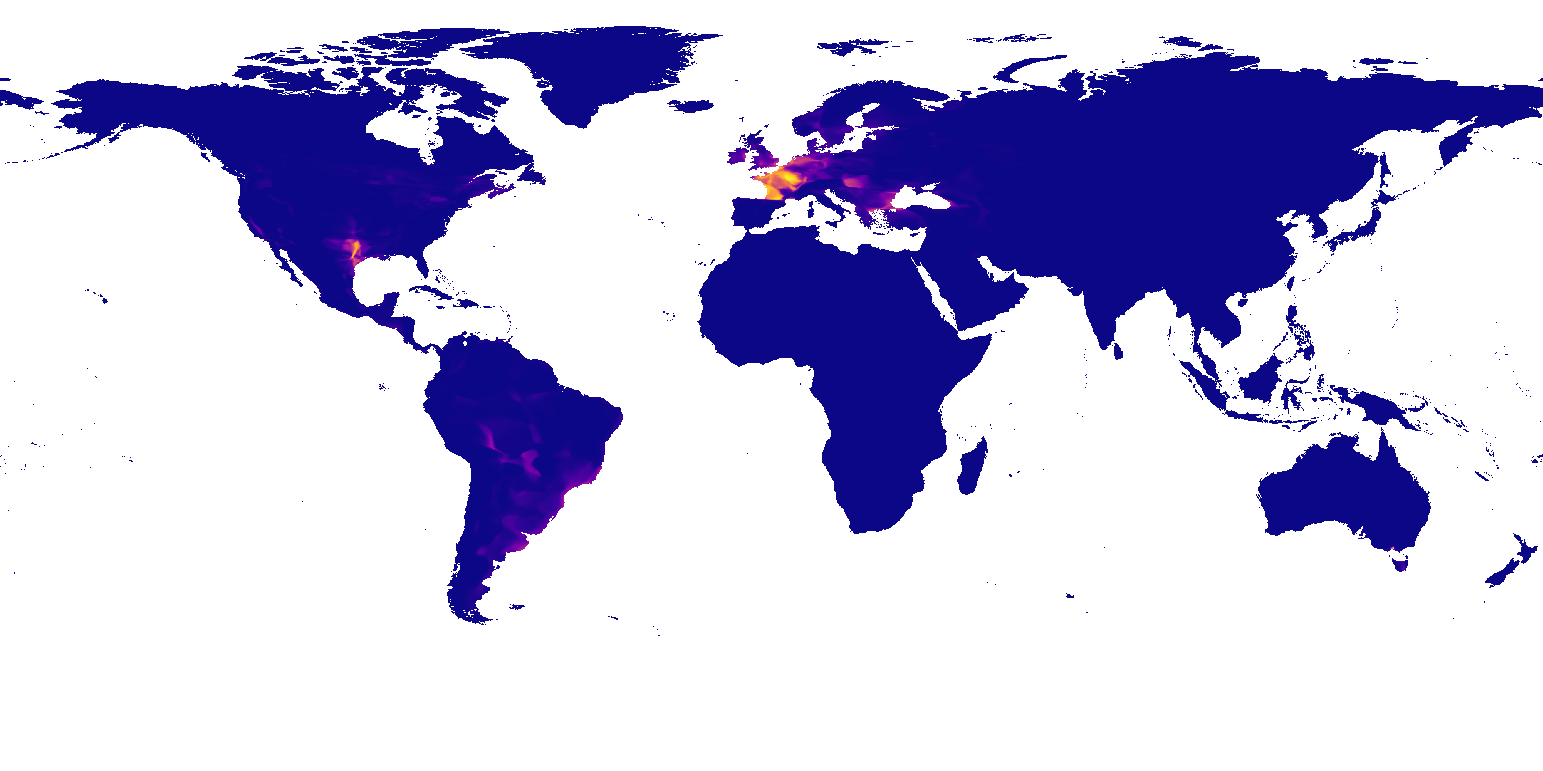}
        \put(0,0){%
          {%
            \setlength\fboxsep{0pt}%
            \setlength\fboxrule{1pt}%
            \fbox{%
              \includegraphics[width=0.25\linewidth]{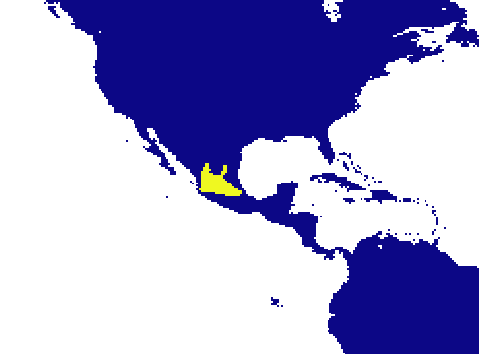}%
            }%
          }%
        }
    \end{overpic}
    \end{minipage}%
    \hspace{0.5em}
    \begin{minipage}{0.29\textwidth}
        \centering
        \includegraphics[width=\linewidth]{figs/raw_map_sm.png}
    \end{minipage}%
    \hspace{0.5em}
    \begin{minipage}{0.29\textwidth}
        \centering
        \includegraphics[width=\linewidth]{figs/raw_map_sm.png}
    \end{minipage}

    \vspace{1em}

    \begin{minipage}{0.04\textwidth}
        \rotatebox{90}{\tiny\textbf{1 Context}}
    \end{minipage}%
    \hspace{0.5em}
    \begin{minipage}{0.29\textwidth}
        \centering
        \includegraphics[width=\linewidth]{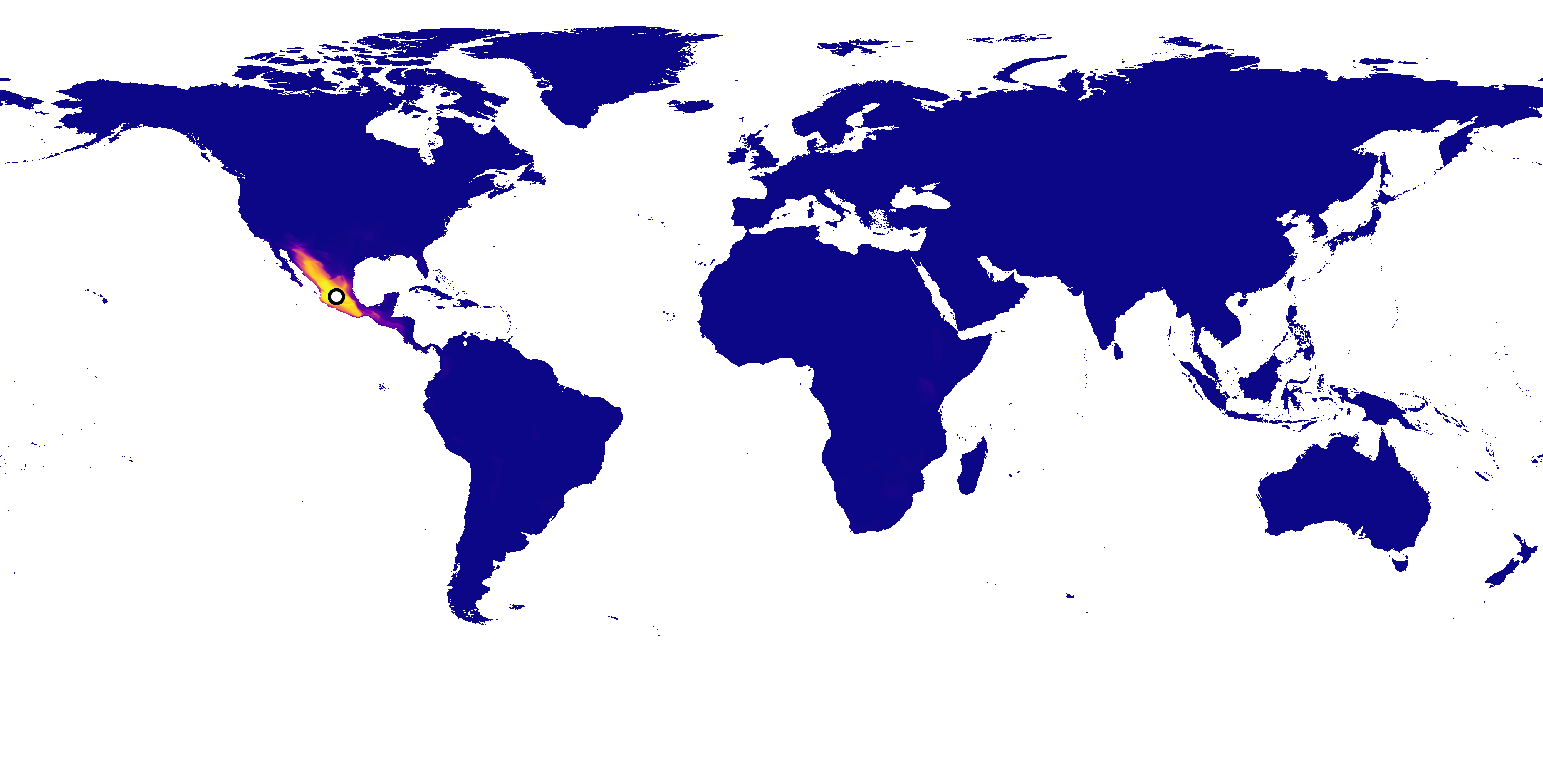}
    \end{minipage}%
    \hspace{0.5em}
    \begin{minipage}{0.29\textwidth}
        \centering
        \includegraphics[width=\linewidth]{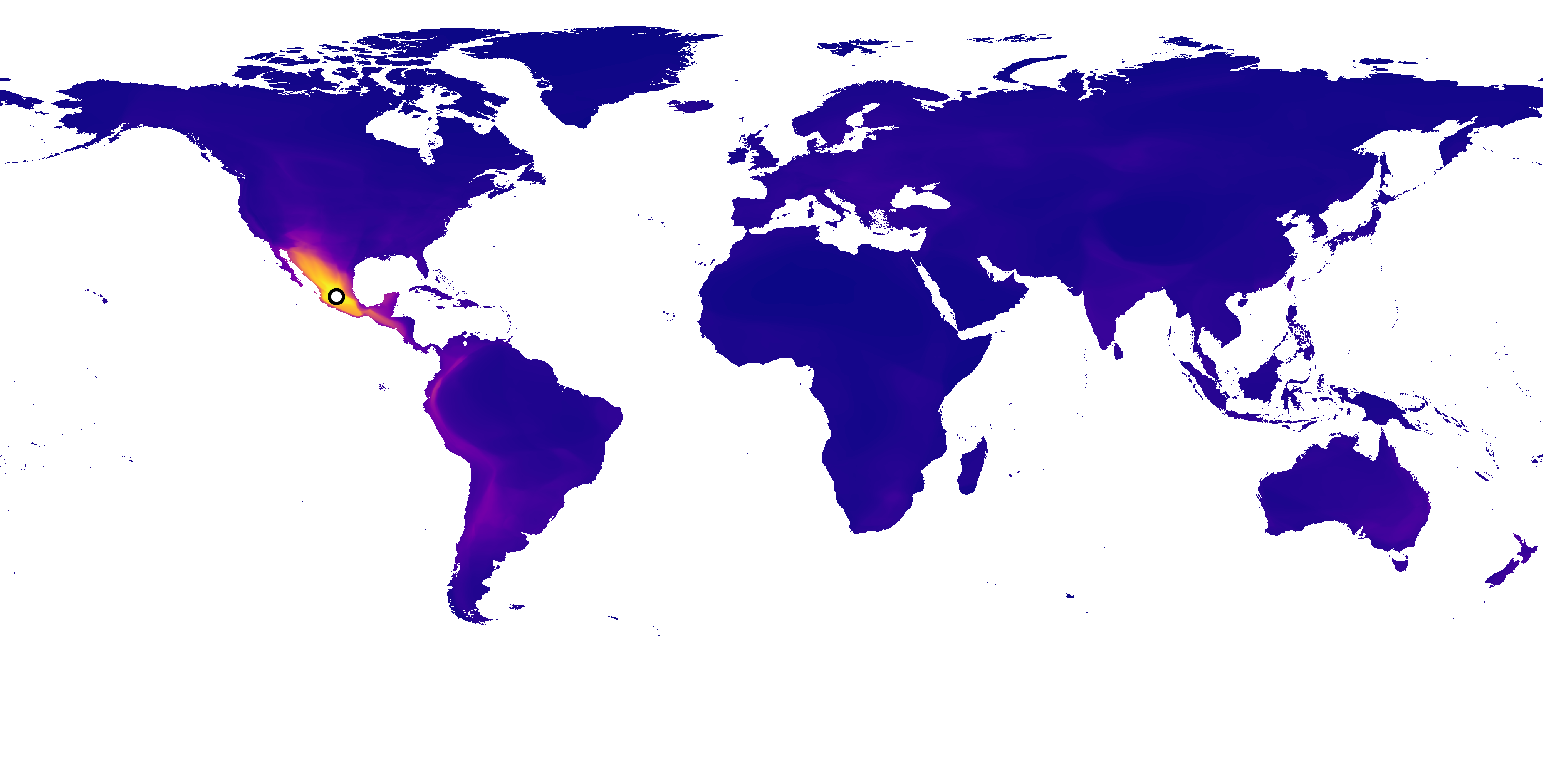}
    \end{minipage}%
    \hspace{0.5em}
    \begin{minipage}{0.29\textwidth}
        \centering
        \includegraphics[width=\linewidth]{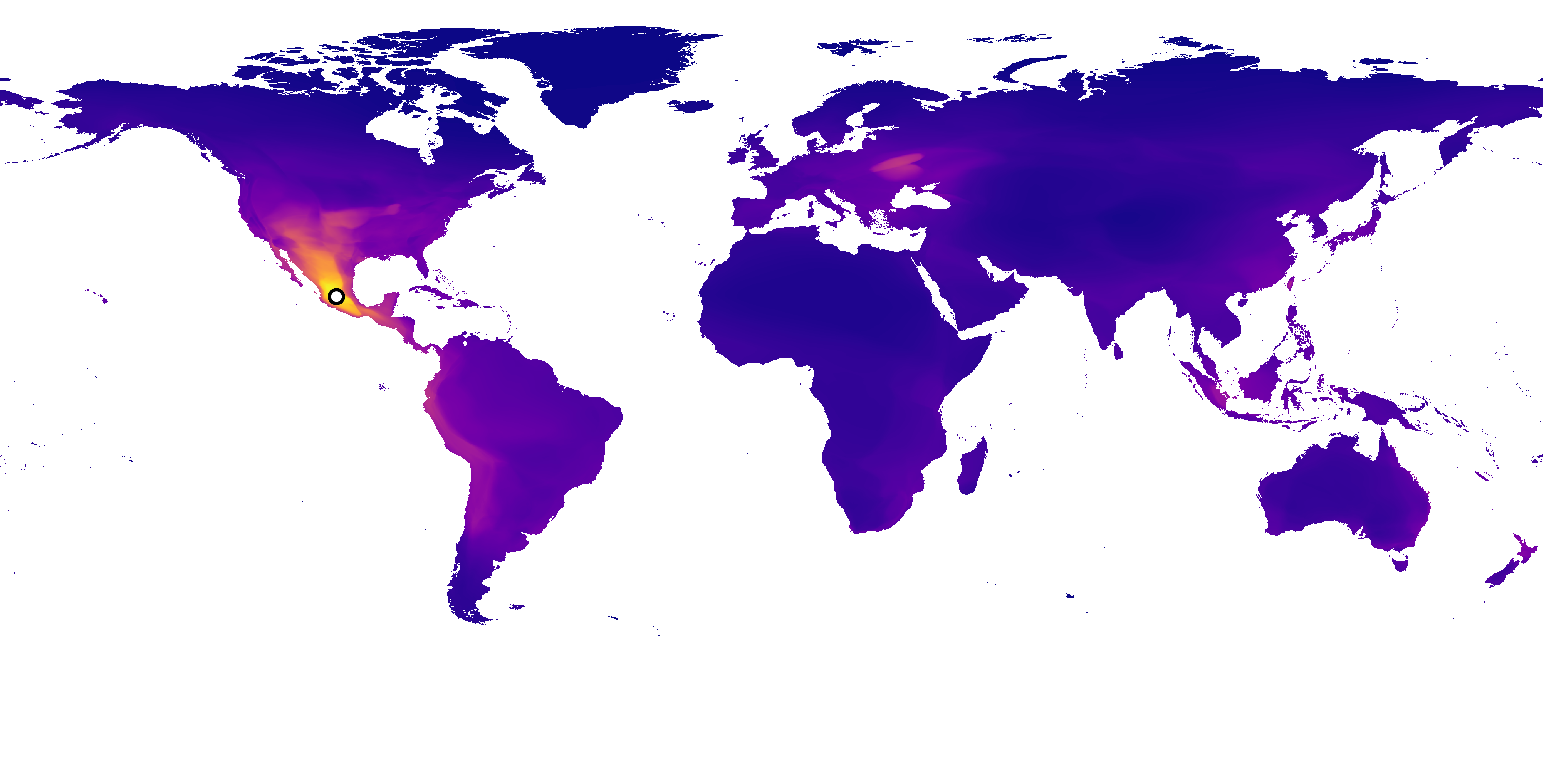}
    \end{minipage}

    \vspace{1em}

    \begin{minipage}{0.04\textwidth}
        \rotatebox{90}{\tiny\textbf{2 Context}}
    \end{minipage}%
    \hspace{0.5em}
    \begin{minipage}{0.29\textwidth}
        \centering
        \includegraphics[width=\linewidth]{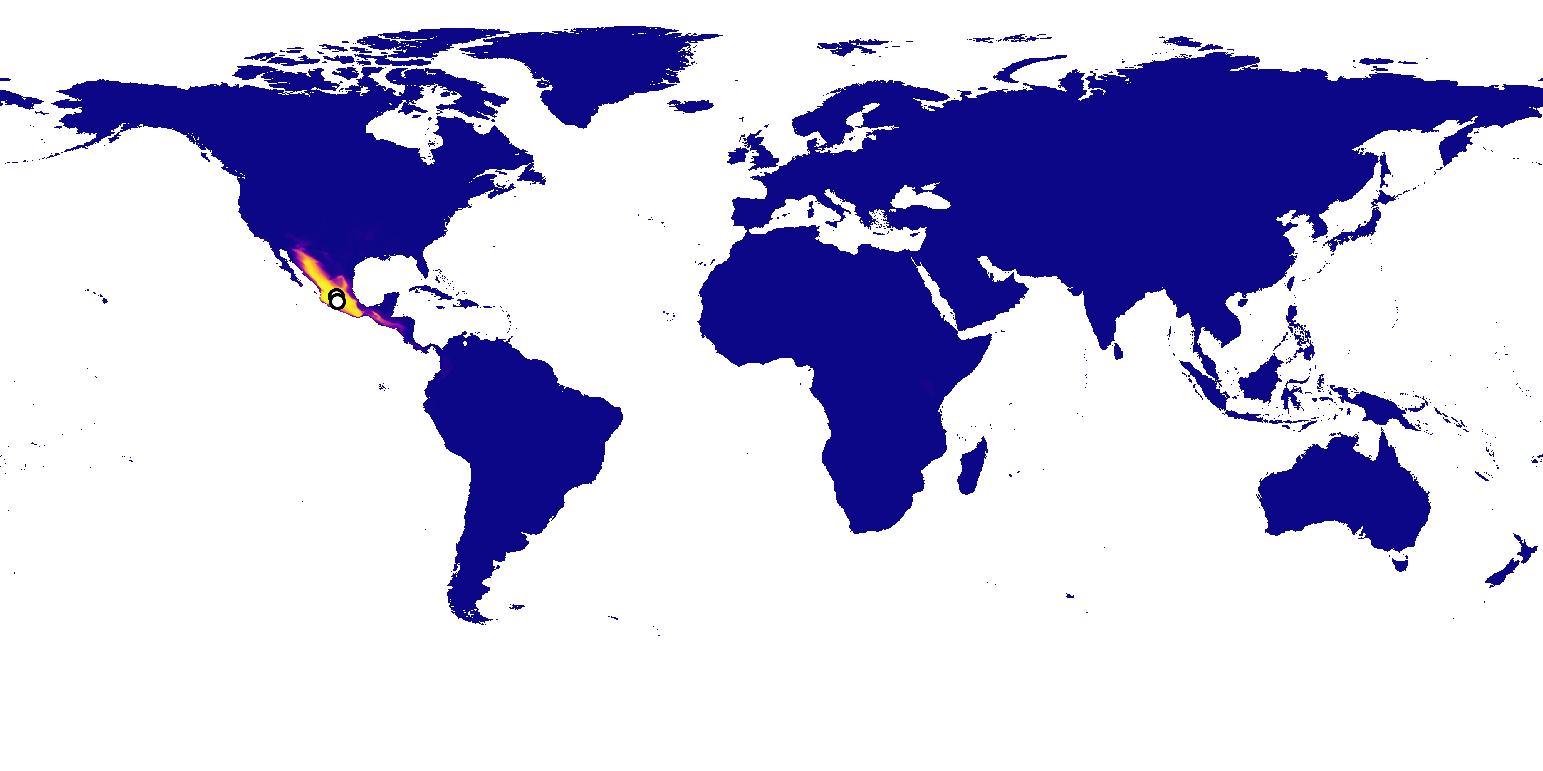}
    \end{minipage}%
    \hspace{0.5em}
    \begin{minipage}{0.29\textwidth}
        \centering
        \includegraphics[width=\linewidth]{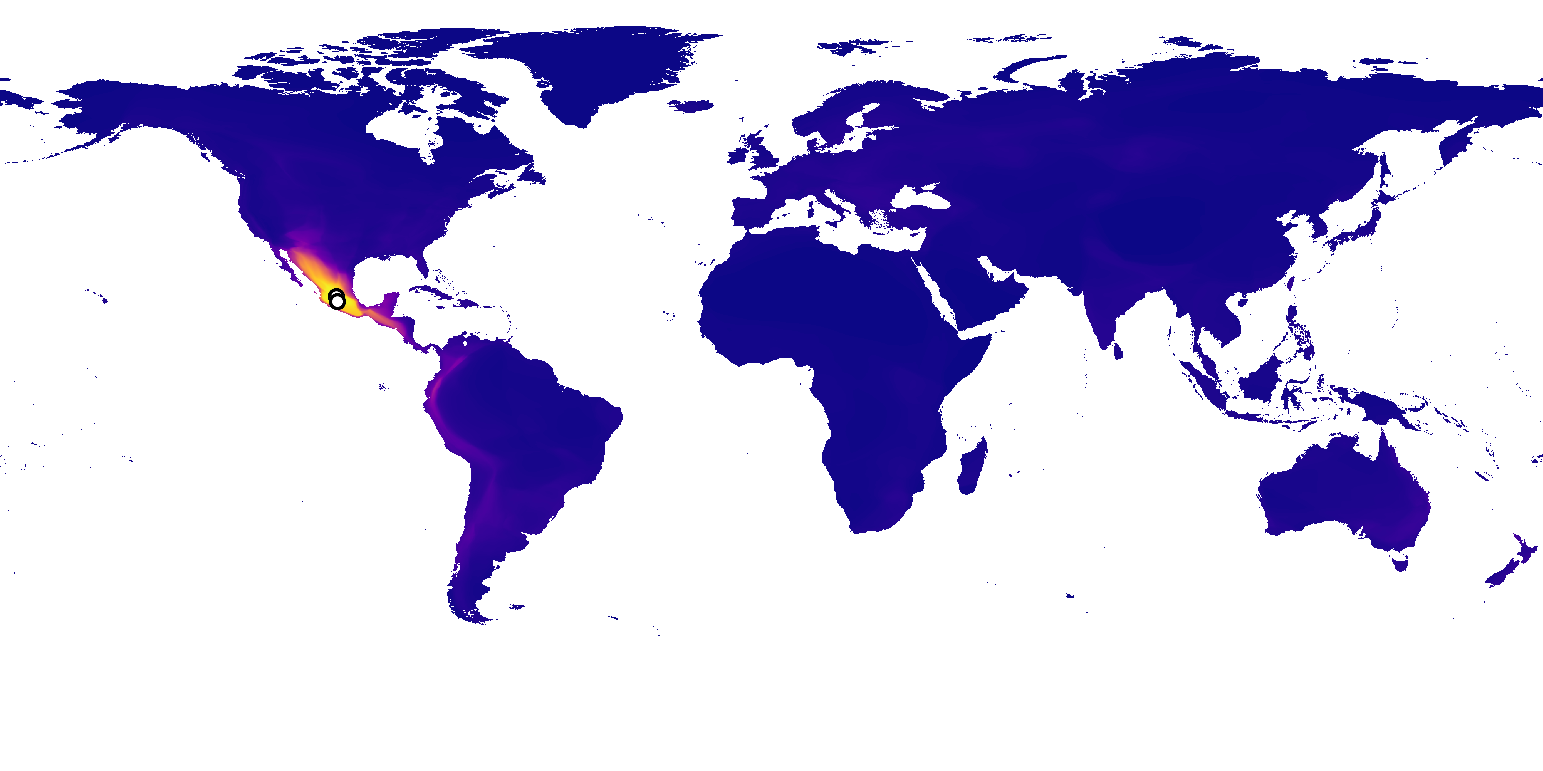}
    \end{minipage}%
    \hspace{0.5em}
    \begin{minipage}{0.29\textwidth}
        \centering
        \includegraphics[width=\linewidth]{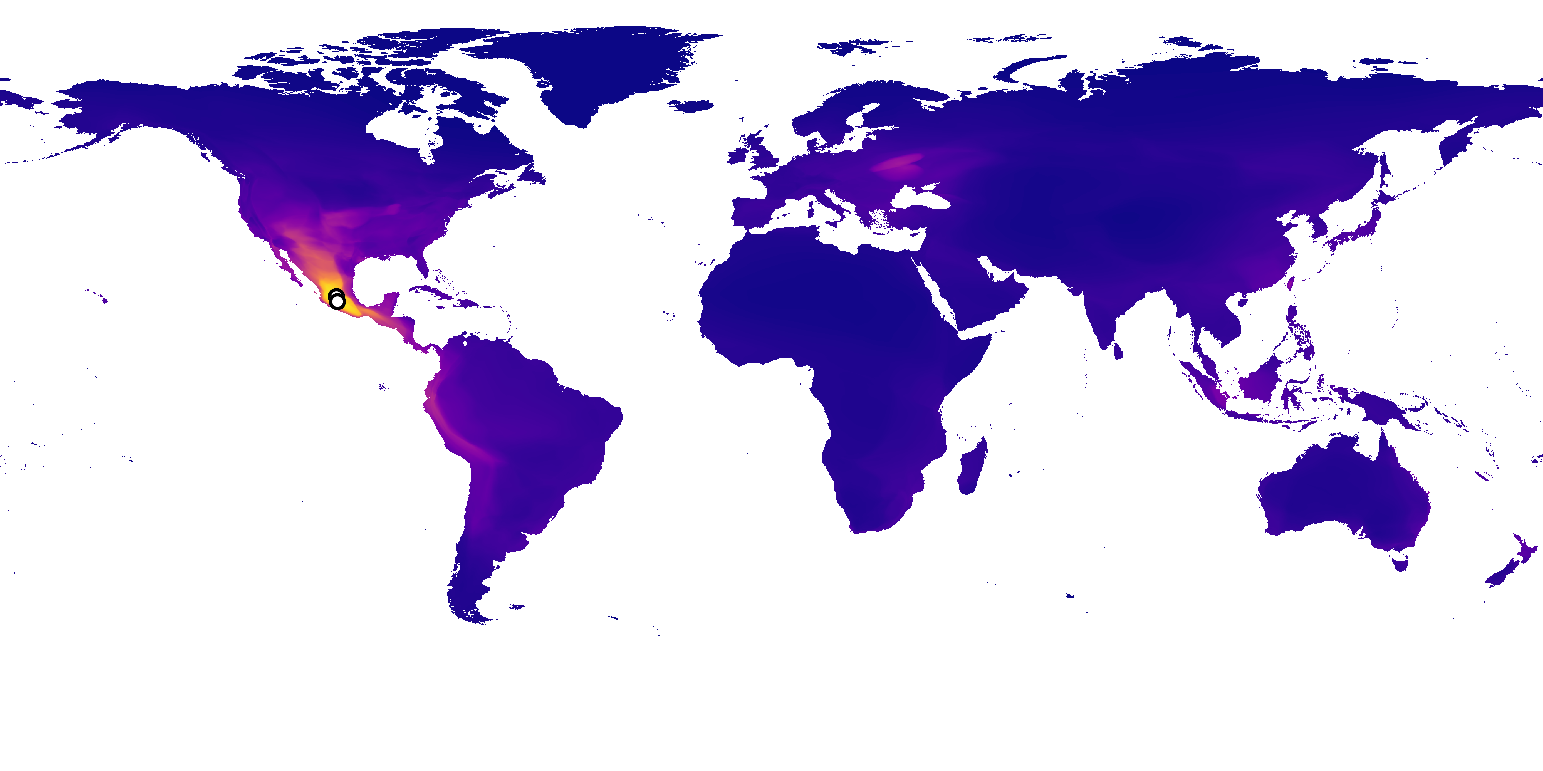}
    \end{minipage}

    \vspace{1em}

    \begin{minipage}{0.04\textwidth}
        \rotatebox{90}{\tiny\textbf{5 Context}}
    \end{minipage}%
    \hspace{0.5em}
    \begin{minipage}{0.29\textwidth}
        \centering
        \includegraphics[width=\linewidth]{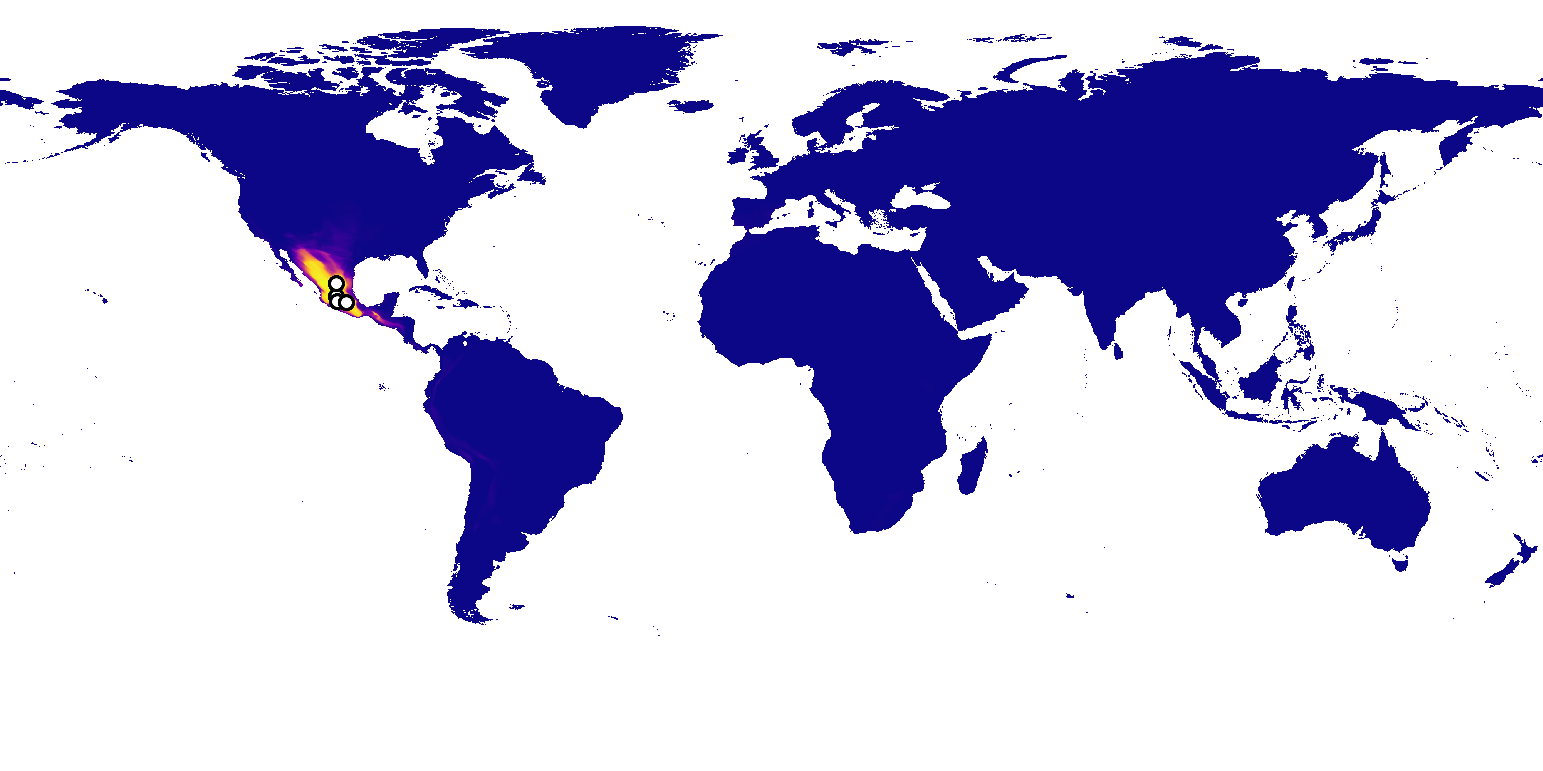}
    \end{minipage}%
    \hspace{0.5em}
    \begin{minipage}{0.29\textwidth}
        \centering
        \includegraphics[width=\linewidth]{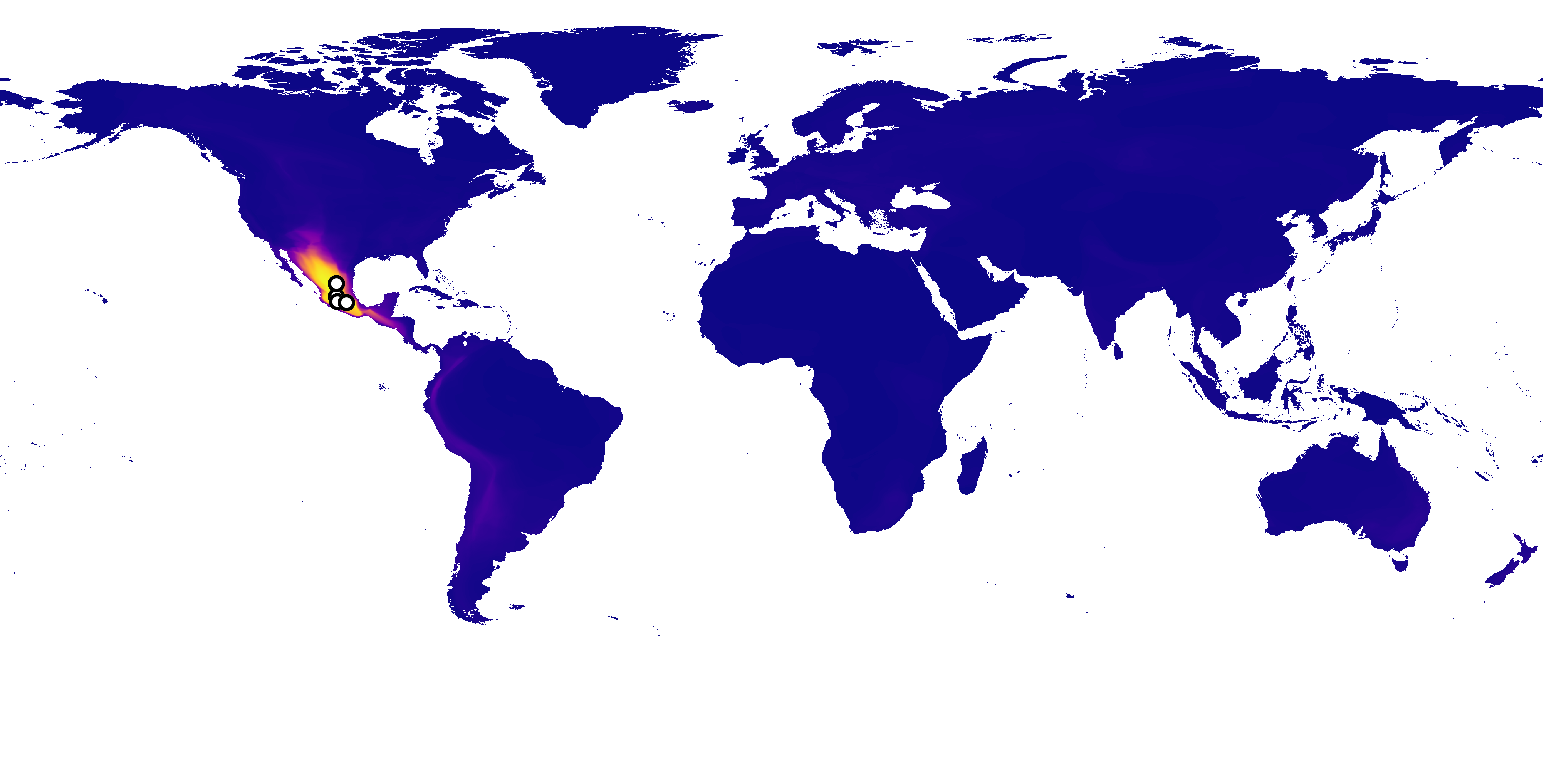}
    \end{minipage}%
    \hspace{0.5em}
    \begin{minipage}{0.29\textwidth}
        \centering
        \includegraphics[width=\linewidth]{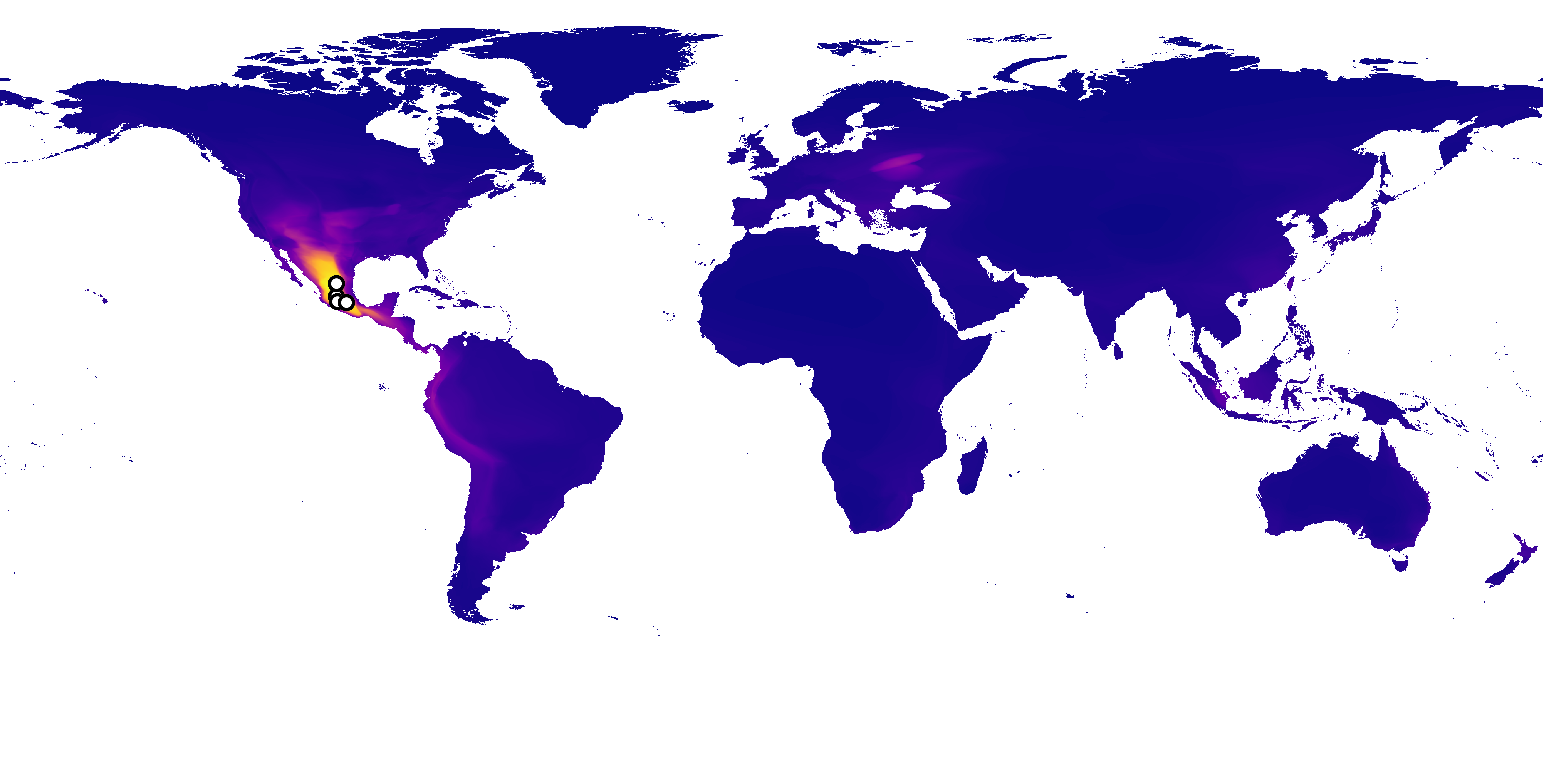}
    \end{minipage}

    \vspace{1em}

    \begin{minipage}{0.04\textwidth}
        \rotatebox{90}{\tiny\textbf{20 Context}}
    \end{minipage}%
    \hspace{0.5em}
    \begin{minipage}{0.29\textwidth}
        \centering
        \includegraphics[width=\linewidth]{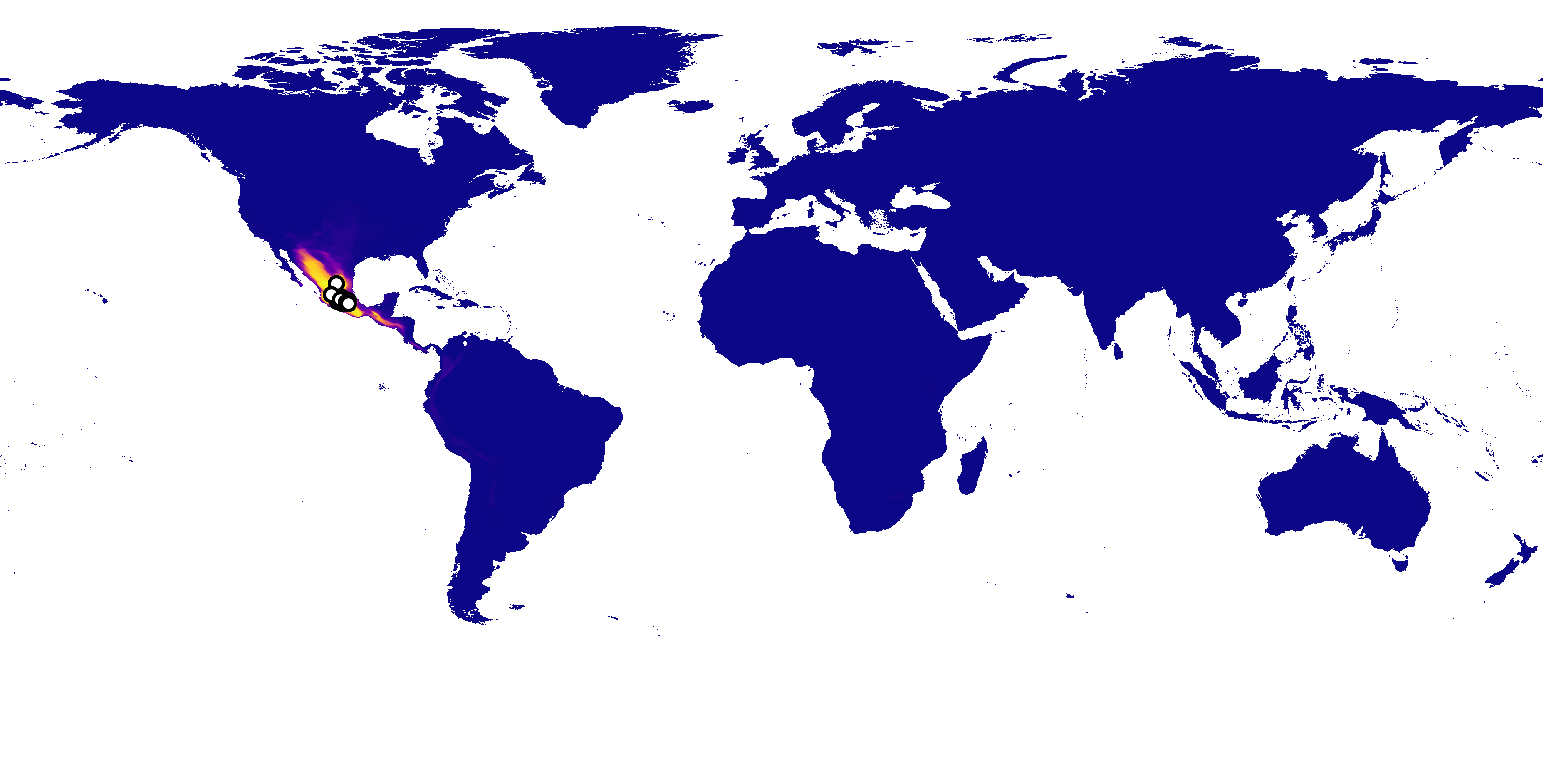}
    \end{minipage}%
    \hspace{0.5em}
    \begin{minipage}{0.29\textwidth}
        \centering
        \includegraphics[width=\linewidth]{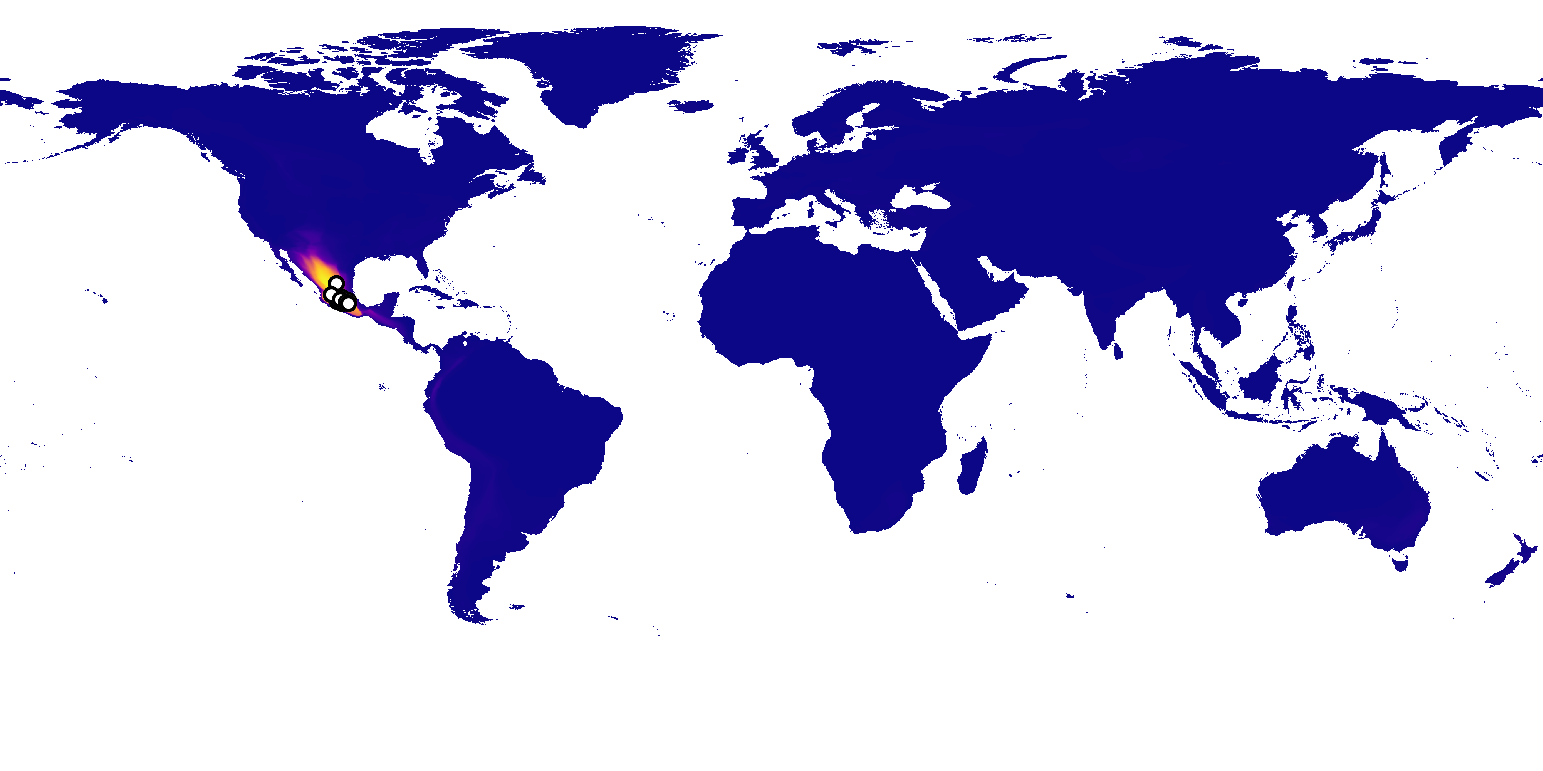}
    \end{minipage}%
    \hspace{0.5em}
    \begin{minipage}{0.29\textwidth}
        \centering
        \includegraphics[width=\linewidth]{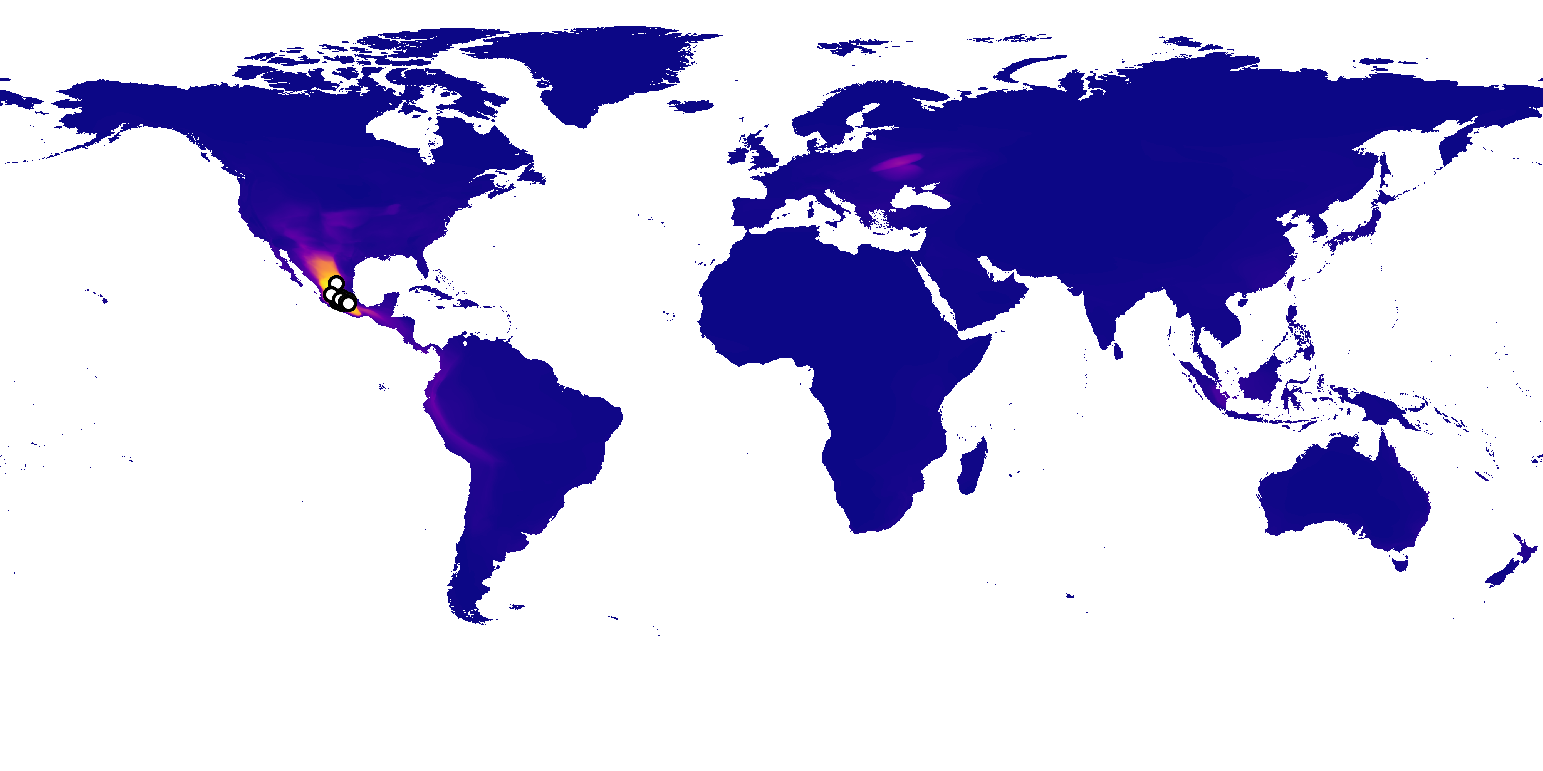}
    \end{minipage}
    \vspace{-15pt}
    \caption{\change{\textbf{Comparing estimated ranges across models.}
    Here we see few-shot range estimates produced by FS-SINR, LE-SINR, and SINR for the \texttt{Crevice Swift} lizard, with expert-derived range in Mexico inset.
    No text is provided and so no sensible zero-shot prediction can be made for any model.
    However while LE-SINR and SINR cannot produce an output for this and so we show a blank map, FS-SINR can generate a predicted range just from feeding the learned CLS and register tokens with no other information into the transformer encoder.
    The range that is produced is contained within the model or the learned tokens itself rather than from any further inputs.
    Absent additional information, the model seems to guide predictions towards areas where it as seen many species during training \eg Europe and North America.
    This may be an unhelpful bias when attempting to model novel species. SINR again produces more diffuse ranges than the other methods, though all approaches struggle to model these small ranges, as seen in \cref{range_size_section}.}}
    \label{fig:qualitative_different_models_no_text}
\end{figure}

In \cref{fig:qualitative_different_models_range} we show range estimates for the \texttt{Brown-banded Watersnake}, using `range' text for FS-SINR and LE-SINR approaches.
In \cref{fig:qualitative_different_models_habitat} we show range estimates for the \texttt{Brown-headed Honeyeater}, using `habitat' text for FS-SINR and LE-SINR approaches.
Finally in \cref{fig:qualitative_different_models_no_text} we show range estimates for the \texttt{Crevice Swift}, without providing text.
Overall, SINR produces more diffuse ranges and requires more locations to narrow down the range.
LE-SINR and FS-SINR appear to have very different zero-shot behaviors, with LE-SINR frequently seeming to predict presence in almost no locations at all, while FS-SINR tends to produce a zero-shot range that is too large.

\begin{figure}[h]
    \centering
    \begin{minipage}{0.04\textwidth}
    \end{minipage}%
    \hspace{0.5em}
    \begin{minipage}{0.29\textwidth}
        \centering \textbf{Run 1}
    \end{minipage}%
    \hspace{0.5em}
    \begin{minipage}{0.29\textwidth}
        \centering \textbf{Run 2}
    \end{minipage}%
    \hspace{0.5em}
    \begin{minipage}{0.29\textwidth}
        \centering \textbf{Run 3}
    \end{minipage}
    
    \vspace{1em}

    \begin{minipage}{0.04\textwidth}
        \rotatebox{90}{\textbf{Range}}
    \end{minipage}%
    \hspace{0.5em}
    \begin{minipage}{0.29\textwidth}
        \centering
                            \begin{overpic}[trim={0 0 0 0},clip,width=\linewidth]{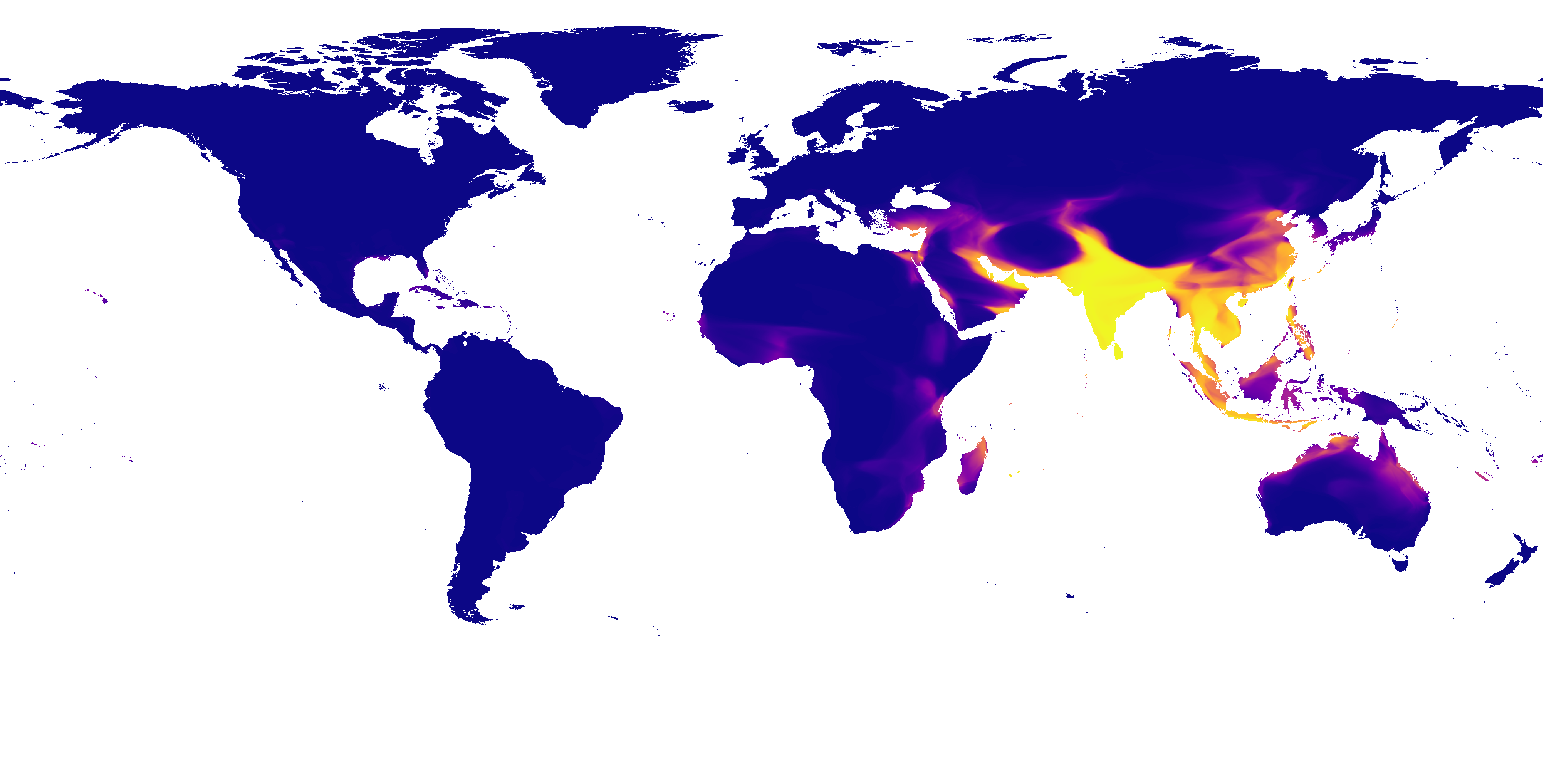}
        \put(0,0){%
          {%
            \setlength\fboxsep{0pt}%
            \setlength\fboxrule{1pt}%
            \fbox{%
              \includegraphics[width=0.25\linewidth]{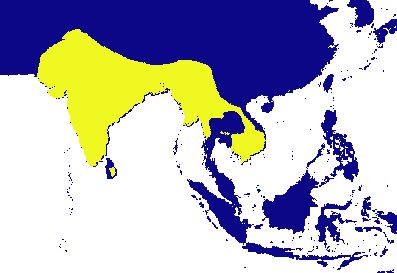}%
            }%
          }%
        }
    \end{overpic}
    \end{minipage}%
    \hspace{0.5em}
    \begin{minipage}{0.29\textwidth}
        \centering
        \includegraphics[width=\linewidth]{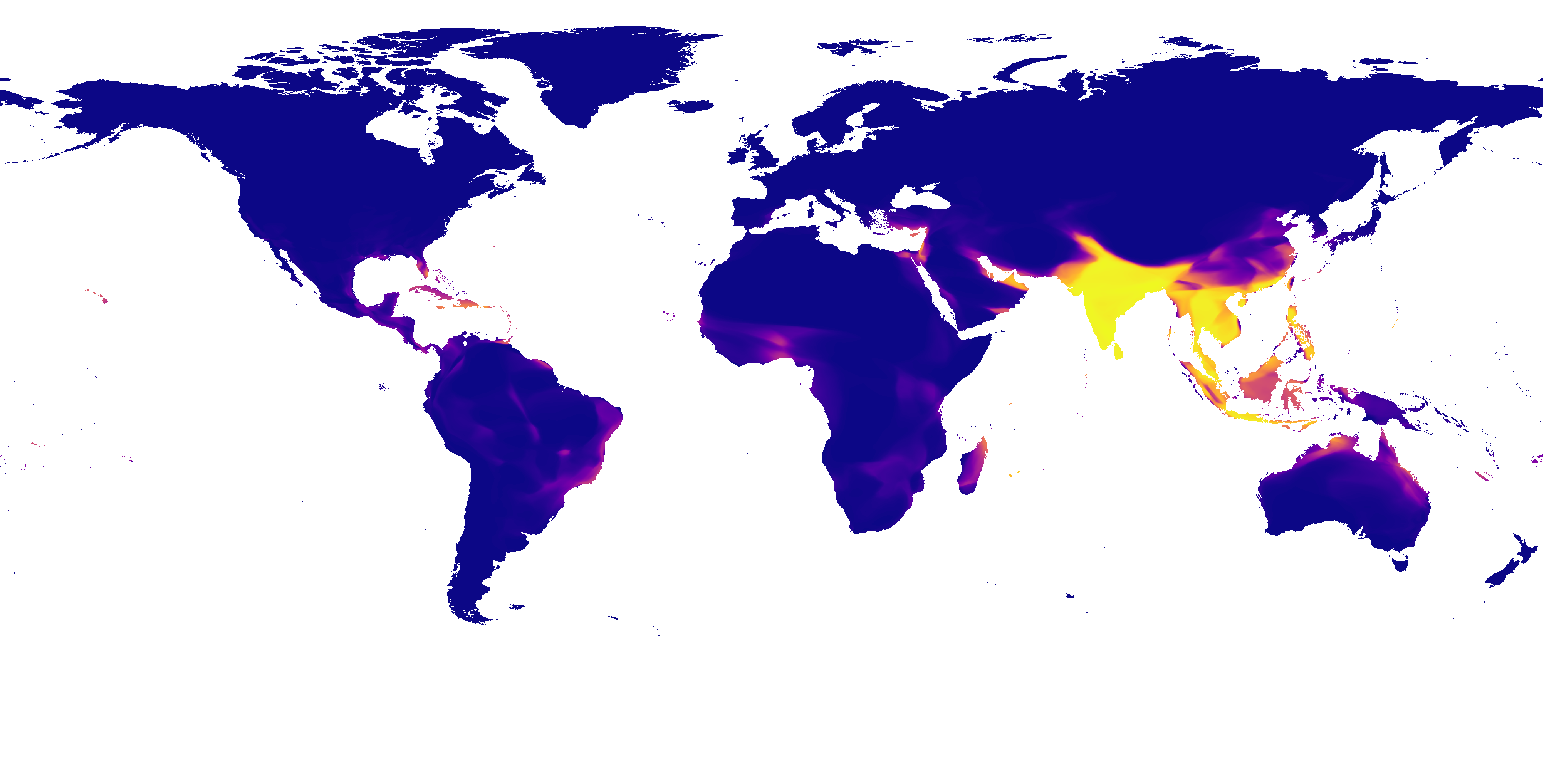}
    \end{minipage}%
    \hspace{0.5em}
    \begin{minipage}{0.29\textwidth}
        \centering
        \includegraphics[width=\linewidth]{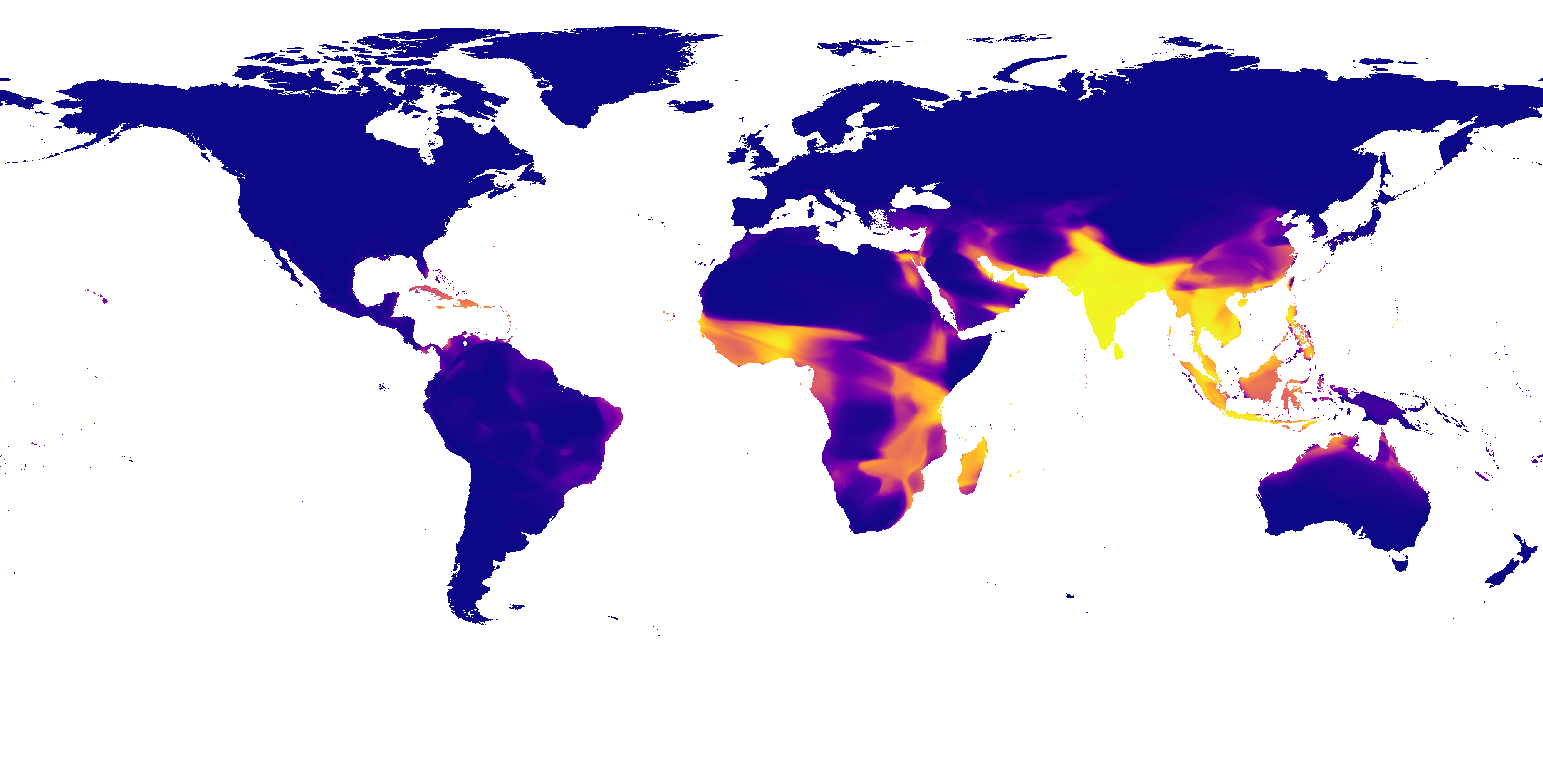}
    \end{minipage}

    \vspace{1em}

    \begin{minipage}{0.04\textwidth}
        \rotatebox{90}{\textbf{Habitat}}
    \end{minipage}%
    \hspace{0.5em}
    \begin{minipage}{0.29\textwidth}
        \centering
        \includegraphics[width=\linewidth]{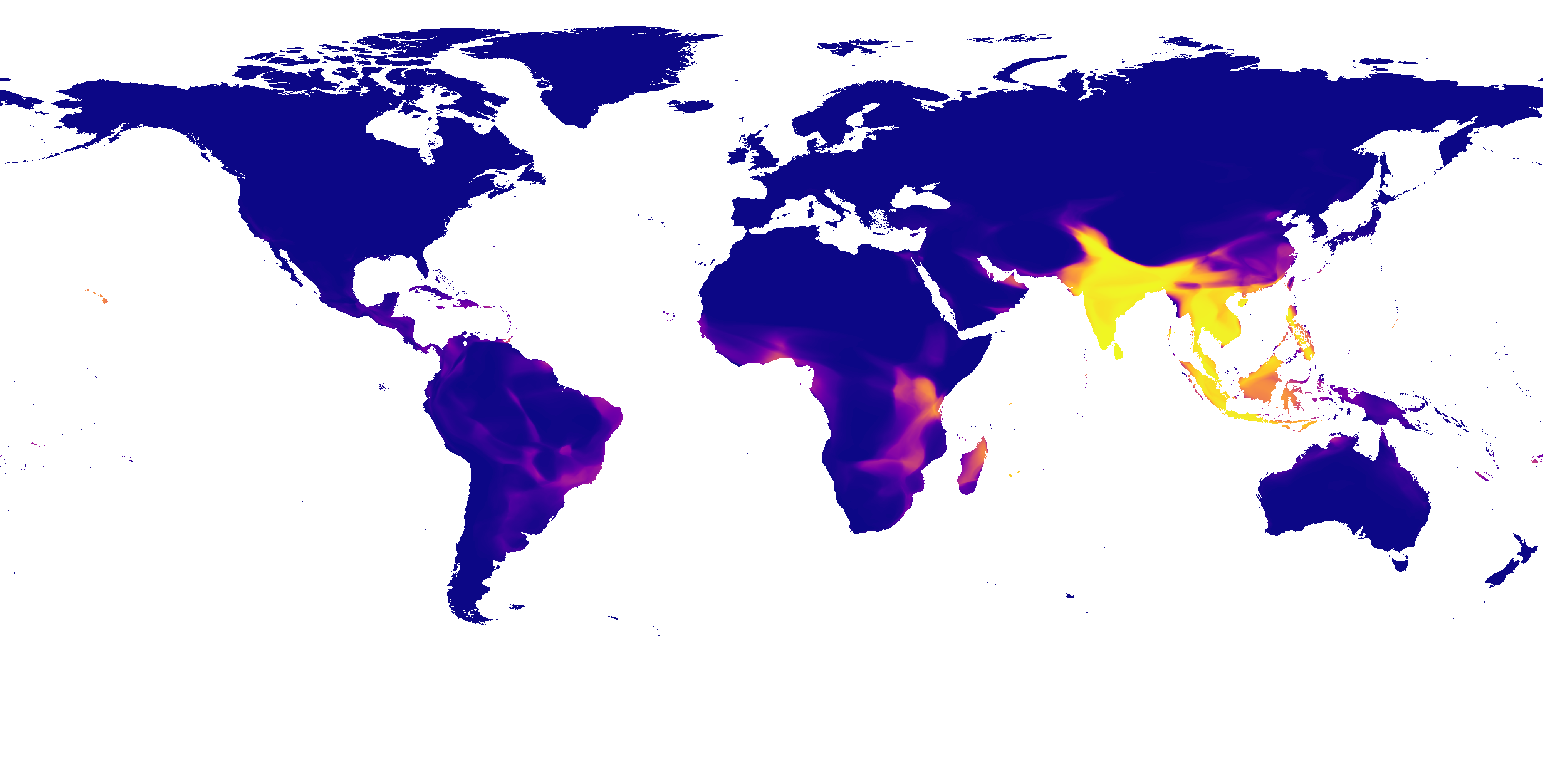}
    \end{minipage}%
    \hspace{0.5em}
    \begin{minipage}{0.29\textwidth}
        \centering
        \includegraphics[width=\linewidth]{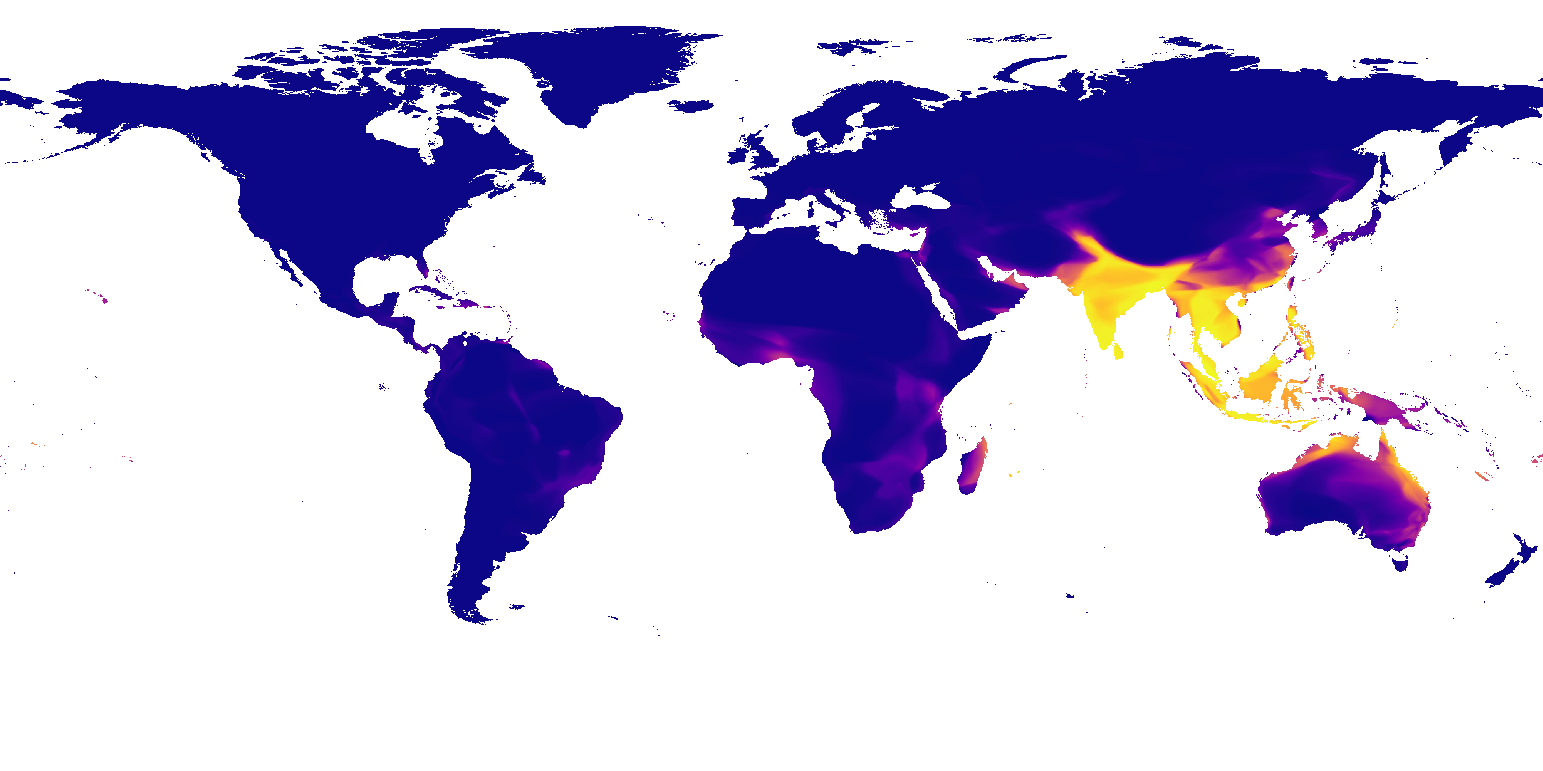}
    \end{minipage}%
    \hspace{0.5em}
    \begin{minipage}{0.29\textwidth}
        \centering
        \includegraphics[width=\linewidth]{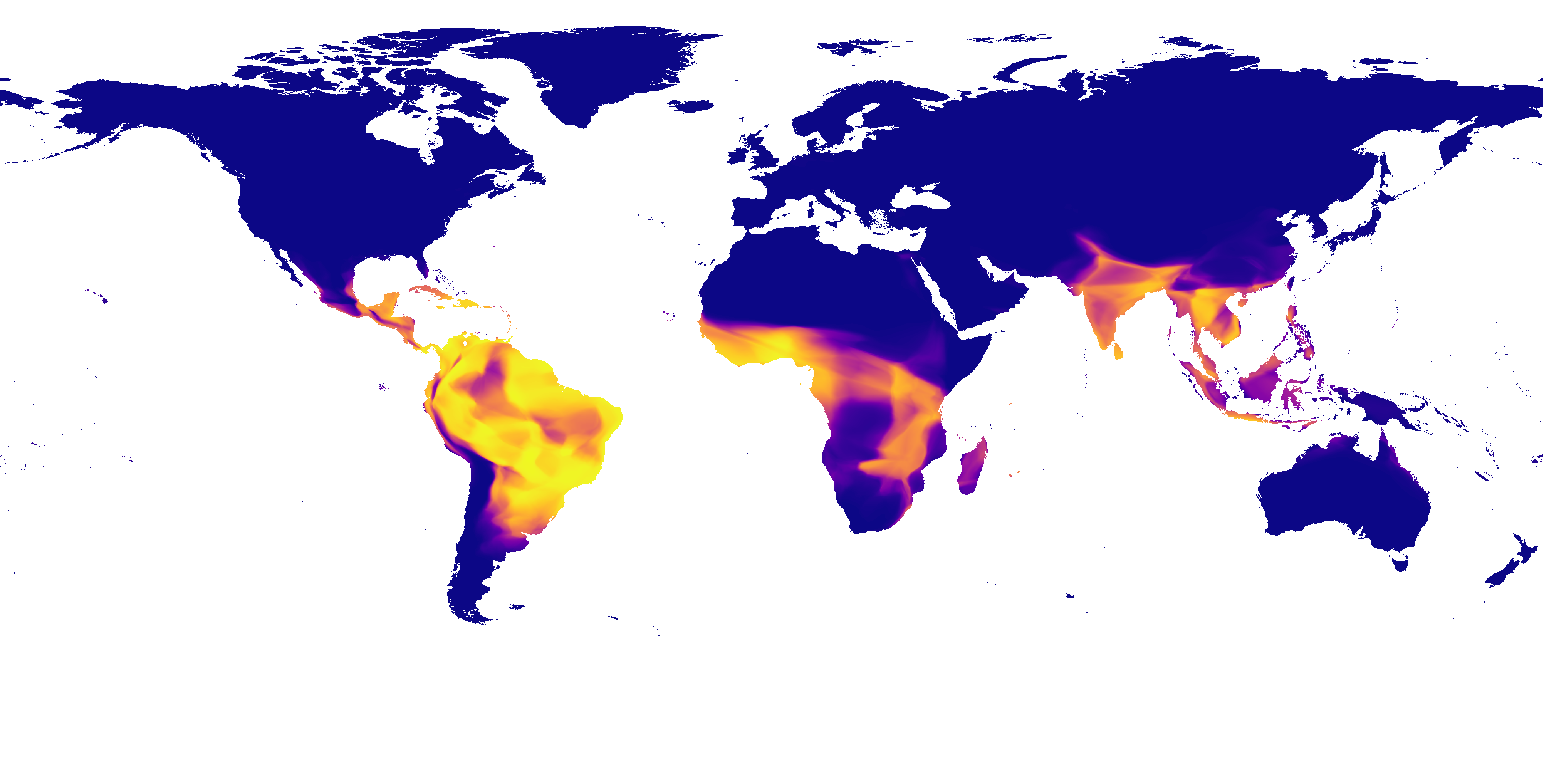}
    \end{minipage}

    \vspace{1em}

    \begin{minipage}{0.04\textwidth}
        \rotatebox{90}{\textbf{No Text}}
    \end{minipage}%
    \hspace{0.5em}
    \begin{minipage}{0.29\textwidth}
        \centering
        \includegraphics[width=\linewidth]{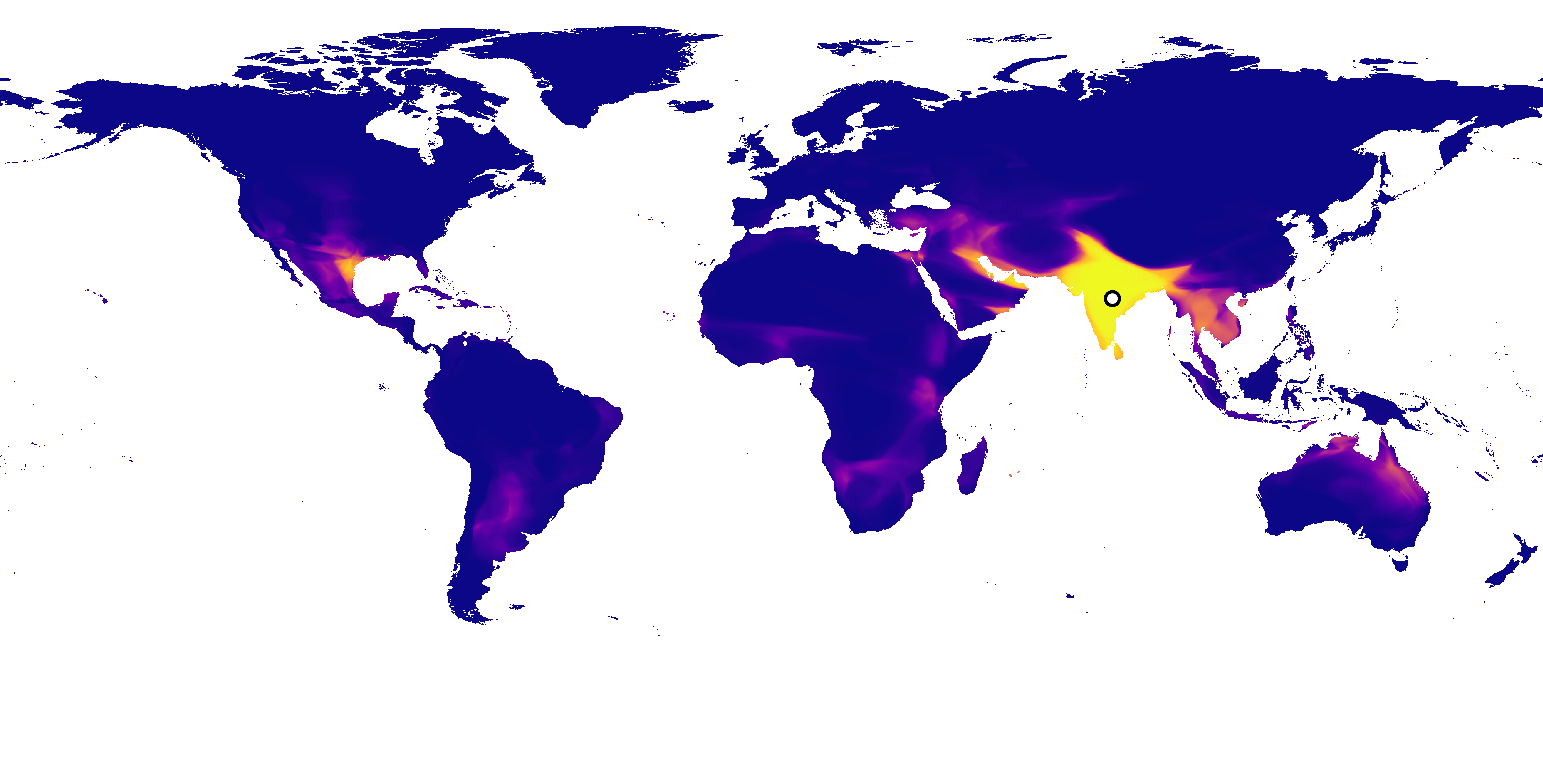}
    \end{minipage}%
    \hspace{0.5em}
    \begin{minipage}{0.29\textwidth}
        \centering
        \includegraphics[width=\linewidth]{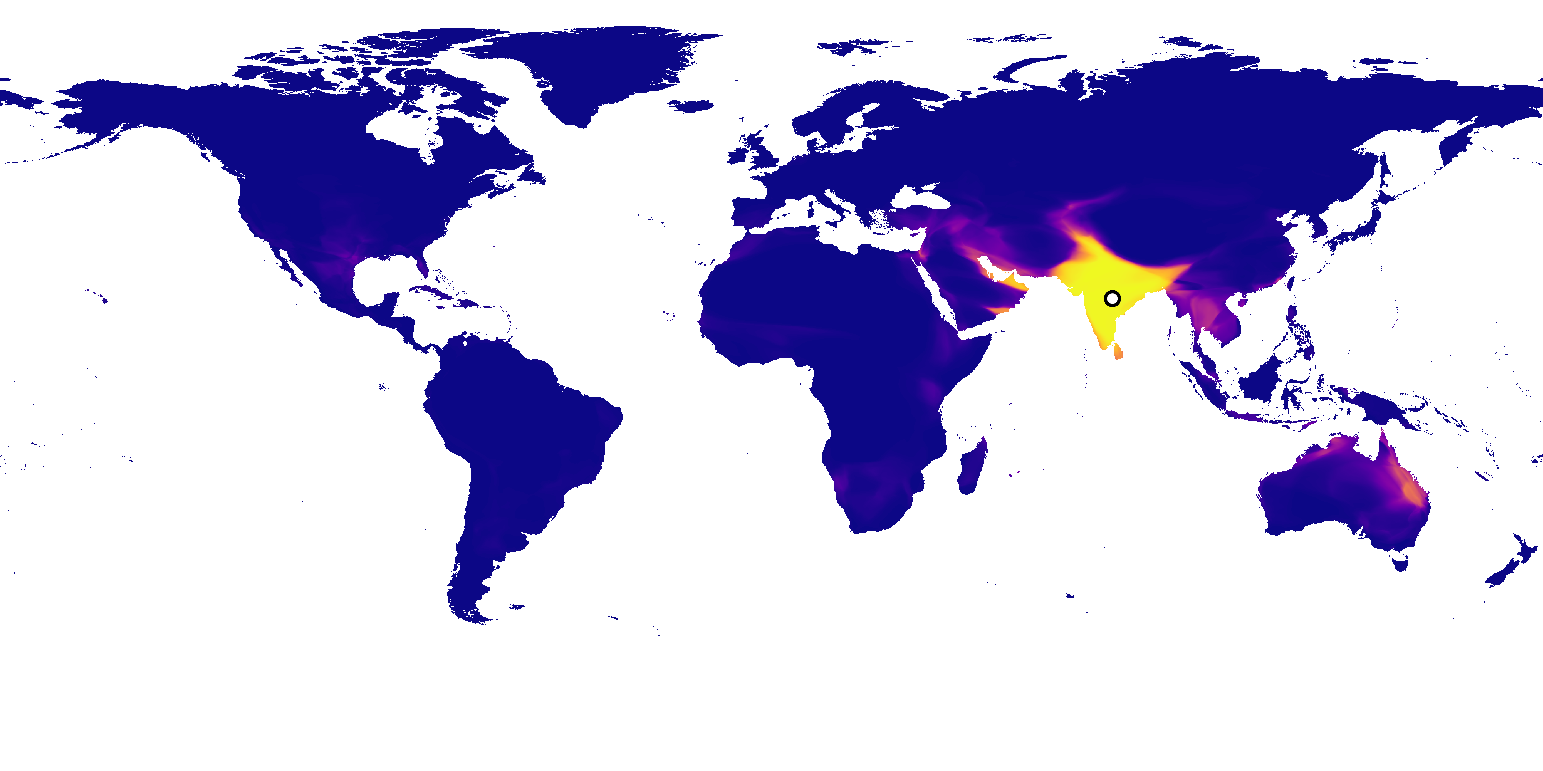}
    \end{minipage}%
    \hspace{0.5em}
    \begin{minipage}{0.29\textwidth}
        \centering
        \includegraphics[width=\linewidth]{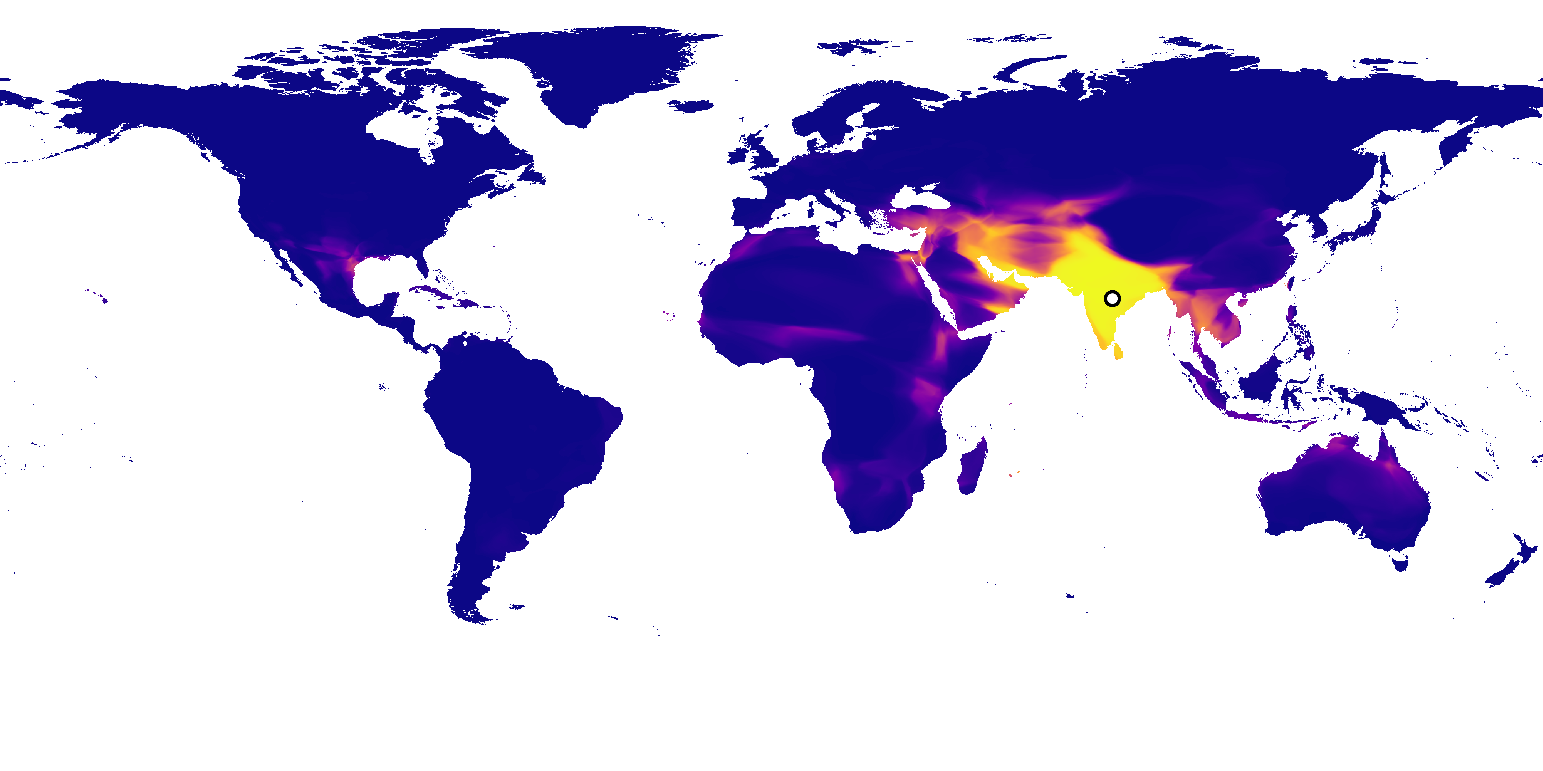}
    \end{minipage}
    \vspace{-15pt}
    \caption{\change{\textbf{Impact of random initialization on \modelname.}
    Here we display range estimates for the \texttt{Yellow-footed Green Pigeon} from three different \modelname models where different random seeds were used to initialize each model during training.
    We show zero-shot results using `range text' (top) and `habitat text' (middle), and also few-shot results using one context location with no text (bottom).
    The IUCN expert-derived range is shown inset. We see that even when provided with the same inputs, different models can perform very differently when this input is very sparse (\eg just text or one context location).
    While most of the Indian part of the actual range is included for all input types and runs, there is significant variability across the runs in other geographic areas.\\ 
    \textit{Range Text:} ``The yellow-footed green pigeon is found in the Indian subcontinent and parts of Southeast Asia. It is the state bird of Maharashtra."\\
    \textit{Habitat Text:} ``The species is a habitat generalist, preferring dense forest areas with emergent trees, especially Banyan trees, but can also be spotted in natural remnants in urban areas. They forage in flocks and are often seen sunning on the tops of trees in the early morning.'' 
    }}
    \label{fig:different_seeds}
\end{figure}

In \cref{fig:different_seeds} we visualize FS-SINR range predictions for the \texttt{Yellow-footed Green Pigeon} for models that have had different random initializations (\ie different random seeds).
We observe that there is a relatively large amount of variance in the outputs produced given the same input data. 
The same set of input context locations could represent many different possible output ranges, and thus being able to represent this variety is advantageous.
We also utilize these different predictions from different seeds for ensembling and uncertainty quantification in \cref{sec:addtional_experiments}.

\end{document}